\documentclass[10pt,twocolumn,letterpaper]{article}

\usepackage[pagenumbers]{preprint}

\newcommand{\chexbottomfloat}{\captionsetup{belowskip=0pt}}
\newcommand{\chexraggedpage}{5}
\makeatletter
\def\@textbottom{\ifnum\c@page=\chexraggedpage\relax\vskip\z@\@plus.0001fil\fi}
\makeatother
\newlength{\chexpagefivegap}
\newcommand{\chexcompactrules}{\setlength{\aboverulesep}{0.3ex}\setlength{\belowrulesep}{0.55ex}}

\usepackage{chexresults}
\usepackage{stfloats}   %

\makeatletter
\newcommand{\chexexternal}[1]{\IfFileExists{#1.tex}{\input{#1}}{}}
\newcommand{\chexxrdef}[3]{\expandafter\gdef\csname chexxr@#1\endcsname{{#2}{#3}}}
\newcommand{\chex@xr}[2]{%
  \begingroup
  \edef\chex@xrlist{\zap@space#2 \@empty}%
  \let\chex@xrext\@empty \let\chex@xrint\@empty \count@\z@
  \@for\chex@xrk:=\chex@xrlist\do{%
    \ifcsname chexxr@\chex@xrk\endcsname
      \advance\count@\@ne
      \ifx\chex@xrext\@empty \let\chex@xrkey\chex@xrk \fi
      \edef\chex@xrext{\ifx\chex@xrext\@empty\else\chex@xrext,\fi\chex@xrk}%
    \else
      \edef\chex@xrint{\ifx\chex@xrint\@empty\else\chex@xrint,\fi\chex@xrk}%
    \fi}%
  \ifnum\count@=\z@
    #1{#2}%
  \else
    \edef\chex@xrtype{\expandafter\expandafter\expandafter\@firstoftwo\csname chexxr@\chex@xrkey\endcsname}%
    \csname\chex@xrpre @\chex@xrtype @name\ifnum\count@>\@ne @plural\fi\endcsname~%
    \@tempcnta\z@
    \@for\chex@xrk:=\chex@xrext\do{%
      \advance\@tempcnta\@ne
      \ifnum\@tempcnta>\@ne
        \ifnum\count@=\tw@ \crefpairconjunction
        \else\ifnum\@tempcnta=\count@ \creflastconjunction\else\crefmiddleconjunction\fi\fi
      \fi
      \expandafter\expandafter\expandafter\@secondoftwo\csname chexxr@\chex@xrk\endcsname}%
    \ifx\chex@xrint\@empty\else \crefpairconjunction \expandafter#1\expandafter{\chex@xrint}\fi
  \fi
  \endgroup}
\AtBeginDocument{%
  \expandafter\let\expandafter\chex@cref\csname cref \endcsname
  \expandafter\let\expandafter\chex@Cref\csname Cref \endcsname
  \DeclareRobustCommand{\cref}{\@ifstar{\chex@cref*}{\def\chex@xrpre{cref}\chex@xr\chex@cref}}%
  \DeclareRobustCommand{\Cref}{\@ifstar{\chex@Cref*}{\def\chex@xrpre{Cref}\chex@xr\chex@Cref}}}
\makeatother

\makeatletter
\@ifundefined{testFirstNumberedPage}{}{%
  \newif\ifchex@LNrestarted
  \def\testFirstNumberedPage#1{\ifnum#1>\c@linenumber
      \ifchex@LNrestarted
        \let\nextLN\@gobble \def\pageLN{\gotNumberedPage{#1}}%
      \else
        \global\chex@LNrestartedtrue
        \def\nextLN##1{\testNextNumberedPage\LN@Pfirst}%
      \fi
    \else
      \let\nextLN\@gobble \def\pageLN{\gotNumberedPage{#1}}%
    \fi}
  \def\getLineNumber{\global\chex@LNrestartedfalse \testNumberedPage}}
\makeatother
\usepackage{afterpage}
\newcommand{\chexpinpage}{7}
\newcommand{\chexlatefloatsep}{22pt}
\newcommand{\chexlatepage}{5}
\newcommand{\chexdatafigsep}{5pt}
\makeatletter
\newcommand{\chexlatefloats}{%
  \global\dbltextfloatsep=\dimexpr\chexlatefloatsep\relax plus 2pt minus 2pt\relax
  \afterpage{\chex@late}}
\newcommand{\chex@late}{%
  \ifnum\value{page}<\chexlatepage\relax
    \afterpage{\chex@late}%
  \else
    \global\textfloatsep=\dimexpr\chexlatefloatsep\relax plus 2pt minus 2pt\relax
  \fi}
\newcommand{\chexpinfloats}{\afterpage{\chex@pin}}
\newcommand{\chex@pin}{%
  \ifnum\value{page}<\chexpinpage\relax
    \afterpage{\chex@pin}%
  \else
    \begin{table}[t]
\centering
\scriptsize
\setlength{\tabcolsep}{1.5pt}
\caption{\textbf{Fusing finding instances across images.} All rows fuse the same candidates and are scored by detection F1 against the reference parse. A study without a lateral keeps its frontal image.\chexshadenote{ Shading is as in \cref{tab:core-multi-image}, against the row Single image.}}
\label{tab:core-fusion}
\resizebox{\linewidth}{!}{%
\begin{tabular}{@{}l cc cc cc@{}}
\toprule
 & \multicolumn{2}{c}{Single frontal} & \multicolumn{2}{c}{Frontal+lateral} & \multicolumn{2}{c}{All images} \\
\cmidrule(lr){2-3}\cmidrule(lr){4-5}\cmidrule(l){6-7}
Fusion & miF1$\uparrow$ & maF1$\uparrow$ & miF1$\uparrow$ & maF1$\uparrow$ & miF1$\uparrow$ & maF1$\uparrow$ \\
\midrule
\chexnorankrow Single image & \chexciu{0.488}{.471}{.503} & \chexci{0.200}{.183}{.206} & \chexci{0.488}{.471}{.503} & \chexci{0.200}{.183}{.206} & \chexci{0.488}{.471}{.503} & \chexci{0.200}{.183}{.206} \\
\chexnorankrow Lateral if available & \chexciu{0.488}{.471}{.503} & \chexci{0.200}{.183}{.206} & \chexvsdown{28} \chexci{0.478}{.462}{.494} & \chexvsdown{28} \chexci{0.182}{.168}{.188} & \chexvsdown{28} \chexci{0.478}{.462}{.494} & \chexvsdown{28} \chexci{0.182}{.168}{.188} \\
\midrule
Concatenation & \chexciu{0.488}{.471}{.503} & \chexci{0.200}{.183}{.206} & \chexvsdown{60} \chexci{0.440}{.424}{.454} & \chexvsdown{60} \chexci{0.159}{.146}{.165} & \chexvsdown{60} \chexci{0.429}{.414}{.443} & \chexvsdown{60} \chexci{0.155}{.142}{.161} \\
\chexrowrule{2-7}
Union & \chexciu{0.488}{.471}{.503} & \chexci{0.200}{.183}{.206} & \chexvsdown{60} \chexci{0.462}{.446}{.477} & \chexvsdown{60} \chexci{0.177}{.162}{.184} & \chexvsdown{60} \chexci{0.463}{.447}{.479} & \chexvsdown{60} \chexci{0.169}{.156}{.176} \\
\chexrowrule{2-7}
Union per name & \chexciu{0.488}{.470}{.503} & \chexci{0.200}{.183}{.207} & \chexvsdown{28} \chexci{0.484}{.466}{.500} & \chexci{0.195}{.180}{.201} & \chexci{0.485}{.468}{.501} & \chexci{0.198}{.183}{.205} \\
\chexrowrule{2-7}
Voting & \chexciu{0.488}{.470}{.503} & \chexci{0.200}{.183}{.207} & \chexvsdown{28} \chexci{0.483}{.466}{.500} & \chexci{0.192}{.177}{.199} & \chexci{0.485}{.468}{.501} & \chexci{0.195}{.179}{.202} \\
\chexrowrule{2-7}
WBF~\cite{solovyev2021wbf} & \chexci{0.487}{.470}{.503} & \chexvsdown{28} \chexci{0.198}{.182}{.205} & \chexvsdown{28} \chexci{0.484}{.468}{.499} & \chexci{0.202}{.186}{.208} & \chexci{0.486}{.470}{.501} & \chexci{0.205}{.189}{.210} \\
\chexrowrule{2-7}
ProbEn~\cite{chen2022proben} & \chexciu{0.488}{.470}{.503} & \chexci{0.200}{.183}{.207} & \chexci{0.485}{.468}{.501} & \chexvsdown{28} \chexci{0.191}{.174}{.197} & \chexci{0.486}{.470}{.502} & \chexci{0.194}{.177}{.200} \\
\midrule
Stacking~\cite{wolpert1992stacked} & \chexciu{0.488}{.471}{.503} & \chexci{0.199}{.182}{.206} & \chexci{0.489}{.472}{.504} & \chexvsup{28} \chexci{0.209}{.191}{.217} & \chexciu{0.490}{.474}{.506} & \chexvsup{28} \chexci{0.213}{.196}{.221} \\
\chexrowrule{2-7}
Learned NMS~\cite{hosang2017nms,hu2018relation} & \chexvsdown{28} \chexci{0.482}{.465}{.498} & \chexcib{0.207}{.190}{.214} & \chexvsdown{28} \chexci{0.476}{.459}{.491} & \chexvsup{28} \chexciu{0.210}{.194}{.218} & \chexvsdown{28} \chexci{0.479}{.462}{.493} & \chexci{0.203}{.185}{.210} \\
\chexrowrule{2-7}
Attention MIL~\cite{ilse2018attention} & \chexcib{0.489}{.471}{.505} & \chexciu{0.204}{.187}{.211} & \chexciu{0.491}{.473}{.507} & \chexvsup{28} \chexci{0.207}{.191}{.214} & \chexciu{0.490}{.472}{.506} & \chexvsup{28} \chexcib{0.216}{.200}{.224} \\
\chexrowrule{2-7}
Linking~\cite{varlamova2024mammnet} & \chexcib{0.489}{.472}{.504} & \chexvsdown{28} \chexci{0.198}{.181}{.205} & \chexvsdown{28} \chexci{0.478}{.461}{.495} & \chexvsdown{60} \chexci{0.169}{.155}{.175} & \chexvsdown{28} \chexci{0.480}{.463}{.496} & \chexvsdown{60} \chexci{0.174}{.159}{.181} \\
\midrule
\textbf{Ours} & \chexcib{0.489}{.472}{.505} & \chexci{0.203}{.187}{.211} & \chexvsup{28} \chexcib{0.493}{.476}{.509} & \chexvsup{28} \chexcib{0.211}{.194}{.218} & \chexvsup{28} \chexcib{0.494}{.478}{.510} & \chexvsup{28} \chexciu{0.214}{.197}{.221} \\
\bottomrule
\end{tabular}%
}
\end{table}
    \begin{table}[b]
\chexbottomfloat
\centering
\scriptsize
\setlength{\tabcolsep}{1.5pt}
\caption{\textbf{Localised findings.} $^\dagger$Phrase-grounded per sentence. The shaded row places each category at its usual location.}
\label{tab:core-grounding}
\resizebox{\linewidth}{!}{%
\begin{tabular}{@{}l ccc cccc@{}}
\toprule
 & \multicolumn{3}{c}{MS-CXR~\cite{boecking2022biovil}} & \multicolumn{4}{c}{VinDr-CXR~\cite{nguyen2022vindrcxr}} \\
\cmidrule(lr){2-4}\cmidrule(l){5-8}
Method & Stated$\uparrow$ & Hit$\uparrow$ & Acc\textsubscript{[0,.95]}$\uparrow$ & Stated$\uparrow$ & Hit$\uparrow$ & Acc\textsubscript{[0,.95]}$\uparrow$ & F1$\uparrow$ \\
\midrule
\chexnorankrow \textcolor{gray}{Oracle, prior box} & \chexci{1.000}{1.000}{1.000} & \chexci{0.936}{.893}{.974} & \chexci{0.478}{.439}{.515} & \chexci{1.000}{1.000}{1.000} & \chexci{0.650}{.628}{.672} & \chexci{0.235}{.226}{.245} & -- \\
\midrule
RGRG~\cite{tanida2023rgrg} & \chexciu{0.745}{.659}{.821} & \chexci{0.886}{.825}{.934} & \chexci{0.406}{.350}{.458} & \chexci{0.257}{.239}{.278} & \chexci{0.768}{.735}{.799} & \chexci{0.103}{.094}{.113} & \chexci{0.207}{.194}{.223} \\
\chexrowrule{2-8}
MAIRA-2~\cite{bannur2024maira2} & \chexci{0.713}{.636}{.780} & \chexci{0.910}{.853}{.964} & \chexciu{0.454}{.396}{.509} & \chexci{0.312}{.293}{.332} & \chexciu{0.865}{.838}{.891} & \chexci{0.098}{.090}{.107} & \chexciu{0.342}{.324}{.363} \\
\chexrowrule{2-8}
CheXagent-2$^\dagger$~\cite{chen2024chexagent} & \chexci{0.670}{.587}{.753} & \chexci{0.762}{.669}{.844} & \chexci{0.284}{.231}{.330} & \chexci{0.282}{.261}{.304} & \chexci{0.795}{.763}{.830} & \chexciu{0.124}{.112}{.137} & \chexci{0.341}{.320}{.363} \\
\chexrowrule{2-8}
CheXOne$^\dagger$~\cite{chexone} & \chexci{0.660}{.569}{.736} & \chexciu{0.944}{.897}{.981} & \chexci{0.368}{.309}{.423} & \chexciu{0.315}{.297}{.333} & \chexci{0.793}{.764}{.822} & \chexci{0.107}{.098}{.116} & \chexci{0.333}{.316}{.352} \\
\chexrowrule{2-8}
ChEX~\cite{muller2024chex} & \chexcib{0.798}{.739}{.852} & \chexci{0.853}{.784}{.913} & \chexci{0.386}{.343}{.423} & \chexci{0.243}{.225}{.261} & \chexci{0.709}{.671}{.745} & \chexci{0.096}{.087}{.105} & \chexci{0.140}{.129}{.152} \\
\midrule
\textbf{Ours} & \chexcib{0.798}{.718}{.862} & \chexcib{1.000}{1.000}{1.000} & \chexcib{0.528}{.457}{.592} & \chexcib{0.387}{.368}{.406} & \chexcib{0.963}{.950}{.976} & \chexcib{0.126}{.117}{.135} & \chexcib{0.359}{.343}{.376} \\
\bottomrule
\end{tabular}%
}
\end{table}
    \if@firstcolumn
      \afterpage{\begin{table}[t]
\centering
\footnotesize
\setlength{\tabcolsep}{3pt}
\caption{\textbf{Design choices and the finding set.} Each + row adds to the row above\chextrinote{, and \chextriup/\chextridown{} mark a significant gain or loss against it}. Strict also requires presence and exact regions. R$_{\mathrm{rare}}$ is the recall on names with fewer than 1,000 training mentions.}
\label{tab:core-det-ablation}
\resizebox{\linewidth}{!}{%
\begin{tabular}{@{}l cccc@{}}
\toprule
 & miF1$\uparrow$ & maF1$\uparrow$ & Strict$\uparrow$ & R$_{\mathrm{rare}}$$\uparrow$ \\
\midrule
\chexnorankrow image-blind prior & \chextrislot{}\chexci{0.256}{.240}{.272} & \chextrislot{}\chexci{0.015}{.014}{.015} & n/a & \chextrislot{}\chexci{0.000}{.000}{.000} \\
\chexnorankrow kNN retrieval & \chextrislot{}\chexci{0.413}{.395}{.429} & \chextrislot{}\chexci{0.103}{.092}{.110} & \chextrislot{}\chexci{0.258}{.242}{.273} & \chextrislot{}\chexci{0.007}{.002}{.012} \\
\midrule
end-to-end control, parsed & \chextrislot{}\chexci{0.364}{.348}{.382} & \chextrislot{}\chexci{0.166}{.149}{.174} & \chextrislot{}\chexci{0.253}{.238}{.271} & \chextrislot{}\chexci{0.036}{.024}{.050} \\
\midrule
multi-label classifier (30k) & \chextrislot{}\chexci{0.480}{.463}{.496} & \chextrislot{}\chexci{0.188}{.172}{.195} & n/a & \chextrislot{}\chexci{0.026}{.015}{.040} \\
\chexrowrule{2-5}
+ location, attribute heads & \chextrislot{}\chexci{0.479}{.461}{.495} & \chextrislot{\chextridown}\chexci{0.177}{.161}{.185} & \chextrislot{}\chexci{0.287}{.269}{.306} & \chextrislot{\chextridown}\chexci{0.021}{.012}{.032} \\
\midrule
set predictor (30k) & \chextrislot{}\chexci{0.484}{.465}{.500} & \chextrislot{}\chexci{0.172}{.158}{.179} & n/a & \chextrislot{}\chexci{0.032}{.013}{.053} \\
\chexrowrule{2-5}
+ location & \chextrislot{}\chexci{0.482}{.464}{.498} & \chextrislot{\chextriup}\chexci{0.179}{.166}{.185} & \chextrislot{}\chexci{0.285}{.268}{.303} & \chextrislot{}\chexci{0.037}{.019}{.058} \\
\chexrowrule{2-5}
+ presence, cert., sev. & \chextrislot{\chextriup}\chexci{0.485}{.467}{.500} & \chextrislot{}\chexci{0.182}{.167}{.188} & \chextrislot{\chextriup}\chexci{0.292}{.274}{.311} & \chextrislot{}\chexci{0.038}{.018}{.061} \\
\chexrowrule{2-5}
+ caption & \chextrislot{}\chexci{0.484}{.467}{.500} & \chextrislot{}\chexci{0.180}{.166}{.186} & \chextrislot{\chextriup}\chexci{0.296}{.279}{.314} & \chextrislot{}\chexci{0.040}{.019}{.062} \\
\chexrowrule{2-5}
+ mask & \chextrislot{}\chexci{0.483}{.465}{.498} & \chextrislot{\chextriup}\chexci{0.189}{.174}{.195} & \chextrislot{\chextriup}\chexci{0.308}{.291}{.326} & \chextrislot{\chextriup}\chexciu{0.047}{.025}{.071} \\
\midrule
+ 10k iterations (40k) & \chextrislot{}\chexciu{0.486}{.468}{.502} & \chextrislot{}\chexciu{0.196}{.180}{.204} & \chextrislot{\chextriup}\chexciu{0.321}{.304}{.339} & \chextrislot{}\chexci{0.046}{.030}{.065} \\
\chexrowrule{2-5}
+ 20k iterations (60k) & \chextrislot{}\chexcib{0.488}{.471}{.503} & \chextrislot{}\chexcib{0.200}{.183}{.207} & \chextrislot{\chextriup}\chexcib{0.328}{.313}{.344} & \chextrislot{}\chexcib{0.053}{.034}{.075} \\
\bottomrule
\end{tabular}%
}
\end{table}
\begin{table*}[t]
\centering
\footnotesize
\setlength{\tabcolsep}{3pt}
\caption{\textbf{Writer}, from the reference parse (upper bound) and from predicted findings. The rows below Ref.\ input also train on predicted findings, and indented rows retrain the last unindented row without one input.}
\label{tab:core-verbaliser}
\begin{tabular}{@{}l ccccc ccccc@{}}
\toprule
 & \multicolumn{5}{c}{Reference parse} & \multicolumn{5}{c}{Predicted findings} \\
\cmidrule(lr){2-6}\cmidrule(l){7-11}
 & GREEN$\uparrow$ & RG\textsubscript{ER}$\uparrow$ & miF1$\uparrow$ & maF1$\uparrow$ & RadCliQ$\downarrow$ & GREEN$\uparrow$ & RG\textsubscript{ER}$\uparrow$ & miF1$\uparrow$ & maF1$\uparrow$ & RadCliQ$\downarrow$ \\
\midrule
\chexnorankrow Template & \chexci{0.254}{.246}{.262} & \chexci{0.258}{.252}{.265} & \chexci{0.800}{.790}{.810} & \chexci{0.750}{.732}{.763} & \chexci{1.15}{1.12}{1.17} & \chexci{0.144}{.136}{.153} & \chexci{0.168}{.161}{.174} & \chexci{0.598}{.585}{.611} & \chexci{0.414}{.393}{.432} & \chexci{1.50}{1.48}{1.52} \\
\chexnorankrow + negatives & \chexci{0.357}{.346}{.369} & \chexci{0.309}{.303}{.315} & \chexci{0.801}{.791}{.811} & \chexci{0.748}{.730}{.762} & \chexci{1.05}{1.03}{1.07} & \chexci{0.274}{.259}{.287} & \chexci{0.228}{.223}{.234} & \chexci{0.596}{.581}{.609} & \chexciu{0.415}{.394}{.433} & \chexci{1.34}{1.31}{1.36} \\
\chexnorankrow Tuned (\textbf{ours}) & \chexciu{0.422}{.410}{.434} & \chexcib{0.360}{.353}{.367} & \chexcib{0.852}{.843}{.861} & \chexcib{0.766}{.749}{.780} & \chexcib{0.76}{.73}{.78} & \chexcib{0.289}{.278}{.300} & \chexcib{0.254}{.247}{.261} & \chexcib{0.626}{.611}{.640} & \chexci{0.409}{.389}{.426} & \chexcib{1.14}{1.11}{1.17} \\
\chexnorankrow \quad w/o loc. & \chexcib{0.424}{.410}{.437} & \chexci{0.291}{.282}{.301} & \chexciu{0.842}{.832}{.852} & \chexciu{0.752}{.736}{.765} & \chexci{0.88}{.85}{.91} & \chexciu{0.288}{.276}{.300} & \chexci{0.213}{.205}{.221} & \chexciu{0.623}{.608}{.637} & \chexci{0.406}{.386}{.424} & \chexci{1.21}{1.18}{1.24} \\
\chexnorankrow \quad w/o cert. \& sev. & \chexci{0.407}{.395}{.418} & \chexciu{0.354}{.347}{.361} & \chexci{0.840}{.830}{.848} & \chexci{0.751}{.733}{.765} & \chexciu{0.78}{.76}{.81} & \chexci{0.287}{.276}{.298} & \chexciu{0.252}{.245}{.259} & \chexcib{0.626}{.611}{.640} & \chexci{0.409}{.389}{.427} & \chexciu{1.15}{1.12}{1.18} \\
\chexnorankrow \quad names only & \chexci{0.402}{.389}{.415} & \chexci{0.273}{.264}{.283} & \chexci{0.801}{.790}{.812} & \chexci{0.717}{.700}{.731} & \chexci{0.95}{.92}{.98} & \chexci{0.283}{.271}{.295} & \chexci{0.208}{.200}{.217} & \chexciu{0.623}{.608}{.636} & \chexcib{0.421}{.398}{.439} & \chexci{1.23}{1.19}{1.26} \\
\midrule
Ref. input (\textbf{ours}) & \chexcib{0.494}{.480}{.507} & \chexcib{0.402}{.395}{.409} & \chexci{0.825}{.814}{.836} & \chexci{0.739}{.720}{.753} & \chexcib{0.63}{.60}{.66} & \chexcib{0.320}{.307}{.334} & \chexci{0.265}{.258}{.273} & \chexci{0.606}{.590}{.623} & \chexci{0.403}{.383}{.422} & \chexcib{1.07}{1.04}{1.10} \\
\midrule
+ pred. input & \chexciu{0.469}{.455}{.484} & \chexciu{0.389}{.381}{.397} & \chexcib{0.828}{.817}{.838} & \chexcib{0.747}{.730}{.760} & \chexciu{0.67}{.64}{.69} & \chexciu{0.308}{.294}{.322} & \chexcib{0.269}{.262}{.277} & \chexci{0.621}{.607}{.635} & \chexciu{0.416}{.394}{.434} & \chexcib{1.07}{1.04}{1.10} \\
\chexrowrule{2-11}
+ alts. & \chexci{0.466}{.451}{.480} & \chexci{0.387}{.379}{.394} & \chexci{0.823}{.812}{.833} & \chexci{0.745}{.728}{.758} & \chexciu{0.67}{.64}{.70} & \chexci{0.305}{.291}{.321} & \chexci{0.266}{.258}{.274} & \chexci{0.623}{.609}{.636} & \chexciu{0.416}{.394}{.435} & \chexciu{1.08}{1.04}{1.11} \\
\chexrowrule{2-11}
\quad w/o loc. & \chexci{0.418}{.403}{.434} & \chexci{0.358}{.351}{.366} & \chexciu{0.826}{.817}{.836} & \chexci{0.738}{.720}{.754} & \chexci{0.75}{.72}{.78} & \chexci{0.294}{.280}{.310} & \chexci{0.259}{.252}{.267} & \chexci{0.623}{.609}{.637} & \chexci{0.413}{.392}{.433} & \chexci{1.09}{1.06}{1.11} \\
\chexrowrule{2-11}
\quad w/o cert. \& sev. & \chexci{0.436}{.420}{.451} & \chexci{0.374}{.366}{.382} & \chexci{0.814}{.802}{.824} & \chexci{0.730}{.711}{.745} & \chexci{0.72}{.69}{.74} & \chexci{0.304}{.289}{.319} & \chexci{0.266}{.258}{.274} & \chexci{0.622}{.607}{.636} & \chexci{0.414}{.393}{.434} & \chexci{1.09}{1.05}{1.11} \\
\chexrowrule{2-11}
\quad w/o conf. & \chexci{0.464}{.449}{.478} & \chexci{0.385}{.378}{.393} & \chexci{0.821}{.810}{.832} & \chexciu{0.746}{.727}{.760} & \chexci{0.68}{.65}{.71} & \chexci{0.305}{.292}{.321} & \chexciu{0.268}{.261}{.276} & \chexciu{0.624}{.610}{.637} & \chexcib{0.420}{.398}{.440} & \chexcib{1.07}{1.04}{1.10} \\
\chexrowrule{2-11}
+ captions & \chexci{0.466}{.451}{.481} & \chexci{0.383}{.375}{.390} & \chexci{0.812}{.801}{.822} & \chexci{0.732}{.716}{.745} & \chexci{0.69}{.66}{.71} & \chexci{0.299}{.285}{.313} & \chexci{0.265}{.258}{.272} & \chexcib{0.625}{.610}{.639} & \chexci{0.409}{.388}{.429} & \chexcib{1.07}{1.04}{1.09} \\
\bottomrule
\end{tabular}
\end{table*}
}%
    \else
    \fi
  \fi}
\makeatother
\input{generated/results}

\chexdefmeta{critical45_mimic_random-report}{backend}{hf}
\chexdefmeta{critical45_mimic_random-report}{dataset}{mimic}
\chexdefmeta{critical45_mimic_random-report}{model}{random-report}
\chexdefmeta{critical45_mimic_random-report}{nconfigs}{9}
\chexdefmeta{critical45_mimic_random-report}{revision}{random-report-v1}
\chexdefmeta{critical45_mimic_random-report}{seed}{0}
\chexdefmeta{critical45_mimic_random-report}{split}{test}
\chexdefmeta{critical45_mimic_random-report}{temperature}{0.0}
\chexdefmeta{critical_mimic_chexagent-2-3b-native}{backend}{hf}
\chexdefmeta{critical_mimic_chexagent-2-3b-native}{dataset}{mimic}
\chexdefmeta{critical_mimic_chexagent-2-3b-native}{model}{chexagent-2-3b-native}
\chexdefmeta{critical_mimic_chexagent-2-3b-native}{nconfigs}{9}
\chexdefmeta{critical_mimic_chexagent-2-3b-native}{revision}{8f19b53a2eceda4c33b0acec6c81fbc293ad80d0}
\chexdefmeta{critical_mimic_chexagent-2-3b-native}{seed}{0}
\chexdefmeta{critical_mimic_chexagent-2-3b-native}{split}{test+val}
\chexdefmeta{critical_mimic_chexagent-2-3b-native}{temperature}{0.0}
\chexdefmeta{critical_mimic_chexagent-2-3b-native-notags}{backend}{hf}
\chexdefmeta{critical_mimic_chexagent-2-3b-native-notags}{dataset}{mimic}
\chexdefmeta{critical_mimic_chexagent-2-3b-native-notags}{model}{chexagent-2-3b-native}
\chexdefmeta{critical_mimic_chexagent-2-3b-native-notags}{nconfigs}{9}
\chexdefmeta{critical_mimic_chexagent-2-3b-native-notags}{revision}{8f19b53a2eceda4c33b0acec6c81fbc293ad80d0}
\chexdefmeta{critical_mimic_chexagent-2-3b-native-notags}{seed}{0}
\chexdefmeta{critical_mimic_chexagent-2-3b-native-notags}{split}{test+val}
\chexdefmeta{critical_mimic_chexagent-2-3b-native-notags}{temperature}{0.0}
\chexdefmeta{critical_mimic_chexone-instruct}{backend}{vllm}
\chexdefmeta{critical_mimic_chexone-instruct}{dataset}{mimic}
\chexdefmeta{critical_mimic_chexone-instruct}{maxtokens}{2048}
\chexdefmeta{critical_mimic_chexone-instruct}{model}{chexone-instruct}
\chexdefmeta{critical_mimic_chexone-instruct}{nconfigs}{9}
\chexdefmeta{critical_mimic_chexone-instruct}{revision}{0c350e6852ea08f9d9baf3b7595c1a10d4849927}
\chexdefmeta{critical_mimic_chexone-instruct}{seed}{0}
\chexdefmeta{critical_mimic_chexone-instruct}{split}{test+val}
\chexdefmeta{critical_mimic_chexone-instruct}{temperature}{0.0}
\chexdefmeta{critical_mimic_chexone-reasoning}{backend}{vllm}
\chexdefmeta{critical_mimic_chexone-reasoning}{dataset}{mimic}
\chexdefmeta{critical_mimic_chexone-reasoning}{maxtokens}{2048}
\chexdefmeta{critical_mimic_chexone-reasoning}{model}{chexone-reasoning}
\chexdefmeta{critical_mimic_chexone-reasoning}{nconfigs}{9}
\chexdefmeta{critical_mimic_chexone-reasoning}{revision}{0c350e6852ea08f9d9baf3b7595c1a10d4849927}
\chexdefmeta{critical_mimic_chexone-reasoning}{seed}{0}
\chexdefmeta{critical_mimic_chexone-reasoning}{split}{test+val}
\chexdefmeta{critical_mimic_chexone-reasoning}{temperature}{0.0}
\chexdefmeta{critical_mimic_constant-normal}{backend}{hf}
\chexdefmeta{critical_mimic_constant-normal}{dataset}{mimic}
\chexdefmeta{critical_mimic_constant-normal}{model}{constant-normal}
\chexdefmeta{critical_mimic_constant-normal}{nconfigs}{9}
\chexdefmeta{critical_mimic_constant-normal}{revision}{constant-normal-v1}
\chexdefmeta{critical_mimic_constant-normal}{seed}{0}
\chexdefmeta{critical_mimic_constant-normal}{split}{test+val}
\chexdefmeta{critical_mimic_constant-normal}{temperature}{0.0}
\chexdefmeta{critical_mimic_copy-indication}{backend}{hf}
\chexdefmeta{critical_mimic_copy-indication}{dataset}{mimic}
\chexdefmeta{critical_mimic_copy-indication}{model}{copy-indication}
\chexdefmeta{critical_mimic_copy-indication}{nconfigs}{9}
\chexdefmeta{critical_mimic_copy-indication}{revision}{copy-indication-v1}
\chexdefmeta{critical_mimic_copy-indication}{seed}{0}
\chexdefmeta{critical_mimic_copy-indication}{split}{test+val}
\chexdefmeta{critical_mimic_copy-indication}{temperature}{0.0}
\chexdefmeta{critical_mimic_copy-prior}{backend}{hf}
\chexdefmeta{critical_mimic_copy-prior}{dataset}{mimic}
\chexdefmeta{critical_mimic_copy-prior}{model}{copy-prior}
\chexdefmeta{critical_mimic_copy-prior}{nconfigs}{9}
\chexdefmeta{critical_mimic_copy-prior}{revision}{copy-prior-v1}
\chexdefmeta{critical_mimic_copy-prior}{seed}{0}
\chexdefmeta{critical_mimic_copy-prior}{split}{test+val}
\chexdefmeta{critical_mimic_copy-prior}{temperature}{0.0}
\chexdefmeta{critical_mimic_cxr-llava-v2}{backend}{hf}
\chexdefmeta{critical_mimic_cxr-llava-v2}{dataset}{mimic}
\chexdefmeta{critical_mimic_cxr-llava-v2}{model}{cxr-llava-v2}
\chexdefmeta{critical_mimic_cxr-llava-v2}{nconfigs}{9}
\chexdefmeta{critical_mimic_cxr-llava-v2}{revision}{b2224786bb90d54b1e1291171866706cfbb44e2b}
\chexdefmeta{critical_mimic_cxr-llava-v2}{seed}{0}
\chexdefmeta{critical_mimic_cxr-llava-v2}{split}{test+val}
\chexdefmeta{critical_mimic_cxr-llava-v2}{temperature}{0.0}
\chexdefmeta{critical_mimic_empty}{backend}{hf}
\chexdefmeta{critical_mimic_empty}{dataset}{mimic}
\chexdefmeta{critical_mimic_empty}{model}{empty}
\chexdefmeta{critical_mimic_empty}{nconfigs}{9}
\chexdefmeta{critical_mimic_empty}{revision}{empty-v1}
\chexdefmeta{critical_mimic_empty}{seed}{0}
\chexdefmeta{critical_mimic_empty}{split}{test+val}
\chexdefmeta{critical_mimic_empty}{temperature}{0.0}
\chexdefmeta{critical_mimic_internvl3-8b}{backend}{vllm}
\chexdefmeta{critical_mimic_internvl3-8b}{dataset}{mimic}
\chexdefmeta{critical_mimic_internvl3-8b}{maxtokens}{768}
\chexdefmeta{critical_mimic_internvl3-8b}{model}{internvl3-8b}
\chexdefmeta{critical_mimic_internvl3-8b}{nconfigs}{9}
\chexdefmeta{critical_mimic_internvl3-8b}{revision}{853e3a797a661694b1b8ece0cb72dc2b23e3dac9}
\chexdefmeta{critical_mimic_internvl3-8b}{seed}{0}
\chexdefmeta{critical_mimic_internvl3-8b}{split}{test+val}
\chexdefmeta{critical_mimic_internvl3-8b}{temperature}{0.0}
\chexdefmeta{critical_mimic_llava-med}{backend}{hf}
\chexdefmeta{critical_mimic_llava-med}{dataset}{mimic}
\chexdefmeta{critical_mimic_llava-med}{model}{llava-med}
\chexdefmeta{critical_mimic_llava-med}{nconfigs}{9}
\chexdefmeta{critical_mimic_llava-med}{revision}{91bb16c122001ddc9cf1fd36ce1dae09448943a2}
\chexdefmeta{critical_mimic_llava-med}{seed}{0}
\chexdefmeta{critical_mimic_llava-med}{split}{test+val}
\chexdefmeta{critical_mimic_llava-med}{temperature}{0.0}
\chexdefmeta{critical_mimic_llava-rad}{backend}{hf}
\chexdefmeta{critical_mimic_llava-rad}{dataset}{mimic}
\chexdefmeta{critical_mimic_llava-rad}{model}{llava-rad}
\chexdefmeta{critical_mimic_llava-rad}{nconfigs}{9}
\chexdefmeta{critical_mimic_llava-rad}{revision}{dcdbc6caf6806c9acb66e12a827a39c86e00f6a7}
\chexdefmeta{critical_mimic_llava-rad}{seed}{0}
\chexdefmeta{critical_mimic_llava-rad}{split}{test+val}
\chexdefmeta{critical_mimic_llava-rad}{temperature}{0.0}
\chexdefmeta{critical_mimic_maira-2}{backend}{hf}
\chexdefmeta{critical_mimic_maira-2}{dataset}{mimic}
\chexdefmeta{critical_mimic_maira-2}{model}{maira-2}
\chexdefmeta{critical_mimic_maira-2}{nconfigs}{9}
\chexdefmeta{critical_mimic_maira-2}{revision}{795a2b1cd4a310624b4e3d14b5a23e41fd273deb}
\chexdefmeta{critical_mimic_maira-2}{seed}{0}
\chexdefmeta{critical_mimic_maira-2}{split}{test+val}
\chexdefmeta{critical_mimic_maira-2}{temperature}{0.0}
\chexdefmeta{critical_mimic_medgemma-4b}{backend}{vllm}
\chexdefmeta{critical_mimic_medgemma-4b}{dataset}{mimic}
\chexdefmeta{critical_mimic_medgemma-4b}{maxtokens}{768}
\chexdefmeta{critical_mimic_medgemma-4b}{model}{medgemma-4b}
\chexdefmeta{critical_mimic_medgemma-4b}{nconfigs}{9}
\chexdefmeta{critical_mimic_medgemma-4b}{revision}{290cda5eeccbee130f987c4ad74a59ae6f196408}
\chexdefmeta{critical_mimic_medgemma-4b}{seed}{0}
\chexdefmeta{critical_mimic_medgemma-4b}{split}{test+val}
\chexdefmeta{critical_mimic_medgemma-4b}{temperature}{0.0}
\chexdefmeta{critical_mimic_most-common-report}{backend}{hf}
\chexdefmeta{critical_mimic_most-common-report}{dataset}{mimic}
\chexdefmeta{critical_mimic_most-common-report}{model}{most-common-report}
\chexdefmeta{critical_mimic_most-common-report}{nconfigs}{9}
\chexdefmeta{critical_mimic_most-common-report}{revision}{most-common-report-v1}
\chexdefmeta{critical_mimic_most-common-report}{seed}{0}
\chexdefmeta{critical_mimic_most-common-report}{split}{test+val}
\chexdefmeta{critical_mimic_most-common-report}{temperature}{0.0}
\chexdefmeta{critical_mimic_oracle-struct-a0-full}{backend}{hf}
\chexdefmeta{critical_mimic_oracle-struct-a0-full}{dataset}{mimic}
\chexdefmeta{critical_mimic_oracle-struct-a0-full}{model}{empty}
\chexdefmeta{critical_mimic_oracle-struct-a0-full}{nconfigs}{9}
\chexdefmeta{critical_mimic_oracle-struct-a0-full}{revision}{empty-v1}
\chexdefmeta{critical_mimic_oracle-struct-a0-full}{seed}{0}
\chexdefmeta{critical_mimic_oracle-struct-a0-full}{split}{test+val}
\chexdefmeta{critical_mimic_oracle-struct-a0-full}{temperature}{0.0}
\chexdefmeta{critical_mimic_oracle-struct-a1-no-anatomy}{backend}{hf}
\chexdefmeta{critical_mimic_oracle-struct-a1-no-anatomy}{dataset}{mimic}
\chexdefmeta{critical_mimic_oracle-struct-a1-no-anatomy}{model}{empty}
\chexdefmeta{critical_mimic_oracle-struct-a1-no-anatomy}{nconfigs}{9}
\chexdefmeta{critical_mimic_oracle-struct-a1-no-anatomy}{revision}{empty-v1}
\chexdefmeta{critical_mimic_oracle-struct-a1-no-anatomy}{seed}{0}
\chexdefmeta{critical_mimic_oracle-struct-a1-no-anatomy}{split}{test+val}
\chexdefmeta{critical_mimic_oracle-struct-a1-no-anatomy}{temperature}{0.0}
\chexdefmeta{critical_mimic_oracle-struct-a2-positives}{backend}{hf}
\chexdefmeta{critical_mimic_oracle-struct-a2-positives}{dataset}{mimic}
\chexdefmeta{critical_mimic_oracle-struct-a2-positives}{model}{empty}
\chexdefmeta{critical_mimic_oracle-struct-a2-positives}{nconfigs}{9}
\chexdefmeta{critical_mimic_oracle-struct-a2-positives}{revision}{empty-v1}
\chexdefmeta{critical_mimic_oracle-struct-a2-positives}{seed}{0}
\chexdefmeta{critical_mimic_oracle-struct-a2-positives}{split}{test+val}
\chexdefmeta{critical_mimic_oracle-struct-a2-positives}{temperature}{0.0}
\chexdefmeta{critical_mimic_oracle-struct-a3-no-certainty}{backend}{hf}
\chexdefmeta{critical_mimic_oracle-struct-a3-no-certainty}{dataset}{mimic}
\chexdefmeta{critical_mimic_oracle-struct-a3-no-certainty}{model}{empty}
\chexdefmeta{critical_mimic_oracle-struct-a3-no-certainty}{nconfigs}{9}
\chexdefmeta{critical_mimic_oracle-struct-a3-no-certainty}{revision}{empty-v1}
\chexdefmeta{critical_mimic_oracle-struct-a3-no-certainty}{seed}{0}
\chexdefmeta{critical_mimic_oracle-struct-a3-no-certainty}{split}{test+val}
\chexdefmeta{critical_mimic_oracle-struct-a3-no-certainty}{temperature}{0.0}
\chexdefmeta{critical_mimic_oracle-struct-a4-no-severity}{backend}{hf}
\chexdefmeta{critical_mimic_oracle-struct-a4-no-severity}{dataset}{mimic}
\chexdefmeta{critical_mimic_oracle-struct-a4-no-severity}{model}{empty}
\chexdefmeta{critical_mimic_oracle-struct-a4-no-severity}{nconfigs}{9}
\chexdefmeta{critical_mimic_oracle-struct-a4-no-severity}{revision}{empty-v1}
\chexdefmeta{critical_mimic_oracle-struct-a4-no-severity}{seed}{0}
\chexdefmeta{critical_mimic_oracle-struct-a4-no-severity}{split}{test+val}
\chexdefmeta{critical_mimic_oracle-struct-a4-no-severity}{temperature}{0.0}
\chexdefmeta{critical_mimic_oracle-struct-a5-plus-change}{backend}{hf}
\chexdefmeta{critical_mimic_oracle-struct-a5-plus-change}{dataset}{mimic}
\chexdefmeta{critical_mimic_oracle-struct-a5-plus-change}{model}{empty}
\chexdefmeta{critical_mimic_oracle-struct-a5-plus-change}{nconfigs}{9}
\chexdefmeta{critical_mimic_oracle-struct-a5-plus-change}{revision}{empty-v1}
\chexdefmeta{critical_mimic_oracle-struct-a5-plus-change}{seed}{0}
\chexdefmeta{critical_mimic_oracle-struct-a5-plus-change}{split}{test+val}
\chexdefmeta{critical_mimic_oracle-struct-a5-plus-change}{temperature}{0.0}
\chexdefmeta{critical_mimic_oracle-struct-a6-names-only}{backend}{hf}
\chexdefmeta{critical_mimic_oracle-struct-a6-names-only}{dataset}{mimic}
\chexdefmeta{critical_mimic_oracle-struct-a6-names-only}{model}{empty}
\chexdefmeta{critical_mimic_oracle-struct-a6-names-only}{nconfigs}{9}
\chexdefmeta{critical_mimic_oracle-struct-a6-names-only}{revision}{empty-v1}
\chexdefmeta{critical_mimic_oracle-struct-a6-names-only}{seed}{0}
\chexdefmeta{critical_mimic_oracle-struct-a6-names-only}{split}{test+val}
\chexdefmeta{critical_mimic_oracle-struct-a6-names-only}{temperature}{0.0}
\chexdefmeta{critical_mimic_oracle-struct-a7-chexpert14}{backend}{hf}
\chexdefmeta{critical_mimic_oracle-struct-a7-chexpert14}{dataset}{mimic}
\chexdefmeta{critical_mimic_oracle-struct-a7-chexpert14}{model}{empty}
\chexdefmeta{critical_mimic_oracle-struct-a7-chexpert14}{nconfigs}{9}
\chexdefmeta{critical_mimic_oracle-struct-a7-chexpert14}{revision}{empty-v1}
\chexdefmeta{critical_mimic_oracle-struct-a7-chexpert14}{seed}{0}
\chexdefmeta{critical_mimic_oracle-struct-a7-chexpert14}{split}{test+val}
\chexdefmeta{critical_mimic_oracle-struct-a7-chexpert14}{temperature}{0.0}
\chexdefmeta{critical_mimic_oracle-template-a0-full}{backend}{hf}
\chexdefmeta{critical_mimic_oracle-template-a0-full}{dataset}{mimic}
\chexdefmeta{critical_mimic_oracle-template-a0-full}{model}{empty}
\chexdefmeta{critical_mimic_oracle-template-a0-full}{nconfigs}{9}
\chexdefmeta{critical_mimic_oracle-template-a0-full}{revision}{empty-v1}
\chexdefmeta{critical_mimic_oracle-template-a0-full}{seed}{0}
\chexdefmeta{critical_mimic_oracle-template-a0-full}{split}{test+val}
\chexdefmeta{critical_mimic_oracle-template-a0-full}{temperature}{0.0}
\chexdefmeta{critical_mimic_oracle-template-a1-no-anatomy}{backend}{hf}
\chexdefmeta{critical_mimic_oracle-template-a1-no-anatomy}{dataset}{mimic}
\chexdefmeta{critical_mimic_oracle-template-a1-no-anatomy}{model}{empty}
\chexdefmeta{critical_mimic_oracle-template-a1-no-anatomy}{nconfigs}{9}
\chexdefmeta{critical_mimic_oracle-template-a1-no-anatomy}{revision}{empty-v1}
\chexdefmeta{critical_mimic_oracle-template-a1-no-anatomy}{seed}{0}
\chexdefmeta{critical_mimic_oracle-template-a1-no-anatomy}{split}{test+val}
\chexdefmeta{critical_mimic_oracle-template-a1-no-anatomy}{temperature}{0.0}
\chexdefmeta{critical_mimic_oracle-template-a2-positives}{backend}{hf}
\chexdefmeta{critical_mimic_oracle-template-a2-positives}{dataset}{mimic}
\chexdefmeta{critical_mimic_oracle-template-a2-positives}{model}{empty}
\chexdefmeta{critical_mimic_oracle-template-a2-positives}{nconfigs}{9}
\chexdefmeta{critical_mimic_oracle-template-a2-positives}{revision}{empty-v1}
\chexdefmeta{critical_mimic_oracle-template-a2-positives}{seed}{0}
\chexdefmeta{critical_mimic_oracle-template-a2-positives}{split}{test+val}
\chexdefmeta{critical_mimic_oracle-template-a2-positives}{temperature}{0.0}
\chexdefmeta{critical_mimic_oracle-template-a3-no-certainty}{backend}{hf}
\chexdefmeta{critical_mimic_oracle-template-a3-no-certainty}{dataset}{mimic}
\chexdefmeta{critical_mimic_oracle-template-a3-no-certainty}{model}{empty}
\chexdefmeta{critical_mimic_oracle-template-a3-no-certainty}{nconfigs}{9}
\chexdefmeta{critical_mimic_oracle-template-a3-no-certainty}{revision}{empty-v1}
\chexdefmeta{critical_mimic_oracle-template-a3-no-certainty}{seed}{0}
\chexdefmeta{critical_mimic_oracle-template-a3-no-certainty}{split}{test+val}
\chexdefmeta{critical_mimic_oracle-template-a3-no-certainty}{temperature}{0.0}
\chexdefmeta{critical_mimic_oracle-template-a4-no-severity}{backend}{hf}
\chexdefmeta{critical_mimic_oracle-template-a4-no-severity}{dataset}{mimic}
\chexdefmeta{critical_mimic_oracle-template-a4-no-severity}{model}{empty}
\chexdefmeta{critical_mimic_oracle-template-a4-no-severity}{nconfigs}{9}
\chexdefmeta{critical_mimic_oracle-template-a4-no-severity}{revision}{empty-v1}
\chexdefmeta{critical_mimic_oracle-template-a4-no-severity}{seed}{0}
\chexdefmeta{critical_mimic_oracle-template-a4-no-severity}{split}{test+val}
\chexdefmeta{critical_mimic_oracle-template-a4-no-severity}{temperature}{0.0}
\chexdefmeta{critical_mimic_oracle-template-a5-plus-change}{backend}{hf}
\chexdefmeta{critical_mimic_oracle-template-a5-plus-change}{dataset}{mimic}
\chexdefmeta{critical_mimic_oracle-template-a5-plus-change}{model}{empty}
\chexdefmeta{critical_mimic_oracle-template-a5-plus-change}{nconfigs}{9}
\chexdefmeta{critical_mimic_oracle-template-a5-plus-change}{revision}{empty-v1}
\chexdefmeta{critical_mimic_oracle-template-a5-plus-change}{seed}{0}
\chexdefmeta{critical_mimic_oracle-template-a5-plus-change}{split}{test+val}
\chexdefmeta{critical_mimic_oracle-template-a5-plus-change}{temperature}{0.0}
\chexdefmeta{critical_mimic_oracle-template-a6-names-only}{backend}{hf}
\chexdefmeta{critical_mimic_oracle-template-a6-names-only}{dataset}{mimic}
\chexdefmeta{critical_mimic_oracle-template-a6-names-only}{model}{empty}
\chexdefmeta{critical_mimic_oracle-template-a6-names-only}{nconfigs}{9}
\chexdefmeta{critical_mimic_oracle-template-a6-names-only}{revision}{empty-v1}
\chexdefmeta{critical_mimic_oracle-template-a6-names-only}{seed}{0}
\chexdefmeta{critical_mimic_oracle-template-a6-names-only}{split}{test+val}
\chexdefmeta{critical_mimic_oracle-template-a6-names-only}{temperature}{0.0}
\chexdefmeta{critical_mimic_oracle-template-a7-chexpert14}{backend}{hf}
\chexdefmeta{critical_mimic_oracle-template-a7-chexpert14}{dataset}{mimic}
\chexdefmeta{critical_mimic_oracle-template-a7-chexpert14}{model}{empty}
\chexdefmeta{critical_mimic_oracle-template-a7-chexpert14}{nconfigs}{9}
\chexdefmeta{critical_mimic_oracle-template-a7-chexpert14}{revision}{empty-v1}
\chexdefmeta{critical_mimic_oracle-template-a7-chexpert14}{seed}{0}
\chexdefmeta{critical_mimic_oracle-template-a7-chexpert14}{split}{test+val}
\chexdefmeta{critical_mimic_oracle-template-a7-chexpert14}{temperature}{0.0}
\chexdefmeta{critical_mimic_qwen2.5-vl-3b}{backend}{vllm}
\chexdefmeta{critical_mimic_qwen2.5-vl-3b}{dataset}{mimic}
\chexdefmeta{critical_mimic_qwen2.5-vl-3b}{maxtokens}{768}
\chexdefmeta{critical_mimic_qwen2.5-vl-3b}{model}{qwen2.5-vl-3b}
\chexdefmeta{critical_mimic_qwen2.5-vl-3b}{nconfigs}{9}
\chexdefmeta{critical_mimic_qwen2.5-vl-3b}{revision}{66285546d2b821cf421d4f5eb2576359d3770cd3}
\chexdefmeta{critical_mimic_qwen2.5-vl-3b}{seed}{0}
\chexdefmeta{critical_mimic_qwen2.5-vl-3b}{split}{test+val}
\chexdefmeta{critical_mimic_qwen2.5-vl-3b}{temperature}{0.0}
\chexdefmeta{critical_mimic_qwen2.5-vl-7b}{backend}{vllm}
\chexdefmeta{critical_mimic_qwen2.5-vl-7b}{dataset}{mimic}
\chexdefmeta{critical_mimic_qwen2.5-vl-7b}{maxtokens}{768}
\chexdefmeta{critical_mimic_qwen2.5-vl-7b}{model}{qwen2.5-vl-7b}
\chexdefmeta{critical_mimic_qwen2.5-vl-7b}{nconfigs}{9}
\chexdefmeta{critical_mimic_qwen2.5-vl-7b}{revision}{cc594898137f460bfe9f0759e9844b3ce807cfb5}
\chexdefmeta{critical_mimic_qwen2.5-vl-7b}{seed}{0}
\chexdefmeta{critical_mimic_qwen2.5-vl-7b}{split}{test+val}
\chexdefmeta{critical_mimic_qwen2.5-vl-7b}{temperature}{0.0}
\chexdefmeta{critical_mimic_raddialog}{backend}{hf}
\chexdefmeta{critical_mimic_raddialog}{dataset}{mimic}
\chexdefmeta{critical_mimic_raddialog}{model}{raddialog}
\chexdefmeta{critical_mimic_raddialog}{nconfigs}{9}
\chexdefmeta{critical_mimic_raddialog}{revision}{a0c5cb0da00d528500ce56f25e78d908a76a46a9}
\chexdefmeta{critical_mimic_raddialog}{seed}{0}
\chexdefmeta{critical_mimic_raddialog}{split}{test+val}
\chexdefmeta{critical_mimic_raddialog}{temperature}{0.0}
\chexdefmeta{critical_mimic_raddialog-prefixed}{backend}{hf}
\chexdefmeta{critical_mimic_raddialog-prefixed}{dataset}{mimic}
\chexdefmeta{critical_mimic_raddialog-prefixed}{model}{raddialog-prefixed}
\chexdefmeta{critical_mimic_raddialog-prefixed}{nconfigs}{9}
\chexdefmeta{critical_mimic_raddialog-prefixed}{revision}{a0c5cb0da00d528500ce56f25e78d908a76a46a9}
\chexdefmeta{critical_mimic_raddialog-prefixed}{seed}{0}
\chexdefmeta{critical_mimic_raddialog-prefixed}{split}{test+val}
\chexdefmeta{critical_mimic_raddialog-prefixed}{temperature}{0.0}
\chexdefmeta{critical_mimic_random-report}{backend}{hf}
\chexdefmeta{critical_mimic_random-report}{dataset}{mimic}
\chexdefmeta{critical_mimic_random-report}{model}{random-report}
\chexdefmeta{critical_mimic_random-report}{nconfigs}{9}
\chexdefmeta{critical_mimic_random-report}{revision}{random-report-v1}
\chexdefmeta{critical_mimic_random-report}{seed}{0}
\chexdefmeta{critical_mimic_random-report}{split}{test+val}
\chexdefmeta{critical_mimic_random-report}{temperature}{0.0}
\chexdefmeta{critical_mimic_rgrg}{backend}{hf}
\chexdefmeta{critical_mimic_rgrg}{dataset}{mimic}
\chexdefmeta{critical_mimic_rgrg}{model}{rgrg}
\chexdefmeta{critical_mimic_rgrg}{nconfigs}{2}
\chexdefmeta{critical_mimic_rgrg}{revision}{rgrg@9520b6d+gdrive:1rDxqzOhjqydsOrITJrX0Rj1PAdMeP7Wy}
\chexdefmeta{critical_mimic_rgrg}{seed}{0}
\chexdefmeta{critical_mimic_rgrg}{split}{test+val}
\chexdefmeta{critical_mimic_rgrg}{temperature}{0.0}
\chexdefmeta{critical_mimic_upper-bound}{backend}{hf}
\chexdefmeta{critical_mimic_upper-bound}{dataset}{mimic}
\chexdefmeta{critical_mimic_upper-bound}{model}{empty}
\chexdefmeta{critical_mimic_upper-bound}{nconfigs}{9}
\chexdefmeta{critical_mimic_upper-bound}{revision}{empty-v1}
\chexdefmeta{critical_mimic_upper-bound}{seed}{0}
\chexdefmeta{critical_mimic_upper-bound}{split}{test}
\chexdefmeta{critical_mimic_upper-bound}{temperature}{0.0}
\chexdefmeta{critical_openi_chexagent-2-3b-native}{backend}{hf}
\chexdefmeta{critical_openi_chexagent-2-3b-native}{dataset}{openi}
\chexdefmeta{critical_openi_chexagent-2-3b-native}{model}{chexagent-2-3b-native}
\chexdefmeta{critical_openi_chexagent-2-3b-native}{nconfigs}{9}
\chexdefmeta{critical_openi_chexagent-2-3b-native}{revision}{8f19b53a2eceda4c33b0acec6c81fbc293ad80d0}
\chexdefmeta{critical_openi_chexagent-2-3b-native}{seed}{0}
\chexdefmeta{critical_openi_chexagent-2-3b-native}{split}{test}
\chexdefmeta{critical_openi_chexagent-2-3b-native}{temperature}{0.0}
\chexdefmeta{critical_openi_chexagent-2-3b-native-notags}{backend}{hf}
\chexdefmeta{critical_openi_chexagent-2-3b-native-notags}{dataset}{openi}
\chexdefmeta{critical_openi_chexagent-2-3b-native-notags}{model}{chexagent-2-3b-native}
\chexdefmeta{critical_openi_chexagent-2-3b-native-notags}{nconfigs}{9}
\chexdefmeta{critical_openi_chexagent-2-3b-native-notags}{revision}{8f19b53a2eceda4c33b0acec6c81fbc293ad80d0}
\chexdefmeta{critical_openi_chexagent-2-3b-native-notags}{seed}{0}
\chexdefmeta{critical_openi_chexagent-2-3b-native-notags}{split}{test}
\chexdefmeta{critical_openi_chexagent-2-3b-native-notags}{temperature}{0.0}
\chexdefmeta{critical_openi_chexone-instruct}{backend}{vllm}
\chexdefmeta{critical_openi_chexone-instruct}{dataset}{openi}
\chexdefmeta{critical_openi_chexone-instruct}{maxtokens}{2048}
\chexdefmeta{critical_openi_chexone-instruct}{model}{chexone-instruct}
\chexdefmeta{critical_openi_chexone-instruct}{nconfigs}{9}
\chexdefmeta{critical_openi_chexone-instruct}{revision}{0c350e6852ea08f9d9baf3b7595c1a10d4849927}
\chexdefmeta{critical_openi_chexone-instruct}{seed}{0}
\chexdefmeta{critical_openi_chexone-instruct}{split}{test}
\chexdefmeta{critical_openi_chexone-instruct}{temperature}{0.0}
\chexdefmeta{critical_openi_chexone-reasoning}{backend}{vllm}
\chexdefmeta{critical_openi_chexone-reasoning}{dataset}{openi}
\chexdefmeta{critical_openi_chexone-reasoning}{maxtokens}{2048}
\chexdefmeta{critical_openi_chexone-reasoning}{model}{chexone-reasoning}
\chexdefmeta{critical_openi_chexone-reasoning}{nconfigs}{9}
\chexdefmeta{critical_openi_chexone-reasoning}{revision}{0c350e6852ea08f9d9baf3b7595c1a10d4849927}
\chexdefmeta{critical_openi_chexone-reasoning}{seed}{0}
\chexdefmeta{critical_openi_chexone-reasoning}{split}{test}
\chexdefmeta{critical_openi_chexone-reasoning}{temperature}{0.0}
\chexdefmeta{critical_openi_claude-fable-5-1}{backend}{claude-cli}
\chexdefmeta{critical_openi_claude-fable-5-1}{dataset}{openi}
\chexdefmeta{critical_openi_claude-fable-5-1}{model}{claude-fable-5-1}
\chexdefmeta{critical_openi_claude-fable-5-1}{nconfigs}{9}
\chexdefmeta{critical_openi_claude-fable-5-1}{revision}{claude-fable-5-1}
\chexdefmeta{critical_openi_claude-fable-5-1}{split}{test}
\chexdefmeta{critical_openi_claude-opus-5-5}{backend}{claude-cli}
\chexdefmeta{critical_openi_claude-opus-5-5}{dataset}{openi}
\chexdefmeta{critical_openi_claude-opus-5-5}{model}{claude-opus-5-5}
\chexdefmeta{critical_openi_claude-opus-5-5}{nconfigs}{9}
\chexdefmeta{critical_openi_claude-opus-5-5}{revision}{claude-opus-5-5}
\chexdefmeta{critical_openi_claude-opus-5-5}{split}{test}
\chexdefmeta{critical_openi_constant-normal}{backend}{hf}
\chexdefmeta{critical_openi_constant-normal}{dataset}{openi}
\chexdefmeta{critical_openi_constant-normal}{model}{constant-normal}
\chexdefmeta{critical_openi_constant-normal}{nconfigs}{9}
\chexdefmeta{critical_openi_constant-normal}{revision}{constant-normal-v1}
\chexdefmeta{critical_openi_constant-normal}{seed}{0}
\chexdefmeta{critical_openi_constant-normal}{split}{test}
\chexdefmeta{critical_openi_constant-normal}{temperature}{0.0}
\chexdefmeta{critical_openi_copy-indication}{backend}{hf}
\chexdefmeta{critical_openi_copy-indication}{dataset}{openi}
\chexdefmeta{critical_openi_copy-indication}{model}{copy-indication}
\chexdefmeta{critical_openi_copy-indication}{nconfigs}{9}
\chexdefmeta{critical_openi_copy-indication}{revision}{copy-indication-v1}
\chexdefmeta{critical_openi_copy-indication}{seed}{0}
\chexdefmeta{critical_openi_copy-indication}{split}{test}
\chexdefmeta{critical_openi_copy-indication}{temperature}{0.0}
\chexdefmeta{critical_openi_copy-prior}{backend}{hf}
\chexdefmeta{critical_openi_copy-prior}{dataset}{openi}
\chexdefmeta{critical_openi_copy-prior}{model}{copy-prior}
\chexdefmeta{critical_openi_copy-prior}{nconfigs}{9}
\chexdefmeta{critical_openi_copy-prior}{revision}{copy-prior-v1}
\chexdefmeta{critical_openi_copy-prior}{seed}{0}
\chexdefmeta{critical_openi_copy-prior}{split}{test}
\chexdefmeta{critical_openi_copy-prior}{temperature}{0.0}
\chexdefmeta{critical_openi_cxr-llava-v2}{backend}{hf}
\chexdefmeta{critical_openi_cxr-llava-v2}{dataset}{openi}
\chexdefmeta{critical_openi_cxr-llava-v2}{model}{cxr-llava-v2}
\chexdefmeta{critical_openi_cxr-llava-v2}{nconfigs}{9}
\chexdefmeta{critical_openi_cxr-llava-v2}{revision}{b2224786bb90d54b1e1291171866706cfbb44e2b}
\chexdefmeta{critical_openi_cxr-llava-v2}{seed}{0}
\chexdefmeta{critical_openi_cxr-llava-v2}{split}{test}
\chexdefmeta{critical_openi_cxr-llava-v2}{temperature}{0.0}
\chexdefmeta{critical_openi_empty}{backend}{hf}
\chexdefmeta{critical_openi_empty}{dataset}{openi}
\chexdefmeta{critical_openi_empty}{model}{empty}
\chexdefmeta{critical_openi_empty}{nconfigs}{9}
\chexdefmeta{critical_openi_empty}{revision}{empty-v1}
\chexdefmeta{critical_openi_empty}{seed}{0}
\chexdefmeta{critical_openi_empty}{split}{test}
\chexdefmeta{critical_openi_empty}{temperature}{0.0}
\chexdefmeta{critical_openi_internvl3-8b}{backend}{vllm}
\chexdefmeta{critical_openi_internvl3-8b}{dataset}{openi}
\chexdefmeta{critical_openi_internvl3-8b}{maxtokens}{768}
\chexdefmeta{critical_openi_internvl3-8b}{model}{internvl3-8b}
\chexdefmeta{critical_openi_internvl3-8b}{nconfigs}{9}
\chexdefmeta{critical_openi_internvl3-8b}{revision}{853e3a797a661694b1b8ece0cb72dc2b23e3dac9}
\chexdefmeta{critical_openi_internvl3-8b}{seed}{0}
\chexdefmeta{critical_openi_internvl3-8b}{split}{test}
\chexdefmeta{critical_openi_internvl3-8b}{temperature}{0.0}
\chexdefmeta{critical_openi_llava-med}{backend}{hf}
\chexdefmeta{critical_openi_llava-med}{dataset}{openi}
\chexdefmeta{critical_openi_llava-med}{model}{llava-med}
\chexdefmeta{critical_openi_llava-med}{nconfigs}{9}
\chexdefmeta{critical_openi_llava-med}{revision}{91bb16c122001ddc9cf1fd36ce1dae09448943a2}
\chexdefmeta{critical_openi_llava-med}{seed}{0}
\chexdefmeta{critical_openi_llava-med}{split}{test}
\chexdefmeta{critical_openi_llava-med}{temperature}{0.0}
\chexdefmeta{critical_openi_llava-rad}{backend}{hf}
\chexdefmeta{critical_openi_llava-rad}{dataset}{openi}
\chexdefmeta{critical_openi_llava-rad}{model}{llava-rad}
\chexdefmeta{critical_openi_llava-rad}{nconfigs}{9}
\chexdefmeta{critical_openi_llava-rad}{revision}{dcdbc6caf6806c9acb66e12a827a39c86e00f6a7}
\chexdefmeta{critical_openi_llava-rad}{seed}{0}
\chexdefmeta{critical_openi_llava-rad}{split}{test}
\chexdefmeta{critical_openi_llava-rad}{temperature}{0.0}
\chexdefmeta{critical_openi_maira-2}{backend}{hf}
\chexdefmeta{critical_openi_maira-2}{dataset}{openi}
\chexdefmeta{critical_openi_maira-2}{model}{maira-2}
\chexdefmeta{critical_openi_maira-2}{nconfigs}{9}
\chexdefmeta{critical_openi_maira-2}{revision}{795a2b1cd4a310624b4e3d14b5a23e41fd273deb}
\chexdefmeta{critical_openi_maira-2}{seed}{0}
\chexdefmeta{critical_openi_maira-2}{split}{test}
\chexdefmeta{critical_openi_maira-2}{temperature}{0.0}
\chexdefmeta{critical_openi_medgemma-4b}{backend}{vllm}
\chexdefmeta{critical_openi_medgemma-4b}{dataset}{openi}
\chexdefmeta{critical_openi_medgemma-4b}{maxtokens}{768}
\chexdefmeta{critical_openi_medgemma-4b}{model}{medgemma-4b}
\chexdefmeta{critical_openi_medgemma-4b}{nconfigs}{9}
\chexdefmeta{critical_openi_medgemma-4b}{revision}{290cda5eeccbee130f987c4ad74a59ae6f196408}
\chexdefmeta{critical_openi_medgemma-4b}{seed}{0}
\chexdefmeta{critical_openi_medgemma-4b}{split}{test}
\chexdefmeta{critical_openi_medgemma-4b}{temperature}{0.0}
\chexdefmeta{critical_openi_most-common-report}{backend}{hf}
\chexdefmeta{critical_openi_most-common-report}{dataset}{openi}
\chexdefmeta{critical_openi_most-common-report}{model}{most-common-report}
\chexdefmeta{critical_openi_most-common-report}{nconfigs}{9}
\chexdefmeta{critical_openi_most-common-report}{revision}{most-common-report-v1}
\chexdefmeta{critical_openi_most-common-report}{seed}{0}
\chexdefmeta{critical_openi_most-common-report}{split}{test}
\chexdefmeta{critical_openi_most-common-report}{temperature}{0.0}
\chexdefmeta{critical_openi_qwen2.5-vl-3b}{backend}{vllm}
\chexdefmeta{critical_openi_qwen2.5-vl-3b}{dataset}{openi}
\chexdefmeta{critical_openi_qwen2.5-vl-3b}{maxtokens}{768}
\chexdefmeta{critical_openi_qwen2.5-vl-3b}{model}{qwen2.5-vl-3b}
\chexdefmeta{critical_openi_qwen2.5-vl-3b}{nconfigs}{9}
\chexdefmeta{critical_openi_qwen2.5-vl-3b}{revision}{66285546d2b821cf421d4f5eb2576359d3770cd3}
\chexdefmeta{critical_openi_qwen2.5-vl-3b}{seed}{0}
\chexdefmeta{critical_openi_qwen2.5-vl-3b}{split}{test}
\chexdefmeta{critical_openi_qwen2.5-vl-3b}{temperature}{0.0}
\chexdefmeta{critical_openi_qwen2.5-vl-7b}{backend}{vllm}
\chexdefmeta{critical_openi_qwen2.5-vl-7b}{dataset}{openi}
\chexdefmeta{critical_openi_qwen2.5-vl-7b}{maxtokens}{768}
\chexdefmeta{critical_openi_qwen2.5-vl-7b}{model}{qwen2.5-vl-7b}
\chexdefmeta{critical_openi_qwen2.5-vl-7b}{nconfigs}{9}
\chexdefmeta{critical_openi_qwen2.5-vl-7b}{revision}{cc594898137f460bfe9f0759e9844b3ce807cfb5}
\chexdefmeta{critical_openi_qwen2.5-vl-7b}{seed}{0}
\chexdefmeta{critical_openi_qwen2.5-vl-7b}{split}{test}
\chexdefmeta{critical_openi_qwen2.5-vl-7b}{temperature}{0.0}
\chexdefmeta{critical_openi_raddialog}{backend}{hf}
\chexdefmeta{critical_openi_raddialog}{dataset}{openi}
\chexdefmeta{critical_openi_raddialog}{model}{raddialog}
\chexdefmeta{critical_openi_raddialog}{nconfigs}{9}
\chexdefmeta{critical_openi_raddialog}{revision}{a0c5cb0da00d528500ce56f25e78d908a76a46a9}
\chexdefmeta{critical_openi_raddialog}{seed}{0}
\chexdefmeta{critical_openi_raddialog}{split}{test}
\chexdefmeta{critical_openi_raddialog}{temperature}{0.0}
\chexdefmeta{critical_openi_raddialog-prefixed}{backend}{hf}
\chexdefmeta{critical_openi_raddialog-prefixed}{dataset}{openi}
\chexdefmeta{critical_openi_raddialog-prefixed}{model}{raddialog-prefixed}
\chexdefmeta{critical_openi_raddialog-prefixed}{nconfigs}{9}
\chexdefmeta{critical_openi_raddialog-prefixed}{revision}{a0c5cb0da00d528500ce56f25e78d908a76a46a9}
\chexdefmeta{critical_openi_raddialog-prefixed}{seed}{0}
\chexdefmeta{critical_openi_raddialog-prefixed}{split}{test}
\chexdefmeta{critical_openi_raddialog-prefixed}{temperature}{0.0}
\chexdefmeta{critical_openi_random-report}{backend}{hf}
\chexdefmeta{critical_openi_random-report}{dataset}{openi}
\chexdefmeta{critical_openi_random-report}{model}{random-report}
\chexdefmeta{critical_openi_random-report}{nconfigs}{9}
\chexdefmeta{critical_openi_random-report}{revision}{random-report-v1}
\chexdefmeta{critical_openi_random-report}{seed}{0}
\chexdefmeta{critical_openi_random-report}{split}{test}
\chexdefmeta{critical_openi_random-report}{temperature}{0.0}
\chexdefmeta{critical_openi_rgrg}{backend}{hf}
\chexdefmeta{critical_openi_rgrg}{dataset}{openi}
\chexdefmeta{critical_openi_rgrg}{model}{rgrg}
\chexdefmeta{critical_openi_rgrg}{nconfigs}{2}
\chexdefmeta{critical_openi_rgrg}{revision}{rgrg@9520b6d+gdrive:1rDxqzOhjqydsOrITJrX0Rj1PAdMeP7Wy}
\chexdefmeta{critical_openi_rgrg}{seed}{0}
\chexdefmeta{critical_openi_rgrg}{split}{test}
\chexdefmeta{critical_openi_rgrg}{temperature}{0.0}
\chexdefmeta{critical_openi_upper-bound}{backend}{hf}
\chexdefmeta{critical_openi_upper-bound}{dataset}{openi}
\chexdefmeta{critical_openi_upper-bound}{model}{empty}
\chexdefmeta{critical_openi_upper-bound}{nconfigs}{9}
\chexdefmeta{critical_openi_upper-bound}{revision}{empty-v1}
\chexdefmeta{critical_openi_upper-bound}{seed}{0}
\chexdefmeta{critical_openi_upper-bound}{split}{test}
\chexdefmeta{critical_openi_upper-bound}{temperature}{0.0}
\chexdefmeta{settings_mimic_chexagent-2-3b-native}{backend}{hf}
\chexdefmeta{settings_mimic_chexagent-2-3b-native}{dataset}{mimic}
\chexdefmeta{settings_mimic_chexagent-2-3b-native}{model}{chexagent-2-3b-native}
\chexdefmeta{settings_mimic_chexagent-2-3b-native}{nconfigs}{9}
\chexdefmeta{settings_mimic_chexagent-2-3b-native}{revision}{8f19b53a2eceda4c33b0acec6c81fbc293ad80d0}
\chexdefmeta{settings_mimic_chexagent-2-3b-native}{seed}{0}
\chexdefmeta{settings_mimic_chexagent-2-3b-native}{split}{test}
\chexdefmeta{settings_mimic_chexagent-2-3b-native}{temperature}{0.0}
\chexdefmeta{settings_mimic_chexagent-2-3b-native-notags}{backend}{hf}
\chexdefmeta{settings_mimic_chexagent-2-3b-native-notags}{dataset}{mimic}
\chexdefmeta{settings_mimic_chexagent-2-3b-native-notags}{model}{chexagent-2-3b-native}
\chexdefmeta{settings_mimic_chexagent-2-3b-native-notags}{nconfigs}{9}
\chexdefmeta{settings_mimic_chexagent-2-3b-native-notags}{revision}{8f19b53a2eceda4c33b0acec6c81fbc293ad80d0}
\chexdefmeta{settings_mimic_chexagent-2-3b-native-notags}{seed}{0}
\chexdefmeta{settings_mimic_chexagent-2-3b-native-notags}{split}{test}
\chexdefmeta{settings_mimic_chexagent-2-3b-native-notags}{temperature}{0.0}
\chexdefmeta{settings_mimic_chexone-instruct}{backend}{vllm}
\chexdefmeta{settings_mimic_chexone-instruct}{dataset}{mimic}
\chexdefmeta{settings_mimic_chexone-instruct}{maxtokens}{2048}
\chexdefmeta{settings_mimic_chexone-instruct}{model}{chexone-instruct}
\chexdefmeta{settings_mimic_chexone-instruct}{nconfigs}{9}
\chexdefmeta{settings_mimic_chexone-instruct}{revision}{0c350e6852ea08f9d9baf3b7595c1a10d4849927}
\chexdefmeta{settings_mimic_chexone-instruct}{seed}{0}
\chexdefmeta{settings_mimic_chexone-instruct}{split}{test}
\chexdefmeta{settings_mimic_chexone-instruct}{temperature}{0.0}
\chexdefmeta{settings_mimic_chexone-reasoning}{backend}{vllm}
\chexdefmeta{settings_mimic_chexone-reasoning}{dataset}{mimic}
\chexdefmeta{settings_mimic_chexone-reasoning}{maxtokens}{2048}
\chexdefmeta{settings_mimic_chexone-reasoning}{model}{chexone-reasoning}
\chexdefmeta{settings_mimic_chexone-reasoning}{nconfigs}{9}
\chexdefmeta{settings_mimic_chexone-reasoning}{revision}{0c350e6852ea08f9d9baf3b7595c1a10d4849927}
\chexdefmeta{settings_mimic_chexone-reasoning}{seed}{0}
\chexdefmeta{settings_mimic_chexone-reasoning}{split}{test}
\chexdefmeta{settings_mimic_chexone-reasoning}{temperature}{0.0}
\chexdefmeta{settings_mimic_constant-normal}{backend}{hf}
\chexdefmeta{settings_mimic_constant-normal}{dataset}{mimic}
\chexdefmeta{settings_mimic_constant-normal}{model}{constant-normal}
\chexdefmeta{settings_mimic_constant-normal}{nconfigs}{9}
\chexdefmeta{settings_mimic_constant-normal}{revision}{constant-normal-v1}
\chexdefmeta{settings_mimic_constant-normal}{seed}{0}
\chexdefmeta{settings_mimic_constant-normal}{split}{test}
\chexdefmeta{settings_mimic_constant-normal}{temperature}{0.0}
\chexdefmeta{settings_mimic_copy-indication}{backend}{hf}
\chexdefmeta{settings_mimic_copy-indication}{dataset}{mimic}
\chexdefmeta{settings_mimic_copy-indication}{model}{copy-indication}
\chexdefmeta{settings_mimic_copy-indication}{nconfigs}{9}
\chexdefmeta{settings_mimic_copy-indication}{revision}{copy-indication-v1}
\chexdefmeta{settings_mimic_copy-indication}{seed}{0}
\chexdefmeta{settings_mimic_copy-indication}{split}{test}
\chexdefmeta{settings_mimic_copy-indication}{temperature}{0.0}
\chexdefmeta{settings_mimic_copy-prior}{backend}{hf}
\chexdefmeta{settings_mimic_copy-prior}{dataset}{mimic}
\chexdefmeta{settings_mimic_copy-prior}{model}{copy-prior}
\chexdefmeta{settings_mimic_copy-prior}{nconfigs}{9}
\chexdefmeta{settings_mimic_copy-prior}{revision}{copy-prior-v1}
\chexdefmeta{settings_mimic_copy-prior}{seed}{0}
\chexdefmeta{settings_mimic_copy-prior}{split}{test}
\chexdefmeta{settings_mimic_copy-prior}{temperature}{0.0}
\chexdefmeta{settings_mimic_cxr-llava-v2}{backend}{hf}
\chexdefmeta{settings_mimic_cxr-llava-v2}{dataset}{mimic}
\chexdefmeta{settings_mimic_cxr-llava-v2}{model}{cxr-llava-v2}
\chexdefmeta{settings_mimic_cxr-llava-v2}{nconfigs}{9}
\chexdefmeta{settings_mimic_cxr-llava-v2}{revision}{b2224786bb90d54b1e1291171866706cfbb44e2b}
\chexdefmeta{settings_mimic_cxr-llava-v2}{seed}{0}
\chexdefmeta{settings_mimic_cxr-llava-v2}{split}{test}
\chexdefmeta{settings_mimic_cxr-llava-v2}{temperature}{0.0}
\chexdefmeta{settings_mimic_empty}{backend}{hf}
\chexdefmeta{settings_mimic_empty}{dataset}{mimic}
\chexdefmeta{settings_mimic_empty}{model}{empty}
\chexdefmeta{settings_mimic_empty}{nconfigs}{9}
\chexdefmeta{settings_mimic_empty}{revision}{empty-v1}
\chexdefmeta{settings_mimic_empty}{seed}{0}
\chexdefmeta{settings_mimic_empty}{split}{test}
\chexdefmeta{settings_mimic_empty}{temperature}{0.0}
\chexdefmeta{settings_mimic_fsp-devB-c40-f21-tv3}{backend}{hf}
\chexdefmeta{settings_mimic_fsp-devB-c40-f21-tv3}{dataset}{mimic}
\chexdefmeta{settings_mimic_fsp-devB-c40-f21-tv3}{model}{empty}
\chexdefmeta{settings_mimic_fsp-devB-c40-f21-tv3}{nconfigs}{9}
\chexdefmeta{settings_mimic_fsp-devB-c40-f21-tv3}{revision}{empty-v1}
\chexdefmeta{settings_mimic_fsp-devB-c40-f21-tv3}{seed}{0}
\chexdefmeta{settings_mimic_fsp-devB-c40-f21-tv3}{split}{test}
\chexdefmeta{settings_mimic_fsp-devB-c40-f21-tv3}{temperature}{0.0}
\chexdefmeta{settings_mimic_internvl3-8b}{backend}{vllm}
\chexdefmeta{settings_mimic_internvl3-8b}{dataset}{mimic}
\chexdefmeta{settings_mimic_internvl3-8b}{maxtokens}{768}
\chexdefmeta{settings_mimic_internvl3-8b}{model}{internvl3-8b}
\chexdefmeta{settings_mimic_internvl3-8b}{nconfigs}{9}
\chexdefmeta{settings_mimic_internvl3-8b}{revision}{853e3a797a661694b1b8ece0cb72dc2b23e3dac9}
\chexdefmeta{settings_mimic_internvl3-8b}{seed}{0}
\chexdefmeta{settings_mimic_internvl3-8b}{split}{test}
\chexdefmeta{settings_mimic_internvl3-8b}{temperature}{0.0}
\chexdefmeta{settings_mimic_llava-med}{backend}{hf}
\chexdefmeta{settings_mimic_llava-med}{dataset}{mimic}
\chexdefmeta{settings_mimic_llava-med}{model}{llava-med}
\chexdefmeta{settings_mimic_llava-med}{nconfigs}{9}
\chexdefmeta{settings_mimic_llava-med}{revision}{91bb16c122001ddc9cf1fd36ce1dae09448943a2}
\chexdefmeta{settings_mimic_llava-med}{seed}{0}
\chexdefmeta{settings_mimic_llava-med}{split}{test}
\chexdefmeta{settings_mimic_llava-med}{temperature}{0.0}
\chexdefmeta{settings_mimic_llava-rad}{backend}{hf}
\chexdefmeta{settings_mimic_llava-rad}{dataset}{mimic}
\chexdefmeta{settings_mimic_llava-rad}{model}{llava-rad}
\chexdefmeta{settings_mimic_llava-rad}{nconfigs}{9}
\chexdefmeta{settings_mimic_llava-rad}{revision}{dcdbc6caf6806c9acb66e12a827a39c86e00f6a7}
\chexdefmeta{settings_mimic_llava-rad}{seed}{0}
\chexdefmeta{settings_mimic_llava-rad}{split}{test}
\chexdefmeta{settings_mimic_llava-rad}{temperature}{0.0}
\chexdefmeta{settings_mimic_maira-2}{backend}{hf}
\chexdefmeta{settings_mimic_maira-2}{dataset}{mimic}
\chexdefmeta{settings_mimic_maira-2}{model}{maira-2}
\chexdefmeta{settings_mimic_maira-2}{nconfigs}{9}
\chexdefmeta{settings_mimic_maira-2}{revision}{795a2b1cd4a310624b4e3d14b5a23e41fd273deb}
\chexdefmeta{settings_mimic_maira-2}{seed}{0}
\chexdefmeta{settings_mimic_maira-2}{split}{test}
\chexdefmeta{settings_mimic_maira-2}{temperature}{0.0}
\chexdefmeta{settings_mimic_medgemma-4b}{backend}{vllm}
\chexdefmeta{settings_mimic_medgemma-4b}{dataset}{mimic}
\chexdefmeta{settings_mimic_medgemma-4b}{maxtokens}{768}
\chexdefmeta{settings_mimic_medgemma-4b}{model}{medgemma-4b}
\chexdefmeta{settings_mimic_medgemma-4b}{nconfigs}{9}
\chexdefmeta{settings_mimic_medgemma-4b}{revision}{290cda5eeccbee130f987c4ad74a59ae6f196408}
\chexdefmeta{settings_mimic_medgemma-4b}{seed}{0}
\chexdefmeta{settings_mimic_medgemma-4b}{split}{test}
\chexdefmeta{settings_mimic_medgemma-4b}{temperature}{0.0}
\chexdefmeta{settings_mimic_most-common-report}{backend}{hf}
\chexdefmeta{settings_mimic_most-common-report}{dataset}{mimic}
\chexdefmeta{settings_mimic_most-common-report}{model}{most-common-report}
\chexdefmeta{settings_mimic_most-common-report}{nconfigs}{9}
\chexdefmeta{settings_mimic_most-common-report}{revision}{most-common-report-v1}
\chexdefmeta{settings_mimic_most-common-report}{seed}{0}
\chexdefmeta{settings_mimic_most-common-report}{split}{test}
\chexdefmeta{settings_mimic_most-common-report}{temperature}{0.0}
\chexdefmeta{settings_mimic_oracle-struct-a0-full}{backend}{hf}
\chexdefmeta{settings_mimic_oracle-struct-a0-full}{dataset}{mimic}
\chexdefmeta{settings_mimic_oracle-struct-a0-full}{model}{empty}
\chexdefmeta{settings_mimic_oracle-struct-a0-full}{nconfigs}{9}
\chexdefmeta{settings_mimic_oracle-struct-a0-full}{revision}{empty-v1}
\chexdefmeta{settings_mimic_oracle-struct-a0-full}{seed}{0}
\chexdefmeta{settings_mimic_oracle-struct-a0-full}{split}{test}
\chexdefmeta{settings_mimic_oracle-struct-a0-full}{temperature}{0.0}
\chexdefmeta{settings_mimic_oracle-struct-a1-no-anatomy}{backend}{hf}
\chexdefmeta{settings_mimic_oracle-struct-a1-no-anatomy}{dataset}{mimic}
\chexdefmeta{settings_mimic_oracle-struct-a1-no-anatomy}{model}{empty}
\chexdefmeta{settings_mimic_oracle-struct-a1-no-anatomy}{nconfigs}{9}
\chexdefmeta{settings_mimic_oracle-struct-a1-no-anatomy}{revision}{empty-v1}
\chexdefmeta{settings_mimic_oracle-struct-a1-no-anatomy}{seed}{0}
\chexdefmeta{settings_mimic_oracle-struct-a1-no-anatomy}{split}{test}
\chexdefmeta{settings_mimic_oracle-struct-a1-no-anatomy}{temperature}{0.0}
\chexdefmeta{settings_mimic_oracle-struct-a2-positives}{backend}{hf}
\chexdefmeta{settings_mimic_oracle-struct-a2-positives}{dataset}{mimic}
\chexdefmeta{settings_mimic_oracle-struct-a2-positives}{model}{empty}
\chexdefmeta{settings_mimic_oracle-struct-a2-positives}{nconfigs}{9}
\chexdefmeta{settings_mimic_oracle-struct-a2-positives}{revision}{empty-v1}
\chexdefmeta{settings_mimic_oracle-struct-a2-positives}{seed}{0}
\chexdefmeta{settings_mimic_oracle-struct-a2-positives}{split}{test}
\chexdefmeta{settings_mimic_oracle-struct-a2-positives}{temperature}{0.0}
\chexdefmeta{settings_mimic_oracle-struct-a3-no-certainty}{backend}{hf}
\chexdefmeta{settings_mimic_oracle-struct-a3-no-certainty}{dataset}{mimic}
\chexdefmeta{settings_mimic_oracle-struct-a3-no-certainty}{model}{empty}
\chexdefmeta{settings_mimic_oracle-struct-a3-no-certainty}{nconfigs}{9}
\chexdefmeta{settings_mimic_oracle-struct-a3-no-certainty}{revision}{empty-v1}
\chexdefmeta{settings_mimic_oracle-struct-a3-no-certainty}{seed}{0}
\chexdefmeta{settings_mimic_oracle-struct-a3-no-certainty}{split}{test}
\chexdefmeta{settings_mimic_oracle-struct-a3-no-certainty}{temperature}{0.0}
\chexdefmeta{settings_mimic_oracle-struct-a4-no-severity}{backend}{hf}
\chexdefmeta{settings_mimic_oracle-struct-a4-no-severity}{dataset}{mimic}
\chexdefmeta{settings_mimic_oracle-struct-a4-no-severity}{model}{empty}
\chexdefmeta{settings_mimic_oracle-struct-a4-no-severity}{nconfigs}{9}
\chexdefmeta{settings_mimic_oracle-struct-a4-no-severity}{revision}{empty-v1}
\chexdefmeta{settings_mimic_oracle-struct-a4-no-severity}{seed}{0}
\chexdefmeta{settings_mimic_oracle-struct-a4-no-severity}{split}{test}
\chexdefmeta{settings_mimic_oracle-struct-a4-no-severity}{temperature}{0.0}
\chexdefmeta{settings_mimic_oracle-struct-a5-plus-change}{backend}{hf}
\chexdefmeta{settings_mimic_oracle-struct-a5-plus-change}{dataset}{mimic}
\chexdefmeta{settings_mimic_oracle-struct-a5-plus-change}{model}{empty}
\chexdefmeta{settings_mimic_oracle-struct-a5-plus-change}{nconfigs}{9}
\chexdefmeta{settings_mimic_oracle-struct-a5-plus-change}{revision}{empty-v1}
\chexdefmeta{settings_mimic_oracle-struct-a5-plus-change}{seed}{0}
\chexdefmeta{settings_mimic_oracle-struct-a5-plus-change}{split}{test}
\chexdefmeta{settings_mimic_oracle-struct-a5-plus-change}{temperature}{0.0}
\chexdefmeta{settings_mimic_oracle-struct-a6-names-only}{backend}{hf}
\chexdefmeta{settings_mimic_oracle-struct-a6-names-only}{dataset}{mimic}
\chexdefmeta{settings_mimic_oracle-struct-a6-names-only}{model}{empty}
\chexdefmeta{settings_mimic_oracle-struct-a6-names-only}{nconfigs}{9}
\chexdefmeta{settings_mimic_oracle-struct-a6-names-only}{revision}{empty-v1}
\chexdefmeta{settings_mimic_oracle-struct-a6-names-only}{seed}{0}
\chexdefmeta{settings_mimic_oracle-struct-a6-names-only}{split}{test}
\chexdefmeta{settings_mimic_oracle-struct-a6-names-only}{temperature}{0.0}
\chexdefmeta{settings_mimic_oracle-struct-a7-chexpert14}{backend}{hf}
\chexdefmeta{settings_mimic_oracle-struct-a7-chexpert14}{dataset}{mimic}
\chexdefmeta{settings_mimic_oracle-struct-a7-chexpert14}{model}{empty}
\chexdefmeta{settings_mimic_oracle-struct-a7-chexpert14}{nconfigs}{9}
\chexdefmeta{settings_mimic_oracle-struct-a7-chexpert14}{revision}{empty-v1}
\chexdefmeta{settings_mimic_oracle-struct-a7-chexpert14}{seed}{0}
\chexdefmeta{settings_mimic_oracle-struct-a7-chexpert14}{split}{test}
\chexdefmeta{settings_mimic_oracle-struct-a7-chexpert14}{temperature}{0.0}
\chexdefmeta{settings_mimic_oracle-template-a0-full}{backend}{hf}
\chexdefmeta{settings_mimic_oracle-template-a0-full}{dataset}{mimic}
\chexdefmeta{settings_mimic_oracle-template-a0-full}{model}{empty}
\chexdefmeta{settings_mimic_oracle-template-a0-full}{nconfigs}{9}
\chexdefmeta{settings_mimic_oracle-template-a0-full}{revision}{empty-v1}
\chexdefmeta{settings_mimic_oracle-template-a0-full}{seed}{0}
\chexdefmeta{settings_mimic_oracle-template-a0-full}{split}{test}
\chexdefmeta{settings_mimic_oracle-template-a0-full}{temperature}{0.0}
\chexdefmeta{settings_mimic_oracle-template-a1-no-anatomy}{backend}{hf}
\chexdefmeta{settings_mimic_oracle-template-a1-no-anatomy}{dataset}{mimic}
\chexdefmeta{settings_mimic_oracle-template-a1-no-anatomy}{model}{empty}
\chexdefmeta{settings_mimic_oracle-template-a1-no-anatomy}{nconfigs}{9}
\chexdefmeta{settings_mimic_oracle-template-a1-no-anatomy}{revision}{empty-v1}
\chexdefmeta{settings_mimic_oracle-template-a1-no-anatomy}{seed}{0}
\chexdefmeta{settings_mimic_oracle-template-a1-no-anatomy}{split}{test}
\chexdefmeta{settings_mimic_oracle-template-a1-no-anatomy}{temperature}{0.0}
\chexdefmeta{settings_mimic_oracle-template-a2-positives}{backend}{hf}
\chexdefmeta{settings_mimic_oracle-template-a2-positives}{dataset}{mimic}
\chexdefmeta{settings_mimic_oracle-template-a2-positives}{model}{empty}
\chexdefmeta{settings_mimic_oracle-template-a2-positives}{nconfigs}{9}
\chexdefmeta{settings_mimic_oracle-template-a2-positives}{revision}{empty-v1}
\chexdefmeta{settings_mimic_oracle-template-a2-positives}{seed}{0}
\chexdefmeta{settings_mimic_oracle-template-a2-positives}{split}{test}
\chexdefmeta{settings_mimic_oracle-template-a2-positives}{temperature}{0.0}
\chexdefmeta{settings_mimic_oracle-template-a3-no-certainty}{backend}{hf}
\chexdefmeta{settings_mimic_oracle-template-a3-no-certainty}{dataset}{mimic}
\chexdefmeta{settings_mimic_oracle-template-a3-no-certainty}{model}{empty}
\chexdefmeta{settings_mimic_oracle-template-a3-no-certainty}{nconfigs}{9}
\chexdefmeta{settings_mimic_oracle-template-a3-no-certainty}{revision}{empty-v1}
\chexdefmeta{settings_mimic_oracle-template-a3-no-certainty}{seed}{0}
\chexdefmeta{settings_mimic_oracle-template-a3-no-certainty}{split}{test}
\chexdefmeta{settings_mimic_oracle-template-a3-no-certainty}{temperature}{0.0}
\chexdefmeta{settings_mimic_oracle-template-a4-no-severity}{backend}{hf}
\chexdefmeta{settings_mimic_oracle-template-a4-no-severity}{dataset}{mimic}
\chexdefmeta{settings_mimic_oracle-template-a4-no-severity}{model}{empty}
\chexdefmeta{settings_mimic_oracle-template-a4-no-severity}{nconfigs}{9}
\chexdefmeta{settings_mimic_oracle-template-a4-no-severity}{revision}{empty-v1}
\chexdefmeta{settings_mimic_oracle-template-a4-no-severity}{seed}{0}
\chexdefmeta{settings_mimic_oracle-template-a4-no-severity}{split}{test}
\chexdefmeta{settings_mimic_oracle-template-a4-no-severity}{temperature}{0.0}
\chexdefmeta{settings_mimic_oracle-template-a5-plus-change}{backend}{hf}
\chexdefmeta{settings_mimic_oracle-template-a5-plus-change}{dataset}{mimic}
\chexdefmeta{settings_mimic_oracle-template-a5-plus-change}{model}{empty}
\chexdefmeta{settings_mimic_oracle-template-a5-plus-change}{nconfigs}{9}
\chexdefmeta{settings_mimic_oracle-template-a5-plus-change}{revision}{empty-v1}
\chexdefmeta{settings_mimic_oracle-template-a5-plus-change}{seed}{0}
\chexdefmeta{settings_mimic_oracle-template-a5-plus-change}{split}{test}
\chexdefmeta{settings_mimic_oracle-template-a5-plus-change}{temperature}{0.0}
\chexdefmeta{settings_mimic_oracle-template-a6-names-only}{backend}{hf}
\chexdefmeta{settings_mimic_oracle-template-a6-names-only}{dataset}{mimic}
\chexdefmeta{settings_mimic_oracle-template-a6-names-only}{model}{empty}
\chexdefmeta{settings_mimic_oracle-template-a6-names-only}{nconfigs}{9}
\chexdefmeta{settings_mimic_oracle-template-a6-names-only}{revision}{empty-v1}
\chexdefmeta{settings_mimic_oracle-template-a6-names-only}{seed}{0}
\chexdefmeta{settings_mimic_oracle-template-a6-names-only}{split}{test}
\chexdefmeta{settings_mimic_oracle-template-a6-names-only}{temperature}{0.0}
\chexdefmeta{settings_mimic_oracle-template-a7-chexpert14}{backend}{hf}
\chexdefmeta{settings_mimic_oracle-template-a7-chexpert14}{dataset}{mimic}
\chexdefmeta{settings_mimic_oracle-template-a7-chexpert14}{model}{empty}
\chexdefmeta{settings_mimic_oracle-template-a7-chexpert14}{nconfigs}{9}
\chexdefmeta{settings_mimic_oracle-template-a7-chexpert14}{revision}{empty-v1}
\chexdefmeta{settings_mimic_oracle-template-a7-chexpert14}{seed}{0}
\chexdefmeta{settings_mimic_oracle-template-a7-chexpert14}{split}{test}
\chexdefmeta{settings_mimic_oracle-template-a7-chexpert14}{temperature}{0.0}
\chexdefmeta{settings_mimic_qwen2.5-vl-3b}{backend}{vllm}
\chexdefmeta{settings_mimic_qwen2.5-vl-3b}{dataset}{mimic}
\chexdefmeta{settings_mimic_qwen2.5-vl-3b}{maxtokens}{768}
\chexdefmeta{settings_mimic_qwen2.5-vl-3b}{model}{qwen2.5-vl-3b}
\chexdefmeta{settings_mimic_qwen2.5-vl-3b}{nconfigs}{9}
\chexdefmeta{settings_mimic_qwen2.5-vl-3b}{revision}{66285546d2b821cf421d4f5eb2576359d3770cd3}
\chexdefmeta{settings_mimic_qwen2.5-vl-3b}{seed}{0}
\chexdefmeta{settings_mimic_qwen2.5-vl-3b}{split}{test}
\chexdefmeta{settings_mimic_qwen2.5-vl-3b}{temperature}{0.0}
\chexdefmeta{settings_mimic_qwen2.5-vl-7b}{backend}{vllm}
\chexdefmeta{settings_mimic_qwen2.5-vl-7b}{dataset}{mimic}
\chexdefmeta{settings_mimic_qwen2.5-vl-7b}{maxtokens}{768}
\chexdefmeta{settings_mimic_qwen2.5-vl-7b}{model}{qwen2.5-vl-7b}
\chexdefmeta{settings_mimic_qwen2.5-vl-7b}{nconfigs}{9}
\chexdefmeta{settings_mimic_qwen2.5-vl-7b}{revision}{cc594898137f460bfe9f0759e9844b3ce807cfb5}
\chexdefmeta{settings_mimic_qwen2.5-vl-7b}{seed}{0}
\chexdefmeta{settings_mimic_qwen2.5-vl-7b}{split}{test}
\chexdefmeta{settings_mimic_qwen2.5-vl-7b}{temperature}{0.0}
\chexdefmeta{settings_mimic_raddialog}{backend}{hf}
\chexdefmeta{settings_mimic_raddialog}{dataset}{mimic}
\chexdefmeta{settings_mimic_raddialog}{model}{raddialog}
\chexdefmeta{settings_mimic_raddialog}{nconfigs}{9}
\chexdefmeta{settings_mimic_raddialog}{revision}{a0c5cb0da00d528500ce56f25e78d908a76a46a9}
\chexdefmeta{settings_mimic_raddialog}{seed}{0}
\chexdefmeta{settings_mimic_raddialog}{split}{test}
\chexdefmeta{settings_mimic_raddialog}{temperature}{0.0}
\chexdefmeta{settings_mimic_raddialog-prefixed}{backend}{hf}
\chexdefmeta{settings_mimic_raddialog-prefixed}{dataset}{mimic}
\chexdefmeta{settings_mimic_raddialog-prefixed}{model}{raddialog-prefixed}
\chexdefmeta{settings_mimic_raddialog-prefixed}{nconfigs}{9}
\chexdefmeta{settings_mimic_raddialog-prefixed}{revision}{a0c5cb0da00d528500ce56f25e78d908a76a46a9}
\chexdefmeta{settings_mimic_raddialog-prefixed}{seed}{0}
\chexdefmeta{settings_mimic_raddialog-prefixed}{split}{test}
\chexdefmeta{settings_mimic_raddialog-prefixed}{temperature}{0.0}
\chexdefmeta{settings_mimic_random-report}{backend}{hf}
\chexdefmeta{settings_mimic_random-report}{dataset}{mimic}
\chexdefmeta{settings_mimic_random-report}{model}{random-report}
\chexdefmeta{settings_mimic_random-report}{nconfigs}{9}
\chexdefmeta{settings_mimic_random-report}{revision}{random-report-v1}
\chexdefmeta{settings_mimic_random-report}{seed}{0}
\chexdefmeta{settings_mimic_random-report}{split}{test}
\chexdefmeta{settings_mimic_random-report}{temperature}{0.0}
\chexdefmeta{settings_mimic_rgrg}{backend}{hf}
\chexdefmeta{settings_mimic_rgrg}{dataset}{mimic}
\chexdefmeta{settings_mimic_rgrg}{model}{rgrg}
\chexdefmeta{settings_mimic_rgrg}{nconfigs}{2}
\chexdefmeta{settings_mimic_rgrg}{revision}{rgrg@9520b6d+gdrive:1rDxqzOhjqydsOrITJrX0Rj1PAdMeP7Wy}
\chexdefmeta{settings_mimic_rgrg}{seed}{0}
\chexdefmeta{settings_mimic_rgrg}{split}{test}
\chexdefmeta{settings_mimic_rgrg}{temperature}{0.0}
\chexdefmeta{settings_mimic_upper-bound}{backend}{hf}
\chexdefmeta{settings_mimic_upper-bound}{dataset}{mimic}
\chexdefmeta{settings_mimic_upper-bound}{model}{empty}
\chexdefmeta{settings_mimic_upper-bound}{nconfigs}{9}
\chexdefmeta{settings_mimic_upper-bound}{revision}{empty-v1}
\chexdefmeta{settings_mimic_upper-bound}{seed}{0}
\chexdefmeta{settings_mimic_upper-bound}{split}{test}
\chexdefmeta{settings_mimic_upper-bound}{temperature}{0.0}
\chexdefmeta{settings_openi_chexagent-2-3b-native}{backend}{hf}
\chexdefmeta{settings_openi_chexagent-2-3b-native}{dataset}{openi}
\chexdefmeta{settings_openi_chexagent-2-3b-native}{model}{chexagent-2-3b-native}
\chexdefmeta{settings_openi_chexagent-2-3b-native}{nconfigs}{9}
\chexdefmeta{settings_openi_chexagent-2-3b-native}{revision}{8f19b53a2eceda4c33b0acec6c81fbc293ad80d0}
\chexdefmeta{settings_openi_chexagent-2-3b-native}{seed}{0}
\chexdefmeta{settings_openi_chexagent-2-3b-native}{split}{test}
\chexdefmeta{settings_openi_chexagent-2-3b-native}{temperature}{0.0}
\chexdefmeta{settings_openi_chexagent-2-3b-native-notags}{backend}{hf}
\chexdefmeta{settings_openi_chexagent-2-3b-native-notags}{dataset}{openi}
\chexdefmeta{settings_openi_chexagent-2-3b-native-notags}{model}{chexagent-2-3b-native}
\chexdefmeta{settings_openi_chexagent-2-3b-native-notags}{nconfigs}{9}
\chexdefmeta{settings_openi_chexagent-2-3b-native-notags}{revision}{8f19b53a2eceda4c33b0acec6c81fbc293ad80d0}
\chexdefmeta{settings_openi_chexagent-2-3b-native-notags}{seed}{0}
\chexdefmeta{settings_openi_chexagent-2-3b-native-notags}{split}{test}
\chexdefmeta{settings_openi_chexagent-2-3b-native-notags}{temperature}{0.0}
\chexdefmeta{settings_openi_chexone-instruct}{backend}{vllm}
\chexdefmeta{settings_openi_chexone-instruct}{dataset}{openi}
\chexdefmeta{settings_openi_chexone-instruct}{maxtokens}{2048}
\chexdefmeta{settings_openi_chexone-instruct}{model}{chexone-instruct}
\chexdefmeta{settings_openi_chexone-instruct}{nconfigs}{9}
\chexdefmeta{settings_openi_chexone-instruct}{revision}{0c350e6852ea08f9d9baf3b7595c1a10d4849927}
\chexdefmeta{settings_openi_chexone-instruct}{seed}{0}
\chexdefmeta{settings_openi_chexone-instruct}{split}{test}
\chexdefmeta{settings_openi_chexone-instruct}{temperature}{0.0}
\chexdefmeta{settings_openi_chexone-reasoning}{backend}{vllm}
\chexdefmeta{settings_openi_chexone-reasoning}{dataset}{openi}
\chexdefmeta{settings_openi_chexone-reasoning}{maxtokens}{2048}
\chexdefmeta{settings_openi_chexone-reasoning}{model}{chexone-reasoning}
\chexdefmeta{settings_openi_chexone-reasoning}{nconfigs}{9}
\chexdefmeta{settings_openi_chexone-reasoning}{revision}{0c350e6852ea08f9d9baf3b7595c1a10d4849927}
\chexdefmeta{settings_openi_chexone-reasoning}{seed}{0}
\chexdefmeta{settings_openi_chexone-reasoning}{split}{test}
\chexdefmeta{settings_openi_chexone-reasoning}{temperature}{0.0}
\chexdefmeta{settings_openi_constant-normal}{backend}{hf}
\chexdefmeta{settings_openi_constant-normal}{dataset}{openi}
\chexdefmeta{settings_openi_constant-normal}{model}{constant-normal}
\chexdefmeta{settings_openi_constant-normal}{nconfigs}{9}
\chexdefmeta{settings_openi_constant-normal}{revision}{constant-normal-v1}
\chexdefmeta{settings_openi_constant-normal}{seed}{0}
\chexdefmeta{settings_openi_constant-normal}{split}{test}
\chexdefmeta{settings_openi_constant-normal}{temperature}{0.0}
\chexdefmeta{settings_openi_copy-indication}{backend}{hf}
\chexdefmeta{settings_openi_copy-indication}{dataset}{openi}
\chexdefmeta{settings_openi_copy-indication}{model}{copy-indication}
\chexdefmeta{settings_openi_copy-indication}{nconfigs}{9}
\chexdefmeta{settings_openi_copy-indication}{revision}{copy-indication-v1}
\chexdefmeta{settings_openi_copy-indication}{seed}{0}
\chexdefmeta{settings_openi_copy-indication}{split}{test}
\chexdefmeta{settings_openi_copy-indication}{temperature}{0.0}
\chexdefmeta{settings_openi_copy-prior}{backend}{hf}
\chexdefmeta{settings_openi_copy-prior}{dataset}{openi}
\chexdefmeta{settings_openi_copy-prior}{model}{copy-prior}
\chexdefmeta{settings_openi_copy-prior}{nconfigs}{9}
\chexdefmeta{settings_openi_copy-prior}{revision}{copy-prior-v1}
\chexdefmeta{settings_openi_copy-prior}{seed}{0}
\chexdefmeta{settings_openi_copy-prior}{split}{test}
\chexdefmeta{settings_openi_copy-prior}{temperature}{0.0}
\chexdefmeta{settings_openi_cxr-llava-v2}{backend}{hf}
\chexdefmeta{settings_openi_cxr-llava-v2}{dataset}{openi}
\chexdefmeta{settings_openi_cxr-llava-v2}{model}{cxr-llava-v2}
\chexdefmeta{settings_openi_cxr-llava-v2}{nconfigs}{9}
\chexdefmeta{settings_openi_cxr-llava-v2}{revision}{b2224786bb90d54b1e1291171866706cfbb44e2b}
\chexdefmeta{settings_openi_cxr-llava-v2}{seed}{0}
\chexdefmeta{settings_openi_cxr-llava-v2}{split}{test}
\chexdefmeta{settings_openi_cxr-llava-v2}{temperature}{0.0}
\chexdefmeta{settings_openi_empty}{backend}{hf}
\chexdefmeta{settings_openi_empty}{dataset}{openi}
\chexdefmeta{settings_openi_empty}{model}{empty}
\chexdefmeta{settings_openi_empty}{nconfigs}{9}
\chexdefmeta{settings_openi_empty}{revision}{empty-v1}
\chexdefmeta{settings_openi_empty}{seed}{0}
\chexdefmeta{settings_openi_empty}{split}{test}
\chexdefmeta{settings_openi_empty}{temperature}{0.0}
\chexdefmeta{settings_openi_internvl3-8b}{backend}{vllm}
\chexdefmeta{settings_openi_internvl3-8b}{dataset}{openi}
\chexdefmeta{settings_openi_internvl3-8b}{maxtokens}{768}
\chexdefmeta{settings_openi_internvl3-8b}{model}{internvl3-8b}
\chexdefmeta{settings_openi_internvl3-8b}{nconfigs}{9}
\chexdefmeta{settings_openi_internvl3-8b}{revision}{853e3a797a661694b1b8ece0cb72dc2b23e3dac9}
\chexdefmeta{settings_openi_internvl3-8b}{seed}{0}
\chexdefmeta{settings_openi_internvl3-8b}{split}{test}
\chexdefmeta{settings_openi_internvl3-8b}{temperature}{0.0}
\chexdefmeta{settings_openi_llava-med}{backend}{hf}
\chexdefmeta{settings_openi_llava-med}{dataset}{openi}
\chexdefmeta{settings_openi_llava-med}{model}{llava-med}
\chexdefmeta{settings_openi_llava-med}{nconfigs}{9}
\chexdefmeta{settings_openi_llava-med}{revision}{91bb16c122001ddc9cf1fd36ce1dae09448943a2}
\chexdefmeta{settings_openi_llava-med}{seed}{0}
\chexdefmeta{settings_openi_llava-med}{split}{test}
\chexdefmeta{settings_openi_llava-med}{temperature}{0.0}
\chexdefmeta{settings_openi_llava-rad}{backend}{hf}
\chexdefmeta{settings_openi_llava-rad}{dataset}{openi}
\chexdefmeta{settings_openi_llava-rad}{model}{llava-rad}
\chexdefmeta{settings_openi_llava-rad}{nconfigs}{9}
\chexdefmeta{settings_openi_llava-rad}{revision}{dcdbc6caf6806c9acb66e12a827a39c86e00f6a7}
\chexdefmeta{settings_openi_llava-rad}{seed}{0}
\chexdefmeta{settings_openi_llava-rad}{split}{test}
\chexdefmeta{settings_openi_llava-rad}{temperature}{0.0}
\chexdefmeta{settings_openi_maira-2}{backend}{hf}
\chexdefmeta{settings_openi_maira-2}{dataset}{openi}
\chexdefmeta{settings_openi_maira-2}{model}{maira-2}
\chexdefmeta{settings_openi_maira-2}{nconfigs}{9}
\chexdefmeta{settings_openi_maira-2}{revision}{795a2b1cd4a310624b4e3d14b5a23e41fd273deb}
\chexdefmeta{settings_openi_maira-2}{seed}{0}
\chexdefmeta{settings_openi_maira-2}{split}{test}
\chexdefmeta{settings_openi_maira-2}{temperature}{0.0}
\chexdefmeta{settings_openi_medgemma-4b}{backend}{vllm}
\chexdefmeta{settings_openi_medgemma-4b}{dataset}{openi}
\chexdefmeta{settings_openi_medgemma-4b}{maxtokens}{768}
\chexdefmeta{settings_openi_medgemma-4b}{model}{medgemma-4b}
\chexdefmeta{settings_openi_medgemma-4b}{nconfigs}{9}
\chexdefmeta{settings_openi_medgemma-4b}{revision}{290cda5eeccbee130f987c4ad74a59ae6f196408}
\chexdefmeta{settings_openi_medgemma-4b}{seed}{0}
\chexdefmeta{settings_openi_medgemma-4b}{split}{test}
\chexdefmeta{settings_openi_medgemma-4b}{temperature}{0.0}
\chexdefmeta{settings_openi_most-common-report}{backend}{hf}
\chexdefmeta{settings_openi_most-common-report}{dataset}{openi}
\chexdefmeta{settings_openi_most-common-report}{model}{most-common-report}
\chexdefmeta{settings_openi_most-common-report}{nconfigs}{9}
\chexdefmeta{settings_openi_most-common-report}{revision}{most-common-report-v1}
\chexdefmeta{settings_openi_most-common-report}{seed}{0}
\chexdefmeta{settings_openi_most-common-report}{split}{test}
\chexdefmeta{settings_openi_most-common-report}{temperature}{0.0}
\chexdefmeta{settings_openi_qwen2.5-vl-3b}{backend}{vllm}
\chexdefmeta{settings_openi_qwen2.5-vl-3b}{dataset}{openi}
\chexdefmeta{settings_openi_qwen2.5-vl-3b}{maxtokens}{768}
\chexdefmeta{settings_openi_qwen2.5-vl-3b}{model}{qwen2.5-vl-3b}
\chexdefmeta{settings_openi_qwen2.5-vl-3b}{nconfigs}{9}
\chexdefmeta{settings_openi_qwen2.5-vl-3b}{revision}{66285546d2b821cf421d4f5eb2576359d3770cd3}
\chexdefmeta{settings_openi_qwen2.5-vl-3b}{seed}{0}
\chexdefmeta{settings_openi_qwen2.5-vl-3b}{split}{test}
\chexdefmeta{settings_openi_qwen2.5-vl-3b}{temperature}{0.0}
\chexdefmeta{settings_openi_qwen2.5-vl-7b}{backend}{vllm}
\chexdefmeta{settings_openi_qwen2.5-vl-7b}{dataset}{openi}
\chexdefmeta{settings_openi_qwen2.5-vl-7b}{maxtokens}{768}
\chexdefmeta{settings_openi_qwen2.5-vl-7b}{model}{qwen2.5-vl-7b}
\chexdefmeta{settings_openi_qwen2.5-vl-7b}{nconfigs}{9}
\chexdefmeta{settings_openi_qwen2.5-vl-7b}{revision}{cc594898137f460bfe9f0759e9844b3ce807cfb5}
\chexdefmeta{settings_openi_qwen2.5-vl-7b}{seed}{0}
\chexdefmeta{settings_openi_qwen2.5-vl-7b}{split}{test}
\chexdefmeta{settings_openi_qwen2.5-vl-7b}{temperature}{0.0}
\chexdefmeta{settings_openi_raddialog}{backend}{hf}
\chexdefmeta{settings_openi_raddialog}{dataset}{openi}
\chexdefmeta{settings_openi_raddialog}{model}{raddialog}
\chexdefmeta{settings_openi_raddialog}{nconfigs}{9}
\chexdefmeta{settings_openi_raddialog}{revision}{a0c5cb0da00d528500ce56f25e78d908a76a46a9}
\chexdefmeta{settings_openi_raddialog}{seed}{0}
\chexdefmeta{settings_openi_raddialog}{split}{test}
\chexdefmeta{settings_openi_raddialog}{temperature}{0.0}
\chexdefmeta{settings_openi_raddialog-prefixed}{backend}{hf}
\chexdefmeta{settings_openi_raddialog-prefixed}{dataset}{openi}
\chexdefmeta{settings_openi_raddialog-prefixed}{model}{raddialog-prefixed}
\chexdefmeta{settings_openi_raddialog-prefixed}{nconfigs}{9}
\chexdefmeta{settings_openi_raddialog-prefixed}{revision}{a0c5cb0da00d528500ce56f25e78d908a76a46a9}
\chexdefmeta{settings_openi_raddialog-prefixed}{seed}{0}
\chexdefmeta{settings_openi_raddialog-prefixed}{split}{test}
\chexdefmeta{settings_openi_raddialog-prefixed}{temperature}{0.0}
\chexdefmeta{settings_openi_random-report}{backend}{hf}
\chexdefmeta{settings_openi_random-report}{dataset}{openi}
\chexdefmeta{settings_openi_random-report}{model}{random-report}
\chexdefmeta{settings_openi_random-report}{nconfigs}{9}
\chexdefmeta{settings_openi_random-report}{revision}{random-report-v1}
\chexdefmeta{settings_openi_random-report}{seed}{0}
\chexdefmeta{settings_openi_random-report}{split}{test}
\chexdefmeta{settings_openi_random-report}{temperature}{0.0}
\chexdefmeta{settings_openi_rgrg}{backend}{hf}
\chexdefmeta{settings_openi_rgrg}{dataset}{openi}
\chexdefmeta{settings_openi_rgrg}{model}{rgrg}
\chexdefmeta{settings_openi_rgrg}{nconfigs}{2}
\chexdefmeta{settings_openi_rgrg}{revision}{rgrg@9520b6d+gdrive:1rDxqzOhjqydsOrITJrX0Rj1PAdMeP7Wy}
\chexdefmeta{settings_openi_rgrg}{seed}{0}
\chexdefmeta{settings_openi_rgrg}{split}{test}
\chexdefmeta{settings_openi_rgrg}{temperature}{0.0}
\chexdefmeta{settings_openi_upper-bound}{backend}{hf}
\chexdefmeta{settings_openi_upper-bound}{dataset}{openi}
\chexdefmeta{settings_openi_upper-bound}{model}{empty}
\chexdefmeta{settings_openi_upper-bound}{nconfigs}{9}
\chexdefmeta{settings_openi_upper-bound}{revision}{empty-v1}
\chexdefmeta{settings_openi_upper-bound}{seed}{0}
\chexdefmeta{settings_openi_upper-bound}{split}{test}
\chexdefmeta{settings_openi_upper-bound}{temperature}{0.0}

\chexdefskip{critical_mimic_chexagent-2-3b-native}{B_f_both}{empty}{0}
\chexdefskip{critical_mimic_chexagent-2-3b-native}{B_f_both}{empty_target}{22}
\chexdefskip{critical_mimic_chexagent-2-3b-native}{B_f_both}{generated}{53}
\chexdefskip{critical_mimic_chexagent-2-3b-native}{B_f_both}{length_truncated}{0}
\chexdefskip{critical_mimic_chexagent-2-3b-native}{B_f_both}{non_english}{0}
\chexdefskip{critical_mimic_chexagent-2-3b-native}{B_f_both}{refused}{0}
\chexdefskip{critical_mimic_chexagent-2-3b-native}{B_f_both}{truncations}{0}
\chexdefskip{critical_mimic_chexagent-2-3b-native}{B_f_findings}{empty}{0}
\chexdefskip{critical_mimic_chexagent-2-3b-native}{B_f_findings}{empty_target}{22}
\chexdefskip{critical_mimic_chexagent-2-3b-native}{B_f_findings}{generated}{53}
\chexdefskip{critical_mimic_chexagent-2-3b-native}{B_f_findings}{length_truncated}{0}
\chexdefskip{critical_mimic_chexagent-2-3b-native}{B_f_findings}{non_english}{0}
\chexdefskip{critical_mimic_chexagent-2-3b-native}{B_f_findings}{refused}{0}
\chexdefskip{critical_mimic_chexagent-2-3b-native}{B_f_findings}{truncations}{0}
\chexdefskip{critical_mimic_chexagent-2-3b-native}{B_f_impression}{empty}{0}
\chexdefskip{critical_mimic_chexagent-2-3b-native}{B_f_impression}{generated}{75}
\chexdefskip{critical_mimic_chexagent-2-3b-native}{B_f_impression}{length_truncated}{0}
\chexdefskip{critical_mimic_chexagent-2-3b-native}{B_f_impression}{non_english}{0}
\chexdefskip{critical_mimic_chexagent-2-3b-native}{B_f_impression}{refused}{0}
\chexdefskip{critical_mimic_chexagent-2-3b-native}{B_f_impression}{truncations}{0}
\chexdefskip{critical_mimic_chexagent-2-3b-native}{P_f_both}{empty}{0}
\chexdefskip{critical_mimic_chexagent-2-3b-native}{P_f_both}{empty_target}{22}
\chexdefskip{critical_mimic_chexagent-2-3b-native}{P_f_both}{generated}{53}
\chexdefskip{critical_mimic_chexagent-2-3b-native}{P_f_both}{length_truncated}{0}
\chexdefskip{critical_mimic_chexagent-2-3b-native}{P_f_both}{non_english}{0}
\chexdefskip{critical_mimic_chexagent-2-3b-native}{P_f_both}{refused}{0}
\chexdefskip{critical_mimic_chexagent-2-3b-native}{P_f_both}{truncations}{0}
\chexdefskip{critical_mimic_chexagent-2-3b-native}{P_f_findings}{empty}{0}
\chexdefskip{critical_mimic_chexagent-2-3b-native}{P_f_findings}{empty_target}{22}
\chexdefskip{critical_mimic_chexagent-2-3b-native}{P_f_findings}{generated}{53}
\chexdefskip{critical_mimic_chexagent-2-3b-native}{P_f_findings}{length_truncated}{0}
\chexdefskip{critical_mimic_chexagent-2-3b-native}{P_f_findings}{non_english}{0}
\chexdefskip{critical_mimic_chexagent-2-3b-native}{P_f_findings}{refused}{0}
\chexdefskip{critical_mimic_chexagent-2-3b-native}{P_f_findings}{truncations}{0}
\chexdefskip{critical_mimic_chexagent-2-3b-native}{P_f_impression}{empty}{0}
\chexdefskip{critical_mimic_chexagent-2-3b-native}{P_f_impression}{generated}{75}
\chexdefskip{critical_mimic_chexagent-2-3b-native}{P_f_impression}{length_truncated}{0}
\chexdefskip{critical_mimic_chexagent-2-3b-native}{P_f_impression}{non_english}{0}
\chexdefskip{critical_mimic_chexagent-2-3b-native}{P_f_impression}{refused}{0}
\chexdefskip{critical_mimic_chexagent-2-3b-native}{P_f_impression}{truncations}{0}
\chexdefskip{critical_mimic_chexagent-2-3b-native}{P_n_both}{empty}{0}
\chexdefskip{critical_mimic_chexagent-2-3b-native}{P_n_both}{generated}{0}
\chexdefskip{critical_mimic_chexagent-2-3b-native}{P_n_both}{length_truncated}{0}
\chexdefskip{critical_mimic_chexagent-2-3b-native}{P_n_both}{non_english}{0}
\chexdefskip{critical_mimic_chexagent-2-3b-native}{P_n_both}{refused}{0}
\chexdefskip{critical_mimic_chexagent-2-3b-native}{P_n_both}{text_only_unsupported}{75}
\chexdefskip{critical_mimic_chexagent-2-3b-native}{P_n_both}{truncations}{0}
\chexdefskip{critical_mimic_chexagent-2-3b-native}{P_n_findings}{empty}{0}
\chexdefskip{critical_mimic_chexagent-2-3b-native}{P_n_findings}{generated}{0}
\chexdefskip{critical_mimic_chexagent-2-3b-native}{P_n_findings}{length_truncated}{0}
\chexdefskip{critical_mimic_chexagent-2-3b-native}{P_n_findings}{non_english}{0}
\chexdefskip{critical_mimic_chexagent-2-3b-native}{P_n_findings}{refused}{0}
\chexdefskip{critical_mimic_chexagent-2-3b-native}{P_n_findings}{text_only_unsupported}{75}
\chexdefskip{critical_mimic_chexagent-2-3b-native}{P_n_findings}{truncations}{0}
\chexdefskip{critical_mimic_chexagent-2-3b-native}{P_n_impression}{empty}{0}
\chexdefskip{critical_mimic_chexagent-2-3b-native}{P_n_impression}{generated}{0}
\chexdefskip{critical_mimic_chexagent-2-3b-native}{P_n_impression}{length_truncated}{0}
\chexdefskip{critical_mimic_chexagent-2-3b-native}{P_n_impression}{non_english}{0}
\chexdefskip{critical_mimic_chexagent-2-3b-native}{P_n_impression}{refused}{0}
\chexdefskip{critical_mimic_chexagent-2-3b-native}{P_n_impression}{text_only_unsupported}{75}
\chexdefskip{critical_mimic_chexagent-2-3b-native}{P_n_impression}{truncations}{0}
\chexdefskip{critical_mimic_chexagent-2-3b-native-notags}{B_f_both}{empty}{0}
\chexdefskip{critical_mimic_chexagent-2-3b-native-notags}{B_f_both}{empty_target}{22}
\chexdefskip{critical_mimic_chexagent-2-3b-native-notags}{B_f_both}{generated}{53}
\chexdefskip{critical_mimic_chexagent-2-3b-native-notags}{B_f_both}{length_truncated}{0}
\chexdefskip{critical_mimic_chexagent-2-3b-native-notags}{B_f_both}{non_english}{0}
\chexdefskip{critical_mimic_chexagent-2-3b-native-notags}{B_f_both}{refused}{0}
\chexdefskip{critical_mimic_chexagent-2-3b-native-notags}{B_f_both}{truncations}{0}
\chexdefskip{critical_mimic_chexagent-2-3b-native-notags}{B_f_findings}{empty}{0}
\chexdefskip{critical_mimic_chexagent-2-3b-native-notags}{B_f_findings}{empty_target}{22}
\chexdefskip{critical_mimic_chexagent-2-3b-native-notags}{B_f_findings}{generated}{53}
\chexdefskip{critical_mimic_chexagent-2-3b-native-notags}{B_f_findings}{length_truncated}{0}
\chexdefskip{critical_mimic_chexagent-2-3b-native-notags}{B_f_findings}{non_english}{0}
\chexdefskip{critical_mimic_chexagent-2-3b-native-notags}{B_f_findings}{refused}{0}
\chexdefskip{critical_mimic_chexagent-2-3b-native-notags}{B_f_findings}{truncations}{0}
\chexdefskip{critical_mimic_chexagent-2-3b-native-notags}{B_f_impression}{empty}{0}
\chexdefskip{critical_mimic_chexagent-2-3b-native-notags}{B_f_impression}{generated}{75}
\chexdefskip{critical_mimic_chexagent-2-3b-native-notags}{B_f_impression}{length_truncated}{0}
\chexdefskip{critical_mimic_chexagent-2-3b-native-notags}{B_f_impression}{non_english}{0}
\chexdefskip{critical_mimic_chexagent-2-3b-native-notags}{B_f_impression}{refused}{0}
\chexdefskip{critical_mimic_chexagent-2-3b-native-notags}{B_f_impression}{truncations}{0}
\chexdefskip{critical_mimic_chexagent-2-3b-native-notags}{P_f_both}{empty}{0}
\chexdefskip{critical_mimic_chexagent-2-3b-native-notags}{P_f_both}{empty_target}{22}
\chexdefskip{critical_mimic_chexagent-2-3b-native-notags}{P_f_both}{generated}{53}
\chexdefskip{critical_mimic_chexagent-2-3b-native-notags}{P_f_both}{length_truncated}{0}
\chexdefskip{critical_mimic_chexagent-2-3b-native-notags}{P_f_both}{non_english}{0}
\chexdefskip{critical_mimic_chexagent-2-3b-native-notags}{P_f_both}{refused}{0}
\chexdefskip{critical_mimic_chexagent-2-3b-native-notags}{P_f_both}{truncations}{0}
\chexdefskip{critical_mimic_chexagent-2-3b-native-notags}{P_f_findings}{empty}{0}
\chexdefskip{critical_mimic_chexagent-2-3b-native-notags}{P_f_findings}{empty_target}{22}
\chexdefskip{critical_mimic_chexagent-2-3b-native-notags}{P_f_findings}{generated}{53}
\chexdefskip{critical_mimic_chexagent-2-3b-native-notags}{P_f_findings}{length_truncated}{0}
\chexdefskip{critical_mimic_chexagent-2-3b-native-notags}{P_f_findings}{non_english}{0}
\chexdefskip{critical_mimic_chexagent-2-3b-native-notags}{P_f_findings}{refused}{0}
\chexdefskip{critical_mimic_chexagent-2-3b-native-notags}{P_f_findings}{truncations}{0}
\chexdefskip{critical_mimic_chexagent-2-3b-native-notags}{P_f_impression}{empty}{0}
\chexdefskip{critical_mimic_chexagent-2-3b-native-notags}{P_f_impression}{generated}{75}
\chexdefskip{critical_mimic_chexagent-2-3b-native-notags}{P_f_impression}{length_truncated}{0}
\chexdefskip{critical_mimic_chexagent-2-3b-native-notags}{P_f_impression}{non_english}{0}
\chexdefskip{critical_mimic_chexagent-2-3b-native-notags}{P_f_impression}{refused}{0}
\chexdefskip{critical_mimic_chexagent-2-3b-native-notags}{P_f_impression}{truncations}{0}
\chexdefskip{critical_mimic_chexagent-2-3b-native-notags}{P_n_both}{empty}{0}
\chexdefskip{critical_mimic_chexagent-2-3b-native-notags}{P_n_both}{generated}{0}
\chexdefskip{critical_mimic_chexagent-2-3b-native-notags}{P_n_both}{length_truncated}{0}
\chexdefskip{critical_mimic_chexagent-2-3b-native-notags}{P_n_both}{non_english}{0}
\chexdefskip{critical_mimic_chexagent-2-3b-native-notags}{P_n_both}{refused}{0}
\chexdefskip{critical_mimic_chexagent-2-3b-native-notags}{P_n_both}{text_only_unsupported}{75}
\chexdefskip{critical_mimic_chexagent-2-3b-native-notags}{P_n_both}{truncations}{0}
\chexdefskip{critical_mimic_chexagent-2-3b-native-notags}{P_n_findings}{empty}{0}
\chexdefskip{critical_mimic_chexagent-2-3b-native-notags}{P_n_findings}{generated}{0}
\chexdefskip{critical_mimic_chexagent-2-3b-native-notags}{P_n_findings}{length_truncated}{0}
\chexdefskip{critical_mimic_chexagent-2-3b-native-notags}{P_n_findings}{non_english}{0}
\chexdefskip{critical_mimic_chexagent-2-3b-native-notags}{P_n_findings}{refused}{0}
\chexdefskip{critical_mimic_chexagent-2-3b-native-notags}{P_n_findings}{text_only_unsupported}{75}
\chexdefskip{critical_mimic_chexagent-2-3b-native-notags}{P_n_findings}{truncations}{0}
\chexdefskip{critical_mimic_chexagent-2-3b-native-notags}{P_n_impression}{empty}{0}
\chexdefskip{critical_mimic_chexagent-2-3b-native-notags}{P_n_impression}{generated}{0}
\chexdefskip{critical_mimic_chexagent-2-3b-native-notags}{P_n_impression}{length_truncated}{0}
\chexdefskip{critical_mimic_chexagent-2-3b-native-notags}{P_n_impression}{non_english}{0}
\chexdefskip{critical_mimic_chexagent-2-3b-native-notags}{P_n_impression}{refused}{0}
\chexdefskip{critical_mimic_chexagent-2-3b-native-notags}{P_n_impression}{text_only_unsupported}{75}
\chexdefskip{critical_mimic_chexagent-2-3b-native-notags}{P_n_impression}{truncations}{0}
\chexdefskip{critical_mimic_chexone-instruct}{B_f_both}{empty}{0}
\chexdefskip{critical_mimic_chexone-instruct}{B_f_both}{empty_target}{22}
\chexdefskip{critical_mimic_chexone-instruct}{B_f_both}{generated}{53}
\chexdefskip{critical_mimic_chexone-instruct}{B_f_both}{length_truncated}{3}
\chexdefskip{critical_mimic_chexone-instruct}{B_f_both}{non_english}{0}
\chexdefskip{critical_mimic_chexone-instruct}{B_f_both}{refused}{0}
\chexdefskip{critical_mimic_chexone-instruct}{B_f_both}{truncations}{0}
\chexdefskip{critical_mimic_chexone-instruct}{B_f_findings}{empty}{0}
\chexdefskip{critical_mimic_chexone-instruct}{B_f_findings}{empty_target}{22}
\chexdefskip{critical_mimic_chexone-instruct}{B_f_findings}{generated}{53}
\chexdefskip{critical_mimic_chexone-instruct}{B_f_findings}{length_truncated}{0}
\chexdefskip{critical_mimic_chexone-instruct}{B_f_findings}{non_english}{0}
\chexdefskip{critical_mimic_chexone-instruct}{B_f_findings}{refused}{0}
\chexdefskip{critical_mimic_chexone-instruct}{B_f_findings}{truncations}{0}
\chexdefskip{critical_mimic_chexone-instruct}{B_f_impression}{empty}{0}
\chexdefskip{critical_mimic_chexone-instruct}{B_f_impression}{generated}{75}
\chexdefskip{critical_mimic_chexone-instruct}{B_f_impression}{length_truncated}{4}
\chexdefskip{critical_mimic_chexone-instruct}{B_f_impression}{non_english}{0}
\chexdefskip{critical_mimic_chexone-instruct}{B_f_impression}{refused}{0}
\chexdefskip{critical_mimic_chexone-instruct}{B_f_impression}{truncations}{0}
\chexdefskip{critical_mimic_chexone-instruct}{P_f_both}{empty}{0}
\chexdefskip{critical_mimic_chexone-instruct}{P_f_both}{empty_target}{22}
\chexdefskip{critical_mimic_chexone-instruct}{P_f_both}{generated}{53}
\chexdefskip{critical_mimic_chexone-instruct}{P_f_both}{length_truncated}{2}
\chexdefskip{critical_mimic_chexone-instruct}{P_f_both}{non_english}{0}
\chexdefskip{critical_mimic_chexone-instruct}{P_f_both}{refused}{0}
\chexdefskip{critical_mimic_chexone-instruct}{P_f_both}{truncations}{0}
\chexdefskip{critical_mimic_chexone-instruct}{P_f_findings}{empty}{0}
\chexdefskip{critical_mimic_chexone-instruct}{P_f_findings}{empty_target}{22}
\chexdefskip{critical_mimic_chexone-instruct}{P_f_findings}{generated}{53}
\chexdefskip{critical_mimic_chexone-instruct}{P_f_findings}{length_truncated}{0}
\chexdefskip{critical_mimic_chexone-instruct}{P_f_findings}{non_english}{0}
\chexdefskip{critical_mimic_chexone-instruct}{P_f_findings}{refused}{0}
\chexdefskip{critical_mimic_chexone-instruct}{P_f_findings}{truncations}{0}
\chexdefskip{critical_mimic_chexone-instruct}{P_f_impression}{empty}{0}
\chexdefskip{critical_mimic_chexone-instruct}{P_f_impression}{generated}{75}
\chexdefskip{critical_mimic_chexone-instruct}{P_f_impression}{length_truncated}{2}
\chexdefskip{critical_mimic_chexone-instruct}{P_f_impression}{non_english}{0}
\chexdefskip{critical_mimic_chexone-instruct}{P_f_impression}{refused}{0}
\chexdefskip{critical_mimic_chexone-instruct}{P_f_impression}{truncations}{0}
\chexdefskip{critical_mimic_chexone-instruct}{P_n_both}{empty}{0}
\chexdefskip{critical_mimic_chexone-instruct}{P_n_both}{empty_target}{22}
\chexdefskip{critical_mimic_chexone-instruct}{P_n_both}{generated}{53}
\chexdefskip{critical_mimic_chexone-instruct}{P_n_both}{length_truncated}{1}
\chexdefskip{critical_mimic_chexone-instruct}{P_n_both}{non_english}{0}
\chexdefskip{critical_mimic_chexone-instruct}{P_n_both}{refused}{0}
\chexdefskip{critical_mimic_chexone-instruct}{P_n_both}{truncations}{0}
\chexdefskip{critical_mimic_chexone-instruct}{P_n_findings}{empty}{0}
\chexdefskip{critical_mimic_chexone-instruct}{P_n_findings}{empty_target}{22}
\chexdefskip{critical_mimic_chexone-instruct}{P_n_findings}{generated}{53}
\chexdefskip{critical_mimic_chexone-instruct}{P_n_findings}{length_truncated}{0}
\chexdefskip{critical_mimic_chexone-instruct}{P_n_findings}{non_english}{0}
\chexdefskip{critical_mimic_chexone-instruct}{P_n_findings}{refused}{0}
\chexdefskip{critical_mimic_chexone-instruct}{P_n_findings}{truncations}{0}
\chexdefskip{critical_mimic_chexone-instruct}{P_n_impression}{empty}{0}
\chexdefskip{critical_mimic_chexone-instruct}{P_n_impression}{generated}{75}
\chexdefskip{critical_mimic_chexone-instruct}{P_n_impression}{length_truncated}{1}
\chexdefskip{critical_mimic_chexone-instruct}{P_n_impression}{non_english}{0}
\chexdefskip{critical_mimic_chexone-instruct}{P_n_impression}{refused}{0}
\chexdefskip{critical_mimic_chexone-instruct}{P_n_impression}{truncations}{0}
\chexdefskip{critical_mimic_chexone-reasoning}{B_f_both}{empty}{0}
\chexdefskip{critical_mimic_chexone-reasoning}{B_f_both}{empty_target}{22}
\chexdefskip{critical_mimic_chexone-reasoning}{B_f_both}{generated}{53}
\chexdefskip{critical_mimic_chexone-reasoning}{B_f_both}{length_truncated}{1}
\chexdefskip{critical_mimic_chexone-reasoning}{B_f_both}{non_english}{0}
\chexdefskip{critical_mimic_chexone-reasoning}{B_f_both}{refused}{0}
\chexdefskip{critical_mimic_chexone-reasoning}{B_f_both}{truncations}{0}
\chexdefskip{critical_mimic_chexone-reasoning}{B_f_findings}{empty}{1}
\chexdefskip{critical_mimic_chexone-reasoning}{B_f_findings}{empty_target}{22}
\chexdefskip{critical_mimic_chexone-reasoning}{B_f_findings}{generated}{53}
\chexdefskip{critical_mimic_chexone-reasoning}{B_f_findings}{length_truncated}{1}
\chexdefskip{critical_mimic_chexone-reasoning}{B_f_findings}{non_english}{0}
\chexdefskip{critical_mimic_chexone-reasoning}{B_f_findings}{refused}{0}
\chexdefskip{critical_mimic_chexone-reasoning}{B_f_findings}{truncations}{0}
\chexdefskip{critical_mimic_chexone-reasoning}{B_f_impression}{empty}{0}
\chexdefskip{critical_mimic_chexone-reasoning}{B_f_impression}{generated}{75}
\chexdefskip{critical_mimic_chexone-reasoning}{B_f_impression}{length_truncated}{0}
\chexdefskip{critical_mimic_chexone-reasoning}{B_f_impression}{non_english}{0}
\chexdefskip{critical_mimic_chexone-reasoning}{B_f_impression}{refused}{0}
\chexdefskip{critical_mimic_chexone-reasoning}{B_f_impression}{truncations}{0}
\chexdefskip{critical_mimic_chexone-reasoning}{P_f_both}{empty}{0}
\chexdefskip{critical_mimic_chexone-reasoning}{P_f_both}{empty_target}{22}
\chexdefskip{critical_mimic_chexone-reasoning}{P_f_both}{generated}{53}
\chexdefskip{critical_mimic_chexone-reasoning}{P_f_both}{length_truncated}{0}
\chexdefskip{critical_mimic_chexone-reasoning}{P_f_both}{non_english}{0}
\chexdefskip{critical_mimic_chexone-reasoning}{P_f_both}{refused}{0}
\chexdefskip{critical_mimic_chexone-reasoning}{P_f_both}{truncations}{0}
\chexdefskip{critical_mimic_chexone-reasoning}{P_f_findings}{empty}{0}
\chexdefskip{critical_mimic_chexone-reasoning}{P_f_findings}{empty_target}{22}
\chexdefskip{critical_mimic_chexone-reasoning}{P_f_findings}{generated}{53}
\chexdefskip{critical_mimic_chexone-reasoning}{P_f_findings}{length_truncated}{0}
\chexdefskip{critical_mimic_chexone-reasoning}{P_f_findings}{non_english}{0}
\chexdefskip{critical_mimic_chexone-reasoning}{P_f_findings}{refused}{0}
\chexdefskip{critical_mimic_chexone-reasoning}{P_f_findings}{truncations}{0}
\chexdefskip{critical_mimic_chexone-reasoning}{P_f_impression}{empty}{0}
\chexdefskip{critical_mimic_chexone-reasoning}{P_f_impression}{generated}{75}
\chexdefskip{critical_mimic_chexone-reasoning}{P_f_impression}{length_truncated}{0}
\chexdefskip{critical_mimic_chexone-reasoning}{P_f_impression}{non_english}{0}
\chexdefskip{critical_mimic_chexone-reasoning}{P_f_impression}{refused}{0}
\chexdefskip{critical_mimic_chexone-reasoning}{P_f_impression}{truncations}{0}
\chexdefskip{critical_mimic_chexone-reasoning}{P_n_both}{empty}{0}
\chexdefskip{critical_mimic_chexone-reasoning}{P_n_both}{empty_target}{22}
\chexdefskip{critical_mimic_chexone-reasoning}{P_n_both}{generated}{53}
\chexdefskip{critical_mimic_chexone-reasoning}{P_n_both}{length_truncated}{4}
\chexdefskip{critical_mimic_chexone-reasoning}{P_n_both}{non_english}{0}
\chexdefskip{critical_mimic_chexone-reasoning}{P_n_both}{refused}{0}
\chexdefskip{critical_mimic_chexone-reasoning}{P_n_both}{truncations}{0}
\chexdefskip{critical_mimic_chexone-reasoning}{P_n_findings}{empty}{4}
\chexdefskip{critical_mimic_chexone-reasoning}{P_n_findings}{empty_target}{22}
\chexdefskip{critical_mimic_chexone-reasoning}{P_n_findings}{generated}{53}
\chexdefskip{critical_mimic_chexone-reasoning}{P_n_findings}{length_truncated}{4}
\chexdefskip{critical_mimic_chexone-reasoning}{P_n_findings}{non_english}{0}
\chexdefskip{critical_mimic_chexone-reasoning}{P_n_findings}{refused}{0}
\chexdefskip{critical_mimic_chexone-reasoning}{P_n_findings}{truncations}{0}
\chexdefskip{critical_mimic_chexone-reasoning}{P_n_impression}{empty}{0}
\chexdefskip{critical_mimic_chexone-reasoning}{P_n_impression}{generated}{75}
\chexdefskip{critical_mimic_chexone-reasoning}{P_n_impression}{length_truncated}{0}
\chexdefskip{critical_mimic_chexone-reasoning}{P_n_impression}{non_english}{0}
\chexdefskip{critical_mimic_chexone-reasoning}{P_n_impression}{refused}{0}
\chexdefskip{critical_mimic_chexone-reasoning}{P_n_impression}{truncations}{0}
\chexdefskip{critical_mimic_constant-normal}{B_f_both}{empty}{0}
\chexdefskip{critical_mimic_constant-normal}{B_f_both}{empty_target}{22}
\chexdefskip{critical_mimic_constant-normal}{B_f_both}{generated}{53}
\chexdefskip{critical_mimic_constant-normal}{B_f_both}{length_truncated}{0}
\chexdefskip{critical_mimic_constant-normal}{B_f_both}{non_english}{0}
\chexdefskip{critical_mimic_constant-normal}{B_f_both}{refused}{0}
\chexdefskip{critical_mimic_constant-normal}{B_f_both}{truncations}{0}
\chexdefskip{critical_mimic_constant-normal}{B_f_findings}{empty}{0}
\chexdefskip{critical_mimic_constant-normal}{B_f_findings}{empty_target}{22}
\chexdefskip{critical_mimic_constant-normal}{B_f_findings}{generated}{53}
\chexdefskip{critical_mimic_constant-normal}{B_f_findings}{length_truncated}{0}
\chexdefskip{critical_mimic_constant-normal}{B_f_findings}{non_english}{0}
\chexdefskip{critical_mimic_constant-normal}{B_f_findings}{refused}{0}
\chexdefskip{critical_mimic_constant-normal}{B_f_findings}{truncations}{0}
\chexdefskip{critical_mimic_constant-normal}{B_f_impression}{empty}{0}
\chexdefskip{critical_mimic_constant-normal}{B_f_impression}{generated}{75}
\chexdefskip{critical_mimic_constant-normal}{B_f_impression}{length_truncated}{0}
\chexdefskip{critical_mimic_constant-normal}{B_f_impression}{non_english}{0}
\chexdefskip{critical_mimic_constant-normal}{B_f_impression}{refused}{0}
\chexdefskip{critical_mimic_constant-normal}{B_f_impression}{truncations}{0}
\chexdefskip{critical_mimic_constant-normal}{P_f_both}{empty}{0}
\chexdefskip{critical_mimic_constant-normal}{P_f_both}{empty_target}{22}
\chexdefskip{critical_mimic_constant-normal}{P_f_both}{generated}{53}
\chexdefskip{critical_mimic_constant-normal}{P_f_both}{length_truncated}{0}
\chexdefskip{critical_mimic_constant-normal}{P_f_both}{non_english}{0}
\chexdefskip{critical_mimic_constant-normal}{P_f_both}{refused}{0}
\chexdefskip{critical_mimic_constant-normal}{P_f_both}{truncations}{0}
\chexdefskip{critical_mimic_constant-normal}{P_f_findings}{empty}{0}
\chexdefskip{critical_mimic_constant-normal}{P_f_findings}{empty_target}{22}
\chexdefskip{critical_mimic_constant-normal}{P_f_findings}{generated}{53}
\chexdefskip{critical_mimic_constant-normal}{P_f_findings}{length_truncated}{0}
\chexdefskip{critical_mimic_constant-normal}{P_f_findings}{non_english}{0}
\chexdefskip{critical_mimic_constant-normal}{P_f_findings}{refused}{0}
\chexdefskip{critical_mimic_constant-normal}{P_f_findings}{truncations}{0}
\chexdefskip{critical_mimic_constant-normal}{P_f_impression}{empty}{0}
\chexdefskip{critical_mimic_constant-normal}{P_f_impression}{generated}{75}
\chexdefskip{critical_mimic_constant-normal}{P_f_impression}{length_truncated}{0}
\chexdefskip{critical_mimic_constant-normal}{P_f_impression}{non_english}{0}
\chexdefskip{critical_mimic_constant-normal}{P_f_impression}{refused}{0}
\chexdefskip{critical_mimic_constant-normal}{P_f_impression}{truncations}{0}
\chexdefskip{critical_mimic_constant-normal}{P_n_both}{empty}{0}
\chexdefskip{critical_mimic_constant-normal}{P_n_both}{empty_target}{22}
\chexdefskip{critical_mimic_constant-normal}{P_n_both}{generated}{53}
\chexdefskip{critical_mimic_constant-normal}{P_n_both}{length_truncated}{0}
\chexdefskip{critical_mimic_constant-normal}{P_n_both}{non_english}{0}
\chexdefskip{critical_mimic_constant-normal}{P_n_both}{refused}{0}
\chexdefskip{critical_mimic_constant-normal}{P_n_both}{truncations}{0}
\chexdefskip{critical_mimic_constant-normal}{P_n_findings}{empty}{0}
\chexdefskip{critical_mimic_constant-normal}{P_n_findings}{empty_target}{22}
\chexdefskip{critical_mimic_constant-normal}{P_n_findings}{generated}{53}
\chexdefskip{critical_mimic_constant-normal}{P_n_findings}{length_truncated}{0}
\chexdefskip{critical_mimic_constant-normal}{P_n_findings}{non_english}{0}
\chexdefskip{critical_mimic_constant-normal}{P_n_findings}{refused}{0}
\chexdefskip{critical_mimic_constant-normal}{P_n_findings}{truncations}{0}
\chexdefskip{critical_mimic_constant-normal}{P_n_impression}{empty}{0}
\chexdefskip{critical_mimic_constant-normal}{P_n_impression}{generated}{75}
\chexdefskip{critical_mimic_constant-normal}{P_n_impression}{length_truncated}{0}
\chexdefskip{critical_mimic_constant-normal}{P_n_impression}{non_english}{0}
\chexdefskip{critical_mimic_constant-normal}{P_n_impression}{refused}{0}
\chexdefskip{critical_mimic_constant-normal}{P_n_impression}{truncations}{0}
\chexdefskip{critical_mimic_copy-indication}{B_f_both}{empty}{53}
\chexdefskip{critical_mimic_copy-indication}{B_f_both}{empty_target}{22}
\chexdefskip{critical_mimic_copy-indication}{B_f_both}{generated}{53}
\chexdefskip{critical_mimic_copy-indication}{B_f_both}{length_truncated}{0}
\chexdefskip{critical_mimic_copy-indication}{B_f_both}{non_english}{0}
\chexdefskip{critical_mimic_copy-indication}{B_f_both}{refused}{0}
\chexdefskip{critical_mimic_copy-indication}{B_f_both}{truncations}{0}
\chexdefskip{critical_mimic_copy-indication}{B_f_findings}{empty}{53}
\chexdefskip{critical_mimic_copy-indication}{B_f_findings}{empty_target}{22}
\chexdefskip{critical_mimic_copy-indication}{B_f_findings}{generated}{53}
\chexdefskip{critical_mimic_copy-indication}{B_f_findings}{length_truncated}{0}
\chexdefskip{critical_mimic_copy-indication}{B_f_findings}{non_english}{0}
\chexdefskip{critical_mimic_copy-indication}{B_f_findings}{refused}{0}
\chexdefskip{critical_mimic_copy-indication}{B_f_findings}{truncations}{0}
\chexdefskip{critical_mimic_copy-indication}{B_f_impression}{empty}{75}
\chexdefskip{critical_mimic_copy-indication}{B_f_impression}{generated}{75}
\chexdefskip{critical_mimic_copy-indication}{B_f_impression}{length_truncated}{0}
\chexdefskip{critical_mimic_copy-indication}{B_f_impression}{non_english}{0}
\chexdefskip{critical_mimic_copy-indication}{B_f_impression}{refused}{0}
\chexdefskip{critical_mimic_copy-indication}{B_f_impression}{truncations}{0}
\chexdefskip{critical_mimic_copy-indication}{P_f_both}{empty}{6}
\chexdefskip{critical_mimic_copy-indication}{P_f_both}{empty_target}{22}
\chexdefskip{critical_mimic_copy-indication}{P_f_both}{generated}{53}
\chexdefskip{critical_mimic_copy-indication}{P_f_both}{length_truncated}{0}
\chexdefskip{critical_mimic_copy-indication}{P_f_both}{non_english}{0}
\chexdefskip{critical_mimic_copy-indication}{P_f_both}{refused}{0}
\chexdefskip{critical_mimic_copy-indication}{P_f_both}{truncations}{0}
\chexdefskip{critical_mimic_copy-indication}{P_f_findings}{empty}{6}
\chexdefskip{critical_mimic_copy-indication}{P_f_findings}{empty_target}{22}
\chexdefskip{critical_mimic_copy-indication}{P_f_findings}{generated}{53}
\chexdefskip{critical_mimic_copy-indication}{P_f_findings}{length_truncated}{0}
\chexdefskip{critical_mimic_copy-indication}{P_f_findings}{non_english}{0}
\chexdefskip{critical_mimic_copy-indication}{P_f_findings}{refused}{0}
\chexdefskip{critical_mimic_copy-indication}{P_f_findings}{truncations}{0}
\chexdefskip{critical_mimic_copy-indication}{P_f_impression}{empty}{6}
\chexdefskip{critical_mimic_copy-indication}{P_f_impression}{generated}{75}
\chexdefskip{critical_mimic_copy-indication}{P_f_impression}{length_truncated}{0}
\chexdefskip{critical_mimic_copy-indication}{P_f_impression}{non_english}{0}
\chexdefskip{critical_mimic_copy-indication}{P_f_impression}{refused}{0}
\chexdefskip{critical_mimic_copy-indication}{P_f_impression}{truncations}{0}
\chexdefskip{critical_mimic_copy-indication}{P_n_both}{empty}{6}
\chexdefskip{critical_mimic_copy-indication}{P_n_both}{empty_target}{22}
\chexdefskip{critical_mimic_copy-indication}{P_n_both}{generated}{53}
\chexdefskip{critical_mimic_copy-indication}{P_n_both}{length_truncated}{0}
\chexdefskip{critical_mimic_copy-indication}{P_n_both}{non_english}{0}
\chexdefskip{critical_mimic_copy-indication}{P_n_both}{refused}{0}
\chexdefskip{critical_mimic_copy-indication}{P_n_both}{truncations}{0}
\chexdefskip{critical_mimic_copy-indication}{P_n_findings}{empty}{6}
\chexdefskip{critical_mimic_copy-indication}{P_n_findings}{empty_target}{22}
\chexdefskip{critical_mimic_copy-indication}{P_n_findings}{generated}{53}
\chexdefskip{critical_mimic_copy-indication}{P_n_findings}{length_truncated}{0}
\chexdefskip{critical_mimic_copy-indication}{P_n_findings}{non_english}{0}
\chexdefskip{critical_mimic_copy-indication}{P_n_findings}{refused}{0}
\chexdefskip{critical_mimic_copy-indication}{P_n_findings}{truncations}{0}
\chexdefskip{critical_mimic_copy-indication}{P_n_impression}{empty}{6}
\chexdefskip{critical_mimic_copy-indication}{P_n_impression}{generated}{75}
\chexdefskip{critical_mimic_copy-indication}{P_n_impression}{length_truncated}{0}
\chexdefskip{critical_mimic_copy-indication}{P_n_impression}{non_english}{0}
\chexdefskip{critical_mimic_copy-indication}{P_n_impression}{refused}{0}
\chexdefskip{critical_mimic_copy-indication}{P_n_impression}{truncations}{0}
\chexdefskip{critical_mimic_copy-prior}{B_f_both}{empty}{53}
\chexdefskip{critical_mimic_copy-prior}{B_f_both}{empty_target}{22}
\chexdefskip{critical_mimic_copy-prior}{B_f_both}{generated}{53}
\chexdefskip{critical_mimic_copy-prior}{B_f_both}{length_truncated}{0}
\chexdefskip{critical_mimic_copy-prior}{B_f_both}{non_english}{0}
\chexdefskip{critical_mimic_copy-prior}{B_f_both}{refused}{0}
\chexdefskip{critical_mimic_copy-prior}{B_f_both}{truncations}{0}
\chexdefskip{critical_mimic_copy-prior}{B_f_findings}{empty}{53}
\chexdefskip{critical_mimic_copy-prior}{B_f_findings}{empty_target}{22}
\chexdefskip{critical_mimic_copy-prior}{B_f_findings}{generated}{53}
\chexdefskip{critical_mimic_copy-prior}{B_f_findings}{length_truncated}{0}
\chexdefskip{critical_mimic_copy-prior}{B_f_findings}{non_english}{0}
\chexdefskip{critical_mimic_copy-prior}{B_f_findings}{refused}{0}
\chexdefskip{critical_mimic_copy-prior}{B_f_findings}{truncations}{0}
\chexdefskip{critical_mimic_copy-prior}{B_f_impression}{empty}{75}
\chexdefskip{critical_mimic_copy-prior}{B_f_impression}{generated}{75}
\chexdefskip{critical_mimic_copy-prior}{B_f_impression}{length_truncated}{0}
\chexdefskip{critical_mimic_copy-prior}{B_f_impression}{non_english}{0}
\chexdefskip{critical_mimic_copy-prior}{B_f_impression}{refused}{0}
\chexdefskip{critical_mimic_copy-prior}{B_f_impression}{truncations}{0}
\chexdefskip{critical_mimic_copy-prior}{P_f_both}{empty}{53}
\chexdefskip{critical_mimic_copy-prior}{P_f_both}{empty_target}{22}
\chexdefskip{critical_mimic_copy-prior}{P_f_both}{generated}{53}
\chexdefskip{critical_mimic_copy-prior}{P_f_both}{length_truncated}{0}
\chexdefskip{critical_mimic_copy-prior}{P_f_both}{non_english}{0}
\chexdefskip{critical_mimic_copy-prior}{P_f_both}{refused}{0}
\chexdefskip{critical_mimic_copy-prior}{P_f_both}{truncations}{0}
\chexdefskip{critical_mimic_copy-prior}{P_f_findings}{empty}{53}
\chexdefskip{critical_mimic_copy-prior}{P_f_findings}{empty_target}{22}
\chexdefskip{critical_mimic_copy-prior}{P_f_findings}{generated}{53}
\chexdefskip{critical_mimic_copy-prior}{P_f_findings}{length_truncated}{0}
\chexdefskip{critical_mimic_copy-prior}{P_f_findings}{non_english}{0}
\chexdefskip{critical_mimic_copy-prior}{P_f_findings}{refused}{0}
\chexdefskip{critical_mimic_copy-prior}{P_f_findings}{truncations}{0}
\chexdefskip{critical_mimic_copy-prior}{P_f_impression}{empty}{75}
\chexdefskip{critical_mimic_copy-prior}{P_f_impression}{generated}{75}
\chexdefskip{critical_mimic_copy-prior}{P_f_impression}{length_truncated}{0}
\chexdefskip{critical_mimic_copy-prior}{P_f_impression}{non_english}{0}
\chexdefskip{critical_mimic_copy-prior}{P_f_impression}{refused}{0}
\chexdefskip{critical_mimic_copy-prior}{P_f_impression}{truncations}{0}
\chexdefskip{critical_mimic_copy-prior}{P_n_both}{empty}{53}
\chexdefskip{critical_mimic_copy-prior}{P_n_both}{empty_target}{22}
\chexdefskip{critical_mimic_copy-prior}{P_n_both}{generated}{53}
\chexdefskip{critical_mimic_copy-prior}{P_n_both}{length_truncated}{0}
\chexdefskip{critical_mimic_copy-prior}{P_n_both}{non_english}{0}
\chexdefskip{critical_mimic_copy-prior}{P_n_both}{refused}{0}
\chexdefskip{critical_mimic_copy-prior}{P_n_both}{truncations}{0}
\chexdefskip{critical_mimic_copy-prior}{P_n_findings}{empty}{53}
\chexdefskip{critical_mimic_copy-prior}{P_n_findings}{empty_target}{22}
\chexdefskip{critical_mimic_copy-prior}{P_n_findings}{generated}{53}
\chexdefskip{critical_mimic_copy-prior}{P_n_findings}{length_truncated}{0}
\chexdefskip{critical_mimic_copy-prior}{P_n_findings}{non_english}{0}
\chexdefskip{critical_mimic_copy-prior}{P_n_findings}{refused}{0}
\chexdefskip{critical_mimic_copy-prior}{P_n_findings}{truncations}{0}
\chexdefskip{critical_mimic_copy-prior}{P_n_impression}{empty}{75}
\chexdefskip{critical_mimic_copy-prior}{P_n_impression}{generated}{75}
\chexdefskip{critical_mimic_copy-prior}{P_n_impression}{length_truncated}{0}
\chexdefskip{critical_mimic_copy-prior}{P_n_impression}{non_english}{0}
\chexdefskip{critical_mimic_copy-prior}{P_n_impression}{refused}{0}
\chexdefskip{critical_mimic_copy-prior}{P_n_impression}{truncations}{0}
\chexdefskip{critical_mimic_cxr-llava-v2}{B_f_both}{empty}{0}
\chexdefskip{critical_mimic_cxr-llava-v2}{B_f_both}{empty_target}{22}
\chexdefskip{critical_mimic_cxr-llava-v2}{B_f_both}{generated}{53}
\chexdefskip{critical_mimic_cxr-llava-v2}{B_f_both}{length_truncated}{0}
\chexdefskip{critical_mimic_cxr-llava-v2}{B_f_both}{non_english}{0}
\chexdefskip{critical_mimic_cxr-llava-v2}{B_f_both}{refused}{0}
\chexdefskip{critical_mimic_cxr-llava-v2}{B_f_both}{truncations}{0}
\chexdefskip{critical_mimic_cxr-llava-v2}{B_f_findings}{empty}{0}
\chexdefskip{critical_mimic_cxr-llava-v2}{B_f_findings}{empty_target}{22}
\chexdefskip{critical_mimic_cxr-llava-v2}{B_f_findings}{generated}{53}
\chexdefskip{critical_mimic_cxr-llava-v2}{B_f_findings}{length_truncated}{0}
\chexdefskip{critical_mimic_cxr-llava-v2}{B_f_findings}{non_english}{0}
\chexdefskip{critical_mimic_cxr-llava-v2}{B_f_findings}{refused}{0}
\chexdefskip{critical_mimic_cxr-llava-v2}{B_f_findings}{truncations}{0}
\chexdefskip{critical_mimic_cxr-llava-v2}{B_f_impression}{empty}{0}
\chexdefskip{critical_mimic_cxr-llava-v2}{B_f_impression}{generated}{75}
\chexdefskip{critical_mimic_cxr-llava-v2}{B_f_impression}{length_truncated}{0}
\chexdefskip{critical_mimic_cxr-llava-v2}{B_f_impression}{non_english}{0}
\chexdefskip{critical_mimic_cxr-llava-v2}{B_f_impression}{refused}{0}
\chexdefskip{critical_mimic_cxr-llava-v2}{B_f_impression}{truncations}{0}
\chexdefskip{critical_mimic_cxr-llava-v2}{P_f_both}{empty}{0}
\chexdefskip{critical_mimic_cxr-llava-v2}{P_f_both}{empty_target}{22}
\chexdefskip{critical_mimic_cxr-llava-v2}{P_f_both}{generated}{53}
\chexdefskip{critical_mimic_cxr-llava-v2}{P_f_both}{length_truncated}{0}
\chexdefskip{critical_mimic_cxr-llava-v2}{P_f_both}{non_english}{0}
\chexdefskip{critical_mimic_cxr-llava-v2}{P_f_both}{refused}{0}
\chexdefskip{critical_mimic_cxr-llava-v2}{P_f_both}{truncations}{0}
\chexdefskip{critical_mimic_cxr-llava-v2}{P_f_findings}{empty}{0}
\chexdefskip{critical_mimic_cxr-llava-v2}{P_f_findings}{empty_target}{22}
\chexdefskip{critical_mimic_cxr-llava-v2}{P_f_findings}{generated}{53}
\chexdefskip{critical_mimic_cxr-llava-v2}{P_f_findings}{length_truncated}{0}
\chexdefskip{critical_mimic_cxr-llava-v2}{P_f_findings}{non_english}{0}
\chexdefskip{critical_mimic_cxr-llava-v2}{P_f_findings}{refused}{0}
\chexdefskip{critical_mimic_cxr-llava-v2}{P_f_findings}{truncations}{0}
\chexdefskip{critical_mimic_cxr-llava-v2}{P_f_impression}{empty}{0}
\chexdefskip{critical_mimic_cxr-llava-v2}{P_f_impression}{generated}{75}
\chexdefskip{critical_mimic_cxr-llava-v2}{P_f_impression}{length_truncated}{0}
\chexdefskip{critical_mimic_cxr-llava-v2}{P_f_impression}{non_english}{0}
\chexdefskip{critical_mimic_cxr-llava-v2}{P_f_impression}{refused}{0}
\chexdefskip{critical_mimic_cxr-llava-v2}{P_f_impression}{truncations}{0}
\chexdefskip{critical_mimic_cxr-llava-v2}{P_n_both}{empty}{0}
\chexdefskip{critical_mimic_cxr-llava-v2}{P_n_both}{generated}{0}
\chexdefskip{critical_mimic_cxr-llava-v2}{P_n_both}{length_truncated}{0}
\chexdefskip{critical_mimic_cxr-llava-v2}{P_n_both}{non_english}{0}
\chexdefskip{critical_mimic_cxr-llava-v2}{P_n_both}{refused}{0}
\chexdefskip{critical_mimic_cxr-llava-v2}{P_n_both}{text_only_unsupported}{75}
\chexdefskip{critical_mimic_cxr-llava-v2}{P_n_both}{truncations}{0}
\chexdefskip{critical_mimic_cxr-llava-v2}{P_n_findings}{empty}{0}
\chexdefskip{critical_mimic_cxr-llava-v2}{P_n_findings}{generated}{0}
\chexdefskip{critical_mimic_cxr-llava-v2}{P_n_findings}{length_truncated}{0}
\chexdefskip{critical_mimic_cxr-llava-v2}{P_n_findings}{non_english}{0}
\chexdefskip{critical_mimic_cxr-llava-v2}{P_n_findings}{refused}{0}
\chexdefskip{critical_mimic_cxr-llava-v2}{P_n_findings}{text_only_unsupported}{75}
\chexdefskip{critical_mimic_cxr-llava-v2}{P_n_findings}{truncations}{0}
\chexdefskip{critical_mimic_cxr-llava-v2}{P_n_impression}{empty}{0}
\chexdefskip{critical_mimic_cxr-llava-v2}{P_n_impression}{generated}{0}
\chexdefskip{critical_mimic_cxr-llava-v2}{P_n_impression}{length_truncated}{0}
\chexdefskip{critical_mimic_cxr-llava-v2}{P_n_impression}{non_english}{0}
\chexdefskip{critical_mimic_cxr-llava-v2}{P_n_impression}{refused}{0}
\chexdefskip{critical_mimic_cxr-llava-v2}{P_n_impression}{text_only_unsupported}{75}
\chexdefskip{critical_mimic_cxr-llava-v2}{P_n_impression}{truncations}{0}
\chexdefskip{critical_mimic_empty}{B_f_both}{empty}{53}
\chexdefskip{critical_mimic_empty}{B_f_both}{empty_target}{22}
\chexdefskip{critical_mimic_empty}{B_f_both}{generated}{53}
\chexdefskip{critical_mimic_empty}{B_f_both}{length_truncated}{0}
\chexdefskip{critical_mimic_empty}{B_f_both}{non_english}{0}
\chexdefskip{critical_mimic_empty}{B_f_both}{refused}{0}
\chexdefskip{critical_mimic_empty}{B_f_both}{truncations}{0}
\chexdefskip{critical_mimic_empty}{B_f_findings}{empty}{53}
\chexdefskip{critical_mimic_empty}{B_f_findings}{empty_target}{22}
\chexdefskip{critical_mimic_empty}{B_f_findings}{generated}{53}
\chexdefskip{critical_mimic_empty}{B_f_findings}{length_truncated}{0}
\chexdefskip{critical_mimic_empty}{B_f_findings}{non_english}{0}
\chexdefskip{critical_mimic_empty}{B_f_findings}{refused}{0}
\chexdefskip{critical_mimic_empty}{B_f_findings}{truncations}{0}
\chexdefskip{critical_mimic_empty}{B_f_impression}{empty}{75}
\chexdefskip{critical_mimic_empty}{B_f_impression}{generated}{75}
\chexdefskip{critical_mimic_empty}{B_f_impression}{length_truncated}{0}
\chexdefskip{critical_mimic_empty}{B_f_impression}{non_english}{0}
\chexdefskip{critical_mimic_empty}{B_f_impression}{refused}{0}
\chexdefskip{critical_mimic_empty}{B_f_impression}{truncations}{0}
\chexdefskip{critical_mimic_empty}{P_f_both}{empty}{53}
\chexdefskip{critical_mimic_empty}{P_f_both}{empty_target}{22}
\chexdefskip{critical_mimic_empty}{P_f_both}{generated}{53}
\chexdefskip{critical_mimic_empty}{P_f_both}{length_truncated}{0}
\chexdefskip{critical_mimic_empty}{P_f_both}{non_english}{0}
\chexdefskip{critical_mimic_empty}{P_f_both}{refused}{0}
\chexdefskip{critical_mimic_empty}{P_f_both}{truncations}{0}
\chexdefskip{critical_mimic_empty}{P_f_findings}{empty}{53}
\chexdefskip{critical_mimic_empty}{P_f_findings}{empty_target}{22}
\chexdefskip{critical_mimic_empty}{P_f_findings}{generated}{53}
\chexdefskip{critical_mimic_empty}{P_f_findings}{length_truncated}{0}
\chexdefskip{critical_mimic_empty}{P_f_findings}{non_english}{0}
\chexdefskip{critical_mimic_empty}{P_f_findings}{refused}{0}
\chexdefskip{critical_mimic_empty}{P_f_findings}{truncations}{0}
\chexdefskip{critical_mimic_empty}{P_f_impression}{empty}{75}
\chexdefskip{critical_mimic_empty}{P_f_impression}{generated}{75}
\chexdefskip{critical_mimic_empty}{P_f_impression}{length_truncated}{0}
\chexdefskip{critical_mimic_empty}{P_f_impression}{non_english}{0}
\chexdefskip{critical_mimic_empty}{P_f_impression}{refused}{0}
\chexdefskip{critical_mimic_empty}{P_f_impression}{truncations}{0}
\chexdefskip{critical_mimic_empty}{P_n_both}{empty}{53}
\chexdefskip{critical_mimic_empty}{P_n_both}{empty_target}{22}
\chexdefskip{critical_mimic_empty}{P_n_both}{generated}{53}
\chexdefskip{critical_mimic_empty}{P_n_both}{length_truncated}{0}
\chexdefskip{critical_mimic_empty}{P_n_both}{non_english}{0}
\chexdefskip{critical_mimic_empty}{P_n_both}{refused}{0}
\chexdefskip{critical_mimic_empty}{P_n_both}{truncations}{0}
\chexdefskip{critical_mimic_empty}{P_n_findings}{empty}{53}
\chexdefskip{critical_mimic_empty}{P_n_findings}{empty_target}{22}
\chexdefskip{critical_mimic_empty}{P_n_findings}{generated}{53}
\chexdefskip{critical_mimic_empty}{P_n_findings}{length_truncated}{0}
\chexdefskip{critical_mimic_empty}{P_n_findings}{non_english}{0}
\chexdefskip{critical_mimic_empty}{P_n_findings}{refused}{0}
\chexdefskip{critical_mimic_empty}{P_n_findings}{truncations}{0}
\chexdefskip{critical_mimic_empty}{P_n_impression}{empty}{75}
\chexdefskip{critical_mimic_empty}{P_n_impression}{generated}{75}
\chexdefskip{critical_mimic_empty}{P_n_impression}{length_truncated}{0}
\chexdefskip{critical_mimic_empty}{P_n_impression}{non_english}{0}
\chexdefskip{critical_mimic_empty}{P_n_impression}{refused}{0}
\chexdefskip{critical_mimic_empty}{P_n_impression}{truncations}{0}
\chexdefskip{critical_mimic_internvl3-8b}{B_f_both}{empty}{0}
\chexdefskip{critical_mimic_internvl3-8b}{B_f_both}{empty_target}{22}
\chexdefskip{critical_mimic_internvl3-8b}{B_f_both}{generated}{53}
\chexdefskip{critical_mimic_internvl3-8b}{B_f_both}{length_truncated}{0}
\chexdefskip{critical_mimic_internvl3-8b}{B_f_both}{non_english}{0}
\chexdefskip{critical_mimic_internvl3-8b}{B_f_both}{refused}{0}
\chexdefskip{critical_mimic_internvl3-8b}{B_f_both}{truncations}{0}
\chexdefskip{critical_mimic_internvl3-8b}{B_f_findings}{empty}{0}
\chexdefskip{critical_mimic_internvl3-8b}{B_f_findings}{empty_target}{22}
\chexdefskip{critical_mimic_internvl3-8b}{B_f_findings}{generated}{53}
\chexdefskip{critical_mimic_internvl3-8b}{B_f_findings}{length_truncated}{0}
\chexdefskip{critical_mimic_internvl3-8b}{B_f_findings}{non_english}{0}
\chexdefskip{critical_mimic_internvl3-8b}{B_f_findings}{refused}{0}
\chexdefskip{critical_mimic_internvl3-8b}{B_f_findings}{truncations}{0}
\chexdefskip{critical_mimic_internvl3-8b}{B_f_impression}{empty}{0}
\chexdefskip{critical_mimic_internvl3-8b}{B_f_impression}{generated}{75}
\chexdefskip{critical_mimic_internvl3-8b}{B_f_impression}{length_truncated}{0}
\chexdefskip{critical_mimic_internvl3-8b}{B_f_impression}{non_english}{0}
\chexdefskip{critical_mimic_internvl3-8b}{B_f_impression}{refused}{0}
\chexdefskip{critical_mimic_internvl3-8b}{B_f_impression}{truncations}{0}
\chexdefskip{critical_mimic_internvl3-8b}{P_f_both}{empty}{0}
\chexdefskip{critical_mimic_internvl3-8b}{P_f_both}{empty_target}{22}
\chexdefskip{critical_mimic_internvl3-8b}{P_f_both}{generated}{53}
\chexdefskip{critical_mimic_internvl3-8b}{P_f_both}{length_truncated}{0}
\chexdefskip{critical_mimic_internvl3-8b}{P_f_both}{non_english}{0}
\chexdefskip{critical_mimic_internvl3-8b}{P_f_both}{refused}{0}
\chexdefskip{critical_mimic_internvl3-8b}{P_f_both}{truncations}{0}
\chexdefskip{critical_mimic_internvl3-8b}{P_f_findings}{empty}{0}
\chexdefskip{critical_mimic_internvl3-8b}{P_f_findings}{empty_target}{22}
\chexdefskip{critical_mimic_internvl3-8b}{P_f_findings}{generated}{53}
\chexdefskip{critical_mimic_internvl3-8b}{P_f_findings}{length_truncated}{0}
\chexdefskip{critical_mimic_internvl3-8b}{P_f_findings}{non_english}{0}
\chexdefskip{critical_mimic_internvl3-8b}{P_f_findings}{refused}{0}
\chexdefskip{critical_mimic_internvl3-8b}{P_f_findings}{truncations}{0}
\chexdefskip{critical_mimic_internvl3-8b}{P_f_impression}{empty}{0}
\chexdefskip{critical_mimic_internvl3-8b}{P_f_impression}{generated}{75}
\chexdefskip{critical_mimic_internvl3-8b}{P_f_impression}{length_truncated}{0}
\chexdefskip{critical_mimic_internvl3-8b}{P_f_impression}{non_english}{0}
\chexdefskip{critical_mimic_internvl3-8b}{P_f_impression}{refused}{0}
\chexdefskip{critical_mimic_internvl3-8b}{P_f_impression}{truncations}{0}
\chexdefskip{critical_mimic_internvl3-8b}{P_n_both}{empty}{0}
\chexdefskip{critical_mimic_internvl3-8b}{P_n_both}{empty_target}{22}
\chexdefskip{critical_mimic_internvl3-8b}{P_n_both}{generated}{53}
\chexdefskip{critical_mimic_internvl3-8b}{P_n_both}{length_truncated}{0}
\chexdefskip{critical_mimic_internvl3-8b}{P_n_both}{non_english}{0}
\chexdefskip{critical_mimic_internvl3-8b}{P_n_both}{refused}{0}
\chexdefskip{critical_mimic_internvl3-8b}{P_n_both}{truncations}{0}
\chexdefskip{critical_mimic_internvl3-8b}{P_n_findings}{empty}{0}
\chexdefskip{critical_mimic_internvl3-8b}{P_n_findings}{empty_target}{22}
\chexdefskip{critical_mimic_internvl3-8b}{P_n_findings}{generated}{53}
\chexdefskip{critical_mimic_internvl3-8b}{P_n_findings}{length_truncated}{0}
\chexdefskip{critical_mimic_internvl3-8b}{P_n_findings}{non_english}{0}
\chexdefskip{critical_mimic_internvl3-8b}{P_n_findings}{refused}{0}
\chexdefskip{critical_mimic_internvl3-8b}{P_n_findings}{truncations}{0}
\chexdefskip{critical_mimic_internvl3-8b}{P_n_impression}{empty}{0}
\chexdefskip{critical_mimic_internvl3-8b}{P_n_impression}{generated}{75}
\chexdefskip{critical_mimic_internvl3-8b}{P_n_impression}{length_truncated}{0}
\chexdefskip{critical_mimic_internvl3-8b}{P_n_impression}{non_english}{0}
\chexdefskip{critical_mimic_internvl3-8b}{P_n_impression}{refused}{0}
\chexdefskip{critical_mimic_internvl3-8b}{P_n_impression}{truncations}{0}
\chexdefskip{critical_mimic_llava-med}{B_f_both}{empty}{0}
\chexdefskip{critical_mimic_llava-med}{B_f_both}{empty_target}{22}
\chexdefskip{critical_mimic_llava-med}{B_f_both}{generated}{53}
\chexdefskip{critical_mimic_llava-med}{B_f_both}{length_truncated}{0}
\chexdefskip{critical_mimic_llava-med}{B_f_both}{non_english}{0}
\chexdefskip{critical_mimic_llava-med}{B_f_both}{refused}{0}
\chexdefskip{critical_mimic_llava-med}{B_f_both}{truncations}{0}
\chexdefskip{critical_mimic_llava-med}{B_f_findings}{empty}{0}
\chexdefskip{critical_mimic_llava-med}{B_f_findings}{empty_target}{22}
\chexdefskip{critical_mimic_llava-med}{B_f_findings}{generated}{53}
\chexdefskip{critical_mimic_llava-med}{B_f_findings}{length_truncated}{0}
\chexdefskip{critical_mimic_llava-med}{B_f_findings}{non_english}{0}
\chexdefskip{critical_mimic_llava-med}{B_f_findings}{refused}{0}
\chexdefskip{critical_mimic_llava-med}{B_f_findings}{truncations}{0}
\chexdefskip{critical_mimic_llava-med}{B_f_impression}{empty}{0}
\chexdefskip{critical_mimic_llava-med}{B_f_impression}{generated}{75}
\chexdefskip{critical_mimic_llava-med}{B_f_impression}{length_truncated}{0}
\chexdefskip{critical_mimic_llava-med}{B_f_impression}{non_english}{0}
\chexdefskip{critical_mimic_llava-med}{B_f_impression}{refused}{0}
\chexdefskip{critical_mimic_llava-med}{B_f_impression}{truncations}{0}
\chexdefskip{critical_mimic_llava-med}{P_f_both}{empty}{0}
\chexdefskip{critical_mimic_llava-med}{P_f_both}{empty_target}{22}
\chexdefskip{critical_mimic_llava-med}{P_f_both}{generated}{53}
\chexdefskip{critical_mimic_llava-med}{P_f_both}{length_truncated}{0}
\chexdefskip{critical_mimic_llava-med}{P_f_both}{non_english}{0}
\chexdefskip{critical_mimic_llava-med}{P_f_both}{refused}{0}
\chexdefskip{critical_mimic_llava-med}{P_f_both}{truncations}{0}
\chexdefskip{critical_mimic_llava-med}{P_f_findings}{empty}{2}
\chexdefskip{critical_mimic_llava-med}{P_f_findings}{empty_target}{22}
\chexdefskip{critical_mimic_llava-med}{P_f_findings}{generated}{53}
\chexdefskip{critical_mimic_llava-med}{P_f_findings}{length_truncated}{0}
\chexdefskip{critical_mimic_llava-med}{P_f_findings}{non_english}{0}
\chexdefskip{critical_mimic_llava-med}{P_f_findings}{refused}{0}
\chexdefskip{critical_mimic_llava-med}{P_f_findings}{truncations}{0}
\chexdefskip{critical_mimic_llava-med}{P_f_impression}{empty}{8}
\chexdefskip{critical_mimic_llava-med}{P_f_impression}{generated}{75}
\chexdefskip{critical_mimic_llava-med}{P_f_impression}{length_truncated}{0}
\chexdefskip{critical_mimic_llava-med}{P_f_impression}{non_english}{0}
\chexdefskip{critical_mimic_llava-med}{P_f_impression}{refused}{0}
\chexdefskip{critical_mimic_llava-med}{P_f_impression}{truncations}{0}
\chexdefskip{critical_mimic_llava-med}{P_n_both}{empty}{0}
\chexdefskip{critical_mimic_llava-med}{P_n_both}{generated}{0}
\chexdefskip{critical_mimic_llava-med}{P_n_both}{length_truncated}{0}
\chexdefskip{critical_mimic_llava-med}{P_n_both}{non_english}{0}
\chexdefskip{critical_mimic_llava-med}{P_n_both}{refused}{0}
\chexdefskip{critical_mimic_llava-med}{P_n_both}{text_only_unsupported}{75}
\chexdefskip{critical_mimic_llava-med}{P_n_both}{truncations}{0}
\chexdefskip{critical_mimic_llava-med}{P_n_findings}{empty}{0}
\chexdefskip{critical_mimic_llava-med}{P_n_findings}{generated}{0}
\chexdefskip{critical_mimic_llava-med}{P_n_findings}{length_truncated}{0}
\chexdefskip{critical_mimic_llava-med}{P_n_findings}{non_english}{0}
\chexdefskip{critical_mimic_llava-med}{P_n_findings}{refused}{0}
\chexdefskip{critical_mimic_llava-med}{P_n_findings}{text_only_unsupported}{75}
\chexdefskip{critical_mimic_llava-med}{P_n_findings}{truncations}{0}
\chexdefskip{critical_mimic_llava-med}{P_n_impression}{empty}{0}
\chexdefskip{critical_mimic_llava-med}{P_n_impression}{generated}{0}
\chexdefskip{critical_mimic_llava-med}{P_n_impression}{length_truncated}{0}
\chexdefskip{critical_mimic_llava-med}{P_n_impression}{non_english}{0}
\chexdefskip{critical_mimic_llava-med}{P_n_impression}{refused}{0}
\chexdefskip{critical_mimic_llava-med}{P_n_impression}{text_only_unsupported}{75}
\chexdefskip{critical_mimic_llava-med}{P_n_impression}{truncations}{0}
\chexdefskip{critical_mimic_llava-rad}{B_f_both}{empty}{0}
\chexdefskip{critical_mimic_llava-rad}{B_f_both}{empty_target}{22}
\chexdefskip{critical_mimic_llava-rad}{B_f_both}{generated}{53}
\chexdefskip{critical_mimic_llava-rad}{B_f_both}{length_truncated}{0}
\chexdefskip{critical_mimic_llava-rad}{B_f_both}{non_english}{0}
\chexdefskip{critical_mimic_llava-rad}{B_f_both}{refused}{0}
\chexdefskip{critical_mimic_llava-rad}{B_f_both}{truncations}{0}
\chexdefskip{critical_mimic_llava-rad}{B_f_findings}{empty}{0}
\chexdefskip{critical_mimic_llava-rad}{B_f_findings}{empty_target}{22}
\chexdefskip{critical_mimic_llava-rad}{B_f_findings}{generated}{53}
\chexdefskip{critical_mimic_llava-rad}{B_f_findings}{length_truncated}{0}
\chexdefskip{critical_mimic_llava-rad}{B_f_findings}{non_english}{0}
\chexdefskip{critical_mimic_llava-rad}{B_f_findings}{refused}{0}
\chexdefskip{critical_mimic_llava-rad}{B_f_findings}{truncations}{0}
\chexdefskip{critical_mimic_llava-rad}{B_f_impression}{empty}{0}
\chexdefskip{critical_mimic_llava-rad}{B_f_impression}{generated}{75}
\chexdefskip{critical_mimic_llava-rad}{B_f_impression}{length_truncated}{0}
\chexdefskip{critical_mimic_llava-rad}{B_f_impression}{non_english}{0}
\chexdefskip{critical_mimic_llava-rad}{B_f_impression}{refused}{0}
\chexdefskip{critical_mimic_llava-rad}{B_f_impression}{truncations}{0}
\chexdefskip{critical_mimic_llava-rad}{P_f_both}{empty}{0}
\chexdefskip{critical_mimic_llava-rad}{P_f_both}{empty_target}{22}
\chexdefskip{critical_mimic_llava-rad}{P_f_both}{generated}{53}
\chexdefskip{critical_mimic_llava-rad}{P_f_both}{length_truncated}{0}
\chexdefskip{critical_mimic_llava-rad}{P_f_both}{non_english}{0}
\chexdefskip{critical_mimic_llava-rad}{P_f_both}{refused}{0}
\chexdefskip{critical_mimic_llava-rad}{P_f_both}{truncations}{0}
\chexdefskip{critical_mimic_llava-rad}{P_f_findings}{empty}{0}
\chexdefskip{critical_mimic_llava-rad}{P_f_findings}{empty_target}{22}
\chexdefskip{critical_mimic_llava-rad}{P_f_findings}{generated}{53}
\chexdefskip{critical_mimic_llava-rad}{P_f_findings}{length_truncated}{0}
\chexdefskip{critical_mimic_llava-rad}{P_f_findings}{non_english}{0}
\chexdefskip{critical_mimic_llava-rad}{P_f_findings}{refused}{0}
\chexdefskip{critical_mimic_llava-rad}{P_f_findings}{truncations}{0}
\chexdefskip{critical_mimic_llava-rad}{P_f_impression}{empty}{0}
\chexdefskip{critical_mimic_llava-rad}{P_f_impression}{generated}{75}
\chexdefskip{critical_mimic_llava-rad}{P_f_impression}{length_truncated}{0}
\chexdefskip{critical_mimic_llava-rad}{P_f_impression}{non_english}{0}
\chexdefskip{critical_mimic_llava-rad}{P_f_impression}{refused}{0}
\chexdefskip{critical_mimic_llava-rad}{P_f_impression}{truncations}{0}
\chexdefskip{critical_mimic_llava-rad}{P_n_both}{empty}{0}
\chexdefskip{critical_mimic_llava-rad}{P_n_both}{generated}{0}
\chexdefskip{critical_mimic_llava-rad}{P_n_both}{length_truncated}{0}
\chexdefskip{critical_mimic_llava-rad}{P_n_both}{non_english}{0}
\chexdefskip{critical_mimic_llava-rad}{P_n_both}{refused}{0}
\chexdefskip{critical_mimic_llava-rad}{P_n_both}{text_only_unsupported}{75}
\chexdefskip{critical_mimic_llava-rad}{P_n_both}{truncations}{0}
\chexdefskip{critical_mimic_llava-rad}{P_n_findings}{empty}{0}
\chexdefskip{critical_mimic_llava-rad}{P_n_findings}{generated}{0}
\chexdefskip{critical_mimic_llava-rad}{P_n_findings}{length_truncated}{0}
\chexdefskip{critical_mimic_llava-rad}{P_n_findings}{non_english}{0}
\chexdefskip{critical_mimic_llava-rad}{P_n_findings}{refused}{0}
\chexdefskip{critical_mimic_llava-rad}{P_n_findings}{text_only_unsupported}{75}
\chexdefskip{critical_mimic_llava-rad}{P_n_findings}{truncations}{0}
\chexdefskip{critical_mimic_llava-rad}{P_n_impression}{empty}{0}
\chexdefskip{critical_mimic_llava-rad}{P_n_impression}{generated}{0}
\chexdefskip{critical_mimic_llava-rad}{P_n_impression}{length_truncated}{0}
\chexdefskip{critical_mimic_llava-rad}{P_n_impression}{non_english}{0}
\chexdefskip{critical_mimic_llava-rad}{P_n_impression}{refused}{0}
\chexdefskip{critical_mimic_llava-rad}{P_n_impression}{text_only_unsupported}{75}
\chexdefskip{critical_mimic_llava-rad}{P_n_impression}{truncations}{0}
\chexdefskip{critical_mimic_maira-2}{B_f_both}{empty}{0}
\chexdefskip{critical_mimic_maira-2}{B_f_both}{empty_target}{22}
\chexdefskip{critical_mimic_maira-2}{B_f_both}{generated}{53}
\chexdefskip{critical_mimic_maira-2}{B_f_both}{length_truncated}{0}
\chexdefskip{critical_mimic_maira-2}{B_f_both}{non_english}{0}
\chexdefskip{critical_mimic_maira-2}{B_f_both}{refused}{0}
\chexdefskip{critical_mimic_maira-2}{B_f_both}{truncations}{0}
\chexdefskip{critical_mimic_maira-2}{B_f_findings}{empty}{0}
\chexdefskip{critical_mimic_maira-2}{B_f_findings}{empty_target}{22}
\chexdefskip{critical_mimic_maira-2}{B_f_findings}{generated}{53}
\chexdefskip{critical_mimic_maira-2}{B_f_findings}{length_truncated}{0}
\chexdefskip{critical_mimic_maira-2}{B_f_findings}{non_english}{0}
\chexdefskip{critical_mimic_maira-2}{B_f_findings}{refused}{0}
\chexdefskip{critical_mimic_maira-2}{B_f_findings}{truncations}{0}
\chexdefskip{critical_mimic_maira-2}{B_f_impression}{empty}{0}
\chexdefskip{critical_mimic_maira-2}{B_f_impression}{generated}{75}
\chexdefskip{critical_mimic_maira-2}{B_f_impression}{length_truncated}{0}
\chexdefskip{critical_mimic_maira-2}{B_f_impression}{non_english}{0}
\chexdefskip{critical_mimic_maira-2}{B_f_impression}{refused}{0}
\chexdefskip{critical_mimic_maira-2}{B_f_impression}{truncations}{0}
\chexdefskip{critical_mimic_maira-2}{P_f_both}{empty}{0}
\chexdefskip{critical_mimic_maira-2}{P_f_both}{empty_target}{22}
\chexdefskip{critical_mimic_maira-2}{P_f_both}{generated}{53}
\chexdefskip{critical_mimic_maira-2}{P_f_both}{length_truncated}{0}
\chexdefskip{critical_mimic_maira-2}{P_f_both}{non_english}{0}
\chexdefskip{critical_mimic_maira-2}{P_f_both}{refused}{0}
\chexdefskip{critical_mimic_maira-2}{P_f_both}{truncations}{0}
\chexdefskip{critical_mimic_maira-2}{P_f_findings}{empty}{0}
\chexdefskip{critical_mimic_maira-2}{P_f_findings}{empty_target}{22}
\chexdefskip{critical_mimic_maira-2}{P_f_findings}{generated}{53}
\chexdefskip{critical_mimic_maira-2}{P_f_findings}{length_truncated}{0}
\chexdefskip{critical_mimic_maira-2}{P_f_findings}{non_english}{0}
\chexdefskip{critical_mimic_maira-2}{P_f_findings}{refused}{0}
\chexdefskip{critical_mimic_maira-2}{P_f_findings}{truncations}{0}
\chexdefskip{critical_mimic_maira-2}{P_f_impression}{empty}{0}
\chexdefskip{critical_mimic_maira-2}{P_f_impression}{generated}{75}
\chexdefskip{critical_mimic_maira-2}{P_f_impression}{length_truncated}{0}
\chexdefskip{critical_mimic_maira-2}{P_f_impression}{non_english}{0}
\chexdefskip{critical_mimic_maira-2}{P_f_impression}{refused}{0}
\chexdefskip{critical_mimic_maira-2}{P_f_impression}{truncations}{0}
\chexdefskip{critical_mimic_maira-2}{P_n_both}{empty}{0}
\chexdefskip{critical_mimic_maira-2}{P_n_both}{generated}{0}
\chexdefskip{critical_mimic_maira-2}{P_n_both}{length_truncated}{0}
\chexdefskip{critical_mimic_maira-2}{P_n_both}{non_english}{0}
\chexdefskip{critical_mimic_maira-2}{P_n_both}{refused}{0}
\chexdefskip{critical_mimic_maira-2}{P_n_both}{text_only_unsupported}{75}
\chexdefskip{critical_mimic_maira-2}{P_n_both}{truncations}{0}
\chexdefskip{critical_mimic_maira-2}{P_n_findings}{empty}{0}
\chexdefskip{critical_mimic_maira-2}{P_n_findings}{generated}{0}
\chexdefskip{critical_mimic_maira-2}{P_n_findings}{length_truncated}{0}
\chexdefskip{critical_mimic_maira-2}{P_n_findings}{non_english}{0}
\chexdefskip{critical_mimic_maira-2}{P_n_findings}{refused}{0}
\chexdefskip{critical_mimic_maira-2}{P_n_findings}{text_only_unsupported}{75}
\chexdefskip{critical_mimic_maira-2}{P_n_findings}{truncations}{0}
\chexdefskip{critical_mimic_maira-2}{P_n_impression}{empty}{0}
\chexdefskip{critical_mimic_maira-2}{P_n_impression}{generated}{0}
\chexdefskip{critical_mimic_maira-2}{P_n_impression}{length_truncated}{0}
\chexdefskip{critical_mimic_maira-2}{P_n_impression}{non_english}{0}
\chexdefskip{critical_mimic_maira-2}{P_n_impression}{refused}{0}
\chexdefskip{critical_mimic_maira-2}{P_n_impression}{text_only_unsupported}{75}
\chexdefskip{critical_mimic_maira-2}{P_n_impression}{truncations}{0}
\chexdefskip{critical_mimic_medgemma-4b}{B_f_both}{empty}{0}
\chexdefskip{critical_mimic_medgemma-4b}{B_f_both}{empty_target}{22}
\chexdefskip{critical_mimic_medgemma-4b}{B_f_both}{generated}{53}
\chexdefskip{critical_mimic_medgemma-4b}{B_f_both}{length_truncated}{0}
\chexdefskip{critical_mimic_medgemma-4b}{B_f_both}{non_english}{0}
\chexdefskip{critical_mimic_medgemma-4b}{B_f_both}{refused}{0}
\chexdefskip{critical_mimic_medgemma-4b}{B_f_both}{truncations}{0}
\chexdefskip{critical_mimic_medgemma-4b}{B_f_findings}{empty}{0}
\chexdefskip{critical_mimic_medgemma-4b}{B_f_findings}{empty_target}{22}
\chexdefskip{critical_mimic_medgemma-4b}{B_f_findings}{generated}{53}
\chexdefskip{critical_mimic_medgemma-4b}{B_f_findings}{length_truncated}{0}
\chexdefskip{critical_mimic_medgemma-4b}{B_f_findings}{non_english}{0}
\chexdefskip{critical_mimic_medgemma-4b}{B_f_findings}{refused}{0}
\chexdefskip{critical_mimic_medgemma-4b}{B_f_findings}{truncations}{0}
\chexdefskip{critical_mimic_medgemma-4b}{B_f_impression}{empty}{0}
\chexdefskip{critical_mimic_medgemma-4b}{B_f_impression}{generated}{75}
\chexdefskip{critical_mimic_medgemma-4b}{B_f_impression}{length_truncated}{0}
\chexdefskip{critical_mimic_medgemma-4b}{B_f_impression}{non_english}{0}
\chexdefskip{critical_mimic_medgemma-4b}{B_f_impression}{refused}{0}
\chexdefskip{critical_mimic_medgemma-4b}{B_f_impression}{truncations}{0}
\chexdefskip{critical_mimic_medgemma-4b}{P_f_both}{empty}{0}
\chexdefskip{critical_mimic_medgemma-4b}{P_f_both}{empty_target}{22}
\chexdefskip{critical_mimic_medgemma-4b}{P_f_both}{generated}{53}
\chexdefskip{critical_mimic_medgemma-4b}{P_f_both}{length_truncated}{0}
\chexdefskip{critical_mimic_medgemma-4b}{P_f_both}{non_english}{0}
\chexdefskip{critical_mimic_medgemma-4b}{P_f_both}{refused}{0}
\chexdefskip{critical_mimic_medgemma-4b}{P_f_both}{truncations}{0}
\chexdefskip{critical_mimic_medgemma-4b}{P_f_findings}{empty}{0}
\chexdefskip{critical_mimic_medgemma-4b}{P_f_findings}{empty_target}{22}
\chexdefskip{critical_mimic_medgemma-4b}{P_f_findings}{generated}{53}
\chexdefskip{critical_mimic_medgemma-4b}{P_f_findings}{length_truncated}{0}
\chexdefskip{critical_mimic_medgemma-4b}{P_f_findings}{non_english}{0}
\chexdefskip{critical_mimic_medgemma-4b}{P_f_findings}{refused}{0}
\chexdefskip{critical_mimic_medgemma-4b}{P_f_findings}{truncations}{0}
\chexdefskip{critical_mimic_medgemma-4b}{P_f_impression}{empty}{0}
\chexdefskip{critical_mimic_medgemma-4b}{P_f_impression}{generated}{75}
\chexdefskip{critical_mimic_medgemma-4b}{P_f_impression}{length_truncated}{0}
\chexdefskip{critical_mimic_medgemma-4b}{P_f_impression}{non_english}{0}
\chexdefskip{critical_mimic_medgemma-4b}{P_f_impression}{refused}{0}
\chexdefskip{critical_mimic_medgemma-4b}{P_f_impression}{truncations}{0}
\chexdefskip{critical_mimic_medgemma-4b}{P_n_both}{empty}{0}
\chexdefskip{critical_mimic_medgemma-4b}{P_n_both}{empty_target}{22}
\chexdefskip{critical_mimic_medgemma-4b}{P_n_both}{generated}{53}
\chexdefskip{critical_mimic_medgemma-4b}{P_n_both}{length_truncated}{0}
\chexdefskip{critical_mimic_medgemma-4b}{P_n_both}{non_english}{0}
\chexdefskip{critical_mimic_medgemma-4b}{P_n_both}{refused}{0}
\chexdefskip{critical_mimic_medgemma-4b}{P_n_both}{truncations}{0}
\chexdefskip{critical_mimic_medgemma-4b}{P_n_findings}{empty}{0}
\chexdefskip{critical_mimic_medgemma-4b}{P_n_findings}{empty_target}{22}
\chexdefskip{critical_mimic_medgemma-4b}{P_n_findings}{generated}{53}
\chexdefskip{critical_mimic_medgemma-4b}{P_n_findings}{length_truncated}{0}
\chexdefskip{critical_mimic_medgemma-4b}{P_n_findings}{non_english}{0}
\chexdefskip{critical_mimic_medgemma-4b}{P_n_findings}{refused}{0}
\chexdefskip{critical_mimic_medgemma-4b}{P_n_findings}{truncations}{0}
\chexdefskip{critical_mimic_medgemma-4b}{P_n_impression}{empty}{0}
\chexdefskip{critical_mimic_medgemma-4b}{P_n_impression}{generated}{75}
\chexdefskip{critical_mimic_medgemma-4b}{P_n_impression}{length_truncated}{0}
\chexdefskip{critical_mimic_medgemma-4b}{P_n_impression}{non_english}{0}
\chexdefskip{critical_mimic_medgemma-4b}{P_n_impression}{refused}{0}
\chexdefskip{critical_mimic_medgemma-4b}{P_n_impression}{truncations}{0}
\chexdefskip{critical_mimic_most-common-report}{B_f_both}{empty}{0}
\chexdefskip{critical_mimic_most-common-report}{B_f_both}{empty_target}{22}
\chexdefskip{critical_mimic_most-common-report}{B_f_both}{generated}{53}
\chexdefskip{critical_mimic_most-common-report}{B_f_both}{length_truncated}{0}
\chexdefskip{critical_mimic_most-common-report}{B_f_both}{non_english}{0}
\chexdefskip{critical_mimic_most-common-report}{B_f_both}{refused}{0}
\chexdefskip{critical_mimic_most-common-report}{B_f_both}{truncations}{0}
\chexdefskip{critical_mimic_most-common-report}{B_f_findings}{empty}{0}
\chexdefskip{critical_mimic_most-common-report}{B_f_findings}{empty_target}{22}
\chexdefskip{critical_mimic_most-common-report}{B_f_findings}{generated}{53}
\chexdefskip{critical_mimic_most-common-report}{B_f_findings}{length_truncated}{0}
\chexdefskip{critical_mimic_most-common-report}{B_f_findings}{non_english}{0}
\chexdefskip{critical_mimic_most-common-report}{B_f_findings}{refused}{0}
\chexdefskip{critical_mimic_most-common-report}{B_f_findings}{truncations}{0}
\chexdefskip{critical_mimic_most-common-report}{B_f_impression}{empty}{0}
\chexdefskip{critical_mimic_most-common-report}{B_f_impression}{generated}{75}
\chexdefskip{critical_mimic_most-common-report}{B_f_impression}{length_truncated}{0}
\chexdefskip{critical_mimic_most-common-report}{B_f_impression}{non_english}{0}
\chexdefskip{critical_mimic_most-common-report}{B_f_impression}{refused}{0}
\chexdefskip{critical_mimic_most-common-report}{B_f_impression}{truncations}{0}
\chexdefskip{critical_mimic_most-common-report}{P_f_both}{empty}{0}
\chexdefskip{critical_mimic_most-common-report}{P_f_both}{empty_target}{22}
\chexdefskip{critical_mimic_most-common-report}{P_f_both}{generated}{53}
\chexdefskip{critical_mimic_most-common-report}{P_f_both}{length_truncated}{0}
\chexdefskip{critical_mimic_most-common-report}{P_f_both}{non_english}{0}
\chexdefskip{critical_mimic_most-common-report}{P_f_both}{refused}{0}
\chexdefskip{critical_mimic_most-common-report}{P_f_both}{truncations}{0}
\chexdefskip{critical_mimic_most-common-report}{P_f_findings}{empty}{0}
\chexdefskip{critical_mimic_most-common-report}{P_f_findings}{empty_target}{22}
\chexdefskip{critical_mimic_most-common-report}{P_f_findings}{generated}{53}
\chexdefskip{critical_mimic_most-common-report}{P_f_findings}{length_truncated}{0}
\chexdefskip{critical_mimic_most-common-report}{P_f_findings}{non_english}{0}
\chexdefskip{critical_mimic_most-common-report}{P_f_findings}{refused}{0}
\chexdefskip{critical_mimic_most-common-report}{P_f_findings}{truncations}{0}
\chexdefskip{critical_mimic_most-common-report}{P_f_impression}{empty}{0}
\chexdefskip{critical_mimic_most-common-report}{P_f_impression}{generated}{75}
\chexdefskip{critical_mimic_most-common-report}{P_f_impression}{length_truncated}{0}
\chexdefskip{critical_mimic_most-common-report}{P_f_impression}{non_english}{0}
\chexdefskip{critical_mimic_most-common-report}{P_f_impression}{refused}{0}
\chexdefskip{critical_mimic_most-common-report}{P_f_impression}{truncations}{0}
\chexdefskip{critical_mimic_most-common-report}{P_n_both}{empty}{0}
\chexdefskip{critical_mimic_most-common-report}{P_n_both}{empty_target}{22}
\chexdefskip{critical_mimic_most-common-report}{P_n_both}{generated}{53}
\chexdefskip{critical_mimic_most-common-report}{P_n_both}{length_truncated}{0}
\chexdefskip{critical_mimic_most-common-report}{P_n_both}{non_english}{0}
\chexdefskip{critical_mimic_most-common-report}{P_n_both}{refused}{0}
\chexdefskip{critical_mimic_most-common-report}{P_n_both}{truncations}{0}
\chexdefskip{critical_mimic_most-common-report}{P_n_findings}{empty}{0}
\chexdefskip{critical_mimic_most-common-report}{P_n_findings}{empty_target}{22}
\chexdefskip{critical_mimic_most-common-report}{P_n_findings}{generated}{53}
\chexdefskip{critical_mimic_most-common-report}{P_n_findings}{length_truncated}{0}
\chexdefskip{critical_mimic_most-common-report}{P_n_findings}{non_english}{0}
\chexdefskip{critical_mimic_most-common-report}{P_n_findings}{refused}{0}
\chexdefskip{critical_mimic_most-common-report}{P_n_findings}{truncations}{0}
\chexdefskip{critical_mimic_most-common-report}{P_n_impression}{empty}{0}
\chexdefskip{critical_mimic_most-common-report}{P_n_impression}{generated}{75}
\chexdefskip{critical_mimic_most-common-report}{P_n_impression}{length_truncated}{0}
\chexdefskip{critical_mimic_most-common-report}{P_n_impression}{non_english}{0}
\chexdefskip{critical_mimic_most-common-report}{P_n_impression}{refused}{0}
\chexdefskip{critical_mimic_most-common-report}{P_n_impression}{truncations}{0}
\chexdefskip{critical_mimic_qwen2.5-vl-3b}{B_f_both}{empty}{0}
\chexdefskip{critical_mimic_qwen2.5-vl-3b}{B_f_both}{empty_target}{22}
\chexdefskip{critical_mimic_qwen2.5-vl-3b}{B_f_both}{generated}{53}
\chexdefskip{critical_mimic_qwen2.5-vl-3b}{B_f_both}{length_truncated}{0}
\chexdefskip{critical_mimic_qwen2.5-vl-3b}{B_f_both}{non_english}{0}
\chexdefskip{critical_mimic_qwen2.5-vl-3b}{B_f_both}{refused}{0}
\chexdefskip{critical_mimic_qwen2.5-vl-3b}{B_f_both}{truncations}{0}
\chexdefskip{critical_mimic_qwen2.5-vl-3b}{B_f_findings}{empty}{0}
\chexdefskip{critical_mimic_qwen2.5-vl-3b}{B_f_findings}{empty_target}{22}
\chexdefskip{critical_mimic_qwen2.5-vl-3b}{B_f_findings}{generated}{53}
\chexdefskip{critical_mimic_qwen2.5-vl-3b}{B_f_findings}{length_truncated}{1}
\chexdefskip{critical_mimic_qwen2.5-vl-3b}{B_f_findings}{non_english}{0}
\chexdefskip{critical_mimic_qwen2.5-vl-3b}{B_f_findings}{refused}{0}
\chexdefskip{critical_mimic_qwen2.5-vl-3b}{B_f_findings}{truncations}{0}
\chexdefskip{critical_mimic_qwen2.5-vl-3b}{B_f_impression}{empty}{0}
\chexdefskip{critical_mimic_qwen2.5-vl-3b}{B_f_impression}{generated}{75}
\chexdefskip{critical_mimic_qwen2.5-vl-3b}{B_f_impression}{length_truncated}{0}
\chexdefskip{critical_mimic_qwen2.5-vl-3b}{B_f_impression}{non_english}{0}
\chexdefskip{critical_mimic_qwen2.5-vl-3b}{B_f_impression}{refused}{0}
\chexdefskip{critical_mimic_qwen2.5-vl-3b}{B_f_impression}{truncations}{0}
\chexdefskip{critical_mimic_qwen2.5-vl-3b}{P_f_both}{empty}{0}
\chexdefskip{critical_mimic_qwen2.5-vl-3b}{P_f_both}{empty_target}{22}
\chexdefskip{critical_mimic_qwen2.5-vl-3b}{P_f_both}{generated}{53}
\chexdefskip{critical_mimic_qwen2.5-vl-3b}{P_f_both}{length_truncated}{7}
\chexdefskip{critical_mimic_qwen2.5-vl-3b}{P_f_both}{non_english}{0}
\chexdefskip{critical_mimic_qwen2.5-vl-3b}{P_f_both}{refused}{0}
\chexdefskip{critical_mimic_qwen2.5-vl-3b}{P_f_both}{truncations}{0}
\chexdefskip{critical_mimic_qwen2.5-vl-3b}{P_f_findings}{empty}{0}
\chexdefskip{critical_mimic_qwen2.5-vl-3b}{P_f_findings}{empty_target}{22}
\chexdefskip{critical_mimic_qwen2.5-vl-3b}{P_f_findings}{generated}{53}
\chexdefskip{critical_mimic_qwen2.5-vl-3b}{P_f_findings}{length_truncated}{0}
\chexdefskip{critical_mimic_qwen2.5-vl-3b}{P_f_findings}{non_english}{0}
\chexdefskip{critical_mimic_qwen2.5-vl-3b}{P_f_findings}{refused}{0}
\chexdefskip{critical_mimic_qwen2.5-vl-3b}{P_f_findings}{truncations}{0}
\chexdefskip{critical_mimic_qwen2.5-vl-3b}{P_f_impression}{empty}{0}
\chexdefskip{critical_mimic_qwen2.5-vl-3b}{P_f_impression}{generated}{75}
\chexdefskip{critical_mimic_qwen2.5-vl-3b}{P_f_impression}{length_truncated}{0}
\chexdefskip{critical_mimic_qwen2.5-vl-3b}{P_f_impression}{non_english}{0}
\chexdefskip{critical_mimic_qwen2.5-vl-3b}{P_f_impression}{refused}{0}
\chexdefskip{critical_mimic_qwen2.5-vl-3b}{P_f_impression}{truncations}{0}
\chexdefskip{critical_mimic_qwen2.5-vl-3b}{P_n_both}{empty}{0}
\chexdefskip{critical_mimic_qwen2.5-vl-3b}{P_n_both}{empty_target}{22}
\chexdefskip{critical_mimic_qwen2.5-vl-3b}{P_n_both}{generated}{53}
\chexdefskip{critical_mimic_qwen2.5-vl-3b}{P_n_both}{length_truncated}{1}
\chexdefskip{critical_mimic_qwen2.5-vl-3b}{P_n_both}{non_english}{0}
\chexdefskip{critical_mimic_qwen2.5-vl-3b}{P_n_both}{refused}{0}
\chexdefskip{critical_mimic_qwen2.5-vl-3b}{P_n_both}{truncations}{0}
\chexdefskip{critical_mimic_qwen2.5-vl-3b}{P_n_findings}{empty}{0}
\chexdefskip{critical_mimic_qwen2.5-vl-3b}{P_n_findings}{empty_target}{22}
\chexdefskip{critical_mimic_qwen2.5-vl-3b}{P_n_findings}{generated}{53}
\chexdefskip{critical_mimic_qwen2.5-vl-3b}{P_n_findings}{length_truncated}{2}
\chexdefskip{critical_mimic_qwen2.5-vl-3b}{P_n_findings}{non_english}{0}
\chexdefskip{critical_mimic_qwen2.5-vl-3b}{P_n_findings}{refused}{0}
\chexdefskip{critical_mimic_qwen2.5-vl-3b}{P_n_findings}{truncations}{0}
\chexdefskip{critical_mimic_qwen2.5-vl-3b}{P_n_impression}{empty}{0}
\chexdefskip{critical_mimic_qwen2.5-vl-3b}{P_n_impression}{generated}{75}
\chexdefskip{critical_mimic_qwen2.5-vl-3b}{P_n_impression}{length_truncated}{0}
\chexdefskip{critical_mimic_qwen2.5-vl-3b}{P_n_impression}{non_english}{0}
\chexdefskip{critical_mimic_qwen2.5-vl-3b}{P_n_impression}{refused}{0}
\chexdefskip{critical_mimic_qwen2.5-vl-3b}{P_n_impression}{truncations}{0}
\chexdefskip{critical_mimic_qwen2.5-vl-7b}{B_f_both}{empty}{0}
\chexdefskip{critical_mimic_qwen2.5-vl-7b}{B_f_both}{empty_target}{22}
\chexdefskip{critical_mimic_qwen2.5-vl-7b}{B_f_both}{generated}{53}
\chexdefskip{critical_mimic_qwen2.5-vl-7b}{B_f_both}{length_truncated}{0}
\chexdefskip{critical_mimic_qwen2.5-vl-7b}{B_f_both}{non_english}{0}
\chexdefskip{critical_mimic_qwen2.5-vl-7b}{B_f_both}{refused}{0}
\chexdefskip{critical_mimic_qwen2.5-vl-7b}{B_f_both}{truncations}{0}
\chexdefskip{critical_mimic_qwen2.5-vl-7b}{B_f_findings}{empty}{0}
\chexdefskip{critical_mimic_qwen2.5-vl-7b}{B_f_findings}{empty_target}{22}
\chexdefskip{critical_mimic_qwen2.5-vl-7b}{B_f_findings}{generated}{53}
\chexdefskip{critical_mimic_qwen2.5-vl-7b}{B_f_findings}{length_truncated}{0}
\chexdefskip{critical_mimic_qwen2.5-vl-7b}{B_f_findings}{non_english}{0}
\chexdefskip{critical_mimic_qwen2.5-vl-7b}{B_f_findings}{refused}{0}
\chexdefskip{critical_mimic_qwen2.5-vl-7b}{B_f_findings}{truncations}{0}
\chexdefskip{critical_mimic_qwen2.5-vl-7b}{B_f_impression}{empty}{0}
\chexdefskip{critical_mimic_qwen2.5-vl-7b}{B_f_impression}{generated}{75}
\chexdefskip{critical_mimic_qwen2.5-vl-7b}{B_f_impression}{length_truncated}{0}
\chexdefskip{critical_mimic_qwen2.5-vl-7b}{B_f_impression}{non_english}{0}
\chexdefskip{critical_mimic_qwen2.5-vl-7b}{B_f_impression}{refused}{0}
\chexdefskip{critical_mimic_qwen2.5-vl-7b}{B_f_impression}{truncations}{0}
\chexdefskip{critical_mimic_qwen2.5-vl-7b}{P_f_both}{empty}{0}
\chexdefskip{critical_mimic_qwen2.5-vl-7b}{P_f_both}{empty_target}{22}
\chexdefskip{critical_mimic_qwen2.5-vl-7b}{P_f_both}{generated}{53}
\chexdefskip{critical_mimic_qwen2.5-vl-7b}{P_f_both}{length_truncated}{0}
\chexdefskip{critical_mimic_qwen2.5-vl-7b}{P_f_both}{non_english}{0}
\chexdefskip{critical_mimic_qwen2.5-vl-7b}{P_f_both}{refused}{0}
\chexdefskip{critical_mimic_qwen2.5-vl-7b}{P_f_both}{truncations}{0}
\chexdefskip{critical_mimic_qwen2.5-vl-7b}{P_f_findings}{empty}{0}
\chexdefskip{critical_mimic_qwen2.5-vl-7b}{P_f_findings}{empty_target}{22}
\chexdefskip{critical_mimic_qwen2.5-vl-7b}{P_f_findings}{generated}{53}
\chexdefskip{critical_mimic_qwen2.5-vl-7b}{P_f_findings}{length_truncated}{2}
\chexdefskip{critical_mimic_qwen2.5-vl-7b}{P_f_findings}{non_english}{0}
\chexdefskip{critical_mimic_qwen2.5-vl-7b}{P_f_findings}{refused}{0}
\chexdefskip{critical_mimic_qwen2.5-vl-7b}{P_f_findings}{truncations}{0}
\chexdefskip{critical_mimic_qwen2.5-vl-7b}{P_f_impression}{empty}{0}
\chexdefskip{critical_mimic_qwen2.5-vl-7b}{P_f_impression}{generated}{75}
\chexdefskip{critical_mimic_qwen2.5-vl-7b}{P_f_impression}{length_truncated}{0}
\chexdefskip{critical_mimic_qwen2.5-vl-7b}{P_f_impression}{non_english}{0}
\chexdefskip{critical_mimic_qwen2.5-vl-7b}{P_f_impression}{refused}{0}
\chexdefskip{critical_mimic_qwen2.5-vl-7b}{P_f_impression}{truncations}{0}
\chexdefskip{critical_mimic_qwen2.5-vl-7b}{P_n_both}{empty}{0}
\chexdefskip{critical_mimic_qwen2.5-vl-7b}{P_n_both}{empty_target}{22}
\chexdefskip{critical_mimic_qwen2.5-vl-7b}{P_n_both}{generated}{53}
\chexdefskip{critical_mimic_qwen2.5-vl-7b}{P_n_both}{length_truncated}{1}
\chexdefskip{critical_mimic_qwen2.5-vl-7b}{P_n_both}{non_english}{0}
\chexdefskip{critical_mimic_qwen2.5-vl-7b}{P_n_both}{refused}{0}
\chexdefskip{critical_mimic_qwen2.5-vl-7b}{P_n_both}{truncations}{0}
\chexdefskip{critical_mimic_qwen2.5-vl-7b}{P_n_findings}{empty}{0}
\chexdefskip{critical_mimic_qwen2.5-vl-7b}{P_n_findings}{empty_target}{22}
\chexdefskip{critical_mimic_qwen2.5-vl-7b}{P_n_findings}{generated}{53}
\chexdefskip{critical_mimic_qwen2.5-vl-7b}{P_n_findings}{length_truncated}{1}
\chexdefskip{critical_mimic_qwen2.5-vl-7b}{P_n_findings}{non_english}{0}
\chexdefskip{critical_mimic_qwen2.5-vl-7b}{P_n_findings}{refused}{0}
\chexdefskip{critical_mimic_qwen2.5-vl-7b}{P_n_findings}{truncations}{0}
\chexdefskip{critical_mimic_qwen2.5-vl-7b}{P_n_impression}{empty}{0}
\chexdefskip{critical_mimic_qwen2.5-vl-7b}{P_n_impression}{generated}{75}
\chexdefskip{critical_mimic_qwen2.5-vl-7b}{P_n_impression}{length_truncated}{0}
\chexdefskip{critical_mimic_qwen2.5-vl-7b}{P_n_impression}{non_english}{0}
\chexdefskip{critical_mimic_qwen2.5-vl-7b}{P_n_impression}{refused}{0}
\chexdefskip{critical_mimic_qwen2.5-vl-7b}{P_n_impression}{truncations}{0}
\chexdefskip{critical_mimic_raddialog}{B_f_both}{empty}{0}
\chexdefskip{critical_mimic_raddialog}{B_f_both}{empty_target}{22}
\chexdefskip{critical_mimic_raddialog}{B_f_both}{generated}{53}
\chexdefskip{critical_mimic_raddialog}{B_f_both}{length_truncated}{0}
\chexdefskip{critical_mimic_raddialog}{B_f_both}{non_english}{0}
\chexdefskip{critical_mimic_raddialog}{B_f_both}{refused}{0}
\chexdefskip{critical_mimic_raddialog}{B_f_both}{truncations}{0}
\chexdefskip{critical_mimic_raddialog}{B_f_findings}{empty}{0}
\chexdefskip{critical_mimic_raddialog}{B_f_findings}{empty_target}{22}
\chexdefskip{critical_mimic_raddialog}{B_f_findings}{generated}{53}
\chexdefskip{critical_mimic_raddialog}{B_f_findings}{length_truncated}{0}
\chexdefskip{critical_mimic_raddialog}{B_f_findings}{non_english}{0}
\chexdefskip{critical_mimic_raddialog}{B_f_findings}{refused}{0}
\chexdefskip{critical_mimic_raddialog}{B_f_findings}{truncations}{0}
\chexdefskip{critical_mimic_raddialog}{B_f_impression}{empty}{0}
\chexdefskip{critical_mimic_raddialog}{B_f_impression}{generated}{75}
\chexdefskip{critical_mimic_raddialog}{B_f_impression}{length_truncated}{0}
\chexdefskip{critical_mimic_raddialog}{B_f_impression}{non_english}{0}
\chexdefskip{critical_mimic_raddialog}{B_f_impression}{refused}{0}
\chexdefskip{critical_mimic_raddialog}{B_f_impression}{truncations}{0}
\chexdefskip{critical_mimic_raddialog}{P_f_both}{empty}{0}
\chexdefskip{critical_mimic_raddialog}{P_f_both}{empty_target}{22}
\chexdefskip{critical_mimic_raddialog}{P_f_both}{generated}{53}
\chexdefskip{critical_mimic_raddialog}{P_f_both}{length_truncated}{0}
\chexdefskip{critical_mimic_raddialog}{P_f_both}{non_english}{0}
\chexdefskip{critical_mimic_raddialog}{P_f_both}{refused}{0}
\chexdefskip{critical_mimic_raddialog}{P_f_both}{truncations}{0}
\chexdefskip{critical_mimic_raddialog}{P_f_findings}{empty}{0}
\chexdefskip{critical_mimic_raddialog}{P_f_findings}{empty_target}{22}
\chexdefskip{critical_mimic_raddialog}{P_f_findings}{generated}{53}
\chexdefskip{critical_mimic_raddialog}{P_f_findings}{length_truncated}{0}
\chexdefskip{critical_mimic_raddialog}{P_f_findings}{non_english}{0}
\chexdefskip{critical_mimic_raddialog}{P_f_findings}{refused}{0}
\chexdefskip{critical_mimic_raddialog}{P_f_findings}{truncations}{0}
\chexdefskip{critical_mimic_raddialog}{P_f_impression}{empty}{0}
\chexdefskip{critical_mimic_raddialog}{P_f_impression}{generated}{75}
\chexdefskip{critical_mimic_raddialog}{P_f_impression}{length_truncated}{0}
\chexdefskip{critical_mimic_raddialog}{P_f_impression}{non_english}{0}
\chexdefskip{critical_mimic_raddialog}{P_f_impression}{refused}{0}
\chexdefskip{critical_mimic_raddialog}{P_f_impression}{truncations}{0}
\chexdefskip{critical_mimic_raddialog}{P_n_both}{empty}{0}
\chexdefskip{critical_mimic_raddialog}{P_n_both}{generated}{0}
\chexdefskip{critical_mimic_raddialog}{P_n_both}{length_truncated}{0}
\chexdefskip{critical_mimic_raddialog}{P_n_both}{non_english}{0}
\chexdefskip{critical_mimic_raddialog}{P_n_both}{refused}{0}
\chexdefskip{critical_mimic_raddialog}{P_n_both}{text_only_unsupported}{75}
\chexdefskip{critical_mimic_raddialog}{P_n_both}{truncations}{0}
\chexdefskip{critical_mimic_raddialog}{P_n_findings}{empty}{0}
\chexdefskip{critical_mimic_raddialog}{P_n_findings}{generated}{0}
\chexdefskip{critical_mimic_raddialog}{P_n_findings}{length_truncated}{0}
\chexdefskip{critical_mimic_raddialog}{P_n_findings}{non_english}{0}
\chexdefskip{critical_mimic_raddialog}{P_n_findings}{refused}{0}
\chexdefskip{critical_mimic_raddialog}{P_n_findings}{text_only_unsupported}{75}
\chexdefskip{critical_mimic_raddialog}{P_n_findings}{truncations}{0}
\chexdefskip{critical_mimic_raddialog}{P_n_impression}{empty}{0}
\chexdefskip{critical_mimic_raddialog}{P_n_impression}{generated}{0}
\chexdefskip{critical_mimic_raddialog}{P_n_impression}{length_truncated}{0}
\chexdefskip{critical_mimic_raddialog}{P_n_impression}{non_english}{0}
\chexdefskip{critical_mimic_raddialog}{P_n_impression}{refused}{0}
\chexdefskip{critical_mimic_raddialog}{P_n_impression}{text_only_unsupported}{75}
\chexdefskip{critical_mimic_raddialog}{P_n_impression}{truncations}{0}
\chexdefskip{critical_mimic_raddialog-prefixed}{B_f_both}{empty}{0}
\chexdefskip{critical_mimic_raddialog-prefixed}{B_f_both}{empty_target}{22}
\chexdefskip{critical_mimic_raddialog-prefixed}{B_f_both}{generated}{53}
\chexdefskip{critical_mimic_raddialog-prefixed}{B_f_both}{length_truncated}{0}
\chexdefskip{critical_mimic_raddialog-prefixed}{B_f_both}{non_english}{0}
\chexdefskip{critical_mimic_raddialog-prefixed}{B_f_both}{refused}{0}
\chexdefskip{critical_mimic_raddialog-prefixed}{B_f_both}{truncations}{0}
\chexdefskip{critical_mimic_raddialog-prefixed}{B_f_findings}{empty}{0}
\chexdefskip{critical_mimic_raddialog-prefixed}{B_f_findings}{empty_target}{22}
\chexdefskip{critical_mimic_raddialog-prefixed}{B_f_findings}{generated}{53}
\chexdefskip{critical_mimic_raddialog-prefixed}{B_f_findings}{length_truncated}{0}
\chexdefskip{critical_mimic_raddialog-prefixed}{B_f_findings}{non_english}{0}
\chexdefskip{critical_mimic_raddialog-prefixed}{B_f_findings}{refused}{0}
\chexdefskip{critical_mimic_raddialog-prefixed}{B_f_findings}{truncations}{0}
\chexdefskip{critical_mimic_raddialog-prefixed}{B_f_impression}{empty}{0}
\chexdefskip{critical_mimic_raddialog-prefixed}{B_f_impression}{generated}{75}
\chexdefskip{critical_mimic_raddialog-prefixed}{B_f_impression}{length_truncated}{0}
\chexdefskip{critical_mimic_raddialog-prefixed}{B_f_impression}{non_english}{0}
\chexdefskip{critical_mimic_raddialog-prefixed}{B_f_impression}{refused}{0}
\chexdefskip{critical_mimic_raddialog-prefixed}{B_f_impression}{truncations}{0}
\chexdefskip{critical_mimic_raddialog-prefixed}{P_f_both}{empty}{0}
\chexdefskip{critical_mimic_raddialog-prefixed}{P_f_both}{empty_target}{22}
\chexdefskip{critical_mimic_raddialog-prefixed}{P_f_both}{generated}{53}
\chexdefskip{critical_mimic_raddialog-prefixed}{P_f_both}{length_truncated}{0}
\chexdefskip{critical_mimic_raddialog-prefixed}{P_f_both}{non_english}{0}
\chexdefskip{critical_mimic_raddialog-prefixed}{P_f_both}{refused}{0}
\chexdefskip{critical_mimic_raddialog-prefixed}{P_f_both}{truncations}{0}
\chexdefskip{critical_mimic_raddialog-prefixed}{P_f_findings}{empty}{0}
\chexdefskip{critical_mimic_raddialog-prefixed}{P_f_findings}{empty_target}{22}
\chexdefskip{critical_mimic_raddialog-prefixed}{P_f_findings}{generated}{53}
\chexdefskip{critical_mimic_raddialog-prefixed}{P_f_findings}{length_truncated}{0}
\chexdefskip{critical_mimic_raddialog-prefixed}{P_f_findings}{non_english}{0}
\chexdefskip{critical_mimic_raddialog-prefixed}{P_f_findings}{refused}{0}
\chexdefskip{critical_mimic_raddialog-prefixed}{P_f_findings}{truncations}{0}
\chexdefskip{critical_mimic_raddialog-prefixed}{P_f_impression}{empty}{0}
\chexdefskip{critical_mimic_raddialog-prefixed}{P_f_impression}{generated}{75}
\chexdefskip{critical_mimic_raddialog-prefixed}{P_f_impression}{length_truncated}{0}
\chexdefskip{critical_mimic_raddialog-prefixed}{P_f_impression}{non_english}{0}
\chexdefskip{critical_mimic_raddialog-prefixed}{P_f_impression}{refused}{0}
\chexdefskip{critical_mimic_raddialog-prefixed}{P_f_impression}{truncations}{0}
\chexdefskip{critical_mimic_raddialog-prefixed}{P_n_both}{empty}{0}
\chexdefskip{critical_mimic_raddialog-prefixed}{P_n_both}{generated}{0}
\chexdefskip{critical_mimic_raddialog-prefixed}{P_n_both}{length_truncated}{0}
\chexdefskip{critical_mimic_raddialog-prefixed}{P_n_both}{non_english}{0}
\chexdefskip{critical_mimic_raddialog-prefixed}{P_n_both}{refused}{0}
\chexdefskip{critical_mimic_raddialog-prefixed}{P_n_both}{text_only_unsupported}{75}
\chexdefskip{critical_mimic_raddialog-prefixed}{P_n_both}{truncations}{0}
\chexdefskip{critical_mimic_raddialog-prefixed}{P_n_findings}{empty}{0}
\chexdefskip{critical_mimic_raddialog-prefixed}{P_n_findings}{generated}{0}
\chexdefskip{critical_mimic_raddialog-prefixed}{P_n_findings}{length_truncated}{0}
\chexdefskip{critical_mimic_raddialog-prefixed}{P_n_findings}{non_english}{0}
\chexdefskip{critical_mimic_raddialog-prefixed}{P_n_findings}{refused}{0}
\chexdefskip{critical_mimic_raddialog-prefixed}{P_n_findings}{text_only_unsupported}{75}
\chexdefskip{critical_mimic_raddialog-prefixed}{P_n_findings}{truncations}{0}
\chexdefskip{critical_mimic_raddialog-prefixed}{P_n_impression}{empty}{0}
\chexdefskip{critical_mimic_raddialog-prefixed}{P_n_impression}{generated}{0}
\chexdefskip{critical_mimic_raddialog-prefixed}{P_n_impression}{length_truncated}{0}
\chexdefskip{critical_mimic_raddialog-prefixed}{P_n_impression}{non_english}{0}
\chexdefskip{critical_mimic_raddialog-prefixed}{P_n_impression}{refused}{0}
\chexdefskip{critical_mimic_raddialog-prefixed}{P_n_impression}{text_only_unsupported}{75}
\chexdefskip{critical_mimic_raddialog-prefixed}{P_n_impression}{truncations}{0}
\chexdefskip{critical_mimic_random-report}{B_f_both}{empty}{0}
\chexdefskip{critical_mimic_random-report}{B_f_both}{empty_target}{22}
\chexdefskip{critical_mimic_random-report}{B_f_both}{generated}{53}
\chexdefskip{critical_mimic_random-report}{B_f_both}{length_truncated}{0}
\chexdefskip{critical_mimic_random-report}{B_f_both}{non_english}{0}
\chexdefskip{critical_mimic_random-report}{B_f_both}{refused}{0}
\chexdefskip{critical_mimic_random-report}{B_f_both}{truncations}{0}
\chexdefskip{critical_mimic_random-report}{B_f_findings}{empty}{0}
\chexdefskip{critical_mimic_random-report}{B_f_findings}{empty_target}{22}
\chexdefskip{critical_mimic_random-report}{B_f_findings}{generated}{53}
\chexdefskip{critical_mimic_random-report}{B_f_findings}{length_truncated}{0}
\chexdefskip{critical_mimic_random-report}{B_f_findings}{non_english}{0}
\chexdefskip{critical_mimic_random-report}{B_f_findings}{refused}{0}
\chexdefskip{critical_mimic_random-report}{B_f_findings}{truncations}{0}
\chexdefskip{critical_mimic_random-report}{B_f_impression}{empty}{0}
\chexdefskip{critical_mimic_random-report}{B_f_impression}{generated}{75}
\chexdefskip{critical_mimic_random-report}{B_f_impression}{length_truncated}{0}
\chexdefskip{critical_mimic_random-report}{B_f_impression}{non_english}{0}
\chexdefskip{critical_mimic_random-report}{B_f_impression}{refused}{0}
\chexdefskip{critical_mimic_random-report}{B_f_impression}{truncations}{0}
\chexdefskip{critical_mimic_random-report}{P_f_both}{empty}{0}
\chexdefskip{critical_mimic_random-report}{P_f_both}{empty_target}{22}
\chexdefskip{critical_mimic_random-report}{P_f_both}{generated}{53}
\chexdefskip{critical_mimic_random-report}{P_f_both}{length_truncated}{0}
\chexdefskip{critical_mimic_random-report}{P_f_both}{non_english}{0}
\chexdefskip{critical_mimic_random-report}{P_f_both}{refused}{0}
\chexdefskip{critical_mimic_random-report}{P_f_both}{truncations}{0}
\chexdefskip{critical_mimic_random-report}{P_f_findings}{empty}{0}
\chexdefskip{critical_mimic_random-report}{P_f_findings}{empty_target}{22}
\chexdefskip{critical_mimic_random-report}{P_f_findings}{generated}{53}
\chexdefskip{critical_mimic_random-report}{P_f_findings}{length_truncated}{0}
\chexdefskip{critical_mimic_random-report}{P_f_findings}{non_english}{0}
\chexdefskip{critical_mimic_random-report}{P_f_findings}{refused}{0}
\chexdefskip{critical_mimic_random-report}{P_f_findings}{truncations}{0}
\chexdefskip{critical_mimic_random-report}{P_f_impression}{empty}{0}
\chexdefskip{critical_mimic_random-report}{P_f_impression}{generated}{75}
\chexdefskip{critical_mimic_random-report}{P_f_impression}{length_truncated}{0}
\chexdefskip{critical_mimic_random-report}{P_f_impression}{non_english}{0}
\chexdefskip{critical_mimic_random-report}{P_f_impression}{refused}{1}
\chexdefskip{critical_mimic_random-report}{P_f_impression}{truncations}{0}
\chexdefskip{critical_mimic_random-report}{P_n_both}{empty}{0}
\chexdefskip{critical_mimic_random-report}{P_n_both}{empty_target}{22}
\chexdefskip{critical_mimic_random-report}{P_n_both}{generated}{53}
\chexdefskip{critical_mimic_random-report}{P_n_both}{length_truncated}{0}
\chexdefskip{critical_mimic_random-report}{P_n_both}{non_english}{0}
\chexdefskip{critical_mimic_random-report}{P_n_both}{refused}{0}
\chexdefskip{critical_mimic_random-report}{P_n_both}{truncations}{0}
\chexdefskip{critical_mimic_random-report}{P_n_findings}{empty}{0}
\chexdefskip{critical_mimic_random-report}{P_n_findings}{empty_target}{22}
\chexdefskip{critical_mimic_random-report}{P_n_findings}{generated}{53}
\chexdefskip{critical_mimic_random-report}{P_n_findings}{length_truncated}{0}
\chexdefskip{critical_mimic_random-report}{P_n_findings}{non_english}{0}
\chexdefskip{critical_mimic_random-report}{P_n_findings}{refused}{0}
\chexdefskip{critical_mimic_random-report}{P_n_findings}{truncations}{0}
\chexdefskip{critical_mimic_random-report}{P_n_impression}{empty}{0}
\chexdefskip{critical_mimic_random-report}{P_n_impression}{generated}{75}
\chexdefskip{critical_mimic_random-report}{P_n_impression}{length_truncated}{0}
\chexdefskip{critical_mimic_random-report}{P_n_impression}{non_english}{0}
\chexdefskip{critical_mimic_random-report}{P_n_impression}{refused}{0}
\chexdefskip{critical_mimic_random-report}{P_n_impression}{truncations}{0}
\chexdefskip{critical_mimic_rgrg}{B_f_findings}{empty}{0}
\chexdefskip{critical_mimic_rgrg}{B_f_findings}{empty_target}{22}
\chexdefskip{critical_mimic_rgrg}{B_f_findings}{generated}{53}
\chexdefskip{critical_mimic_rgrg}{B_f_findings}{length_truncated}{0}
\chexdefskip{critical_mimic_rgrg}{B_f_findings}{non_english}{0}
\chexdefskip{critical_mimic_rgrg}{B_f_findings}{refused}{0}
\chexdefskip{critical_mimic_rgrg}{B_f_findings}{truncations}{0}
\chexdefskip{critical_mimic_rgrg}{P_f_findings}{empty}{0}
\chexdefskip{critical_mimic_rgrg}{P_f_findings}{empty_target}{22}
\chexdefskip{critical_mimic_rgrg}{P_f_findings}{generated}{53}
\chexdefskip{critical_mimic_rgrg}{P_f_findings}{length_truncated}{0}
\chexdefskip{critical_mimic_rgrg}{P_f_findings}{non_english}{0}
\chexdefskip{critical_mimic_rgrg}{P_f_findings}{refused}{0}
\chexdefskip{critical_mimic_rgrg}{P_f_findings}{truncations}{0}
\chexdefskip{settings_mimic_chexagent-2-3b-native}{B_f_both}{empty}{0}
\chexdefskip{settings_mimic_chexagent-2-3b-native}{B_f_both}{empty_target}{1642}
\chexdefskip{settings_mimic_chexagent-2-3b-native}{B_f_both}{generated}{1624}
\chexdefskip{settings_mimic_chexagent-2-3b-native}{B_f_both}{length_truncated}{0}
\chexdefskip{settings_mimic_chexagent-2-3b-native}{B_f_both}{no_frontal}{3}
\chexdefskip{settings_mimic_chexagent-2-3b-native}{B_f_both}{non_english}{0}
\chexdefskip{settings_mimic_chexagent-2-3b-native}{B_f_both}{refused}{0}
\chexdefskip{settings_mimic_chexagent-2-3b-native}{B_f_both}{truncations}{0}
\chexdefskip{settings_mimic_chexagent-2-3b-native}{B_f_findings}{empty}{0}
\chexdefskip{settings_mimic_chexagent-2-3b-native}{B_f_findings}{empty_target}{920}
\chexdefskip{settings_mimic_chexagent-2-3b-native}{B_f_findings}{generated}{2346}
\chexdefskip{settings_mimic_chexagent-2-3b-native}{B_f_findings}{length_truncated}{0}
\chexdefskip{settings_mimic_chexagent-2-3b-native}{B_f_findings}{no_frontal}{3}
\chexdefskip{settings_mimic_chexagent-2-3b-native}{B_f_findings}{non_english}{0}
\chexdefskip{settings_mimic_chexagent-2-3b-native}{B_f_findings}{refused}{0}
\chexdefskip{settings_mimic_chexagent-2-3b-native}{B_f_findings}{truncations}{0}
\chexdefskip{settings_mimic_chexagent-2-3b-native}{B_f_impression}{empty}{0}
\chexdefskip{settings_mimic_chexagent-2-3b-native}{B_f_impression}{empty_target}{1043}
\chexdefskip{settings_mimic_chexagent-2-3b-native}{B_f_impression}{generated}{2223}
\chexdefskip{settings_mimic_chexagent-2-3b-native}{B_f_impression}{length_truncated}{0}
\chexdefskip{settings_mimic_chexagent-2-3b-native}{B_f_impression}{no_frontal}{3}
\chexdefskip{settings_mimic_chexagent-2-3b-native}{B_f_impression}{non_english}{0}
\chexdefskip{settings_mimic_chexagent-2-3b-native}{B_f_impression}{refused}{0}
\chexdefskip{settings_mimic_chexagent-2-3b-native}{B_f_impression}{truncations}{0}
\chexdefskip{settings_mimic_chexagent-2-3b-native}{P_f_both}{empty}{0}
\chexdefskip{settings_mimic_chexagent-2-3b-native}{P_f_both}{empty_target}{1642}
\chexdefskip{settings_mimic_chexagent-2-3b-native}{P_f_both}{generated}{1624}
\chexdefskip{settings_mimic_chexagent-2-3b-native}{P_f_both}{length_truncated}{0}
\chexdefskip{settings_mimic_chexagent-2-3b-native}{P_f_both}{no_frontal}{3}
\chexdefskip{settings_mimic_chexagent-2-3b-native}{P_f_both}{non_english}{0}
\chexdefskip{settings_mimic_chexagent-2-3b-native}{P_f_both}{refused}{0}
\chexdefskip{settings_mimic_chexagent-2-3b-native}{P_f_both}{truncations}{0}
\chexdefskip{settings_mimic_chexagent-2-3b-native}{P_f_findings}{empty}{0}
\chexdefskip{settings_mimic_chexagent-2-3b-native}{P_f_findings}{empty_target}{920}
\chexdefskip{settings_mimic_chexagent-2-3b-native}{P_f_findings}{generated}{2346}
\chexdefskip{settings_mimic_chexagent-2-3b-native}{P_f_findings}{length_truncated}{0}
\chexdefskip{settings_mimic_chexagent-2-3b-native}{P_f_findings}{no_frontal}{3}
\chexdefskip{settings_mimic_chexagent-2-3b-native}{P_f_findings}{non_english}{0}
\chexdefskip{settings_mimic_chexagent-2-3b-native}{P_f_findings}{refused}{0}
\chexdefskip{settings_mimic_chexagent-2-3b-native}{P_f_findings}{truncations}{0}
\chexdefskip{settings_mimic_chexagent-2-3b-native}{P_f_impression}{empty}{0}
\chexdefskip{settings_mimic_chexagent-2-3b-native}{P_f_impression}{empty_target}{1043}
\chexdefskip{settings_mimic_chexagent-2-3b-native}{P_f_impression}{generated}{2223}
\chexdefskip{settings_mimic_chexagent-2-3b-native}{P_f_impression}{length_truncated}{0}
\chexdefskip{settings_mimic_chexagent-2-3b-native}{P_f_impression}{no_frontal}{3}
\chexdefskip{settings_mimic_chexagent-2-3b-native}{P_f_impression}{non_english}{0}
\chexdefskip{settings_mimic_chexagent-2-3b-native}{P_f_impression}{refused}{0}
\chexdefskip{settings_mimic_chexagent-2-3b-native}{P_f_impression}{truncations}{0}
\chexdefskip{settings_mimic_chexagent-2-3b-native}{P_n_both}{empty}{0}
\chexdefskip{settings_mimic_chexagent-2-3b-native}{P_n_both}{generated}{0}
\chexdefskip{settings_mimic_chexagent-2-3b-native}{P_n_both}{length_truncated}{0}
\chexdefskip{settings_mimic_chexagent-2-3b-native}{P_n_both}{non_english}{0}
\chexdefskip{settings_mimic_chexagent-2-3b-native}{P_n_both}{refused}{0}
\chexdefskip{settings_mimic_chexagent-2-3b-native}{P_n_both}{text_only_unsupported}{3269}
\chexdefskip{settings_mimic_chexagent-2-3b-native}{P_n_both}{truncations}{0}
\chexdefskip{settings_mimic_chexagent-2-3b-native}{P_n_findings}{empty}{0}
\chexdefskip{settings_mimic_chexagent-2-3b-native}{P_n_findings}{generated}{0}
\chexdefskip{settings_mimic_chexagent-2-3b-native}{P_n_findings}{length_truncated}{0}
\chexdefskip{settings_mimic_chexagent-2-3b-native}{P_n_findings}{non_english}{0}
\chexdefskip{settings_mimic_chexagent-2-3b-native}{P_n_findings}{refused}{0}
\chexdefskip{settings_mimic_chexagent-2-3b-native}{P_n_findings}{text_only_unsupported}{3269}
\chexdefskip{settings_mimic_chexagent-2-3b-native}{P_n_findings}{truncations}{0}
\chexdefskip{settings_mimic_chexagent-2-3b-native}{P_n_impression}{empty}{0}
\chexdefskip{settings_mimic_chexagent-2-3b-native}{P_n_impression}{generated}{0}
\chexdefskip{settings_mimic_chexagent-2-3b-native}{P_n_impression}{length_truncated}{0}
\chexdefskip{settings_mimic_chexagent-2-3b-native}{P_n_impression}{non_english}{0}
\chexdefskip{settings_mimic_chexagent-2-3b-native}{P_n_impression}{refused}{0}
\chexdefskip{settings_mimic_chexagent-2-3b-native}{P_n_impression}{text_only_unsupported}{3269}
\chexdefskip{settings_mimic_chexagent-2-3b-native}{P_n_impression}{truncations}{0}
\chexdefskip{settings_mimic_chexagent-2-3b-native-notags}{B_f_both}{empty}{0}
\chexdefskip{settings_mimic_chexagent-2-3b-native-notags}{B_f_both}{empty_target}{1642}
\chexdefskip{settings_mimic_chexagent-2-3b-native-notags}{B_f_both}{generated}{1624}
\chexdefskip{settings_mimic_chexagent-2-3b-native-notags}{B_f_both}{length_truncated}{0}
\chexdefskip{settings_mimic_chexagent-2-3b-native-notags}{B_f_both}{no_frontal}{3}
\chexdefskip{settings_mimic_chexagent-2-3b-native-notags}{B_f_both}{non_english}{0}
\chexdefskip{settings_mimic_chexagent-2-3b-native-notags}{B_f_both}{refused}{0}
\chexdefskip{settings_mimic_chexagent-2-3b-native-notags}{B_f_both}{truncations}{0}
\chexdefskip{settings_mimic_chexagent-2-3b-native-notags}{B_f_findings}{empty}{0}
\chexdefskip{settings_mimic_chexagent-2-3b-native-notags}{B_f_findings}{empty_target}{920}
\chexdefskip{settings_mimic_chexagent-2-3b-native-notags}{B_f_findings}{generated}{2346}
\chexdefskip{settings_mimic_chexagent-2-3b-native-notags}{B_f_findings}{length_truncated}{0}
\chexdefskip{settings_mimic_chexagent-2-3b-native-notags}{B_f_findings}{no_frontal}{3}
\chexdefskip{settings_mimic_chexagent-2-3b-native-notags}{B_f_findings}{non_english}{0}
\chexdefskip{settings_mimic_chexagent-2-3b-native-notags}{B_f_findings}{refused}{0}
\chexdefskip{settings_mimic_chexagent-2-3b-native-notags}{B_f_findings}{truncations}{0}
\chexdefskip{settings_mimic_chexagent-2-3b-native-notags}{B_f_impression}{empty}{0}
\chexdefskip{settings_mimic_chexagent-2-3b-native-notags}{B_f_impression}{empty_target}{1043}
\chexdefskip{settings_mimic_chexagent-2-3b-native-notags}{B_f_impression}{generated}{2223}
\chexdefskip{settings_mimic_chexagent-2-3b-native-notags}{B_f_impression}{length_truncated}{0}
\chexdefskip{settings_mimic_chexagent-2-3b-native-notags}{B_f_impression}{no_frontal}{3}
\chexdefskip{settings_mimic_chexagent-2-3b-native-notags}{B_f_impression}{non_english}{0}
\chexdefskip{settings_mimic_chexagent-2-3b-native-notags}{B_f_impression}{refused}{0}
\chexdefskip{settings_mimic_chexagent-2-3b-native-notags}{B_f_impression}{truncations}{0}
\chexdefskip{settings_mimic_chexagent-2-3b-native-notags}{P_f_both}{empty}{0}
\chexdefskip{settings_mimic_chexagent-2-3b-native-notags}{P_f_both}{empty_target}{1642}
\chexdefskip{settings_mimic_chexagent-2-3b-native-notags}{P_f_both}{generated}{1624}
\chexdefskip{settings_mimic_chexagent-2-3b-native-notags}{P_f_both}{length_truncated}{0}
\chexdefskip{settings_mimic_chexagent-2-3b-native-notags}{P_f_both}{no_frontal}{3}
\chexdefskip{settings_mimic_chexagent-2-3b-native-notags}{P_f_both}{non_english}{0}
\chexdefskip{settings_mimic_chexagent-2-3b-native-notags}{P_f_both}{refused}{0}
\chexdefskip{settings_mimic_chexagent-2-3b-native-notags}{P_f_both}{truncations}{0}
\chexdefskip{settings_mimic_chexagent-2-3b-native-notags}{P_f_findings}{empty}{0}
\chexdefskip{settings_mimic_chexagent-2-3b-native-notags}{P_f_findings}{empty_target}{920}
\chexdefskip{settings_mimic_chexagent-2-3b-native-notags}{P_f_findings}{generated}{2346}
\chexdefskip{settings_mimic_chexagent-2-3b-native-notags}{P_f_findings}{length_truncated}{0}
\chexdefskip{settings_mimic_chexagent-2-3b-native-notags}{P_f_findings}{no_frontal}{3}
\chexdefskip{settings_mimic_chexagent-2-3b-native-notags}{P_f_findings}{non_english}{0}
\chexdefskip{settings_mimic_chexagent-2-3b-native-notags}{P_f_findings}{refused}{0}
\chexdefskip{settings_mimic_chexagent-2-3b-native-notags}{P_f_findings}{truncations}{0}
\chexdefskip{settings_mimic_chexagent-2-3b-native-notags}{P_f_impression}{empty}{0}
\chexdefskip{settings_mimic_chexagent-2-3b-native-notags}{P_f_impression}{empty_target}{1043}
\chexdefskip{settings_mimic_chexagent-2-3b-native-notags}{P_f_impression}{generated}{2223}
\chexdefskip{settings_mimic_chexagent-2-3b-native-notags}{P_f_impression}{length_truncated}{0}
\chexdefskip{settings_mimic_chexagent-2-3b-native-notags}{P_f_impression}{no_frontal}{3}
\chexdefskip{settings_mimic_chexagent-2-3b-native-notags}{P_f_impression}{non_english}{0}
\chexdefskip{settings_mimic_chexagent-2-3b-native-notags}{P_f_impression}{refused}{0}
\chexdefskip{settings_mimic_chexagent-2-3b-native-notags}{P_f_impression}{truncations}{0}
\chexdefskip{settings_mimic_chexagent-2-3b-native-notags}{P_n_both}{empty}{0}
\chexdefskip{settings_mimic_chexagent-2-3b-native-notags}{P_n_both}{generated}{0}
\chexdefskip{settings_mimic_chexagent-2-3b-native-notags}{P_n_both}{length_truncated}{0}
\chexdefskip{settings_mimic_chexagent-2-3b-native-notags}{P_n_both}{non_english}{0}
\chexdefskip{settings_mimic_chexagent-2-3b-native-notags}{P_n_both}{refused}{0}
\chexdefskip{settings_mimic_chexagent-2-3b-native-notags}{P_n_both}{text_only_unsupported}{3269}
\chexdefskip{settings_mimic_chexagent-2-3b-native-notags}{P_n_both}{truncations}{0}
\chexdefskip{settings_mimic_chexagent-2-3b-native-notags}{P_n_findings}{empty}{0}
\chexdefskip{settings_mimic_chexagent-2-3b-native-notags}{P_n_findings}{generated}{0}
\chexdefskip{settings_mimic_chexagent-2-3b-native-notags}{P_n_findings}{length_truncated}{0}
\chexdefskip{settings_mimic_chexagent-2-3b-native-notags}{P_n_findings}{non_english}{0}
\chexdefskip{settings_mimic_chexagent-2-3b-native-notags}{P_n_findings}{refused}{0}
\chexdefskip{settings_mimic_chexagent-2-3b-native-notags}{P_n_findings}{text_only_unsupported}{3269}
\chexdefskip{settings_mimic_chexagent-2-3b-native-notags}{P_n_findings}{truncations}{0}
\chexdefskip{settings_mimic_chexagent-2-3b-native-notags}{P_n_impression}{empty}{0}
\chexdefskip{settings_mimic_chexagent-2-3b-native-notags}{P_n_impression}{generated}{0}
\chexdefskip{settings_mimic_chexagent-2-3b-native-notags}{P_n_impression}{length_truncated}{0}
\chexdefskip{settings_mimic_chexagent-2-3b-native-notags}{P_n_impression}{non_english}{0}
\chexdefskip{settings_mimic_chexagent-2-3b-native-notags}{P_n_impression}{refused}{0}
\chexdefskip{settings_mimic_chexagent-2-3b-native-notags}{P_n_impression}{text_only_unsupported}{3269}
\chexdefskip{settings_mimic_chexagent-2-3b-native-notags}{P_n_impression}{truncations}{0}
\chexdefskip{settings_mimic_chexone-instruct}{B_f_both}{empty}{0}
\chexdefskip{settings_mimic_chexone-instruct}{B_f_both}{empty_target}{1642}
\chexdefskip{settings_mimic_chexone-instruct}{B_f_both}{generated}{1624}
\chexdefskip{settings_mimic_chexone-instruct}{B_f_both}{length_truncated}{45}
\chexdefskip{settings_mimic_chexone-instruct}{B_f_both}{no_frontal}{3}
\chexdefskip{settings_mimic_chexone-instruct}{B_f_both}{non_english}{0}
\chexdefskip{settings_mimic_chexone-instruct}{B_f_both}{refused}{0}
\chexdefskip{settings_mimic_chexone-instruct}{B_f_both}{truncations}{0}
\chexdefskip{settings_mimic_chexone-instruct}{B_f_findings}{empty}{0}
\chexdefskip{settings_mimic_chexone-instruct}{B_f_findings}{empty_target}{920}
\chexdefskip{settings_mimic_chexone-instruct}{B_f_findings}{generated}{2346}
\chexdefskip{settings_mimic_chexone-instruct}{B_f_findings}{length_truncated}{5}
\chexdefskip{settings_mimic_chexone-instruct}{B_f_findings}{no_frontal}{3}
\chexdefskip{settings_mimic_chexone-instruct}{B_f_findings}{non_english}{0}
\chexdefskip{settings_mimic_chexone-instruct}{B_f_findings}{refused}{0}
\chexdefskip{settings_mimic_chexone-instruct}{B_f_findings}{truncations}{0}
\chexdefskip{settings_mimic_chexone-instruct}{B_f_impression}{empty}{0}
\chexdefskip{settings_mimic_chexone-instruct}{B_f_impression}{empty_target}{1043}
\chexdefskip{settings_mimic_chexone-instruct}{B_f_impression}{generated}{2223}
\chexdefskip{settings_mimic_chexone-instruct}{B_f_impression}{length_truncated}{60}
\chexdefskip{settings_mimic_chexone-instruct}{B_f_impression}{no_frontal}{3}
\chexdefskip{settings_mimic_chexone-instruct}{B_f_impression}{non_english}{0}
\chexdefskip{settings_mimic_chexone-instruct}{B_f_impression}{refused}{0}
\chexdefskip{settings_mimic_chexone-instruct}{B_f_impression}{truncations}{0}
\chexdefskip{settings_mimic_chexone-instruct}{P_f_both}{empty}{0}
\chexdefskip{settings_mimic_chexone-instruct}{P_f_both}{empty_target}{1642}
\chexdefskip{settings_mimic_chexone-instruct}{P_f_both}{generated}{1624}
\chexdefskip{settings_mimic_chexone-instruct}{P_f_both}{length_truncated}{52}
\chexdefskip{settings_mimic_chexone-instruct}{P_f_both}{no_frontal}{3}
\chexdefskip{settings_mimic_chexone-instruct}{P_f_both}{non_english}{0}
\chexdefskip{settings_mimic_chexone-instruct}{P_f_both}{refused}{0}
\chexdefskip{settings_mimic_chexone-instruct}{P_f_both}{truncations}{0}
\chexdefskip{settings_mimic_chexone-instruct}{P_f_findings}{empty}{0}
\chexdefskip{settings_mimic_chexone-instruct}{P_f_findings}{empty_target}{920}
\chexdefskip{settings_mimic_chexone-instruct}{P_f_findings}{generated}{2346}
\chexdefskip{settings_mimic_chexone-instruct}{P_f_findings}{length_truncated}{20}
\chexdefskip{settings_mimic_chexone-instruct}{P_f_findings}{no_frontal}{3}
\chexdefskip{settings_mimic_chexone-instruct}{P_f_findings}{non_english}{0}
\chexdefskip{settings_mimic_chexone-instruct}{P_f_findings}{refused}{0}
\chexdefskip{settings_mimic_chexone-instruct}{P_f_findings}{truncations}{0}
\chexdefskip{settings_mimic_chexone-instruct}{P_f_impression}{empty}{0}
\chexdefskip{settings_mimic_chexone-instruct}{P_f_impression}{empty_target}{1043}
\chexdefskip{settings_mimic_chexone-instruct}{P_f_impression}{generated}{2223}
\chexdefskip{settings_mimic_chexone-instruct}{P_f_impression}{length_truncated}{46}
\chexdefskip{settings_mimic_chexone-instruct}{P_f_impression}{no_frontal}{3}
\chexdefskip{settings_mimic_chexone-instruct}{P_f_impression}{non_english}{0}
\chexdefskip{settings_mimic_chexone-instruct}{P_f_impression}{refused}{0}
\chexdefskip{settings_mimic_chexone-instruct}{P_f_impression}{truncations}{0}
\chexdefskip{settings_mimic_chexone-instruct}{P_n_both}{empty}{0}
\chexdefskip{settings_mimic_chexone-instruct}{P_n_both}{empty_target}{1644}
\chexdefskip{settings_mimic_chexone-instruct}{P_n_both}{generated}{1625}
\chexdefskip{settings_mimic_chexone-instruct}{P_n_both}{length_truncated}{43}
\chexdefskip{settings_mimic_chexone-instruct}{P_n_both}{non_english}{0}
\chexdefskip{settings_mimic_chexone-instruct}{P_n_both}{refused}{0}
\chexdefskip{settings_mimic_chexone-instruct}{P_n_both}{truncations}{0}
\chexdefskip{settings_mimic_chexone-instruct}{P_n_findings}{empty}{0}
\chexdefskip{settings_mimic_chexone-instruct}{P_n_findings}{empty_target}{921}
\chexdefskip{settings_mimic_chexone-instruct}{P_n_findings}{generated}{2348}
\chexdefskip{settings_mimic_chexone-instruct}{P_n_findings}{length_truncated}{29}
\chexdefskip{settings_mimic_chexone-instruct}{P_n_findings}{non_english}{0}
\chexdefskip{settings_mimic_chexone-instruct}{P_n_findings}{refused}{0}
\chexdefskip{settings_mimic_chexone-instruct}{P_n_findings}{truncations}{0}
\chexdefskip{settings_mimic_chexone-instruct}{P_n_impression}{empty}{0}
\chexdefskip{settings_mimic_chexone-instruct}{P_n_impression}{empty_target}{1045}
\chexdefskip{settings_mimic_chexone-instruct}{P_n_impression}{generated}{2224}
\chexdefskip{settings_mimic_chexone-instruct}{P_n_impression}{length_truncated}{32}
\chexdefskip{settings_mimic_chexone-instruct}{P_n_impression}{non_english}{0}
\chexdefskip{settings_mimic_chexone-instruct}{P_n_impression}{refused}{0}
\chexdefskip{settings_mimic_chexone-instruct}{P_n_impression}{truncations}{0}
\chexdefskip{settings_mimic_chexone-reasoning}{B_f_both}{empty}{0}
\chexdefskip{settings_mimic_chexone-reasoning}{B_f_both}{empty_target}{1642}
\chexdefskip{settings_mimic_chexone-reasoning}{B_f_both}{generated}{1624}
\chexdefskip{settings_mimic_chexone-reasoning}{B_f_both}{length_truncated}{8}
\chexdefskip{settings_mimic_chexone-reasoning}{B_f_both}{no_frontal}{3}
\chexdefskip{settings_mimic_chexone-reasoning}{B_f_both}{non_english}{0}
\chexdefskip{settings_mimic_chexone-reasoning}{B_f_both}{refused}{0}
\chexdefskip{settings_mimic_chexone-reasoning}{B_f_both}{truncations}{0}
\chexdefskip{settings_mimic_chexone-reasoning}{B_f_findings}{empty}{13}
\chexdefskip{settings_mimic_chexone-reasoning}{B_f_findings}{empty_target}{920}
\chexdefskip{settings_mimic_chexone-reasoning}{B_f_findings}{generated}{2346}
\chexdefskip{settings_mimic_chexone-reasoning}{B_f_findings}{length_truncated}{13}
\chexdefskip{settings_mimic_chexone-reasoning}{B_f_findings}{no_frontal}{3}
\chexdefskip{settings_mimic_chexone-reasoning}{B_f_findings}{non_english}{0}
\chexdefskip{settings_mimic_chexone-reasoning}{B_f_findings}{refused}{0}
\chexdefskip{settings_mimic_chexone-reasoning}{B_f_findings}{truncations}{0}
\chexdefskip{settings_mimic_chexone-reasoning}{B_f_impression}{empty}{3}
\chexdefskip{settings_mimic_chexone-reasoning}{B_f_impression}{empty_target}{1043}
\chexdefskip{settings_mimic_chexone-reasoning}{B_f_impression}{generated}{2223}
\chexdefskip{settings_mimic_chexone-reasoning}{B_f_impression}{length_truncated}{3}
\chexdefskip{settings_mimic_chexone-reasoning}{B_f_impression}{no_frontal}{3}
\chexdefskip{settings_mimic_chexone-reasoning}{B_f_impression}{non_english}{0}
\chexdefskip{settings_mimic_chexone-reasoning}{B_f_impression}{refused}{0}
\chexdefskip{settings_mimic_chexone-reasoning}{B_f_impression}{truncations}{0}
\chexdefskip{settings_mimic_chexone-reasoning}{P_f_both}{empty}{0}
\chexdefskip{settings_mimic_chexone-reasoning}{P_f_both}{empty_target}{1642}
\chexdefskip{settings_mimic_chexone-reasoning}{P_f_both}{generated}{1624}
\chexdefskip{settings_mimic_chexone-reasoning}{P_f_both}{length_truncated}{13}
\chexdefskip{settings_mimic_chexone-reasoning}{P_f_both}{no_frontal}{3}
\chexdefskip{settings_mimic_chexone-reasoning}{P_f_both}{non_english}{0}
\chexdefskip{settings_mimic_chexone-reasoning}{P_f_both}{refused}{0}
\chexdefskip{settings_mimic_chexone-reasoning}{P_f_both}{truncations}{0}
\chexdefskip{settings_mimic_chexone-reasoning}{P_f_findings}{empty}{12}
\chexdefskip{settings_mimic_chexone-reasoning}{P_f_findings}{empty_target}{920}
\chexdefskip{settings_mimic_chexone-reasoning}{P_f_findings}{generated}{2346}
\chexdefskip{settings_mimic_chexone-reasoning}{P_f_findings}{length_truncated}{12}
\chexdefskip{settings_mimic_chexone-reasoning}{P_f_findings}{no_frontal}{3}
\chexdefskip{settings_mimic_chexone-reasoning}{P_f_findings}{non_english}{0}
\chexdefskip{settings_mimic_chexone-reasoning}{P_f_findings}{refused}{0}
\chexdefskip{settings_mimic_chexone-reasoning}{P_f_findings}{truncations}{0}
\chexdefskip{settings_mimic_chexone-reasoning}{P_f_impression}{empty}{9}
\chexdefskip{settings_mimic_chexone-reasoning}{P_f_impression}{empty_target}{1043}
\chexdefskip{settings_mimic_chexone-reasoning}{P_f_impression}{generated}{2223}
\chexdefskip{settings_mimic_chexone-reasoning}{P_f_impression}{length_truncated}{9}
\chexdefskip{settings_mimic_chexone-reasoning}{P_f_impression}{no_frontal}{3}
\chexdefskip{settings_mimic_chexone-reasoning}{P_f_impression}{non_english}{0}
\chexdefskip{settings_mimic_chexone-reasoning}{P_f_impression}{refused}{0}
\chexdefskip{settings_mimic_chexone-reasoning}{P_f_impression}{truncations}{0}
\chexdefskip{settings_mimic_chexone-reasoning}{P_n_both}{empty}{2}
\chexdefskip{settings_mimic_chexone-reasoning}{P_n_both}{empty_target}{1644}
\chexdefskip{settings_mimic_chexone-reasoning}{P_n_both}{generated}{1625}
\chexdefskip{settings_mimic_chexone-reasoning}{P_n_both}{length_truncated}{92}
\chexdefskip{settings_mimic_chexone-reasoning}{P_n_both}{non_english}{0}
\chexdefskip{settings_mimic_chexone-reasoning}{P_n_both}{refused}{0}
\chexdefskip{settings_mimic_chexone-reasoning}{P_n_both}{truncations}{0}
\chexdefskip{settings_mimic_chexone-reasoning}{P_n_findings}{empty}{114}
\chexdefskip{settings_mimic_chexone-reasoning}{P_n_findings}{empty_target}{921}
\chexdefskip{settings_mimic_chexone-reasoning}{P_n_findings}{generated}{2348}
\chexdefskip{settings_mimic_chexone-reasoning}{P_n_findings}{length_truncated}{112}
\chexdefskip{settings_mimic_chexone-reasoning}{P_n_findings}{non_english}{0}
\chexdefskip{settings_mimic_chexone-reasoning}{P_n_findings}{refused}{0}
\chexdefskip{settings_mimic_chexone-reasoning}{P_n_findings}{truncations}{0}
\chexdefskip{settings_mimic_chexone-reasoning}{P_n_impression}{empty}{45}
\chexdefskip{settings_mimic_chexone-reasoning}{P_n_impression}{empty_target}{1045}
\chexdefskip{settings_mimic_chexone-reasoning}{P_n_impression}{generated}{2224}
\chexdefskip{settings_mimic_chexone-reasoning}{P_n_impression}{length_truncated}{45}
\chexdefskip{settings_mimic_chexone-reasoning}{P_n_impression}{non_english}{0}
\chexdefskip{settings_mimic_chexone-reasoning}{P_n_impression}{refused}{0}
\chexdefskip{settings_mimic_chexone-reasoning}{P_n_impression}{truncations}{0}
\chexdefskip{settings_mimic_constant-normal}{B_f_both}{empty}{0}
\chexdefskip{settings_mimic_constant-normal}{B_f_both}{empty_target}{1642}
\chexdefskip{settings_mimic_constant-normal}{B_f_both}{generated}{1624}
\chexdefskip{settings_mimic_constant-normal}{B_f_both}{length_truncated}{0}
\chexdefskip{settings_mimic_constant-normal}{B_f_both}{no_frontal}{3}
\chexdefskip{settings_mimic_constant-normal}{B_f_both}{non_english}{0}
\chexdefskip{settings_mimic_constant-normal}{B_f_both}{refused}{0}
\chexdefskip{settings_mimic_constant-normal}{B_f_both}{truncations}{0}
\chexdefskip{settings_mimic_constant-normal}{B_f_findings}{empty}{0}
\chexdefskip{settings_mimic_constant-normal}{B_f_findings}{empty_target}{920}
\chexdefskip{settings_mimic_constant-normal}{B_f_findings}{generated}{2346}
\chexdefskip{settings_mimic_constant-normal}{B_f_findings}{length_truncated}{0}
\chexdefskip{settings_mimic_constant-normal}{B_f_findings}{no_frontal}{3}
\chexdefskip{settings_mimic_constant-normal}{B_f_findings}{non_english}{0}
\chexdefskip{settings_mimic_constant-normal}{B_f_findings}{refused}{0}
\chexdefskip{settings_mimic_constant-normal}{B_f_findings}{truncations}{0}
\chexdefskip{settings_mimic_constant-normal}{B_f_impression}{empty}{0}
\chexdefskip{settings_mimic_constant-normal}{B_f_impression}{empty_target}{1043}
\chexdefskip{settings_mimic_constant-normal}{B_f_impression}{generated}{2223}
\chexdefskip{settings_mimic_constant-normal}{B_f_impression}{length_truncated}{0}
\chexdefskip{settings_mimic_constant-normal}{B_f_impression}{no_frontal}{3}
\chexdefskip{settings_mimic_constant-normal}{B_f_impression}{non_english}{0}
\chexdefskip{settings_mimic_constant-normal}{B_f_impression}{refused}{0}
\chexdefskip{settings_mimic_constant-normal}{B_f_impression}{truncations}{0}
\chexdefskip{settings_mimic_constant-normal}{P_f_both}{empty}{0}
\chexdefskip{settings_mimic_constant-normal}{P_f_both}{empty_target}{1642}
\chexdefskip{settings_mimic_constant-normal}{P_f_both}{generated}{1624}
\chexdefskip{settings_mimic_constant-normal}{P_f_both}{length_truncated}{0}
\chexdefskip{settings_mimic_constant-normal}{P_f_both}{no_frontal}{3}
\chexdefskip{settings_mimic_constant-normal}{P_f_both}{non_english}{0}
\chexdefskip{settings_mimic_constant-normal}{P_f_both}{refused}{0}
\chexdefskip{settings_mimic_constant-normal}{P_f_both}{truncations}{0}
\chexdefskip{settings_mimic_constant-normal}{P_f_findings}{empty}{0}
\chexdefskip{settings_mimic_constant-normal}{P_f_findings}{empty_target}{920}
\chexdefskip{settings_mimic_constant-normal}{P_f_findings}{generated}{2346}
\chexdefskip{settings_mimic_constant-normal}{P_f_findings}{length_truncated}{0}
\chexdefskip{settings_mimic_constant-normal}{P_f_findings}{no_frontal}{3}
\chexdefskip{settings_mimic_constant-normal}{P_f_findings}{non_english}{0}
\chexdefskip{settings_mimic_constant-normal}{P_f_findings}{refused}{0}
\chexdefskip{settings_mimic_constant-normal}{P_f_findings}{truncations}{0}
\chexdefskip{settings_mimic_constant-normal}{P_f_impression}{empty}{0}
\chexdefskip{settings_mimic_constant-normal}{P_f_impression}{empty_target}{1043}
\chexdefskip{settings_mimic_constant-normal}{P_f_impression}{generated}{2223}
\chexdefskip{settings_mimic_constant-normal}{P_f_impression}{length_truncated}{0}
\chexdefskip{settings_mimic_constant-normal}{P_f_impression}{no_frontal}{3}
\chexdefskip{settings_mimic_constant-normal}{P_f_impression}{non_english}{0}
\chexdefskip{settings_mimic_constant-normal}{P_f_impression}{refused}{0}
\chexdefskip{settings_mimic_constant-normal}{P_f_impression}{truncations}{0}
\chexdefskip{settings_mimic_constant-normal}{P_n_both}{empty}{0}
\chexdefskip{settings_mimic_constant-normal}{P_n_both}{empty_target}{1644}
\chexdefskip{settings_mimic_constant-normal}{P_n_both}{generated}{1625}
\chexdefskip{settings_mimic_constant-normal}{P_n_both}{length_truncated}{0}
\chexdefskip{settings_mimic_constant-normal}{P_n_both}{non_english}{0}
\chexdefskip{settings_mimic_constant-normal}{P_n_both}{refused}{0}
\chexdefskip{settings_mimic_constant-normal}{P_n_both}{truncations}{0}
\chexdefskip{settings_mimic_constant-normal}{P_n_findings}{empty}{0}
\chexdefskip{settings_mimic_constant-normal}{P_n_findings}{empty_target}{921}
\chexdefskip{settings_mimic_constant-normal}{P_n_findings}{generated}{2348}
\chexdefskip{settings_mimic_constant-normal}{P_n_findings}{length_truncated}{0}
\chexdefskip{settings_mimic_constant-normal}{P_n_findings}{non_english}{0}
\chexdefskip{settings_mimic_constant-normal}{P_n_findings}{refused}{0}
\chexdefskip{settings_mimic_constant-normal}{P_n_findings}{truncations}{0}
\chexdefskip{settings_mimic_constant-normal}{P_n_impression}{empty}{0}
\chexdefskip{settings_mimic_constant-normal}{P_n_impression}{empty_target}{1045}
\chexdefskip{settings_mimic_constant-normal}{P_n_impression}{generated}{2224}
\chexdefskip{settings_mimic_constant-normal}{P_n_impression}{length_truncated}{0}
\chexdefskip{settings_mimic_constant-normal}{P_n_impression}{non_english}{0}
\chexdefskip{settings_mimic_constant-normal}{P_n_impression}{refused}{0}
\chexdefskip{settings_mimic_constant-normal}{P_n_impression}{truncations}{0}
\chexdefskip{settings_mimic_copy-indication}{B_f_both}{empty}{1624}
\chexdefskip{settings_mimic_copy-indication}{B_f_both}{empty_target}{1642}
\chexdefskip{settings_mimic_copy-indication}{B_f_both}{generated}{1624}
\chexdefskip{settings_mimic_copy-indication}{B_f_both}{length_truncated}{0}
\chexdefskip{settings_mimic_copy-indication}{B_f_both}{no_frontal}{3}
\chexdefskip{settings_mimic_copy-indication}{B_f_both}{non_english}{0}
\chexdefskip{settings_mimic_copy-indication}{B_f_both}{refused}{0}
\chexdefskip{settings_mimic_copy-indication}{B_f_both}{truncations}{0}
\chexdefskip{settings_mimic_copy-indication}{B_f_findings}{empty}{2346}
\chexdefskip{settings_mimic_copy-indication}{B_f_findings}{empty_target}{920}
\chexdefskip{settings_mimic_copy-indication}{B_f_findings}{generated}{2346}
\chexdefskip{settings_mimic_copy-indication}{B_f_findings}{length_truncated}{0}
\chexdefskip{settings_mimic_copy-indication}{B_f_findings}{no_frontal}{3}
\chexdefskip{settings_mimic_copy-indication}{B_f_findings}{non_english}{0}
\chexdefskip{settings_mimic_copy-indication}{B_f_findings}{refused}{0}
\chexdefskip{settings_mimic_copy-indication}{B_f_findings}{truncations}{0}
\chexdefskip{settings_mimic_copy-indication}{B_f_impression}{empty}{2223}
\chexdefskip{settings_mimic_copy-indication}{B_f_impression}{empty_target}{1043}
\chexdefskip{settings_mimic_copy-indication}{B_f_impression}{generated}{2223}
\chexdefskip{settings_mimic_copy-indication}{B_f_impression}{length_truncated}{0}
\chexdefskip{settings_mimic_copy-indication}{B_f_impression}{no_frontal}{3}
\chexdefskip{settings_mimic_copy-indication}{B_f_impression}{non_english}{0}
\chexdefskip{settings_mimic_copy-indication}{B_f_impression}{refused}{0}
\chexdefskip{settings_mimic_copy-indication}{B_f_impression}{truncations}{0}
\chexdefskip{settings_mimic_copy-indication}{P_f_both}{empty}{116}
\chexdefskip{settings_mimic_copy-indication}{P_f_both}{empty_target}{1642}
\chexdefskip{settings_mimic_copy-indication}{P_f_both}{generated}{1624}
\chexdefskip{settings_mimic_copy-indication}{P_f_both}{length_truncated}{0}
\chexdefskip{settings_mimic_copy-indication}{P_f_both}{no_frontal}{3}
\chexdefskip{settings_mimic_copy-indication}{P_f_both}{non_english}{0}
\chexdefskip{settings_mimic_copy-indication}{P_f_both}{refused}{0}
\chexdefskip{settings_mimic_copy-indication}{P_f_both}{truncations}{0}
\chexdefskip{settings_mimic_copy-indication}{P_f_findings}{empty}{229}
\chexdefskip{settings_mimic_copy-indication}{P_f_findings}{empty_target}{920}
\chexdefskip{settings_mimic_copy-indication}{P_f_findings}{generated}{2346}
\chexdefskip{settings_mimic_copy-indication}{P_f_findings}{length_truncated}{0}
\chexdefskip{settings_mimic_copy-indication}{P_f_findings}{no_frontal}{3}
\chexdefskip{settings_mimic_copy-indication}{P_f_findings}{non_english}{0}
\chexdefskip{settings_mimic_copy-indication}{P_f_findings}{refused}{0}
\chexdefskip{settings_mimic_copy-indication}{P_f_findings}{truncations}{0}
\chexdefskip{settings_mimic_copy-indication}{P_f_impression}{empty}{146}
\chexdefskip{settings_mimic_copy-indication}{P_f_impression}{empty_target}{1043}
\chexdefskip{settings_mimic_copy-indication}{P_f_impression}{generated}{2223}
\chexdefskip{settings_mimic_copy-indication}{P_f_impression}{length_truncated}{0}
\chexdefskip{settings_mimic_copy-indication}{P_f_impression}{no_frontal}{3}
\chexdefskip{settings_mimic_copy-indication}{P_f_impression}{non_english}{0}
\chexdefskip{settings_mimic_copy-indication}{P_f_impression}{refused}{0}
\chexdefskip{settings_mimic_copy-indication}{P_f_impression}{truncations}{0}
\chexdefskip{settings_mimic_copy-indication}{P_n_both}{empty}{116}
\chexdefskip{settings_mimic_copy-indication}{P_n_both}{empty_target}{1644}
\chexdefskip{settings_mimic_copy-indication}{P_n_both}{generated}{1625}
\chexdefskip{settings_mimic_copy-indication}{P_n_both}{length_truncated}{0}
\chexdefskip{settings_mimic_copy-indication}{P_n_both}{non_english}{0}
\chexdefskip{settings_mimic_copy-indication}{P_n_both}{refused}{0}
\chexdefskip{settings_mimic_copy-indication}{P_n_both}{truncations}{0}
\chexdefskip{settings_mimic_copy-indication}{P_n_findings}{empty}{229}
\chexdefskip{settings_mimic_copy-indication}{P_n_findings}{empty_target}{921}
\chexdefskip{settings_mimic_copy-indication}{P_n_findings}{generated}{2348}
\chexdefskip{settings_mimic_copy-indication}{P_n_findings}{length_truncated}{0}
\chexdefskip{settings_mimic_copy-indication}{P_n_findings}{non_english}{0}
\chexdefskip{settings_mimic_copy-indication}{P_n_findings}{refused}{0}
\chexdefskip{settings_mimic_copy-indication}{P_n_findings}{truncations}{0}
\chexdefskip{settings_mimic_copy-indication}{P_n_impression}{empty}{146}
\chexdefskip{settings_mimic_copy-indication}{P_n_impression}{empty_target}{1045}
\chexdefskip{settings_mimic_copy-indication}{P_n_impression}{generated}{2224}
\chexdefskip{settings_mimic_copy-indication}{P_n_impression}{length_truncated}{0}
\chexdefskip{settings_mimic_copy-indication}{P_n_impression}{non_english}{0}
\chexdefskip{settings_mimic_copy-indication}{P_n_impression}{refused}{0}
\chexdefskip{settings_mimic_copy-indication}{P_n_impression}{truncations}{0}
\chexdefskip{settings_mimic_copy-prior}{B_f_both}{empty}{1624}
\chexdefskip{settings_mimic_copy-prior}{B_f_both}{empty_target}{1642}
\chexdefskip{settings_mimic_copy-prior}{B_f_both}{generated}{1624}
\chexdefskip{settings_mimic_copy-prior}{B_f_both}{length_truncated}{0}
\chexdefskip{settings_mimic_copy-prior}{B_f_both}{no_frontal}{3}
\chexdefskip{settings_mimic_copy-prior}{B_f_both}{non_english}{0}
\chexdefskip{settings_mimic_copy-prior}{B_f_both}{refused}{0}
\chexdefskip{settings_mimic_copy-prior}{B_f_both}{truncations}{0}
\chexdefskip{settings_mimic_copy-prior}{B_f_findings}{empty}{2346}
\chexdefskip{settings_mimic_copy-prior}{B_f_findings}{empty_target}{920}
\chexdefskip{settings_mimic_copy-prior}{B_f_findings}{generated}{2346}
\chexdefskip{settings_mimic_copy-prior}{B_f_findings}{length_truncated}{0}
\chexdefskip{settings_mimic_copy-prior}{B_f_findings}{no_frontal}{3}
\chexdefskip{settings_mimic_copy-prior}{B_f_findings}{non_english}{0}
\chexdefskip{settings_mimic_copy-prior}{B_f_findings}{refused}{0}
\chexdefskip{settings_mimic_copy-prior}{B_f_findings}{truncations}{0}
\chexdefskip{settings_mimic_copy-prior}{B_f_impression}{empty}{2223}
\chexdefskip{settings_mimic_copy-prior}{B_f_impression}{empty_target}{1043}
\chexdefskip{settings_mimic_copy-prior}{B_f_impression}{generated}{2223}
\chexdefskip{settings_mimic_copy-prior}{B_f_impression}{length_truncated}{0}
\chexdefskip{settings_mimic_copy-prior}{B_f_impression}{no_frontal}{3}
\chexdefskip{settings_mimic_copy-prior}{B_f_impression}{non_english}{0}
\chexdefskip{settings_mimic_copy-prior}{B_f_impression}{refused}{0}
\chexdefskip{settings_mimic_copy-prior}{B_f_impression}{truncations}{0}
\chexdefskip{settings_mimic_copy-prior}{P_f_both}{empty}{1624}
\chexdefskip{settings_mimic_copy-prior}{P_f_both}{empty_target}{1642}
\chexdefskip{settings_mimic_copy-prior}{P_f_both}{generated}{1624}
\chexdefskip{settings_mimic_copy-prior}{P_f_both}{length_truncated}{0}
\chexdefskip{settings_mimic_copy-prior}{P_f_both}{no_frontal}{3}
\chexdefskip{settings_mimic_copy-prior}{P_f_both}{non_english}{0}
\chexdefskip{settings_mimic_copy-prior}{P_f_both}{refused}{0}
\chexdefskip{settings_mimic_copy-prior}{P_f_both}{truncations}{0}
\chexdefskip{settings_mimic_copy-prior}{P_f_findings}{empty}{2346}
\chexdefskip{settings_mimic_copy-prior}{P_f_findings}{empty_target}{920}
\chexdefskip{settings_mimic_copy-prior}{P_f_findings}{generated}{2346}
\chexdefskip{settings_mimic_copy-prior}{P_f_findings}{length_truncated}{0}
\chexdefskip{settings_mimic_copy-prior}{P_f_findings}{no_frontal}{3}
\chexdefskip{settings_mimic_copy-prior}{P_f_findings}{non_english}{0}
\chexdefskip{settings_mimic_copy-prior}{P_f_findings}{refused}{0}
\chexdefskip{settings_mimic_copy-prior}{P_f_findings}{truncations}{0}
\chexdefskip{settings_mimic_copy-prior}{P_f_impression}{empty}{2223}
\chexdefskip{settings_mimic_copy-prior}{P_f_impression}{empty_target}{1043}
\chexdefskip{settings_mimic_copy-prior}{P_f_impression}{generated}{2223}
\chexdefskip{settings_mimic_copy-prior}{P_f_impression}{length_truncated}{0}
\chexdefskip{settings_mimic_copy-prior}{P_f_impression}{no_frontal}{3}
\chexdefskip{settings_mimic_copy-prior}{P_f_impression}{non_english}{0}
\chexdefskip{settings_mimic_copy-prior}{P_f_impression}{refused}{0}
\chexdefskip{settings_mimic_copy-prior}{P_f_impression}{truncations}{0}
\chexdefskip{settings_mimic_copy-prior}{P_n_both}{empty}{1625}
\chexdefskip{settings_mimic_copy-prior}{P_n_both}{empty_target}{1644}
\chexdefskip{settings_mimic_copy-prior}{P_n_both}{generated}{1625}
\chexdefskip{settings_mimic_copy-prior}{P_n_both}{length_truncated}{0}
\chexdefskip{settings_mimic_copy-prior}{P_n_both}{non_english}{0}
\chexdefskip{settings_mimic_copy-prior}{P_n_both}{refused}{0}
\chexdefskip{settings_mimic_copy-prior}{P_n_both}{truncations}{0}
\chexdefskip{settings_mimic_copy-prior}{P_n_findings}{empty}{2348}
\chexdefskip{settings_mimic_copy-prior}{P_n_findings}{empty_target}{921}
\chexdefskip{settings_mimic_copy-prior}{P_n_findings}{generated}{2348}
\chexdefskip{settings_mimic_copy-prior}{P_n_findings}{length_truncated}{0}
\chexdefskip{settings_mimic_copy-prior}{P_n_findings}{non_english}{0}
\chexdefskip{settings_mimic_copy-prior}{P_n_findings}{refused}{0}
\chexdefskip{settings_mimic_copy-prior}{P_n_findings}{truncations}{0}
\chexdefskip{settings_mimic_copy-prior}{P_n_impression}{empty}{2224}
\chexdefskip{settings_mimic_copy-prior}{P_n_impression}{empty_target}{1045}
\chexdefskip{settings_mimic_copy-prior}{P_n_impression}{generated}{2224}
\chexdefskip{settings_mimic_copy-prior}{P_n_impression}{length_truncated}{0}
\chexdefskip{settings_mimic_copy-prior}{P_n_impression}{non_english}{0}
\chexdefskip{settings_mimic_copy-prior}{P_n_impression}{refused}{0}
\chexdefskip{settings_mimic_copy-prior}{P_n_impression}{truncations}{0}
\chexdefskip{settings_mimic_cxr-llava-v2}{B_f_both}{empty}{0}
\chexdefskip{settings_mimic_cxr-llava-v2}{B_f_both}{empty_target}{1642}
\chexdefskip{settings_mimic_cxr-llava-v2}{B_f_both}{generated}{1624}
\chexdefskip{settings_mimic_cxr-llava-v2}{B_f_both}{length_truncated}{0}
\chexdefskip{settings_mimic_cxr-llava-v2}{B_f_both}{no_frontal}{3}
\chexdefskip{settings_mimic_cxr-llava-v2}{B_f_both}{non_english}{0}
\chexdefskip{settings_mimic_cxr-llava-v2}{B_f_both}{refused}{0}
\chexdefskip{settings_mimic_cxr-llava-v2}{B_f_both}{truncations}{0}
\chexdefskip{settings_mimic_cxr-llava-v2}{B_f_findings}{empty}{0}
\chexdefskip{settings_mimic_cxr-llava-v2}{B_f_findings}{empty_target}{920}
\chexdefskip{settings_mimic_cxr-llava-v2}{B_f_findings}{generated}{2346}
\chexdefskip{settings_mimic_cxr-llava-v2}{B_f_findings}{length_truncated}{0}
\chexdefskip{settings_mimic_cxr-llava-v2}{B_f_findings}{no_frontal}{3}
\chexdefskip{settings_mimic_cxr-llava-v2}{B_f_findings}{non_english}{0}
\chexdefskip{settings_mimic_cxr-llava-v2}{B_f_findings}{refused}{0}
\chexdefskip{settings_mimic_cxr-llava-v2}{B_f_findings}{truncations}{0}
\chexdefskip{settings_mimic_cxr-llava-v2}{B_f_impression}{empty}{0}
\chexdefskip{settings_mimic_cxr-llava-v2}{B_f_impression}{empty_target}{1043}
\chexdefskip{settings_mimic_cxr-llava-v2}{B_f_impression}{generated}{2223}
\chexdefskip{settings_mimic_cxr-llava-v2}{B_f_impression}{length_truncated}{0}
\chexdefskip{settings_mimic_cxr-llava-v2}{B_f_impression}{no_frontal}{3}
\chexdefskip{settings_mimic_cxr-llava-v2}{B_f_impression}{non_english}{0}
\chexdefskip{settings_mimic_cxr-llava-v2}{B_f_impression}{refused}{0}
\chexdefskip{settings_mimic_cxr-llava-v2}{B_f_impression}{truncations}{0}
\chexdefskip{settings_mimic_cxr-llava-v2}{P_f_both}{empty}{0}
\chexdefskip{settings_mimic_cxr-llava-v2}{P_f_both}{empty_target}{1642}
\chexdefskip{settings_mimic_cxr-llava-v2}{P_f_both}{generated}{1624}
\chexdefskip{settings_mimic_cxr-llava-v2}{P_f_both}{length_truncated}{0}
\chexdefskip{settings_mimic_cxr-llava-v2}{P_f_both}{no_frontal}{3}
\chexdefskip{settings_mimic_cxr-llava-v2}{P_f_both}{non_english}{0}
\chexdefskip{settings_mimic_cxr-llava-v2}{P_f_both}{refused}{0}
\chexdefskip{settings_mimic_cxr-llava-v2}{P_f_both}{truncations}{0}
\chexdefskip{settings_mimic_cxr-llava-v2}{P_f_findings}{empty}{0}
\chexdefskip{settings_mimic_cxr-llava-v2}{P_f_findings}{empty_target}{920}
\chexdefskip{settings_mimic_cxr-llava-v2}{P_f_findings}{generated}{2346}
\chexdefskip{settings_mimic_cxr-llava-v2}{P_f_findings}{length_truncated}{0}
\chexdefskip{settings_mimic_cxr-llava-v2}{P_f_findings}{no_frontal}{3}
\chexdefskip{settings_mimic_cxr-llava-v2}{P_f_findings}{non_english}{0}
\chexdefskip{settings_mimic_cxr-llava-v2}{P_f_findings}{refused}{0}
\chexdefskip{settings_mimic_cxr-llava-v2}{P_f_findings}{truncations}{0}
\chexdefskip{settings_mimic_cxr-llava-v2}{P_f_impression}{empty}{0}
\chexdefskip{settings_mimic_cxr-llava-v2}{P_f_impression}{empty_target}{1043}
\chexdefskip{settings_mimic_cxr-llava-v2}{P_f_impression}{generated}{2223}
\chexdefskip{settings_mimic_cxr-llava-v2}{P_f_impression}{length_truncated}{0}
\chexdefskip{settings_mimic_cxr-llava-v2}{P_f_impression}{no_frontal}{3}
\chexdefskip{settings_mimic_cxr-llava-v2}{P_f_impression}{non_english}{0}
\chexdefskip{settings_mimic_cxr-llava-v2}{P_f_impression}{refused}{0}
\chexdefskip{settings_mimic_cxr-llava-v2}{P_f_impression}{truncations}{0}
\chexdefskip{settings_mimic_cxr-llava-v2}{P_n_both}{empty}{0}
\chexdefskip{settings_mimic_cxr-llava-v2}{P_n_both}{generated}{0}
\chexdefskip{settings_mimic_cxr-llava-v2}{P_n_both}{length_truncated}{0}
\chexdefskip{settings_mimic_cxr-llava-v2}{P_n_both}{non_english}{0}
\chexdefskip{settings_mimic_cxr-llava-v2}{P_n_both}{refused}{0}
\chexdefskip{settings_mimic_cxr-llava-v2}{P_n_both}{text_only_unsupported}{3269}
\chexdefskip{settings_mimic_cxr-llava-v2}{P_n_both}{truncations}{0}
\chexdefskip{settings_mimic_cxr-llava-v2}{P_n_findings}{empty}{0}
\chexdefskip{settings_mimic_cxr-llava-v2}{P_n_findings}{generated}{0}
\chexdefskip{settings_mimic_cxr-llava-v2}{P_n_findings}{length_truncated}{0}
\chexdefskip{settings_mimic_cxr-llava-v2}{P_n_findings}{non_english}{0}
\chexdefskip{settings_mimic_cxr-llava-v2}{P_n_findings}{refused}{0}
\chexdefskip{settings_mimic_cxr-llava-v2}{P_n_findings}{text_only_unsupported}{3269}
\chexdefskip{settings_mimic_cxr-llava-v2}{P_n_findings}{truncations}{0}
\chexdefskip{settings_mimic_cxr-llava-v2}{P_n_impression}{empty}{0}
\chexdefskip{settings_mimic_cxr-llava-v2}{P_n_impression}{generated}{0}
\chexdefskip{settings_mimic_cxr-llava-v2}{P_n_impression}{length_truncated}{0}
\chexdefskip{settings_mimic_cxr-llava-v2}{P_n_impression}{non_english}{0}
\chexdefskip{settings_mimic_cxr-llava-v2}{P_n_impression}{refused}{0}
\chexdefskip{settings_mimic_cxr-llava-v2}{P_n_impression}{text_only_unsupported}{3269}
\chexdefskip{settings_mimic_cxr-llava-v2}{P_n_impression}{truncations}{0}
\chexdefskip{settings_mimic_empty}{B_f_both}{empty}{1624}
\chexdefskip{settings_mimic_empty}{B_f_both}{empty_target}{1642}
\chexdefskip{settings_mimic_empty}{B_f_both}{generated}{1624}
\chexdefskip{settings_mimic_empty}{B_f_both}{length_truncated}{0}
\chexdefskip{settings_mimic_empty}{B_f_both}{no_frontal}{3}
\chexdefskip{settings_mimic_empty}{B_f_both}{non_english}{0}
\chexdefskip{settings_mimic_empty}{B_f_both}{refused}{0}
\chexdefskip{settings_mimic_empty}{B_f_both}{truncations}{0}
\chexdefskip{settings_mimic_empty}{B_f_findings}{empty}{2346}
\chexdefskip{settings_mimic_empty}{B_f_findings}{empty_target}{920}
\chexdefskip{settings_mimic_empty}{B_f_findings}{generated}{2346}
\chexdefskip{settings_mimic_empty}{B_f_findings}{length_truncated}{0}
\chexdefskip{settings_mimic_empty}{B_f_findings}{no_frontal}{3}
\chexdefskip{settings_mimic_empty}{B_f_findings}{non_english}{0}
\chexdefskip{settings_mimic_empty}{B_f_findings}{refused}{0}
\chexdefskip{settings_mimic_empty}{B_f_findings}{truncations}{0}
\chexdefskip{settings_mimic_empty}{B_f_impression}{empty}{2223}
\chexdefskip{settings_mimic_empty}{B_f_impression}{empty_target}{1043}
\chexdefskip{settings_mimic_empty}{B_f_impression}{generated}{2223}
\chexdefskip{settings_mimic_empty}{B_f_impression}{length_truncated}{0}
\chexdefskip{settings_mimic_empty}{B_f_impression}{no_frontal}{3}
\chexdefskip{settings_mimic_empty}{B_f_impression}{non_english}{0}
\chexdefskip{settings_mimic_empty}{B_f_impression}{refused}{0}
\chexdefskip{settings_mimic_empty}{B_f_impression}{truncations}{0}
\chexdefskip{settings_mimic_empty}{P_f_both}{empty}{1624}
\chexdefskip{settings_mimic_empty}{P_f_both}{empty_target}{1642}
\chexdefskip{settings_mimic_empty}{P_f_both}{generated}{1624}
\chexdefskip{settings_mimic_empty}{P_f_both}{length_truncated}{0}
\chexdefskip{settings_mimic_empty}{P_f_both}{no_frontal}{3}
\chexdefskip{settings_mimic_empty}{P_f_both}{non_english}{0}
\chexdefskip{settings_mimic_empty}{P_f_both}{refused}{0}
\chexdefskip{settings_mimic_empty}{P_f_both}{truncations}{0}
\chexdefskip{settings_mimic_empty}{P_f_findings}{empty}{2346}
\chexdefskip{settings_mimic_empty}{P_f_findings}{empty_target}{920}
\chexdefskip{settings_mimic_empty}{P_f_findings}{generated}{2346}
\chexdefskip{settings_mimic_empty}{P_f_findings}{length_truncated}{0}
\chexdefskip{settings_mimic_empty}{P_f_findings}{no_frontal}{3}
\chexdefskip{settings_mimic_empty}{P_f_findings}{non_english}{0}
\chexdefskip{settings_mimic_empty}{P_f_findings}{refused}{0}
\chexdefskip{settings_mimic_empty}{P_f_findings}{truncations}{0}
\chexdefskip{settings_mimic_empty}{P_f_impression}{empty}{2223}
\chexdefskip{settings_mimic_empty}{P_f_impression}{empty_target}{1043}
\chexdefskip{settings_mimic_empty}{P_f_impression}{generated}{2223}
\chexdefskip{settings_mimic_empty}{P_f_impression}{length_truncated}{0}
\chexdefskip{settings_mimic_empty}{P_f_impression}{no_frontal}{3}
\chexdefskip{settings_mimic_empty}{P_f_impression}{non_english}{0}
\chexdefskip{settings_mimic_empty}{P_f_impression}{refused}{0}
\chexdefskip{settings_mimic_empty}{P_f_impression}{truncations}{0}
\chexdefskip{settings_mimic_empty}{P_n_both}{empty}{1625}
\chexdefskip{settings_mimic_empty}{P_n_both}{empty_target}{1644}
\chexdefskip{settings_mimic_empty}{P_n_both}{generated}{1625}
\chexdefskip{settings_mimic_empty}{P_n_both}{length_truncated}{0}
\chexdefskip{settings_mimic_empty}{P_n_both}{non_english}{0}
\chexdefskip{settings_mimic_empty}{P_n_both}{refused}{0}
\chexdefskip{settings_mimic_empty}{P_n_both}{truncations}{0}
\chexdefskip{settings_mimic_empty}{P_n_findings}{empty}{2348}
\chexdefskip{settings_mimic_empty}{P_n_findings}{empty_target}{921}
\chexdefskip{settings_mimic_empty}{P_n_findings}{generated}{2348}
\chexdefskip{settings_mimic_empty}{P_n_findings}{length_truncated}{0}
\chexdefskip{settings_mimic_empty}{P_n_findings}{non_english}{0}
\chexdefskip{settings_mimic_empty}{P_n_findings}{refused}{0}
\chexdefskip{settings_mimic_empty}{P_n_findings}{truncations}{0}
\chexdefskip{settings_mimic_empty}{P_n_impression}{empty}{2224}
\chexdefskip{settings_mimic_empty}{P_n_impression}{empty_target}{1045}
\chexdefskip{settings_mimic_empty}{P_n_impression}{generated}{2224}
\chexdefskip{settings_mimic_empty}{P_n_impression}{length_truncated}{0}
\chexdefskip{settings_mimic_empty}{P_n_impression}{non_english}{0}
\chexdefskip{settings_mimic_empty}{P_n_impression}{refused}{0}
\chexdefskip{settings_mimic_empty}{P_n_impression}{truncations}{0}
\chexdefskip{settings_mimic_internvl3-8b}{B_f_both}{empty}{0}
\chexdefskip{settings_mimic_internvl3-8b}{B_f_both}{empty_target}{1642}
\chexdefskip{settings_mimic_internvl3-8b}{B_f_both}{generated}{1624}
\chexdefskip{settings_mimic_internvl3-8b}{B_f_both}{length_truncated}{0}
\chexdefskip{settings_mimic_internvl3-8b}{B_f_both}{no_frontal}{3}
\chexdefskip{settings_mimic_internvl3-8b}{B_f_both}{non_english}{0}
\chexdefskip{settings_mimic_internvl3-8b}{B_f_both}{refused}{0}
\chexdefskip{settings_mimic_internvl3-8b}{B_f_both}{truncations}{0}
\chexdefskip{settings_mimic_internvl3-8b}{B_f_findings}{empty}{0}
\chexdefskip{settings_mimic_internvl3-8b}{B_f_findings}{empty_target}{920}
\chexdefskip{settings_mimic_internvl3-8b}{B_f_findings}{generated}{2346}
\chexdefskip{settings_mimic_internvl3-8b}{B_f_findings}{length_truncated}{0}
\chexdefskip{settings_mimic_internvl3-8b}{B_f_findings}{no_frontal}{3}
\chexdefskip{settings_mimic_internvl3-8b}{B_f_findings}{non_english}{0}
\chexdefskip{settings_mimic_internvl3-8b}{B_f_findings}{refused}{0}
\chexdefskip{settings_mimic_internvl3-8b}{B_f_findings}{truncations}{0}
\chexdefskip{settings_mimic_internvl3-8b}{B_f_impression}{empty}{0}
\chexdefskip{settings_mimic_internvl3-8b}{B_f_impression}{empty_target}{1043}
\chexdefskip{settings_mimic_internvl3-8b}{B_f_impression}{generated}{2223}
\chexdefskip{settings_mimic_internvl3-8b}{B_f_impression}{length_truncated}{0}
\chexdefskip{settings_mimic_internvl3-8b}{B_f_impression}{no_frontal}{3}
\chexdefskip{settings_mimic_internvl3-8b}{B_f_impression}{non_english}{0}
\chexdefskip{settings_mimic_internvl3-8b}{B_f_impression}{refused}{0}
\chexdefskip{settings_mimic_internvl3-8b}{B_f_impression}{truncations}{0}
\chexdefskip{settings_mimic_internvl3-8b}{P_f_both}{empty}{0}
\chexdefskip{settings_mimic_internvl3-8b}{P_f_both}{empty_target}{1642}
\chexdefskip{settings_mimic_internvl3-8b}{P_f_both}{generated}{1624}
\chexdefskip{settings_mimic_internvl3-8b}{P_f_both}{length_truncated}{0}
\chexdefskip{settings_mimic_internvl3-8b}{P_f_both}{no_frontal}{3}
\chexdefskip{settings_mimic_internvl3-8b}{P_f_both}{non_english}{0}
\chexdefskip{settings_mimic_internvl3-8b}{P_f_both}{refused}{0}
\chexdefskip{settings_mimic_internvl3-8b}{P_f_both}{truncations}{0}
\chexdefskip{settings_mimic_internvl3-8b}{P_f_findings}{empty}{0}
\chexdefskip{settings_mimic_internvl3-8b}{P_f_findings}{empty_target}{920}
\chexdefskip{settings_mimic_internvl3-8b}{P_f_findings}{generated}{2346}
\chexdefskip{settings_mimic_internvl3-8b}{P_f_findings}{length_truncated}{0}
\chexdefskip{settings_mimic_internvl3-8b}{P_f_findings}{no_frontal}{3}
\chexdefskip{settings_mimic_internvl3-8b}{P_f_findings}{non_english}{0}
\chexdefskip{settings_mimic_internvl3-8b}{P_f_findings}{refused}{0}
\chexdefskip{settings_mimic_internvl3-8b}{P_f_findings}{truncations}{0}
\chexdefskip{settings_mimic_internvl3-8b}{P_f_impression}{empty}{0}
\chexdefskip{settings_mimic_internvl3-8b}{P_f_impression}{empty_target}{1043}
\chexdefskip{settings_mimic_internvl3-8b}{P_f_impression}{generated}{2223}
\chexdefskip{settings_mimic_internvl3-8b}{P_f_impression}{length_truncated}{0}
\chexdefskip{settings_mimic_internvl3-8b}{P_f_impression}{no_frontal}{3}
\chexdefskip{settings_mimic_internvl3-8b}{P_f_impression}{non_english}{0}
\chexdefskip{settings_mimic_internvl3-8b}{P_f_impression}{refused}{0}
\chexdefskip{settings_mimic_internvl3-8b}{P_f_impression}{truncations}{0}
\chexdefskip{settings_mimic_internvl3-8b}{P_n_both}{empty}{0}
\chexdefskip{settings_mimic_internvl3-8b}{P_n_both}{empty_target}{1644}
\chexdefskip{settings_mimic_internvl3-8b}{P_n_both}{generated}{1625}
\chexdefskip{settings_mimic_internvl3-8b}{P_n_both}{length_truncated}{0}
\chexdefskip{settings_mimic_internvl3-8b}{P_n_both}{non_english}{0}
\chexdefskip{settings_mimic_internvl3-8b}{P_n_both}{refused}{0}
\chexdefskip{settings_mimic_internvl3-8b}{P_n_both}{truncations}{0}
\chexdefskip{settings_mimic_internvl3-8b}{P_n_findings}{empty}{0}
\chexdefskip{settings_mimic_internvl3-8b}{P_n_findings}{empty_target}{921}
\chexdefskip{settings_mimic_internvl3-8b}{P_n_findings}{generated}{2348}
\chexdefskip{settings_mimic_internvl3-8b}{P_n_findings}{length_truncated}{0}
\chexdefskip{settings_mimic_internvl3-8b}{P_n_findings}{non_english}{0}
\chexdefskip{settings_mimic_internvl3-8b}{P_n_findings}{refused}{1}
\chexdefskip{settings_mimic_internvl3-8b}{P_n_findings}{truncations}{0}
\chexdefskip{settings_mimic_internvl3-8b}{P_n_impression}{empty}{0}
\chexdefskip{settings_mimic_internvl3-8b}{P_n_impression}{empty_target}{1045}
\chexdefskip{settings_mimic_internvl3-8b}{P_n_impression}{generated}{2224}
\chexdefskip{settings_mimic_internvl3-8b}{P_n_impression}{length_truncated}{1}
\chexdefskip{settings_mimic_internvl3-8b}{P_n_impression}{non_english}{0}
\chexdefskip{settings_mimic_internvl3-8b}{P_n_impression}{refused}{1}
\chexdefskip{settings_mimic_internvl3-8b}{P_n_impression}{truncations}{0}
\chexdefskip{settings_mimic_llava-med}{B_f_both}{empty}{0}
\chexdefskip{settings_mimic_llava-med}{B_f_both}{empty_target}{1642}
\chexdefskip{settings_mimic_llava-med}{B_f_both}{generated}{1624}
\chexdefskip{settings_mimic_llava-med}{B_f_both}{length_truncated}{0}
\chexdefskip{settings_mimic_llava-med}{B_f_both}{no_frontal}{3}
\chexdefskip{settings_mimic_llava-med}{B_f_both}{non_english}{0}
\chexdefskip{settings_mimic_llava-med}{B_f_both}{refused}{0}
\chexdefskip{settings_mimic_llava-med}{B_f_both}{truncations}{0}
\chexdefskip{settings_mimic_llava-med}{B_f_findings}{empty}{0}
\chexdefskip{settings_mimic_llava-med}{B_f_findings}{empty_target}{920}
\chexdefskip{settings_mimic_llava-med}{B_f_findings}{generated}{2346}
\chexdefskip{settings_mimic_llava-med}{B_f_findings}{length_truncated}{0}
\chexdefskip{settings_mimic_llava-med}{B_f_findings}{no_frontal}{3}
\chexdefskip{settings_mimic_llava-med}{B_f_findings}{non_english}{0}
\chexdefskip{settings_mimic_llava-med}{B_f_findings}{refused}{0}
\chexdefskip{settings_mimic_llava-med}{B_f_findings}{truncations}{0}
\chexdefskip{settings_mimic_llava-med}{B_f_impression}{empty}{0}
\chexdefskip{settings_mimic_llava-med}{B_f_impression}{empty_target}{1043}
\chexdefskip{settings_mimic_llava-med}{B_f_impression}{generated}{2223}
\chexdefskip{settings_mimic_llava-med}{B_f_impression}{length_truncated}{0}
\chexdefskip{settings_mimic_llava-med}{B_f_impression}{no_frontal}{3}
\chexdefskip{settings_mimic_llava-med}{B_f_impression}{non_english}{0}
\chexdefskip{settings_mimic_llava-med}{B_f_impression}{refused}{0}
\chexdefskip{settings_mimic_llava-med}{B_f_impression}{truncations}{0}
\chexdefskip{settings_mimic_llava-med}{P_f_both}{empty}{0}
\chexdefskip{settings_mimic_llava-med}{P_f_both}{empty_target}{1642}
\chexdefskip{settings_mimic_llava-med}{P_f_both}{generated}{1624}
\chexdefskip{settings_mimic_llava-med}{P_f_both}{length_truncated}{0}
\chexdefskip{settings_mimic_llava-med}{P_f_both}{no_frontal}{3}
\chexdefskip{settings_mimic_llava-med}{P_f_both}{non_english}{0}
\chexdefskip{settings_mimic_llava-med}{P_f_both}{refused}{0}
\chexdefskip{settings_mimic_llava-med}{P_f_both}{truncations}{0}
\chexdefskip{settings_mimic_llava-med}{P_f_findings}{empty}{58}
\chexdefskip{settings_mimic_llava-med}{P_f_findings}{empty_target}{920}
\chexdefskip{settings_mimic_llava-med}{P_f_findings}{generated}{2346}
\chexdefskip{settings_mimic_llava-med}{P_f_findings}{length_truncated}{0}
\chexdefskip{settings_mimic_llava-med}{P_f_findings}{no_frontal}{3}
\chexdefskip{settings_mimic_llava-med}{P_f_findings}{non_english}{0}
\chexdefskip{settings_mimic_llava-med}{P_f_findings}{refused}{0}
\chexdefskip{settings_mimic_llava-med}{P_f_findings}{truncations}{0}
\chexdefskip{settings_mimic_llava-med}{P_f_impression}{empty}{204}
\chexdefskip{settings_mimic_llava-med}{P_f_impression}{empty_target}{1043}
\chexdefskip{settings_mimic_llava-med}{P_f_impression}{generated}{2223}
\chexdefskip{settings_mimic_llava-med}{P_f_impression}{length_truncated}{0}
\chexdefskip{settings_mimic_llava-med}{P_f_impression}{no_frontal}{3}
\chexdefskip{settings_mimic_llava-med}{P_f_impression}{non_english}{0}
\chexdefskip{settings_mimic_llava-med}{P_f_impression}{refused}{2}
\chexdefskip{settings_mimic_llava-med}{P_f_impression}{truncations}{0}
\chexdefskip{settings_mimic_llava-med}{P_n_both}{empty}{0}
\chexdefskip{settings_mimic_llava-med}{P_n_both}{generated}{0}
\chexdefskip{settings_mimic_llava-med}{P_n_both}{length_truncated}{0}
\chexdefskip{settings_mimic_llava-med}{P_n_both}{non_english}{0}
\chexdefskip{settings_mimic_llava-med}{P_n_both}{refused}{0}
\chexdefskip{settings_mimic_llava-med}{P_n_both}{text_only_unsupported}{3269}
\chexdefskip{settings_mimic_llava-med}{P_n_both}{truncations}{0}
\chexdefskip{settings_mimic_llava-med}{P_n_findings}{empty}{0}
\chexdefskip{settings_mimic_llava-med}{P_n_findings}{generated}{0}
\chexdefskip{settings_mimic_llava-med}{P_n_findings}{length_truncated}{0}
\chexdefskip{settings_mimic_llava-med}{P_n_findings}{non_english}{0}
\chexdefskip{settings_mimic_llava-med}{P_n_findings}{refused}{0}
\chexdefskip{settings_mimic_llava-med}{P_n_findings}{text_only_unsupported}{3269}
\chexdefskip{settings_mimic_llava-med}{P_n_findings}{truncations}{0}
\chexdefskip{settings_mimic_llava-med}{P_n_impression}{empty}{0}
\chexdefskip{settings_mimic_llava-med}{P_n_impression}{generated}{0}
\chexdefskip{settings_mimic_llava-med}{P_n_impression}{length_truncated}{0}
\chexdefskip{settings_mimic_llava-med}{P_n_impression}{non_english}{0}
\chexdefskip{settings_mimic_llava-med}{P_n_impression}{refused}{0}
\chexdefskip{settings_mimic_llava-med}{P_n_impression}{text_only_unsupported}{3269}
\chexdefskip{settings_mimic_llava-med}{P_n_impression}{truncations}{0}
\chexdefskip{settings_mimic_llava-rad}{B_f_both}{empty}{0}
\chexdefskip{settings_mimic_llava-rad}{B_f_both}{empty_target}{1642}
\chexdefskip{settings_mimic_llava-rad}{B_f_both}{generated}{1624}
\chexdefskip{settings_mimic_llava-rad}{B_f_both}{length_truncated}{0}
\chexdefskip{settings_mimic_llava-rad}{B_f_both}{no_frontal}{3}
\chexdefskip{settings_mimic_llava-rad}{B_f_both}{non_english}{0}
\chexdefskip{settings_mimic_llava-rad}{B_f_both}{refused}{0}
\chexdefskip{settings_mimic_llava-rad}{B_f_both}{truncations}{0}
\chexdefskip{settings_mimic_llava-rad}{B_f_findings}{empty}{0}
\chexdefskip{settings_mimic_llava-rad}{B_f_findings}{empty_target}{920}
\chexdefskip{settings_mimic_llava-rad}{B_f_findings}{generated}{2346}
\chexdefskip{settings_mimic_llava-rad}{B_f_findings}{length_truncated}{0}
\chexdefskip{settings_mimic_llava-rad}{B_f_findings}{no_frontal}{3}
\chexdefskip{settings_mimic_llava-rad}{B_f_findings}{non_english}{0}
\chexdefskip{settings_mimic_llava-rad}{B_f_findings}{refused}{0}
\chexdefskip{settings_mimic_llava-rad}{B_f_findings}{truncations}{0}
\chexdefskip{settings_mimic_llava-rad}{B_f_impression}{empty}{0}
\chexdefskip{settings_mimic_llava-rad}{B_f_impression}{empty_target}{1043}
\chexdefskip{settings_mimic_llava-rad}{B_f_impression}{generated}{2223}
\chexdefskip{settings_mimic_llava-rad}{B_f_impression}{length_truncated}{0}
\chexdefskip{settings_mimic_llava-rad}{B_f_impression}{no_frontal}{3}
\chexdefskip{settings_mimic_llava-rad}{B_f_impression}{non_english}{0}
\chexdefskip{settings_mimic_llava-rad}{B_f_impression}{refused}{0}
\chexdefskip{settings_mimic_llava-rad}{B_f_impression}{truncations}{0}
\chexdefskip{settings_mimic_llava-rad}{P_f_both}{empty}{0}
\chexdefskip{settings_mimic_llava-rad}{P_f_both}{empty_target}{1642}
\chexdefskip{settings_mimic_llava-rad}{P_f_both}{generated}{1624}
\chexdefskip{settings_mimic_llava-rad}{P_f_both}{length_truncated}{0}
\chexdefskip{settings_mimic_llava-rad}{P_f_both}{no_frontal}{3}
\chexdefskip{settings_mimic_llava-rad}{P_f_both}{non_english}{0}
\chexdefskip{settings_mimic_llava-rad}{P_f_both}{refused}{0}
\chexdefskip{settings_mimic_llava-rad}{P_f_both}{truncations}{0}
\chexdefskip{settings_mimic_llava-rad}{P_f_findings}{empty}{0}
\chexdefskip{settings_mimic_llava-rad}{P_f_findings}{empty_target}{920}
\chexdefskip{settings_mimic_llava-rad}{P_f_findings}{generated}{2346}
\chexdefskip{settings_mimic_llava-rad}{P_f_findings}{length_truncated}{0}
\chexdefskip{settings_mimic_llava-rad}{P_f_findings}{no_frontal}{3}
\chexdefskip{settings_mimic_llava-rad}{P_f_findings}{non_english}{0}
\chexdefskip{settings_mimic_llava-rad}{P_f_findings}{refused}{0}
\chexdefskip{settings_mimic_llava-rad}{P_f_findings}{truncations}{0}
\chexdefskip{settings_mimic_llava-rad}{P_f_impression}{empty}{0}
\chexdefskip{settings_mimic_llava-rad}{P_f_impression}{empty_target}{1043}
\chexdefskip{settings_mimic_llava-rad}{P_f_impression}{generated}{2223}
\chexdefskip{settings_mimic_llava-rad}{P_f_impression}{length_truncated}{0}
\chexdefskip{settings_mimic_llava-rad}{P_f_impression}{no_frontal}{3}
\chexdefskip{settings_mimic_llava-rad}{P_f_impression}{non_english}{0}
\chexdefskip{settings_mimic_llava-rad}{P_f_impression}{refused}{0}
\chexdefskip{settings_mimic_llava-rad}{P_f_impression}{truncations}{0}
\chexdefskip{settings_mimic_llava-rad}{P_n_both}{empty}{0}
\chexdefskip{settings_mimic_llava-rad}{P_n_both}{generated}{0}
\chexdefskip{settings_mimic_llava-rad}{P_n_both}{length_truncated}{0}
\chexdefskip{settings_mimic_llava-rad}{P_n_both}{non_english}{0}
\chexdefskip{settings_mimic_llava-rad}{P_n_both}{refused}{0}
\chexdefskip{settings_mimic_llava-rad}{P_n_both}{text_only_unsupported}{3269}
\chexdefskip{settings_mimic_llava-rad}{P_n_both}{truncations}{0}
\chexdefskip{settings_mimic_llava-rad}{P_n_findings}{empty}{0}
\chexdefskip{settings_mimic_llava-rad}{P_n_findings}{generated}{0}
\chexdefskip{settings_mimic_llava-rad}{P_n_findings}{length_truncated}{0}
\chexdefskip{settings_mimic_llava-rad}{P_n_findings}{non_english}{0}
\chexdefskip{settings_mimic_llava-rad}{P_n_findings}{refused}{0}
\chexdefskip{settings_mimic_llava-rad}{P_n_findings}{text_only_unsupported}{3269}
\chexdefskip{settings_mimic_llava-rad}{P_n_findings}{truncations}{0}
\chexdefskip{settings_mimic_llava-rad}{P_n_impression}{empty}{0}
\chexdefskip{settings_mimic_llava-rad}{P_n_impression}{generated}{0}
\chexdefskip{settings_mimic_llava-rad}{P_n_impression}{length_truncated}{0}
\chexdefskip{settings_mimic_llava-rad}{P_n_impression}{non_english}{0}
\chexdefskip{settings_mimic_llava-rad}{P_n_impression}{refused}{0}
\chexdefskip{settings_mimic_llava-rad}{P_n_impression}{text_only_unsupported}{3269}
\chexdefskip{settings_mimic_llava-rad}{P_n_impression}{truncations}{0}
\chexdefskip{settings_mimic_maira-2}{B_f_both}{empty}{0}
\chexdefskip{settings_mimic_maira-2}{B_f_both}{empty_target}{1642}
\chexdefskip{settings_mimic_maira-2}{B_f_both}{generated}{1624}
\chexdefskip{settings_mimic_maira-2}{B_f_both}{length_truncated}{0}
\chexdefskip{settings_mimic_maira-2}{B_f_both}{no_frontal}{3}
\chexdefskip{settings_mimic_maira-2}{B_f_both}{non_english}{0}
\chexdefskip{settings_mimic_maira-2}{B_f_both}{refused}{0}
\chexdefskip{settings_mimic_maira-2}{B_f_both}{truncations}{0}
\chexdefskip{settings_mimic_maira-2}{B_f_findings}{empty}{0}
\chexdefskip{settings_mimic_maira-2}{B_f_findings}{empty_target}{920}
\chexdefskip{settings_mimic_maira-2}{B_f_findings}{generated}{2346}
\chexdefskip{settings_mimic_maira-2}{B_f_findings}{length_truncated}{0}
\chexdefskip{settings_mimic_maira-2}{B_f_findings}{no_frontal}{3}
\chexdefskip{settings_mimic_maira-2}{B_f_findings}{non_english}{0}
\chexdefskip{settings_mimic_maira-2}{B_f_findings}{refused}{0}
\chexdefskip{settings_mimic_maira-2}{B_f_findings}{truncations}{0}
\chexdefskip{settings_mimic_maira-2}{B_f_impression}{empty}{0}
\chexdefskip{settings_mimic_maira-2}{B_f_impression}{empty_target}{1043}
\chexdefskip{settings_mimic_maira-2}{B_f_impression}{generated}{2223}
\chexdefskip{settings_mimic_maira-2}{B_f_impression}{length_truncated}{0}
\chexdefskip{settings_mimic_maira-2}{B_f_impression}{no_frontal}{3}
\chexdefskip{settings_mimic_maira-2}{B_f_impression}{non_english}{0}
\chexdefskip{settings_mimic_maira-2}{B_f_impression}{refused}{0}
\chexdefskip{settings_mimic_maira-2}{B_f_impression}{truncations}{0}
\chexdefskip{settings_mimic_maira-2}{P_f_both}{empty}{0}
\chexdefskip{settings_mimic_maira-2}{P_f_both}{empty_target}{1642}
\chexdefskip{settings_mimic_maira-2}{P_f_both}{generated}{1624}
\chexdefskip{settings_mimic_maira-2}{P_f_both}{length_truncated}{0}
\chexdefskip{settings_mimic_maira-2}{P_f_both}{no_frontal}{3}
\chexdefskip{settings_mimic_maira-2}{P_f_both}{non_english}{0}
\chexdefskip{settings_mimic_maira-2}{P_f_both}{refused}{0}
\chexdefskip{settings_mimic_maira-2}{P_f_both}{truncations}{0}
\chexdefskip{settings_mimic_maira-2}{P_f_findings}{empty}{0}
\chexdefskip{settings_mimic_maira-2}{P_f_findings}{empty_target}{920}
\chexdefskip{settings_mimic_maira-2}{P_f_findings}{generated}{2346}
\chexdefskip{settings_mimic_maira-2}{P_f_findings}{length_truncated}{0}
\chexdefskip{settings_mimic_maira-2}{P_f_findings}{no_frontal}{3}
\chexdefskip{settings_mimic_maira-2}{P_f_findings}{non_english}{0}
\chexdefskip{settings_mimic_maira-2}{P_f_findings}{refused}{0}
\chexdefskip{settings_mimic_maira-2}{P_f_findings}{truncations}{0}
\chexdefskip{settings_mimic_maira-2}{P_f_impression}{empty}{0}
\chexdefskip{settings_mimic_maira-2}{P_f_impression}{empty_target}{1043}
\chexdefskip{settings_mimic_maira-2}{P_f_impression}{generated}{2223}
\chexdefskip{settings_mimic_maira-2}{P_f_impression}{length_truncated}{0}
\chexdefskip{settings_mimic_maira-2}{P_f_impression}{no_frontal}{3}
\chexdefskip{settings_mimic_maira-2}{P_f_impression}{non_english}{0}
\chexdefskip{settings_mimic_maira-2}{P_f_impression}{refused}{0}
\chexdefskip{settings_mimic_maira-2}{P_f_impression}{truncations}{0}
\chexdefskip{settings_mimic_maira-2}{P_n_both}{empty}{0}
\chexdefskip{settings_mimic_maira-2}{P_n_both}{generated}{0}
\chexdefskip{settings_mimic_maira-2}{P_n_both}{length_truncated}{0}
\chexdefskip{settings_mimic_maira-2}{P_n_both}{non_english}{0}
\chexdefskip{settings_mimic_maira-2}{P_n_both}{refused}{0}
\chexdefskip{settings_mimic_maira-2}{P_n_both}{text_only_unsupported}{3269}
\chexdefskip{settings_mimic_maira-2}{P_n_both}{truncations}{0}
\chexdefskip{settings_mimic_maira-2}{P_n_findings}{empty}{0}
\chexdefskip{settings_mimic_maira-2}{P_n_findings}{generated}{0}
\chexdefskip{settings_mimic_maira-2}{P_n_findings}{length_truncated}{0}
\chexdefskip{settings_mimic_maira-2}{P_n_findings}{non_english}{0}
\chexdefskip{settings_mimic_maira-2}{P_n_findings}{refused}{0}
\chexdefskip{settings_mimic_maira-2}{P_n_findings}{text_only_unsupported}{3269}
\chexdefskip{settings_mimic_maira-2}{P_n_findings}{truncations}{0}
\chexdefskip{settings_mimic_maira-2}{P_n_impression}{empty}{0}
\chexdefskip{settings_mimic_maira-2}{P_n_impression}{generated}{0}
\chexdefskip{settings_mimic_maira-2}{P_n_impression}{length_truncated}{0}
\chexdefskip{settings_mimic_maira-2}{P_n_impression}{non_english}{0}
\chexdefskip{settings_mimic_maira-2}{P_n_impression}{refused}{0}
\chexdefskip{settings_mimic_maira-2}{P_n_impression}{text_only_unsupported}{3269}
\chexdefskip{settings_mimic_maira-2}{P_n_impression}{truncations}{0}
\chexdefskip{settings_mimic_medgemma-4b}{B_f_both}{empty}{0}
\chexdefskip{settings_mimic_medgemma-4b}{B_f_both}{empty_target}{1642}
\chexdefskip{settings_mimic_medgemma-4b}{B_f_both}{generated}{1624}
\chexdefskip{settings_mimic_medgemma-4b}{B_f_both}{length_truncated}{0}
\chexdefskip{settings_mimic_medgemma-4b}{B_f_both}{no_frontal}{3}
\chexdefskip{settings_mimic_medgemma-4b}{B_f_both}{non_english}{0}
\chexdefskip{settings_mimic_medgemma-4b}{B_f_both}{refused}{0}
\chexdefskip{settings_mimic_medgemma-4b}{B_f_both}{truncations}{0}
\chexdefskip{settings_mimic_medgemma-4b}{B_f_findings}{empty}{0}
\chexdefskip{settings_mimic_medgemma-4b}{B_f_findings}{empty_target}{920}
\chexdefskip{settings_mimic_medgemma-4b}{B_f_findings}{generated}{2346}
\chexdefskip{settings_mimic_medgemma-4b}{B_f_findings}{length_truncated}{0}
\chexdefskip{settings_mimic_medgemma-4b}{B_f_findings}{no_frontal}{3}
\chexdefskip{settings_mimic_medgemma-4b}{B_f_findings}{non_english}{0}
\chexdefskip{settings_mimic_medgemma-4b}{B_f_findings}{refused}{0}
\chexdefskip{settings_mimic_medgemma-4b}{B_f_findings}{truncations}{0}
\chexdefskip{settings_mimic_medgemma-4b}{B_f_impression}{empty}{0}
\chexdefskip{settings_mimic_medgemma-4b}{B_f_impression}{empty_target}{1043}
\chexdefskip{settings_mimic_medgemma-4b}{B_f_impression}{generated}{2223}
\chexdefskip{settings_mimic_medgemma-4b}{B_f_impression}{length_truncated}{1}
\chexdefskip{settings_mimic_medgemma-4b}{B_f_impression}{no_frontal}{3}
\chexdefskip{settings_mimic_medgemma-4b}{B_f_impression}{non_english}{0}
\chexdefskip{settings_mimic_medgemma-4b}{B_f_impression}{refused}{0}
\chexdefskip{settings_mimic_medgemma-4b}{B_f_impression}{truncations}{0}
\chexdefskip{settings_mimic_medgemma-4b}{P_f_both}{empty}{0}
\chexdefskip{settings_mimic_medgemma-4b}{P_f_both}{empty_target}{1642}
\chexdefskip{settings_mimic_medgemma-4b}{P_f_both}{generated}{1624}
\chexdefskip{settings_mimic_medgemma-4b}{P_f_both}{length_truncated}{0}
\chexdefskip{settings_mimic_medgemma-4b}{P_f_both}{no_frontal}{3}
\chexdefskip{settings_mimic_medgemma-4b}{P_f_both}{non_english}{0}
\chexdefskip{settings_mimic_medgemma-4b}{P_f_both}{refused}{0}
\chexdefskip{settings_mimic_medgemma-4b}{P_f_both}{truncations}{0}
\chexdefskip{settings_mimic_medgemma-4b}{P_f_findings}{empty}{0}
\chexdefskip{settings_mimic_medgemma-4b}{P_f_findings}{empty_target}{920}
\chexdefskip{settings_mimic_medgemma-4b}{P_f_findings}{generated}{2346}
\chexdefskip{settings_mimic_medgemma-4b}{P_f_findings}{length_truncated}{1}
\chexdefskip{settings_mimic_medgemma-4b}{P_f_findings}{no_frontal}{3}
\chexdefskip{settings_mimic_medgemma-4b}{P_f_findings}{non_english}{0}
\chexdefskip{settings_mimic_medgemma-4b}{P_f_findings}{refused}{0}
\chexdefskip{settings_mimic_medgemma-4b}{P_f_findings}{truncations}{0}
\chexdefskip{settings_mimic_medgemma-4b}{P_f_impression}{empty}{0}
\chexdefskip{settings_mimic_medgemma-4b}{P_f_impression}{empty_target}{1043}
\chexdefskip{settings_mimic_medgemma-4b}{P_f_impression}{generated}{2223}
\chexdefskip{settings_mimic_medgemma-4b}{P_f_impression}{length_truncated}{0}
\chexdefskip{settings_mimic_medgemma-4b}{P_f_impression}{no_frontal}{3}
\chexdefskip{settings_mimic_medgemma-4b}{P_f_impression}{non_english}{0}
\chexdefskip{settings_mimic_medgemma-4b}{P_f_impression}{refused}{0}
\chexdefskip{settings_mimic_medgemma-4b}{P_f_impression}{truncations}{0}
\chexdefskip{settings_mimic_medgemma-4b}{P_n_both}{empty}{0}
\chexdefskip{settings_mimic_medgemma-4b}{P_n_both}{empty_target}{1644}
\chexdefskip{settings_mimic_medgemma-4b}{P_n_both}{generated}{1625}
\chexdefskip{settings_mimic_medgemma-4b}{P_n_both}{length_truncated}{0}
\chexdefskip{settings_mimic_medgemma-4b}{P_n_both}{non_english}{0}
\chexdefskip{settings_mimic_medgemma-4b}{P_n_both}{refused}{0}
\chexdefskip{settings_mimic_medgemma-4b}{P_n_both}{truncations}{0}
\chexdefskip{settings_mimic_medgemma-4b}{P_n_findings}{empty}{0}
\chexdefskip{settings_mimic_medgemma-4b}{P_n_findings}{empty_target}{921}
\chexdefskip{settings_mimic_medgemma-4b}{P_n_findings}{generated}{2348}
\chexdefskip{settings_mimic_medgemma-4b}{P_n_findings}{length_truncated}{1}
\chexdefskip{settings_mimic_medgemma-4b}{P_n_findings}{non_english}{0}
\chexdefskip{settings_mimic_medgemma-4b}{P_n_findings}{refused}{0}
\chexdefskip{settings_mimic_medgemma-4b}{P_n_findings}{truncations}{0}
\chexdefskip{settings_mimic_medgemma-4b}{P_n_impression}{empty}{0}
\chexdefskip{settings_mimic_medgemma-4b}{P_n_impression}{empty_target}{1045}
\chexdefskip{settings_mimic_medgemma-4b}{P_n_impression}{generated}{2224}
\chexdefskip{settings_mimic_medgemma-4b}{P_n_impression}{length_truncated}{0}
\chexdefskip{settings_mimic_medgemma-4b}{P_n_impression}{non_english}{0}
\chexdefskip{settings_mimic_medgemma-4b}{P_n_impression}{refused}{0}
\chexdefskip{settings_mimic_medgemma-4b}{P_n_impression}{truncations}{0}
\chexdefskip{settings_mimic_most-common-report}{B_f_both}{empty}{0}
\chexdefskip{settings_mimic_most-common-report}{B_f_both}{empty_target}{1642}
\chexdefskip{settings_mimic_most-common-report}{B_f_both}{generated}{1624}
\chexdefskip{settings_mimic_most-common-report}{B_f_both}{length_truncated}{0}
\chexdefskip{settings_mimic_most-common-report}{B_f_both}{no_frontal}{3}
\chexdefskip{settings_mimic_most-common-report}{B_f_both}{non_english}{0}
\chexdefskip{settings_mimic_most-common-report}{B_f_both}{refused}{0}
\chexdefskip{settings_mimic_most-common-report}{B_f_both}{truncations}{0}
\chexdefskip{settings_mimic_most-common-report}{B_f_findings}{empty}{0}
\chexdefskip{settings_mimic_most-common-report}{B_f_findings}{empty_target}{920}
\chexdefskip{settings_mimic_most-common-report}{B_f_findings}{generated}{2346}
\chexdefskip{settings_mimic_most-common-report}{B_f_findings}{length_truncated}{0}
\chexdefskip{settings_mimic_most-common-report}{B_f_findings}{no_frontal}{3}
\chexdefskip{settings_mimic_most-common-report}{B_f_findings}{non_english}{0}
\chexdefskip{settings_mimic_most-common-report}{B_f_findings}{refused}{0}
\chexdefskip{settings_mimic_most-common-report}{B_f_findings}{truncations}{0}
\chexdefskip{settings_mimic_most-common-report}{B_f_impression}{empty}{0}
\chexdefskip{settings_mimic_most-common-report}{B_f_impression}{empty_target}{1043}
\chexdefskip{settings_mimic_most-common-report}{B_f_impression}{generated}{2223}
\chexdefskip{settings_mimic_most-common-report}{B_f_impression}{length_truncated}{0}
\chexdefskip{settings_mimic_most-common-report}{B_f_impression}{no_frontal}{3}
\chexdefskip{settings_mimic_most-common-report}{B_f_impression}{non_english}{0}
\chexdefskip{settings_mimic_most-common-report}{B_f_impression}{refused}{0}
\chexdefskip{settings_mimic_most-common-report}{B_f_impression}{truncations}{0}
\chexdefskip{settings_mimic_most-common-report}{P_f_both}{empty}{0}
\chexdefskip{settings_mimic_most-common-report}{P_f_both}{empty_target}{1642}
\chexdefskip{settings_mimic_most-common-report}{P_f_both}{generated}{1624}
\chexdefskip{settings_mimic_most-common-report}{P_f_both}{length_truncated}{0}
\chexdefskip{settings_mimic_most-common-report}{P_f_both}{no_frontal}{3}
\chexdefskip{settings_mimic_most-common-report}{P_f_both}{non_english}{0}
\chexdefskip{settings_mimic_most-common-report}{P_f_both}{refused}{0}
\chexdefskip{settings_mimic_most-common-report}{P_f_both}{truncations}{0}
\chexdefskip{settings_mimic_most-common-report}{P_f_findings}{empty}{0}
\chexdefskip{settings_mimic_most-common-report}{P_f_findings}{empty_target}{920}
\chexdefskip{settings_mimic_most-common-report}{P_f_findings}{generated}{2346}
\chexdefskip{settings_mimic_most-common-report}{P_f_findings}{length_truncated}{0}
\chexdefskip{settings_mimic_most-common-report}{P_f_findings}{no_frontal}{3}
\chexdefskip{settings_mimic_most-common-report}{P_f_findings}{non_english}{0}
\chexdefskip{settings_mimic_most-common-report}{P_f_findings}{refused}{0}
\chexdefskip{settings_mimic_most-common-report}{P_f_findings}{truncations}{0}
\chexdefskip{settings_mimic_most-common-report}{P_f_impression}{empty}{0}
\chexdefskip{settings_mimic_most-common-report}{P_f_impression}{empty_target}{1043}
\chexdefskip{settings_mimic_most-common-report}{P_f_impression}{generated}{2223}
\chexdefskip{settings_mimic_most-common-report}{P_f_impression}{length_truncated}{0}
\chexdefskip{settings_mimic_most-common-report}{P_f_impression}{no_frontal}{3}
\chexdefskip{settings_mimic_most-common-report}{P_f_impression}{non_english}{0}
\chexdefskip{settings_mimic_most-common-report}{P_f_impression}{refused}{0}
\chexdefskip{settings_mimic_most-common-report}{P_f_impression}{truncations}{0}
\chexdefskip{settings_mimic_most-common-report}{P_n_both}{empty}{0}
\chexdefskip{settings_mimic_most-common-report}{P_n_both}{empty_target}{1644}
\chexdefskip{settings_mimic_most-common-report}{P_n_both}{generated}{1625}
\chexdefskip{settings_mimic_most-common-report}{P_n_both}{length_truncated}{0}
\chexdefskip{settings_mimic_most-common-report}{P_n_both}{non_english}{0}
\chexdefskip{settings_mimic_most-common-report}{P_n_both}{refused}{0}
\chexdefskip{settings_mimic_most-common-report}{P_n_both}{truncations}{0}
\chexdefskip{settings_mimic_most-common-report}{P_n_findings}{empty}{0}
\chexdefskip{settings_mimic_most-common-report}{P_n_findings}{empty_target}{921}
\chexdefskip{settings_mimic_most-common-report}{P_n_findings}{generated}{2348}
\chexdefskip{settings_mimic_most-common-report}{P_n_findings}{length_truncated}{0}
\chexdefskip{settings_mimic_most-common-report}{P_n_findings}{non_english}{0}
\chexdefskip{settings_mimic_most-common-report}{P_n_findings}{refused}{0}
\chexdefskip{settings_mimic_most-common-report}{P_n_findings}{truncations}{0}
\chexdefskip{settings_mimic_most-common-report}{P_n_impression}{empty}{0}
\chexdefskip{settings_mimic_most-common-report}{P_n_impression}{empty_target}{1045}
\chexdefskip{settings_mimic_most-common-report}{P_n_impression}{generated}{2224}
\chexdefskip{settings_mimic_most-common-report}{P_n_impression}{length_truncated}{0}
\chexdefskip{settings_mimic_most-common-report}{P_n_impression}{non_english}{0}
\chexdefskip{settings_mimic_most-common-report}{P_n_impression}{refused}{0}
\chexdefskip{settings_mimic_most-common-report}{P_n_impression}{truncations}{0}
\chexdefskip{settings_mimic_qwen2.5-vl-3b}{B_f_both}{empty}{0}
\chexdefskip{settings_mimic_qwen2.5-vl-3b}{B_f_both}{empty_target}{1642}
\chexdefskip{settings_mimic_qwen2.5-vl-3b}{B_f_both}{generated}{1624}
\chexdefskip{settings_mimic_qwen2.5-vl-3b}{B_f_both}{length_truncated}{68}
\chexdefskip{settings_mimic_qwen2.5-vl-3b}{B_f_both}{no_frontal}{3}
\chexdefskip{settings_mimic_qwen2.5-vl-3b}{B_f_both}{non_english}{0}
\chexdefskip{settings_mimic_qwen2.5-vl-3b}{B_f_both}{refused}{0}
\chexdefskip{settings_mimic_qwen2.5-vl-3b}{B_f_both}{truncations}{0}
\chexdefskip{settings_mimic_qwen2.5-vl-3b}{B_f_findings}{empty}{0}
\chexdefskip{settings_mimic_qwen2.5-vl-3b}{B_f_findings}{empty_target}{920}
\chexdefskip{settings_mimic_qwen2.5-vl-3b}{B_f_findings}{generated}{2346}
\chexdefskip{settings_mimic_qwen2.5-vl-3b}{B_f_findings}{length_truncated}{86}
\chexdefskip{settings_mimic_qwen2.5-vl-3b}{B_f_findings}{no_frontal}{3}
\chexdefskip{settings_mimic_qwen2.5-vl-3b}{B_f_findings}{non_english}{0}
\chexdefskip{settings_mimic_qwen2.5-vl-3b}{B_f_findings}{refused}{0}
\chexdefskip{settings_mimic_qwen2.5-vl-3b}{B_f_findings}{truncations}{0}
\chexdefskip{settings_mimic_qwen2.5-vl-3b}{B_f_impression}{empty}{0}
\chexdefskip{settings_mimic_qwen2.5-vl-3b}{B_f_impression}{empty_target}{1043}
\chexdefskip{settings_mimic_qwen2.5-vl-3b}{B_f_impression}{generated}{2223}
\chexdefskip{settings_mimic_qwen2.5-vl-3b}{B_f_impression}{length_truncated}{1}
\chexdefskip{settings_mimic_qwen2.5-vl-3b}{B_f_impression}{no_frontal}{3}
\chexdefskip{settings_mimic_qwen2.5-vl-3b}{B_f_impression}{non_english}{0}
\chexdefskip{settings_mimic_qwen2.5-vl-3b}{B_f_impression}{refused}{0}
\chexdefskip{settings_mimic_qwen2.5-vl-3b}{B_f_impression}{truncations}{0}
\chexdefskip{settings_mimic_qwen2.5-vl-3b}{P_f_both}{empty}{0}
\chexdefskip{settings_mimic_qwen2.5-vl-3b}{P_f_both}{empty_target}{1642}
\chexdefskip{settings_mimic_qwen2.5-vl-3b}{P_f_both}{generated}{1624}
\chexdefskip{settings_mimic_qwen2.5-vl-3b}{P_f_both}{length_truncated}{117}
\chexdefskip{settings_mimic_qwen2.5-vl-3b}{P_f_both}{no_frontal}{3}
\chexdefskip{settings_mimic_qwen2.5-vl-3b}{P_f_both}{non_english}{0}
\chexdefskip{settings_mimic_qwen2.5-vl-3b}{P_f_both}{refused}{0}
\chexdefskip{settings_mimic_qwen2.5-vl-3b}{P_f_both}{truncations}{0}
\chexdefskip{settings_mimic_qwen2.5-vl-3b}{P_f_findings}{empty}{0}
\chexdefskip{settings_mimic_qwen2.5-vl-3b}{P_f_findings}{empty_target}{920}
\chexdefskip{settings_mimic_qwen2.5-vl-3b}{P_f_findings}{generated}{2346}
\chexdefskip{settings_mimic_qwen2.5-vl-3b}{P_f_findings}{length_truncated}{33}
\chexdefskip{settings_mimic_qwen2.5-vl-3b}{P_f_findings}{no_frontal}{3}
\chexdefskip{settings_mimic_qwen2.5-vl-3b}{P_f_findings}{non_english}{0}
\chexdefskip{settings_mimic_qwen2.5-vl-3b}{P_f_findings}{refused}{0}
\chexdefskip{settings_mimic_qwen2.5-vl-3b}{P_f_findings}{truncations}{0}
\chexdefskip{settings_mimic_qwen2.5-vl-3b}{P_f_impression}{empty}{0}
\chexdefskip{settings_mimic_qwen2.5-vl-3b}{P_f_impression}{empty_target}{1043}
\chexdefskip{settings_mimic_qwen2.5-vl-3b}{P_f_impression}{generated}{2223}
\chexdefskip{settings_mimic_qwen2.5-vl-3b}{P_f_impression}{length_truncated}{7}
\chexdefskip{settings_mimic_qwen2.5-vl-3b}{P_f_impression}{no_frontal}{3}
\chexdefskip{settings_mimic_qwen2.5-vl-3b}{P_f_impression}{non_english}{0}
\chexdefskip{settings_mimic_qwen2.5-vl-3b}{P_f_impression}{refused}{0}
\chexdefskip{settings_mimic_qwen2.5-vl-3b}{P_f_impression}{truncations}{0}
\chexdefskip{settings_mimic_qwen2.5-vl-3b}{P_n_both}{empty}{0}
\chexdefskip{settings_mimic_qwen2.5-vl-3b}{P_n_both}{empty_target}{1644}
\chexdefskip{settings_mimic_qwen2.5-vl-3b}{P_n_both}{generated}{1625}
\chexdefskip{settings_mimic_qwen2.5-vl-3b}{P_n_both}{length_truncated}{107}
\chexdefskip{settings_mimic_qwen2.5-vl-3b}{P_n_both}{non_english}{0}
\chexdefskip{settings_mimic_qwen2.5-vl-3b}{P_n_both}{refused}{0}
\chexdefskip{settings_mimic_qwen2.5-vl-3b}{P_n_both}{truncations}{0}
\chexdefskip{settings_mimic_qwen2.5-vl-3b}{P_n_findings}{empty}{0}
\chexdefskip{settings_mimic_qwen2.5-vl-3b}{P_n_findings}{empty_target}{921}
\chexdefskip{settings_mimic_qwen2.5-vl-3b}{P_n_findings}{generated}{2348}
\chexdefskip{settings_mimic_qwen2.5-vl-3b}{P_n_findings}{length_truncated}{52}
\chexdefskip{settings_mimic_qwen2.5-vl-3b}{P_n_findings}{non_english}{0}
\chexdefskip{settings_mimic_qwen2.5-vl-3b}{P_n_findings}{refused}{0}
\chexdefskip{settings_mimic_qwen2.5-vl-3b}{P_n_findings}{truncations}{0}
\chexdefskip{settings_mimic_qwen2.5-vl-3b}{P_n_impression}{empty}{0}
\chexdefskip{settings_mimic_qwen2.5-vl-3b}{P_n_impression}{empty_target}{1045}
\chexdefskip{settings_mimic_qwen2.5-vl-3b}{P_n_impression}{generated}{2224}
\chexdefskip{settings_mimic_qwen2.5-vl-3b}{P_n_impression}{length_truncated}{4}
\chexdefskip{settings_mimic_qwen2.5-vl-3b}{P_n_impression}{non_english}{0}
\chexdefskip{settings_mimic_qwen2.5-vl-3b}{P_n_impression}{refused}{0}
\chexdefskip{settings_mimic_qwen2.5-vl-3b}{P_n_impression}{truncations}{0}
\chexdefskip{settings_mimic_qwen2.5-vl-7b}{B_f_both}{empty}{0}
\chexdefskip{settings_mimic_qwen2.5-vl-7b}{B_f_both}{empty_target}{1642}
\chexdefskip{settings_mimic_qwen2.5-vl-7b}{B_f_both}{generated}{1624}
\chexdefskip{settings_mimic_qwen2.5-vl-7b}{B_f_both}{length_truncated}{0}
\chexdefskip{settings_mimic_qwen2.5-vl-7b}{B_f_both}{no_frontal}{3}
\chexdefskip{settings_mimic_qwen2.5-vl-7b}{B_f_both}{non_english}{0}
\chexdefskip{settings_mimic_qwen2.5-vl-7b}{B_f_both}{refused}{0}
\chexdefskip{settings_mimic_qwen2.5-vl-7b}{B_f_both}{truncations}{0}
\chexdefskip{settings_mimic_qwen2.5-vl-7b}{B_f_findings}{empty}{0}
\chexdefskip{settings_mimic_qwen2.5-vl-7b}{B_f_findings}{empty_target}{920}
\chexdefskip{settings_mimic_qwen2.5-vl-7b}{B_f_findings}{generated}{2346}
\chexdefskip{settings_mimic_qwen2.5-vl-7b}{B_f_findings}{length_truncated}{16}
\chexdefskip{settings_mimic_qwen2.5-vl-7b}{B_f_findings}{no_frontal}{3}
\chexdefskip{settings_mimic_qwen2.5-vl-7b}{B_f_findings}{non_english}{0}
\chexdefskip{settings_mimic_qwen2.5-vl-7b}{B_f_findings}{refused}{0}
\chexdefskip{settings_mimic_qwen2.5-vl-7b}{B_f_findings}{truncations}{0}
\chexdefskip{settings_mimic_qwen2.5-vl-7b}{B_f_impression}{empty}{0}
\chexdefskip{settings_mimic_qwen2.5-vl-7b}{B_f_impression}{empty_target}{1043}
\chexdefskip{settings_mimic_qwen2.5-vl-7b}{B_f_impression}{generated}{2223}
\chexdefskip{settings_mimic_qwen2.5-vl-7b}{B_f_impression}{length_truncated}{3}
\chexdefskip{settings_mimic_qwen2.5-vl-7b}{B_f_impression}{no_frontal}{3}
\chexdefskip{settings_mimic_qwen2.5-vl-7b}{B_f_impression}{non_english}{0}
\chexdefskip{settings_mimic_qwen2.5-vl-7b}{B_f_impression}{refused}{0}
\chexdefskip{settings_mimic_qwen2.5-vl-7b}{B_f_impression}{truncations}{0}
\chexdefskip{settings_mimic_qwen2.5-vl-7b}{P_f_both}{empty}{0}
\chexdefskip{settings_mimic_qwen2.5-vl-7b}{P_f_both}{empty_target}{1642}
\chexdefskip{settings_mimic_qwen2.5-vl-7b}{P_f_both}{generated}{1624}
\chexdefskip{settings_mimic_qwen2.5-vl-7b}{P_f_both}{length_truncated}{3}
\chexdefskip{settings_mimic_qwen2.5-vl-7b}{P_f_both}{no_frontal}{3}
\chexdefskip{settings_mimic_qwen2.5-vl-7b}{P_f_both}{non_english}{0}
\chexdefskip{settings_mimic_qwen2.5-vl-7b}{P_f_both}{refused}{0}
\chexdefskip{settings_mimic_qwen2.5-vl-7b}{P_f_both}{truncations}{0}
\chexdefskip{settings_mimic_qwen2.5-vl-7b}{P_f_findings}{empty}{0}
\chexdefskip{settings_mimic_qwen2.5-vl-7b}{P_f_findings}{empty_target}{920}
\chexdefskip{settings_mimic_qwen2.5-vl-7b}{P_f_findings}{generated}{2346}
\chexdefskip{settings_mimic_qwen2.5-vl-7b}{P_f_findings}{length_truncated}{26}
\chexdefskip{settings_mimic_qwen2.5-vl-7b}{P_f_findings}{no_frontal}{3}
\chexdefskip{settings_mimic_qwen2.5-vl-7b}{P_f_findings}{non_english}{0}
\chexdefskip{settings_mimic_qwen2.5-vl-7b}{P_f_findings}{refused}{0}
\chexdefskip{settings_mimic_qwen2.5-vl-7b}{P_f_findings}{truncations}{0}
\chexdefskip{settings_mimic_qwen2.5-vl-7b}{P_f_impression}{empty}{0}
\chexdefskip{settings_mimic_qwen2.5-vl-7b}{P_f_impression}{empty_target}{1043}
\chexdefskip{settings_mimic_qwen2.5-vl-7b}{P_f_impression}{generated}{2223}
\chexdefskip{settings_mimic_qwen2.5-vl-7b}{P_f_impression}{length_truncated}{10}
\chexdefskip{settings_mimic_qwen2.5-vl-7b}{P_f_impression}{no_frontal}{3}
\chexdefskip{settings_mimic_qwen2.5-vl-7b}{P_f_impression}{non_english}{0}
\chexdefskip{settings_mimic_qwen2.5-vl-7b}{P_f_impression}{refused}{0}
\chexdefskip{settings_mimic_qwen2.5-vl-7b}{P_f_impression}{truncations}{0}
\chexdefskip{settings_mimic_qwen2.5-vl-7b}{P_n_both}{empty}{0}
\chexdefskip{settings_mimic_qwen2.5-vl-7b}{P_n_both}{empty_target}{1644}
\chexdefskip{settings_mimic_qwen2.5-vl-7b}{P_n_both}{generated}{1625}
\chexdefskip{settings_mimic_qwen2.5-vl-7b}{P_n_both}{length_truncated}{6}
\chexdefskip{settings_mimic_qwen2.5-vl-7b}{P_n_both}{non_english}{0}
\chexdefskip{settings_mimic_qwen2.5-vl-7b}{P_n_both}{refused}{0}
\chexdefskip{settings_mimic_qwen2.5-vl-7b}{P_n_both}{truncations}{0}
\chexdefskip{settings_mimic_qwen2.5-vl-7b}{P_n_findings}{empty}{0}
\chexdefskip{settings_mimic_qwen2.5-vl-7b}{P_n_findings}{empty_target}{921}
\chexdefskip{settings_mimic_qwen2.5-vl-7b}{P_n_findings}{generated}{2348}
\chexdefskip{settings_mimic_qwen2.5-vl-7b}{P_n_findings}{length_truncated}{41}
\chexdefskip{settings_mimic_qwen2.5-vl-7b}{P_n_findings}{non_english}{1}
\chexdefskip{settings_mimic_qwen2.5-vl-7b}{P_n_findings}{refused}{0}
\chexdefskip{settings_mimic_qwen2.5-vl-7b}{P_n_findings}{truncations}{0}
\chexdefskip{settings_mimic_qwen2.5-vl-7b}{P_n_impression}{empty}{0}
\chexdefskip{settings_mimic_qwen2.5-vl-7b}{P_n_impression}{empty_target}{1045}
\chexdefskip{settings_mimic_qwen2.5-vl-7b}{P_n_impression}{generated}{2224}
\chexdefskip{settings_mimic_qwen2.5-vl-7b}{P_n_impression}{length_truncated}{1}
\chexdefskip{settings_mimic_qwen2.5-vl-7b}{P_n_impression}{non_english}{0}
\chexdefskip{settings_mimic_qwen2.5-vl-7b}{P_n_impression}{refused}{0}
\chexdefskip{settings_mimic_qwen2.5-vl-7b}{P_n_impression}{truncations}{0}
\chexdefskip{settings_mimic_raddialog}{B_f_both}{empty}{0}
\chexdefskip{settings_mimic_raddialog}{B_f_both}{empty_target}{1642}
\chexdefskip{settings_mimic_raddialog}{B_f_both}{generated}{1624}
\chexdefskip{settings_mimic_raddialog}{B_f_both}{length_truncated}{0}
\chexdefskip{settings_mimic_raddialog}{B_f_both}{no_frontal}{3}
\chexdefskip{settings_mimic_raddialog}{B_f_both}{non_english}{0}
\chexdefskip{settings_mimic_raddialog}{B_f_both}{refused}{0}
\chexdefskip{settings_mimic_raddialog}{B_f_both}{truncations}{0}
\chexdefskip{settings_mimic_raddialog}{B_f_findings}{empty}{0}
\chexdefskip{settings_mimic_raddialog}{B_f_findings}{empty_target}{920}
\chexdefskip{settings_mimic_raddialog}{B_f_findings}{generated}{2346}
\chexdefskip{settings_mimic_raddialog}{B_f_findings}{length_truncated}{0}
\chexdefskip{settings_mimic_raddialog}{B_f_findings}{no_frontal}{3}
\chexdefskip{settings_mimic_raddialog}{B_f_findings}{non_english}{0}
\chexdefskip{settings_mimic_raddialog}{B_f_findings}{refused}{0}
\chexdefskip{settings_mimic_raddialog}{B_f_findings}{truncations}{0}
\chexdefskip{settings_mimic_raddialog}{B_f_impression}{empty}{0}
\chexdefskip{settings_mimic_raddialog}{B_f_impression}{empty_target}{1043}
\chexdefskip{settings_mimic_raddialog}{B_f_impression}{generated}{2223}
\chexdefskip{settings_mimic_raddialog}{B_f_impression}{length_truncated}{0}
\chexdefskip{settings_mimic_raddialog}{B_f_impression}{no_frontal}{3}
\chexdefskip{settings_mimic_raddialog}{B_f_impression}{non_english}{0}
\chexdefskip{settings_mimic_raddialog}{B_f_impression}{refused}{0}
\chexdefskip{settings_mimic_raddialog}{B_f_impression}{truncations}{0}
\chexdefskip{settings_mimic_raddialog}{P_f_both}{empty}{0}
\chexdefskip{settings_mimic_raddialog}{P_f_both}{empty_target}{1642}
\chexdefskip{settings_mimic_raddialog}{P_f_both}{generated}{1624}
\chexdefskip{settings_mimic_raddialog}{P_f_both}{length_truncated}{0}
\chexdefskip{settings_mimic_raddialog}{P_f_both}{no_frontal}{3}
\chexdefskip{settings_mimic_raddialog}{P_f_both}{non_english}{0}
\chexdefskip{settings_mimic_raddialog}{P_f_both}{refused}{0}
\chexdefskip{settings_mimic_raddialog}{P_f_both}{truncations}{0}
\chexdefskip{settings_mimic_raddialog}{P_f_findings}{empty}{0}
\chexdefskip{settings_mimic_raddialog}{P_f_findings}{empty_target}{920}
\chexdefskip{settings_mimic_raddialog}{P_f_findings}{generated}{2346}
\chexdefskip{settings_mimic_raddialog}{P_f_findings}{length_truncated}{0}
\chexdefskip{settings_mimic_raddialog}{P_f_findings}{no_frontal}{3}
\chexdefskip{settings_mimic_raddialog}{P_f_findings}{non_english}{0}
\chexdefskip{settings_mimic_raddialog}{P_f_findings}{refused}{0}
\chexdefskip{settings_mimic_raddialog}{P_f_findings}{truncations}{0}
\chexdefskip{settings_mimic_raddialog}{P_f_impression}{empty}{0}
\chexdefskip{settings_mimic_raddialog}{P_f_impression}{empty_target}{1043}
\chexdefskip{settings_mimic_raddialog}{P_f_impression}{generated}{2223}
\chexdefskip{settings_mimic_raddialog}{P_f_impression}{length_truncated}{0}
\chexdefskip{settings_mimic_raddialog}{P_f_impression}{no_frontal}{3}
\chexdefskip{settings_mimic_raddialog}{P_f_impression}{non_english}{0}
\chexdefskip{settings_mimic_raddialog}{P_f_impression}{refused}{0}
\chexdefskip{settings_mimic_raddialog}{P_f_impression}{truncations}{0}
\chexdefskip{settings_mimic_raddialog}{P_n_both}{empty}{0}
\chexdefskip{settings_mimic_raddialog}{P_n_both}{generated}{0}
\chexdefskip{settings_mimic_raddialog}{P_n_both}{length_truncated}{0}
\chexdefskip{settings_mimic_raddialog}{P_n_both}{non_english}{0}
\chexdefskip{settings_mimic_raddialog}{P_n_both}{refused}{0}
\chexdefskip{settings_mimic_raddialog}{P_n_both}{text_only_unsupported}{3269}
\chexdefskip{settings_mimic_raddialog}{P_n_both}{truncations}{0}
\chexdefskip{settings_mimic_raddialog}{P_n_findings}{empty}{0}
\chexdefskip{settings_mimic_raddialog}{P_n_findings}{generated}{0}
\chexdefskip{settings_mimic_raddialog}{P_n_findings}{length_truncated}{0}
\chexdefskip{settings_mimic_raddialog}{P_n_findings}{non_english}{0}
\chexdefskip{settings_mimic_raddialog}{P_n_findings}{refused}{0}
\chexdefskip{settings_mimic_raddialog}{P_n_findings}{text_only_unsupported}{3269}
\chexdefskip{settings_mimic_raddialog}{P_n_findings}{truncations}{0}
\chexdefskip{settings_mimic_raddialog}{P_n_impression}{empty}{0}
\chexdefskip{settings_mimic_raddialog}{P_n_impression}{generated}{0}
\chexdefskip{settings_mimic_raddialog}{P_n_impression}{length_truncated}{0}
\chexdefskip{settings_mimic_raddialog}{P_n_impression}{non_english}{0}
\chexdefskip{settings_mimic_raddialog}{P_n_impression}{refused}{0}
\chexdefskip{settings_mimic_raddialog}{P_n_impression}{text_only_unsupported}{3269}
\chexdefskip{settings_mimic_raddialog}{P_n_impression}{truncations}{0}
\chexdefskip{settings_mimic_raddialog-prefixed}{B_f_both}{empty}{0}
\chexdefskip{settings_mimic_raddialog-prefixed}{B_f_both}{empty_target}{1642}
\chexdefskip{settings_mimic_raddialog-prefixed}{B_f_both}{generated}{1624}
\chexdefskip{settings_mimic_raddialog-prefixed}{B_f_both}{length_truncated}{0}
\chexdefskip{settings_mimic_raddialog-prefixed}{B_f_both}{no_frontal}{3}
\chexdefskip{settings_mimic_raddialog-prefixed}{B_f_both}{non_english}{0}
\chexdefskip{settings_mimic_raddialog-prefixed}{B_f_both}{refused}{0}
\chexdefskip{settings_mimic_raddialog-prefixed}{B_f_both}{truncations}{0}
\chexdefskip{settings_mimic_raddialog-prefixed}{B_f_findings}{empty}{0}
\chexdefskip{settings_mimic_raddialog-prefixed}{B_f_findings}{empty_target}{920}
\chexdefskip{settings_mimic_raddialog-prefixed}{B_f_findings}{generated}{2346}
\chexdefskip{settings_mimic_raddialog-prefixed}{B_f_findings}{length_truncated}{0}
\chexdefskip{settings_mimic_raddialog-prefixed}{B_f_findings}{no_frontal}{3}
\chexdefskip{settings_mimic_raddialog-prefixed}{B_f_findings}{non_english}{0}
\chexdefskip{settings_mimic_raddialog-prefixed}{B_f_findings}{refused}{0}
\chexdefskip{settings_mimic_raddialog-prefixed}{B_f_findings}{truncations}{0}
\chexdefskip{settings_mimic_raddialog-prefixed}{B_f_impression}{empty}{0}
\chexdefskip{settings_mimic_raddialog-prefixed}{B_f_impression}{empty_target}{1043}
\chexdefskip{settings_mimic_raddialog-prefixed}{B_f_impression}{generated}{2223}
\chexdefskip{settings_mimic_raddialog-prefixed}{B_f_impression}{length_truncated}{0}
\chexdefskip{settings_mimic_raddialog-prefixed}{B_f_impression}{no_frontal}{3}
\chexdefskip{settings_mimic_raddialog-prefixed}{B_f_impression}{non_english}{0}
\chexdefskip{settings_mimic_raddialog-prefixed}{B_f_impression}{refused}{0}
\chexdefskip{settings_mimic_raddialog-prefixed}{B_f_impression}{truncations}{0}
\chexdefskip{settings_mimic_raddialog-prefixed}{P_f_both}{empty}{0}
\chexdefskip{settings_mimic_raddialog-prefixed}{P_f_both}{empty_target}{1642}
\chexdefskip{settings_mimic_raddialog-prefixed}{P_f_both}{generated}{1624}
\chexdefskip{settings_mimic_raddialog-prefixed}{P_f_both}{length_truncated}{0}
\chexdefskip{settings_mimic_raddialog-prefixed}{P_f_both}{no_frontal}{3}
\chexdefskip{settings_mimic_raddialog-prefixed}{P_f_both}{non_english}{0}
\chexdefskip{settings_mimic_raddialog-prefixed}{P_f_both}{refused}{0}
\chexdefskip{settings_mimic_raddialog-prefixed}{P_f_both}{truncations}{0}
\chexdefskip{settings_mimic_raddialog-prefixed}{P_f_findings}{empty}{0}
\chexdefskip{settings_mimic_raddialog-prefixed}{P_f_findings}{empty_target}{920}
\chexdefskip{settings_mimic_raddialog-prefixed}{P_f_findings}{generated}{2346}
\chexdefskip{settings_mimic_raddialog-prefixed}{P_f_findings}{length_truncated}{0}
\chexdefskip{settings_mimic_raddialog-prefixed}{P_f_findings}{no_frontal}{3}
\chexdefskip{settings_mimic_raddialog-prefixed}{P_f_findings}{non_english}{0}
\chexdefskip{settings_mimic_raddialog-prefixed}{P_f_findings}{refused}{0}
\chexdefskip{settings_mimic_raddialog-prefixed}{P_f_findings}{truncations}{0}
\chexdefskip{settings_mimic_raddialog-prefixed}{P_f_impression}{empty}{0}
\chexdefskip{settings_mimic_raddialog-prefixed}{P_f_impression}{empty_target}{1043}
\chexdefskip{settings_mimic_raddialog-prefixed}{P_f_impression}{generated}{2223}
\chexdefskip{settings_mimic_raddialog-prefixed}{P_f_impression}{length_truncated}{0}
\chexdefskip{settings_mimic_raddialog-prefixed}{P_f_impression}{no_frontal}{3}
\chexdefskip{settings_mimic_raddialog-prefixed}{P_f_impression}{non_english}{0}
\chexdefskip{settings_mimic_raddialog-prefixed}{P_f_impression}{refused}{0}
\chexdefskip{settings_mimic_raddialog-prefixed}{P_f_impression}{truncations}{0}
\chexdefskip{settings_mimic_raddialog-prefixed}{P_n_both}{empty}{0}
\chexdefskip{settings_mimic_raddialog-prefixed}{P_n_both}{generated}{0}
\chexdefskip{settings_mimic_raddialog-prefixed}{P_n_both}{length_truncated}{0}
\chexdefskip{settings_mimic_raddialog-prefixed}{P_n_both}{non_english}{0}
\chexdefskip{settings_mimic_raddialog-prefixed}{P_n_both}{refused}{0}
\chexdefskip{settings_mimic_raddialog-prefixed}{P_n_both}{text_only_unsupported}{3269}
\chexdefskip{settings_mimic_raddialog-prefixed}{P_n_both}{truncations}{0}
\chexdefskip{settings_mimic_raddialog-prefixed}{P_n_findings}{empty}{0}
\chexdefskip{settings_mimic_raddialog-prefixed}{P_n_findings}{generated}{0}
\chexdefskip{settings_mimic_raddialog-prefixed}{P_n_findings}{length_truncated}{0}
\chexdefskip{settings_mimic_raddialog-prefixed}{P_n_findings}{non_english}{0}
\chexdefskip{settings_mimic_raddialog-prefixed}{P_n_findings}{refused}{0}
\chexdefskip{settings_mimic_raddialog-prefixed}{P_n_findings}{text_only_unsupported}{3269}
\chexdefskip{settings_mimic_raddialog-prefixed}{P_n_findings}{truncations}{0}
\chexdefskip{settings_mimic_raddialog-prefixed}{P_n_impression}{empty}{0}
\chexdefskip{settings_mimic_raddialog-prefixed}{P_n_impression}{generated}{0}
\chexdefskip{settings_mimic_raddialog-prefixed}{P_n_impression}{length_truncated}{0}
\chexdefskip{settings_mimic_raddialog-prefixed}{P_n_impression}{non_english}{0}
\chexdefskip{settings_mimic_raddialog-prefixed}{P_n_impression}{refused}{0}
\chexdefskip{settings_mimic_raddialog-prefixed}{P_n_impression}{text_only_unsupported}{3269}
\chexdefskip{settings_mimic_raddialog-prefixed}{P_n_impression}{truncations}{0}
\chexdefskip{settings_mimic_random-report}{B_f_both}{empty}{0}
\chexdefskip{settings_mimic_random-report}{B_f_both}{empty_target}{1642}
\chexdefskip{settings_mimic_random-report}{B_f_both}{generated}{1624}
\chexdefskip{settings_mimic_random-report}{B_f_both}{length_truncated}{0}
\chexdefskip{settings_mimic_random-report}{B_f_both}{no_frontal}{3}
\chexdefskip{settings_mimic_random-report}{B_f_both}{non_english}{0}
\chexdefskip{settings_mimic_random-report}{B_f_both}{refused}{3}
\chexdefskip{settings_mimic_random-report}{B_f_both}{truncations}{0}
\chexdefskip{settings_mimic_random-report}{B_f_findings}{empty}{0}
\chexdefskip{settings_mimic_random-report}{B_f_findings}{empty_target}{920}
\chexdefskip{settings_mimic_random-report}{B_f_findings}{generated}{2346}
\chexdefskip{settings_mimic_random-report}{B_f_findings}{length_truncated}{0}
\chexdefskip{settings_mimic_random-report}{B_f_findings}{no_frontal}{3}
\chexdefskip{settings_mimic_random-report}{B_f_findings}{non_english}{0}
\chexdefskip{settings_mimic_random-report}{B_f_findings}{refused}{0}
\chexdefskip{settings_mimic_random-report}{B_f_findings}{truncations}{0}
\chexdefskip{settings_mimic_random-report}{B_f_impression}{empty}{0}
\chexdefskip{settings_mimic_random-report}{B_f_impression}{empty_target}{1043}
\chexdefskip{settings_mimic_random-report}{B_f_impression}{generated}{2223}
\chexdefskip{settings_mimic_random-report}{B_f_impression}{length_truncated}{0}
\chexdefskip{settings_mimic_random-report}{B_f_impression}{no_frontal}{3}
\chexdefskip{settings_mimic_random-report}{B_f_impression}{non_english}{0}
\chexdefskip{settings_mimic_random-report}{B_f_impression}{refused}{3}
\chexdefskip{settings_mimic_random-report}{B_f_impression}{truncations}{0}
\chexdefskip{settings_mimic_random-report}{P_f_both}{empty}{0}
\chexdefskip{settings_mimic_random-report}{P_f_both}{empty_target}{1642}
\chexdefskip{settings_mimic_random-report}{P_f_both}{generated}{1624}
\chexdefskip{settings_mimic_random-report}{P_f_both}{length_truncated}{0}
\chexdefskip{settings_mimic_random-report}{P_f_both}{no_frontal}{3}
\chexdefskip{settings_mimic_random-report}{P_f_both}{non_english}{0}
\chexdefskip{settings_mimic_random-report}{P_f_both}{refused}{2}
\chexdefskip{settings_mimic_random-report}{P_f_both}{truncations}{0}
\chexdefskip{settings_mimic_random-report}{P_f_findings}{empty}{0}
\chexdefskip{settings_mimic_random-report}{P_f_findings}{empty_target}{920}
\chexdefskip{settings_mimic_random-report}{P_f_findings}{generated}{2346}
\chexdefskip{settings_mimic_random-report}{P_f_findings}{length_truncated}{0}
\chexdefskip{settings_mimic_random-report}{P_f_findings}{no_frontal}{3}
\chexdefskip{settings_mimic_random-report}{P_f_findings}{non_english}{0}
\chexdefskip{settings_mimic_random-report}{P_f_findings}{refused}{0}
\chexdefskip{settings_mimic_random-report}{P_f_findings}{truncations}{0}
\chexdefskip{settings_mimic_random-report}{P_f_impression}{empty}{0}
\chexdefskip{settings_mimic_random-report}{P_f_impression}{empty_target}{1043}
\chexdefskip{settings_mimic_random-report}{P_f_impression}{generated}{2223}
\chexdefskip{settings_mimic_random-report}{P_f_impression}{length_truncated}{0}
\chexdefskip{settings_mimic_random-report}{P_f_impression}{no_frontal}{3}
\chexdefskip{settings_mimic_random-report}{P_f_impression}{non_english}{0}
\chexdefskip{settings_mimic_random-report}{P_f_impression}{refused}{7}
\chexdefskip{settings_mimic_random-report}{P_f_impression}{truncations}{0}
\chexdefskip{settings_mimic_random-report}{P_n_both}{empty}{0}
\chexdefskip{settings_mimic_random-report}{P_n_both}{empty_target}{1644}
\chexdefskip{settings_mimic_random-report}{P_n_both}{generated}{1625}
\chexdefskip{settings_mimic_random-report}{P_n_both}{length_truncated}{0}
\chexdefskip{settings_mimic_random-report}{P_n_both}{non_english}{0}
\chexdefskip{settings_mimic_random-report}{P_n_both}{refused}{3}
\chexdefskip{settings_mimic_random-report}{P_n_both}{truncations}{0}
\chexdefskip{settings_mimic_random-report}{P_n_findings}{empty}{0}
\chexdefskip{settings_mimic_random-report}{P_n_findings}{empty_target}{921}
\chexdefskip{settings_mimic_random-report}{P_n_findings}{generated}{2348}
\chexdefskip{settings_mimic_random-report}{P_n_findings}{length_truncated}{0}
\chexdefskip{settings_mimic_random-report}{P_n_findings}{non_english}{0}
\chexdefskip{settings_mimic_random-report}{P_n_findings}{refused}{0}
\chexdefskip{settings_mimic_random-report}{P_n_findings}{truncations}{0}
\chexdefskip{settings_mimic_random-report}{P_n_impression}{empty}{0}
\chexdefskip{settings_mimic_random-report}{P_n_impression}{empty_target}{1045}
\chexdefskip{settings_mimic_random-report}{P_n_impression}{generated}{2224}
\chexdefskip{settings_mimic_random-report}{P_n_impression}{length_truncated}{0}
\chexdefskip{settings_mimic_random-report}{P_n_impression}{non_english}{0}
\chexdefskip{settings_mimic_random-report}{P_n_impression}{refused}{6}
\chexdefskip{settings_mimic_random-report}{P_n_impression}{truncations}{0}
\chexdefskip{settings_mimic_rgrg}{B_f_findings}{empty}{0}
\chexdefskip{settings_mimic_rgrg}{B_f_findings}{empty_target}{920}
\chexdefskip{settings_mimic_rgrg}{B_f_findings}{generated}{2346}
\chexdefskip{settings_mimic_rgrg}{B_f_findings}{length_truncated}{0}
\chexdefskip{settings_mimic_rgrg}{B_f_findings}{no_frontal}{3}
\chexdefskip{settings_mimic_rgrg}{B_f_findings}{non_english}{0}
\chexdefskip{settings_mimic_rgrg}{B_f_findings}{refused}{0}
\chexdefskip{settings_mimic_rgrg}{B_f_findings}{truncations}{0}
\chexdefskip{settings_mimic_rgrg}{P_f_findings}{empty}{0}
\chexdefskip{settings_mimic_rgrg}{P_f_findings}{empty_target}{920}
\chexdefskip{settings_mimic_rgrg}{P_f_findings}{generated}{2346}
\chexdefskip{settings_mimic_rgrg}{P_f_findings}{length_truncated}{0}
\chexdefskip{settings_mimic_rgrg}{P_f_findings}{no_frontal}{3}
\chexdefskip{settings_mimic_rgrg}{P_f_findings}{non_english}{0}
\chexdefskip{settings_mimic_rgrg}{P_f_findings}{refused}{0}
\chexdefskip{settings_mimic_rgrg}{P_f_findings}{truncations}{0}
\chexdefskip{settings_openi_chexagent-2-3b-native}{B_f_both}{empty}{0}
\chexdefskip{settings_openi_chexagent-2-3b-native}{B_f_both}{empty_target}{496}
\chexdefskip{settings_openi_chexagent-2-3b-native}{B_f_both}{generated}{3192}
\chexdefskip{settings_openi_chexagent-2-3b-native}{B_f_both}{length_truncated}{0}
\chexdefskip{settings_openi_chexagent-2-3b-native}{B_f_both}{no_frontal}{163}
\chexdefskip{settings_openi_chexagent-2-3b-native}{B_f_both}{non_english}{0}
\chexdefskip{settings_openi_chexagent-2-3b-native}{B_f_both}{refused}{0}
\chexdefskip{settings_openi_chexagent-2-3b-native}{B_f_both}{truncations}{0}
\chexdefskip{settings_openi_chexagent-2-3b-native}{B_f_findings}{empty}{0}
\chexdefskip{settings_openi_chexagent-2-3b-native}{B_f_findings}{empty_target}{490}
\chexdefskip{settings_openi_chexagent-2-3b-native}{B_f_findings}{generated}{3198}
\chexdefskip{settings_openi_chexagent-2-3b-native}{B_f_findings}{length_truncated}{0}
\chexdefskip{settings_openi_chexagent-2-3b-native}{B_f_findings}{no_frontal}{163}
\chexdefskip{settings_openi_chexagent-2-3b-native}{B_f_findings}{non_english}{0}
\chexdefskip{settings_openi_chexagent-2-3b-native}{B_f_findings}{refused}{0}
\chexdefskip{settings_openi_chexagent-2-3b-native}{B_f_findings}{truncations}{0}
\chexdefskip{settings_openi_chexagent-2-3b-native}{B_f_impression}{empty}{0}
\chexdefskip{settings_openi_chexagent-2-3b-native}{B_f_impression}{empty_target}{29}
\chexdefskip{settings_openi_chexagent-2-3b-native}{B_f_impression}{generated}{3659}
\chexdefskip{settings_openi_chexagent-2-3b-native}{B_f_impression}{length_truncated}{0}
\chexdefskip{settings_openi_chexagent-2-3b-native}{B_f_impression}{no_frontal}{163}
\chexdefskip{settings_openi_chexagent-2-3b-native}{B_f_impression}{non_english}{0}
\chexdefskip{settings_openi_chexagent-2-3b-native}{B_f_impression}{refused}{0}
\chexdefskip{settings_openi_chexagent-2-3b-native}{B_f_impression}{truncations}{0}
\chexdefskip{settings_openi_chexagent-2-3b-native}{P_f_both}{empty}{0}
\chexdefskip{settings_openi_chexagent-2-3b-native}{P_f_both}{empty_target}{496}
\chexdefskip{settings_openi_chexagent-2-3b-native}{P_f_both}{generated}{3192}
\chexdefskip{settings_openi_chexagent-2-3b-native}{P_f_both}{length_truncated}{0}
\chexdefskip{settings_openi_chexagent-2-3b-native}{P_f_both}{no_frontal}{163}
\chexdefskip{settings_openi_chexagent-2-3b-native}{P_f_both}{non_english}{0}
\chexdefskip{settings_openi_chexagent-2-3b-native}{P_f_both}{refused}{0}
\chexdefskip{settings_openi_chexagent-2-3b-native}{P_f_both}{truncations}{0}
\chexdefskip{settings_openi_chexagent-2-3b-native}{P_f_findings}{empty}{0}
\chexdefskip{settings_openi_chexagent-2-3b-native}{P_f_findings}{empty_target}{490}
\chexdefskip{settings_openi_chexagent-2-3b-native}{P_f_findings}{generated}{3198}
\chexdefskip{settings_openi_chexagent-2-3b-native}{P_f_findings}{length_truncated}{0}
\chexdefskip{settings_openi_chexagent-2-3b-native}{P_f_findings}{no_frontal}{163}
\chexdefskip{settings_openi_chexagent-2-3b-native}{P_f_findings}{non_english}{0}
\chexdefskip{settings_openi_chexagent-2-3b-native}{P_f_findings}{refused}{0}
\chexdefskip{settings_openi_chexagent-2-3b-native}{P_f_findings}{truncations}{0}
\chexdefskip{settings_openi_chexagent-2-3b-native}{P_f_impression}{empty}{0}
\chexdefskip{settings_openi_chexagent-2-3b-native}{P_f_impression}{empty_target}{29}
\chexdefskip{settings_openi_chexagent-2-3b-native}{P_f_impression}{generated}{3659}
\chexdefskip{settings_openi_chexagent-2-3b-native}{P_f_impression}{length_truncated}{0}
\chexdefskip{settings_openi_chexagent-2-3b-native}{P_f_impression}{no_frontal}{163}
\chexdefskip{settings_openi_chexagent-2-3b-native}{P_f_impression}{non_english}{0}
\chexdefskip{settings_openi_chexagent-2-3b-native}{P_f_impression}{refused}{0}
\chexdefskip{settings_openi_chexagent-2-3b-native}{P_f_impression}{truncations}{0}
\chexdefskip{settings_openi_chexagent-2-3b-native}{P_n_both}{empty}{0}
\chexdefskip{settings_openi_chexagent-2-3b-native}{P_n_both}{generated}{0}
\chexdefskip{settings_openi_chexagent-2-3b-native}{P_n_both}{length_truncated}{0}
\chexdefskip{settings_openi_chexagent-2-3b-native}{P_n_both}{non_english}{0}
\chexdefskip{settings_openi_chexagent-2-3b-native}{P_n_both}{refused}{0}
\chexdefskip{settings_openi_chexagent-2-3b-native}{P_n_both}{text_only_unsupported}{3851}
\chexdefskip{settings_openi_chexagent-2-3b-native}{P_n_both}{truncations}{0}
\chexdefskip{settings_openi_chexagent-2-3b-native}{P_n_findings}{empty}{0}
\chexdefskip{settings_openi_chexagent-2-3b-native}{P_n_findings}{generated}{0}
\chexdefskip{settings_openi_chexagent-2-3b-native}{P_n_findings}{length_truncated}{0}
\chexdefskip{settings_openi_chexagent-2-3b-native}{P_n_findings}{non_english}{0}
\chexdefskip{settings_openi_chexagent-2-3b-native}{P_n_findings}{refused}{0}
\chexdefskip{settings_openi_chexagent-2-3b-native}{P_n_findings}{text_only_unsupported}{3851}
\chexdefskip{settings_openi_chexagent-2-3b-native}{P_n_findings}{truncations}{0}
\chexdefskip{settings_openi_chexagent-2-3b-native}{P_n_impression}{empty}{0}
\chexdefskip{settings_openi_chexagent-2-3b-native}{P_n_impression}{generated}{0}
\chexdefskip{settings_openi_chexagent-2-3b-native}{P_n_impression}{length_truncated}{0}
\chexdefskip{settings_openi_chexagent-2-3b-native}{P_n_impression}{non_english}{0}
\chexdefskip{settings_openi_chexagent-2-3b-native}{P_n_impression}{refused}{0}
\chexdefskip{settings_openi_chexagent-2-3b-native}{P_n_impression}{text_only_unsupported}{3851}
\chexdefskip{settings_openi_chexagent-2-3b-native}{P_n_impression}{truncations}{0}
\chexdefskip{settings_openi_chexagent-2-3b-native-notags}{B_f_both}{empty}{0}
\chexdefskip{settings_openi_chexagent-2-3b-native-notags}{B_f_both}{empty_target}{496}
\chexdefskip{settings_openi_chexagent-2-3b-native-notags}{B_f_both}{generated}{3192}
\chexdefskip{settings_openi_chexagent-2-3b-native-notags}{B_f_both}{length_truncated}{0}
\chexdefskip{settings_openi_chexagent-2-3b-native-notags}{B_f_both}{no_frontal}{163}
\chexdefskip{settings_openi_chexagent-2-3b-native-notags}{B_f_both}{non_english}{0}
\chexdefskip{settings_openi_chexagent-2-3b-native-notags}{B_f_both}{refused}{0}
\chexdefskip{settings_openi_chexagent-2-3b-native-notags}{B_f_both}{truncations}{0}
\chexdefskip{settings_openi_chexagent-2-3b-native-notags}{B_f_findings}{empty}{0}
\chexdefskip{settings_openi_chexagent-2-3b-native-notags}{B_f_findings}{empty_target}{490}
\chexdefskip{settings_openi_chexagent-2-3b-native-notags}{B_f_findings}{generated}{3198}
\chexdefskip{settings_openi_chexagent-2-3b-native-notags}{B_f_findings}{length_truncated}{0}
\chexdefskip{settings_openi_chexagent-2-3b-native-notags}{B_f_findings}{no_frontal}{163}
\chexdefskip{settings_openi_chexagent-2-3b-native-notags}{B_f_findings}{non_english}{0}
\chexdefskip{settings_openi_chexagent-2-3b-native-notags}{B_f_findings}{refused}{0}
\chexdefskip{settings_openi_chexagent-2-3b-native-notags}{B_f_findings}{truncations}{0}
\chexdefskip{settings_openi_chexagent-2-3b-native-notags}{B_f_impression}{empty}{0}
\chexdefskip{settings_openi_chexagent-2-3b-native-notags}{B_f_impression}{empty_target}{29}
\chexdefskip{settings_openi_chexagent-2-3b-native-notags}{B_f_impression}{generated}{3659}
\chexdefskip{settings_openi_chexagent-2-3b-native-notags}{B_f_impression}{length_truncated}{0}
\chexdefskip{settings_openi_chexagent-2-3b-native-notags}{B_f_impression}{no_frontal}{163}
\chexdefskip{settings_openi_chexagent-2-3b-native-notags}{B_f_impression}{non_english}{0}
\chexdefskip{settings_openi_chexagent-2-3b-native-notags}{B_f_impression}{refused}{0}
\chexdefskip{settings_openi_chexagent-2-3b-native-notags}{B_f_impression}{truncations}{0}
\chexdefskip{settings_openi_chexagent-2-3b-native-notags}{P_f_both}{empty}{0}
\chexdefskip{settings_openi_chexagent-2-3b-native-notags}{P_f_both}{empty_target}{496}
\chexdefskip{settings_openi_chexagent-2-3b-native-notags}{P_f_both}{generated}{3192}
\chexdefskip{settings_openi_chexagent-2-3b-native-notags}{P_f_both}{length_truncated}{0}
\chexdefskip{settings_openi_chexagent-2-3b-native-notags}{P_f_both}{no_frontal}{163}
\chexdefskip{settings_openi_chexagent-2-3b-native-notags}{P_f_both}{non_english}{0}
\chexdefskip{settings_openi_chexagent-2-3b-native-notags}{P_f_both}{refused}{0}
\chexdefskip{settings_openi_chexagent-2-3b-native-notags}{P_f_both}{truncations}{0}
\chexdefskip{settings_openi_chexagent-2-3b-native-notags}{P_f_findings}{empty}{0}
\chexdefskip{settings_openi_chexagent-2-3b-native-notags}{P_f_findings}{empty_target}{490}
\chexdefskip{settings_openi_chexagent-2-3b-native-notags}{P_f_findings}{generated}{3198}
\chexdefskip{settings_openi_chexagent-2-3b-native-notags}{P_f_findings}{length_truncated}{0}
\chexdefskip{settings_openi_chexagent-2-3b-native-notags}{P_f_findings}{no_frontal}{163}
\chexdefskip{settings_openi_chexagent-2-3b-native-notags}{P_f_findings}{non_english}{0}
\chexdefskip{settings_openi_chexagent-2-3b-native-notags}{P_f_findings}{refused}{0}
\chexdefskip{settings_openi_chexagent-2-3b-native-notags}{P_f_findings}{truncations}{0}
\chexdefskip{settings_openi_chexagent-2-3b-native-notags}{P_f_impression}{empty}{0}
\chexdefskip{settings_openi_chexagent-2-3b-native-notags}{P_f_impression}{empty_target}{29}
\chexdefskip{settings_openi_chexagent-2-3b-native-notags}{P_f_impression}{generated}{3659}
\chexdefskip{settings_openi_chexagent-2-3b-native-notags}{P_f_impression}{length_truncated}{0}
\chexdefskip{settings_openi_chexagent-2-3b-native-notags}{P_f_impression}{no_frontal}{163}
\chexdefskip{settings_openi_chexagent-2-3b-native-notags}{P_f_impression}{non_english}{0}
\chexdefskip{settings_openi_chexagent-2-3b-native-notags}{P_f_impression}{refused}{0}
\chexdefskip{settings_openi_chexagent-2-3b-native-notags}{P_f_impression}{truncations}{0}
\chexdefskip{settings_openi_chexagent-2-3b-native-notags}{P_n_both}{empty}{0}
\chexdefskip{settings_openi_chexagent-2-3b-native-notags}{P_n_both}{generated}{0}
\chexdefskip{settings_openi_chexagent-2-3b-native-notags}{P_n_both}{length_truncated}{0}
\chexdefskip{settings_openi_chexagent-2-3b-native-notags}{P_n_both}{non_english}{0}
\chexdefskip{settings_openi_chexagent-2-3b-native-notags}{P_n_both}{refused}{0}
\chexdefskip{settings_openi_chexagent-2-3b-native-notags}{P_n_both}{text_only_unsupported}{3851}
\chexdefskip{settings_openi_chexagent-2-3b-native-notags}{P_n_both}{truncations}{0}
\chexdefskip{settings_openi_chexagent-2-3b-native-notags}{P_n_findings}{empty}{0}
\chexdefskip{settings_openi_chexagent-2-3b-native-notags}{P_n_findings}{generated}{0}
\chexdefskip{settings_openi_chexagent-2-3b-native-notags}{P_n_findings}{length_truncated}{0}
\chexdefskip{settings_openi_chexagent-2-3b-native-notags}{P_n_findings}{non_english}{0}
\chexdefskip{settings_openi_chexagent-2-3b-native-notags}{P_n_findings}{refused}{0}
\chexdefskip{settings_openi_chexagent-2-3b-native-notags}{P_n_findings}{text_only_unsupported}{3851}
\chexdefskip{settings_openi_chexagent-2-3b-native-notags}{P_n_findings}{truncations}{0}
\chexdefskip{settings_openi_chexagent-2-3b-native-notags}{P_n_impression}{empty}{0}
\chexdefskip{settings_openi_chexagent-2-3b-native-notags}{P_n_impression}{generated}{0}
\chexdefskip{settings_openi_chexagent-2-3b-native-notags}{P_n_impression}{length_truncated}{0}
\chexdefskip{settings_openi_chexagent-2-3b-native-notags}{P_n_impression}{non_english}{0}
\chexdefskip{settings_openi_chexagent-2-3b-native-notags}{P_n_impression}{refused}{0}
\chexdefskip{settings_openi_chexagent-2-3b-native-notags}{P_n_impression}{text_only_unsupported}{3851}
\chexdefskip{settings_openi_chexagent-2-3b-native-notags}{P_n_impression}{truncations}{0}
\chexdefskip{settings_openi_chexone-instruct}{B_f_both}{empty}{0}
\chexdefskip{settings_openi_chexone-instruct}{B_f_both}{empty_target}{496}
\chexdefskip{settings_openi_chexone-instruct}{B_f_both}{generated}{3192}
\chexdefskip{settings_openi_chexone-instruct}{B_f_both}{length_truncated}{199}
\chexdefskip{settings_openi_chexone-instruct}{B_f_both}{no_frontal}{163}
\chexdefskip{settings_openi_chexone-instruct}{B_f_both}{non_english}{0}
\chexdefskip{settings_openi_chexone-instruct}{B_f_both}{refused}{0}
\chexdefskip{settings_openi_chexone-instruct}{B_f_both}{truncations}{0}
\chexdefskip{settings_openi_chexone-instruct}{B_f_findings}{empty}{1}
\chexdefskip{settings_openi_chexone-instruct}{B_f_findings}{empty_target}{490}
\chexdefskip{settings_openi_chexone-instruct}{B_f_findings}{generated}{3198}
\chexdefskip{settings_openi_chexone-instruct}{B_f_findings}{length_truncated}{2}
\chexdefskip{settings_openi_chexone-instruct}{B_f_findings}{no_frontal}{163}
\chexdefskip{settings_openi_chexone-instruct}{B_f_findings}{non_english}{0}
\chexdefskip{settings_openi_chexone-instruct}{B_f_findings}{refused}{0}
\chexdefskip{settings_openi_chexone-instruct}{B_f_findings}{truncations}{0}
\chexdefskip{settings_openi_chexone-instruct}{B_f_impression}{empty}{0}
\chexdefskip{settings_openi_chexone-instruct}{B_f_impression}{empty_target}{29}
\chexdefskip{settings_openi_chexone-instruct}{B_f_impression}{generated}{3659}
\chexdefskip{settings_openi_chexone-instruct}{B_f_impression}{length_truncated}{225}
\chexdefskip{settings_openi_chexone-instruct}{B_f_impression}{no_frontal}{163}
\chexdefskip{settings_openi_chexone-instruct}{B_f_impression}{non_english}{0}
\chexdefskip{settings_openi_chexone-instruct}{B_f_impression}{refused}{0}
\chexdefskip{settings_openi_chexone-instruct}{B_f_impression}{truncations}{0}
\chexdefskip{settings_openi_chexone-instruct}{P_f_both}{empty}{0}
\chexdefskip{settings_openi_chexone-instruct}{P_f_both}{empty_target}{496}
\chexdefskip{settings_openi_chexone-instruct}{P_f_both}{generated}{3192}
\chexdefskip{settings_openi_chexone-instruct}{P_f_both}{length_truncated}{259}
\chexdefskip{settings_openi_chexone-instruct}{P_f_both}{no_frontal}{163}
\chexdefskip{settings_openi_chexone-instruct}{P_f_both}{non_english}{0}
\chexdefskip{settings_openi_chexone-instruct}{P_f_both}{refused}{0}
\chexdefskip{settings_openi_chexone-instruct}{P_f_both}{truncations}{0}
\chexdefskip{settings_openi_chexone-instruct}{P_f_findings}{empty}{0}
\chexdefskip{settings_openi_chexone-instruct}{P_f_findings}{empty_target}{490}
\chexdefskip{settings_openi_chexone-instruct}{P_f_findings}{generated}{3198}
\chexdefskip{settings_openi_chexone-instruct}{P_f_findings}{length_truncated}{2}
\chexdefskip{settings_openi_chexone-instruct}{P_f_findings}{no_frontal}{163}
\chexdefskip{settings_openi_chexone-instruct}{P_f_findings}{non_english}{0}
\chexdefskip{settings_openi_chexone-instruct}{P_f_findings}{refused}{0}
\chexdefskip{settings_openi_chexone-instruct}{P_f_findings}{truncations}{0}
\chexdefskip{settings_openi_chexone-instruct}{P_f_impression}{empty}{0}
\chexdefskip{settings_openi_chexone-instruct}{P_f_impression}{empty_target}{29}
\chexdefskip{settings_openi_chexone-instruct}{P_f_impression}{generated}{3659}
\chexdefskip{settings_openi_chexone-instruct}{P_f_impression}{length_truncated}{308}
\chexdefskip{settings_openi_chexone-instruct}{P_f_impression}{no_frontal}{163}
\chexdefskip{settings_openi_chexone-instruct}{P_f_impression}{non_english}{0}
\chexdefskip{settings_openi_chexone-instruct}{P_f_impression}{refused}{0}
\chexdefskip{settings_openi_chexone-instruct}{P_f_impression}{truncations}{0}
\chexdefskip{settings_openi_chexone-instruct}{P_n_both}{empty}{0}
\chexdefskip{settings_openi_chexone-instruct}{P_n_both}{empty_target}{520}
\chexdefskip{settings_openi_chexone-instruct}{P_n_both}{generated}{3331}
\chexdefskip{settings_openi_chexone-instruct}{P_n_both}{length_truncated}{70}
\chexdefskip{settings_openi_chexone-instruct}{P_n_both}{non_english}{0}
\chexdefskip{settings_openi_chexone-instruct}{P_n_both}{refused}{0}
\chexdefskip{settings_openi_chexone-instruct}{P_n_both}{truncations}{0}
\chexdefskip{settings_openi_chexone-instruct}{P_n_findings}{empty}{0}
\chexdefskip{settings_openi_chexone-instruct}{P_n_findings}{empty_target}{514}
\chexdefskip{settings_openi_chexone-instruct}{P_n_findings}{generated}{3337}
\chexdefskip{settings_openi_chexone-instruct}{P_n_findings}{length_truncated}{3}
\chexdefskip{settings_openi_chexone-instruct}{P_n_findings}{non_english}{0}
\chexdefskip{settings_openi_chexone-instruct}{P_n_findings}{refused}{0}
\chexdefskip{settings_openi_chexone-instruct}{P_n_findings}{truncations}{0}
\chexdefskip{settings_openi_chexone-instruct}{P_n_impression}{empty}{0}
\chexdefskip{settings_openi_chexone-instruct}{P_n_impression}{empty_target}{31}
\chexdefskip{settings_openi_chexone-instruct}{P_n_impression}{generated}{3820}
\chexdefskip{settings_openi_chexone-instruct}{P_n_impression}{length_truncated}{81}
\chexdefskip{settings_openi_chexone-instruct}{P_n_impression}{non_english}{0}
\chexdefskip{settings_openi_chexone-instruct}{P_n_impression}{refused}{0}
\chexdefskip{settings_openi_chexone-instruct}{P_n_impression}{truncations}{0}
\chexdefskip{settings_openi_chexone-reasoning}{B_f_both}{empty}{0}
\chexdefskip{settings_openi_chexone-reasoning}{B_f_both}{empty_target}{496}
\chexdefskip{settings_openi_chexone-reasoning}{B_f_both}{generated}{3192}
\chexdefskip{settings_openi_chexone-reasoning}{B_f_both}{length_truncated}{8}
\chexdefskip{settings_openi_chexone-reasoning}{B_f_both}{no_frontal}{163}
\chexdefskip{settings_openi_chexone-reasoning}{B_f_both}{non_english}{0}
\chexdefskip{settings_openi_chexone-reasoning}{B_f_both}{refused}{0}
\chexdefskip{settings_openi_chexone-reasoning}{B_f_both}{truncations}{0}
\chexdefskip{settings_openi_chexone-reasoning}{B_f_findings}{empty}{2}
\chexdefskip{settings_openi_chexone-reasoning}{B_f_findings}{empty_target}{490}
\chexdefskip{settings_openi_chexone-reasoning}{B_f_findings}{generated}{3198}
\chexdefskip{settings_openi_chexone-reasoning}{B_f_findings}{length_truncated}{2}
\chexdefskip{settings_openi_chexone-reasoning}{B_f_findings}{no_frontal}{163}
\chexdefskip{settings_openi_chexone-reasoning}{B_f_findings}{non_english}{0}
\chexdefskip{settings_openi_chexone-reasoning}{B_f_findings}{refused}{0}
\chexdefskip{settings_openi_chexone-reasoning}{B_f_findings}{truncations}{0}
\chexdefskip{settings_openi_chexone-reasoning}{B_f_impression}{empty}{6}
\chexdefskip{settings_openi_chexone-reasoning}{B_f_impression}{empty_target}{29}
\chexdefskip{settings_openi_chexone-reasoning}{B_f_impression}{generated}{3659}
\chexdefskip{settings_openi_chexone-reasoning}{B_f_impression}{length_truncated}{6}
\chexdefskip{settings_openi_chexone-reasoning}{B_f_impression}{no_frontal}{163}
\chexdefskip{settings_openi_chexone-reasoning}{B_f_impression}{non_english}{0}
\chexdefskip{settings_openi_chexone-reasoning}{B_f_impression}{refused}{0}
\chexdefskip{settings_openi_chexone-reasoning}{B_f_impression}{truncations}{0}
\chexdefskip{settings_openi_chexone-reasoning}{P_f_both}{empty}{0}
\chexdefskip{settings_openi_chexone-reasoning}{P_f_both}{empty_target}{496}
\chexdefskip{settings_openi_chexone-reasoning}{P_f_both}{generated}{3192}
\chexdefskip{settings_openi_chexone-reasoning}{P_f_both}{length_truncated}{18}
\chexdefskip{settings_openi_chexone-reasoning}{P_f_both}{no_frontal}{163}
\chexdefskip{settings_openi_chexone-reasoning}{P_f_both}{non_english}{0}
\chexdefskip{settings_openi_chexone-reasoning}{P_f_both}{refused}{0}
\chexdefskip{settings_openi_chexone-reasoning}{P_f_both}{truncations}{0}
\chexdefskip{settings_openi_chexone-reasoning}{P_f_findings}{empty}{5}
\chexdefskip{settings_openi_chexone-reasoning}{P_f_findings}{empty_target}{490}
\chexdefskip{settings_openi_chexone-reasoning}{P_f_findings}{generated}{3198}
\chexdefskip{settings_openi_chexone-reasoning}{P_f_findings}{length_truncated}{5}
\chexdefskip{settings_openi_chexone-reasoning}{P_f_findings}{no_frontal}{163}
\chexdefskip{settings_openi_chexone-reasoning}{P_f_findings}{non_english}{0}
\chexdefskip{settings_openi_chexone-reasoning}{P_f_findings}{refused}{0}
\chexdefskip{settings_openi_chexone-reasoning}{P_f_findings}{truncations}{0}
\chexdefskip{settings_openi_chexone-reasoning}{P_f_impression}{empty}{17}
\chexdefskip{settings_openi_chexone-reasoning}{P_f_impression}{empty_target}{29}
\chexdefskip{settings_openi_chexone-reasoning}{P_f_impression}{generated}{3659}
\chexdefskip{settings_openi_chexone-reasoning}{P_f_impression}{length_truncated}{16}
\chexdefskip{settings_openi_chexone-reasoning}{P_f_impression}{no_frontal}{163}
\chexdefskip{settings_openi_chexone-reasoning}{P_f_impression}{non_english}{0}
\chexdefskip{settings_openi_chexone-reasoning}{P_f_impression}{refused}{0}
\chexdefskip{settings_openi_chexone-reasoning}{P_f_impression}{truncations}{0}
\chexdefskip{settings_openi_chexone-reasoning}{P_n_both}{empty}{0}
\chexdefskip{settings_openi_chexone-reasoning}{P_n_both}{empty_target}{520}
\chexdefskip{settings_openi_chexone-reasoning}{P_n_both}{generated}{3331}
\chexdefskip{settings_openi_chexone-reasoning}{P_n_both}{length_truncated}{60}
\chexdefskip{settings_openi_chexone-reasoning}{P_n_both}{non_english}{0}
\chexdefskip{settings_openi_chexone-reasoning}{P_n_both}{refused}{0}
\chexdefskip{settings_openi_chexone-reasoning}{P_n_both}{truncations}{0}
\chexdefskip{settings_openi_chexone-reasoning}{P_n_findings}{empty}{30}
\chexdefskip{settings_openi_chexone-reasoning}{P_n_findings}{empty_target}{514}
\chexdefskip{settings_openi_chexone-reasoning}{P_n_findings}{generated}{3337}
\chexdefskip{settings_openi_chexone-reasoning}{P_n_findings}{length_truncated}{28}
\chexdefskip{settings_openi_chexone-reasoning}{P_n_findings}{non_english}{0}
\chexdefskip{settings_openi_chexone-reasoning}{P_n_findings}{refused}{0}
\chexdefskip{settings_openi_chexone-reasoning}{P_n_findings}{truncations}{0}
\chexdefskip{settings_openi_chexone-reasoning}{P_n_impression}{empty}{44}
\chexdefskip{settings_openi_chexone-reasoning}{P_n_impression}{empty_target}{31}
\chexdefskip{settings_openi_chexone-reasoning}{P_n_impression}{generated}{3820}
\chexdefskip{settings_openi_chexone-reasoning}{P_n_impression}{length_truncated}{38}
\chexdefskip{settings_openi_chexone-reasoning}{P_n_impression}{non_english}{0}
\chexdefskip{settings_openi_chexone-reasoning}{P_n_impression}{refused}{0}
\chexdefskip{settings_openi_chexone-reasoning}{P_n_impression}{truncations}{0}
\chexdefskip{settings_openi_constant-normal}{B_f_both}{empty}{0}
\chexdefskip{settings_openi_constant-normal}{B_f_both}{empty_target}{496}
\chexdefskip{settings_openi_constant-normal}{B_f_both}{generated}{3192}
\chexdefskip{settings_openi_constant-normal}{B_f_both}{length_truncated}{0}
\chexdefskip{settings_openi_constant-normal}{B_f_both}{no_frontal}{163}
\chexdefskip{settings_openi_constant-normal}{B_f_both}{non_english}{0}
\chexdefskip{settings_openi_constant-normal}{B_f_both}{refused}{0}
\chexdefskip{settings_openi_constant-normal}{B_f_both}{truncations}{0}
\chexdefskip{settings_openi_constant-normal}{B_f_findings}{empty}{0}
\chexdefskip{settings_openi_constant-normal}{B_f_findings}{empty_target}{490}
\chexdefskip{settings_openi_constant-normal}{B_f_findings}{generated}{3198}
\chexdefskip{settings_openi_constant-normal}{B_f_findings}{length_truncated}{0}
\chexdefskip{settings_openi_constant-normal}{B_f_findings}{no_frontal}{163}
\chexdefskip{settings_openi_constant-normal}{B_f_findings}{non_english}{0}
\chexdefskip{settings_openi_constant-normal}{B_f_findings}{refused}{0}
\chexdefskip{settings_openi_constant-normal}{B_f_findings}{truncations}{0}
\chexdefskip{settings_openi_constant-normal}{B_f_impression}{empty}{0}
\chexdefskip{settings_openi_constant-normal}{B_f_impression}{empty_target}{29}
\chexdefskip{settings_openi_constant-normal}{B_f_impression}{generated}{3659}
\chexdefskip{settings_openi_constant-normal}{B_f_impression}{length_truncated}{0}
\chexdefskip{settings_openi_constant-normal}{B_f_impression}{no_frontal}{163}
\chexdefskip{settings_openi_constant-normal}{B_f_impression}{non_english}{0}
\chexdefskip{settings_openi_constant-normal}{B_f_impression}{refused}{0}
\chexdefskip{settings_openi_constant-normal}{B_f_impression}{truncations}{0}
\chexdefskip{settings_openi_constant-normal}{P_f_both}{empty}{0}
\chexdefskip{settings_openi_constant-normal}{P_f_both}{empty_target}{496}
\chexdefskip{settings_openi_constant-normal}{P_f_both}{generated}{3192}
\chexdefskip{settings_openi_constant-normal}{P_f_both}{length_truncated}{0}
\chexdefskip{settings_openi_constant-normal}{P_f_both}{no_frontal}{163}
\chexdefskip{settings_openi_constant-normal}{P_f_both}{non_english}{0}
\chexdefskip{settings_openi_constant-normal}{P_f_both}{refused}{0}
\chexdefskip{settings_openi_constant-normal}{P_f_both}{truncations}{0}
\chexdefskip{settings_openi_constant-normal}{P_f_findings}{empty}{0}
\chexdefskip{settings_openi_constant-normal}{P_f_findings}{empty_target}{490}
\chexdefskip{settings_openi_constant-normal}{P_f_findings}{generated}{3198}
\chexdefskip{settings_openi_constant-normal}{P_f_findings}{length_truncated}{0}
\chexdefskip{settings_openi_constant-normal}{P_f_findings}{no_frontal}{163}
\chexdefskip{settings_openi_constant-normal}{P_f_findings}{non_english}{0}
\chexdefskip{settings_openi_constant-normal}{P_f_findings}{refused}{0}
\chexdefskip{settings_openi_constant-normal}{P_f_findings}{truncations}{0}
\chexdefskip{settings_openi_constant-normal}{P_f_impression}{empty}{0}
\chexdefskip{settings_openi_constant-normal}{P_f_impression}{empty_target}{29}
\chexdefskip{settings_openi_constant-normal}{P_f_impression}{generated}{3659}
\chexdefskip{settings_openi_constant-normal}{P_f_impression}{length_truncated}{0}
\chexdefskip{settings_openi_constant-normal}{P_f_impression}{no_frontal}{163}
\chexdefskip{settings_openi_constant-normal}{P_f_impression}{non_english}{0}
\chexdefskip{settings_openi_constant-normal}{P_f_impression}{refused}{0}
\chexdefskip{settings_openi_constant-normal}{P_f_impression}{truncations}{0}
\chexdefskip{settings_openi_constant-normal}{P_n_both}{empty}{0}
\chexdefskip{settings_openi_constant-normal}{P_n_both}{empty_target}{520}
\chexdefskip{settings_openi_constant-normal}{P_n_both}{generated}{3331}
\chexdefskip{settings_openi_constant-normal}{P_n_both}{length_truncated}{0}
\chexdefskip{settings_openi_constant-normal}{P_n_both}{non_english}{0}
\chexdefskip{settings_openi_constant-normal}{P_n_both}{refused}{0}
\chexdefskip{settings_openi_constant-normal}{P_n_both}{truncations}{0}
\chexdefskip{settings_openi_constant-normal}{P_n_findings}{empty}{0}
\chexdefskip{settings_openi_constant-normal}{P_n_findings}{empty_target}{514}
\chexdefskip{settings_openi_constant-normal}{P_n_findings}{generated}{3337}
\chexdefskip{settings_openi_constant-normal}{P_n_findings}{length_truncated}{0}
\chexdefskip{settings_openi_constant-normal}{P_n_findings}{non_english}{0}
\chexdefskip{settings_openi_constant-normal}{P_n_findings}{refused}{0}
\chexdefskip{settings_openi_constant-normal}{P_n_findings}{truncations}{0}
\chexdefskip{settings_openi_constant-normal}{P_n_impression}{empty}{0}
\chexdefskip{settings_openi_constant-normal}{P_n_impression}{empty_target}{31}
\chexdefskip{settings_openi_constant-normal}{P_n_impression}{generated}{3820}
\chexdefskip{settings_openi_constant-normal}{P_n_impression}{length_truncated}{0}
\chexdefskip{settings_openi_constant-normal}{P_n_impression}{non_english}{0}
\chexdefskip{settings_openi_constant-normal}{P_n_impression}{refused}{0}
\chexdefskip{settings_openi_constant-normal}{P_n_impression}{truncations}{0}
\chexdefskip{settings_openi_copy-indication}{B_f_both}{empty}{3192}
\chexdefskip{settings_openi_copy-indication}{B_f_both}{empty_target}{496}
\chexdefskip{settings_openi_copy-indication}{B_f_both}{generated}{3192}
\chexdefskip{settings_openi_copy-indication}{B_f_both}{length_truncated}{0}
\chexdefskip{settings_openi_copy-indication}{B_f_both}{no_frontal}{163}
\chexdefskip{settings_openi_copy-indication}{B_f_both}{non_english}{0}
\chexdefskip{settings_openi_copy-indication}{B_f_both}{refused}{0}
\chexdefskip{settings_openi_copy-indication}{B_f_both}{truncations}{0}
\chexdefskip{settings_openi_copy-indication}{B_f_findings}{empty}{3198}
\chexdefskip{settings_openi_copy-indication}{B_f_findings}{empty_target}{490}
\chexdefskip{settings_openi_copy-indication}{B_f_findings}{generated}{3198}
\chexdefskip{settings_openi_copy-indication}{B_f_findings}{length_truncated}{0}
\chexdefskip{settings_openi_copy-indication}{B_f_findings}{no_frontal}{163}
\chexdefskip{settings_openi_copy-indication}{B_f_findings}{non_english}{0}
\chexdefskip{settings_openi_copy-indication}{B_f_findings}{refused}{0}
\chexdefskip{settings_openi_copy-indication}{B_f_findings}{truncations}{0}
\chexdefskip{settings_openi_copy-indication}{B_f_impression}{empty}{3659}
\chexdefskip{settings_openi_copy-indication}{B_f_impression}{empty_target}{29}
\chexdefskip{settings_openi_copy-indication}{B_f_impression}{generated}{3659}
\chexdefskip{settings_openi_copy-indication}{B_f_impression}{length_truncated}{0}
\chexdefskip{settings_openi_copy-indication}{B_f_impression}{no_frontal}{163}
\chexdefskip{settings_openi_copy-indication}{B_f_impression}{non_english}{0}
\chexdefskip{settings_openi_copy-indication}{B_f_impression}{refused}{0}
\chexdefskip{settings_openi_copy-indication}{B_f_impression}{truncations}{0}
\chexdefskip{settings_openi_copy-indication}{P_f_both}{empty}{52}
\chexdefskip{settings_openi_copy-indication}{P_f_both}{empty_target}{496}
\chexdefskip{settings_openi_copy-indication}{P_f_both}{generated}{3192}
\chexdefskip{settings_openi_copy-indication}{P_f_both}{length_truncated}{0}
\chexdefskip{settings_openi_copy-indication}{P_f_both}{no_frontal}{163}
\chexdefskip{settings_openi_copy-indication}{P_f_both}{non_english}{0}
\chexdefskip{settings_openi_copy-indication}{P_f_both}{refused}{0}
\chexdefskip{settings_openi_copy-indication}{P_f_both}{truncations}{0}
\chexdefskip{settings_openi_copy-indication}{P_f_findings}{empty}{52}
\chexdefskip{settings_openi_copy-indication}{P_f_findings}{empty_target}{490}
\chexdefskip{settings_openi_copy-indication}{P_f_findings}{generated}{3198}
\chexdefskip{settings_openi_copy-indication}{P_f_findings}{length_truncated}{0}
\chexdefskip{settings_openi_copy-indication}{P_f_findings}{no_frontal}{163}
\chexdefskip{settings_openi_copy-indication}{P_f_findings}{non_english}{0}
\chexdefskip{settings_openi_copy-indication}{P_f_findings}{refused}{0}
\chexdefskip{settings_openi_copy-indication}{P_f_findings}{truncations}{0}
\chexdefskip{settings_openi_copy-indication}{P_f_impression}{empty}{60}
\chexdefskip{settings_openi_copy-indication}{P_f_impression}{empty_target}{29}
\chexdefskip{settings_openi_copy-indication}{P_f_impression}{generated}{3659}
\chexdefskip{settings_openi_copy-indication}{P_f_impression}{length_truncated}{0}
\chexdefskip{settings_openi_copy-indication}{P_f_impression}{no_frontal}{163}
\chexdefskip{settings_openi_copy-indication}{P_f_impression}{non_english}{0}
\chexdefskip{settings_openi_copy-indication}{P_f_impression}{refused}{0}
\chexdefskip{settings_openi_copy-indication}{P_f_impression}{truncations}{0}
\chexdefskip{settings_openi_copy-indication}{P_n_both}{empty}{53}
\chexdefskip{settings_openi_copy-indication}{P_n_both}{empty_target}{520}
\chexdefskip{settings_openi_copy-indication}{P_n_both}{generated}{3331}
\chexdefskip{settings_openi_copy-indication}{P_n_both}{length_truncated}{0}
\chexdefskip{settings_openi_copy-indication}{P_n_both}{non_english}{0}
\chexdefskip{settings_openi_copy-indication}{P_n_both}{refused}{0}
\chexdefskip{settings_openi_copy-indication}{P_n_both}{truncations}{0}
\chexdefskip{settings_openi_copy-indication}{P_n_findings}{empty}{53}
\chexdefskip{settings_openi_copy-indication}{P_n_findings}{empty_target}{514}
\chexdefskip{settings_openi_copy-indication}{P_n_findings}{generated}{3337}
\chexdefskip{settings_openi_copy-indication}{P_n_findings}{length_truncated}{0}
\chexdefskip{settings_openi_copy-indication}{P_n_findings}{non_english}{0}
\chexdefskip{settings_openi_copy-indication}{P_n_findings}{refused}{0}
\chexdefskip{settings_openi_copy-indication}{P_n_findings}{truncations}{0}
\chexdefskip{settings_openi_copy-indication}{P_n_impression}{empty}{61}
\chexdefskip{settings_openi_copy-indication}{P_n_impression}{empty_target}{31}
\chexdefskip{settings_openi_copy-indication}{P_n_impression}{generated}{3820}
\chexdefskip{settings_openi_copy-indication}{P_n_impression}{length_truncated}{0}
\chexdefskip{settings_openi_copy-indication}{P_n_impression}{non_english}{0}
\chexdefskip{settings_openi_copy-indication}{P_n_impression}{refused}{0}
\chexdefskip{settings_openi_copy-indication}{P_n_impression}{truncations}{0}
\chexdefskip{settings_openi_copy-prior}{B_f_both}{empty}{3192}
\chexdefskip{settings_openi_copy-prior}{B_f_both}{empty_target}{496}
\chexdefskip{settings_openi_copy-prior}{B_f_both}{generated}{3192}
\chexdefskip{settings_openi_copy-prior}{B_f_both}{length_truncated}{0}
\chexdefskip{settings_openi_copy-prior}{B_f_both}{no_frontal}{163}
\chexdefskip{settings_openi_copy-prior}{B_f_both}{non_english}{0}
\chexdefskip{settings_openi_copy-prior}{B_f_both}{refused}{0}
\chexdefskip{settings_openi_copy-prior}{B_f_both}{truncations}{0}
\chexdefskip{settings_openi_copy-prior}{B_f_findings}{empty}{3198}
\chexdefskip{settings_openi_copy-prior}{B_f_findings}{empty_target}{490}
\chexdefskip{settings_openi_copy-prior}{B_f_findings}{generated}{3198}
\chexdefskip{settings_openi_copy-prior}{B_f_findings}{length_truncated}{0}
\chexdefskip{settings_openi_copy-prior}{B_f_findings}{no_frontal}{163}
\chexdefskip{settings_openi_copy-prior}{B_f_findings}{non_english}{0}
\chexdefskip{settings_openi_copy-prior}{B_f_findings}{refused}{0}
\chexdefskip{settings_openi_copy-prior}{B_f_findings}{truncations}{0}
\chexdefskip{settings_openi_copy-prior}{B_f_impression}{empty}{3659}
\chexdefskip{settings_openi_copy-prior}{B_f_impression}{empty_target}{29}
\chexdefskip{settings_openi_copy-prior}{B_f_impression}{generated}{3659}
\chexdefskip{settings_openi_copy-prior}{B_f_impression}{length_truncated}{0}
\chexdefskip{settings_openi_copy-prior}{B_f_impression}{no_frontal}{163}
\chexdefskip{settings_openi_copy-prior}{B_f_impression}{non_english}{0}
\chexdefskip{settings_openi_copy-prior}{B_f_impression}{refused}{0}
\chexdefskip{settings_openi_copy-prior}{B_f_impression}{truncations}{0}
\chexdefskip{settings_openi_copy-prior}{P_f_both}{empty}{3192}
\chexdefskip{settings_openi_copy-prior}{P_f_both}{empty_target}{496}
\chexdefskip{settings_openi_copy-prior}{P_f_both}{generated}{3192}
\chexdefskip{settings_openi_copy-prior}{P_f_both}{length_truncated}{0}
\chexdefskip{settings_openi_copy-prior}{P_f_both}{no_frontal}{163}
\chexdefskip{settings_openi_copy-prior}{P_f_both}{non_english}{0}
\chexdefskip{settings_openi_copy-prior}{P_f_both}{refused}{0}
\chexdefskip{settings_openi_copy-prior}{P_f_both}{truncations}{0}
\chexdefskip{settings_openi_copy-prior}{P_f_findings}{empty}{3198}
\chexdefskip{settings_openi_copy-prior}{P_f_findings}{empty_target}{490}
\chexdefskip{settings_openi_copy-prior}{P_f_findings}{generated}{3198}
\chexdefskip{settings_openi_copy-prior}{P_f_findings}{length_truncated}{0}
\chexdefskip{settings_openi_copy-prior}{P_f_findings}{no_frontal}{163}
\chexdefskip{settings_openi_copy-prior}{P_f_findings}{non_english}{0}
\chexdefskip{settings_openi_copy-prior}{P_f_findings}{refused}{0}
\chexdefskip{settings_openi_copy-prior}{P_f_findings}{truncations}{0}
\chexdefskip{settings_openi_copy-prior}{P_f_impression}{empty}{3659}
\chexdefskip{settings_openi_copy-prior}{P_f_impression}{empty_target}{29}
\chexdefskip{settings_openi_copy-prior}{P_f_impression}{generated}{3659}
\chexdefskip{settings_openi_copy-prior}{P_f_impression}{length_truncated}{0}
\chexdefskip{settings_openi_copy-prior}{P_f_impression}{no_frontal}{163}
\chexdefskip{settings_openi_copy-prior}{P_f_impression}{non_english}{0}
\chexdefskip{settings_openi_copy-prior}{P_f_impression}{refused}{0}
\chexdefskip{settings_openi_copy-prior}{P_f_impression}{truncations}{0}
\chexdefskip{settings_openi_copy-prior}{P_n_both}{empty}{3331}
\chexdefskip{settings_openi_copy-prior}{P_n_both}{empty_target}{520}
\chexdefskip{settings_openi_copy-prior}{P_n_both}{generated}{3331}
\chexdefskip{settings_openi_copy-prior}{P_n_both}{length_truncated}{0}
\chexdefskip{settings_openi_copy-prior}{P_n_both}{non_english}{0}
\chexdefskip{settings_openi_copy-prior}{P_n_both}{refused}{0}
\chexdefskip{settings_openi_copy-prior}{P_n_both}{truncations}{0}
\chexdefskip{settings_openi_copy-prior}{P_n_findings}{empty}{3337}
\chexdefskip{settings_openi_copy-prior}{P_n_findings}{empty_target}{514}
\chexdefskip{settings_openi_copy-prior}{P_n_findings}{generated}{3337}
\chexdefskip{settings_openi_copy-prior}{P_n_findings}{length_truncated}{0}
\chexdefskip{settings_openi_copy-prior}{P_n_findings}{non_english}{0}
\chexdefskip{settings_openi_copy-prior}{P_n_findings}{refused}{0}
\chexdefskip{settings_openi_copy-prior}{P_n_findings}{truncations}{0}
\chexdefskip{settings_openi_copy-prior}{P_n_impression}{empty}{3820}
\chexdefskip{settings_openi_copy-prior}{P_n_impression}{empty_target}{31}
\chexdefskip{settings_openi_copy-prior}{P_n_impression}{generated}{3820}
\chexdefskip{settings_openi_copy-prior}{P_n_impression}{length_truncated}{0}
\chexdefskip{settings_openi_copy-prior}{P_n_impression}{non_english}{0}
\chexdefskip{settings_openi_copy-prior}{P_n_impression}{refused}{0}
\chexdefskip{settings_openi_copy-prior}{P_n_impression}{truncations}{0}
\chexdefskip{settings_openi_cxr-llava-v2}{B_f_both}{empty}{0}
\chexdefskip{settings_openi_cxr-llava-v2}{B_f_both}{empty_target}{496}
\chexdefskip{settings_openi_cxr-llava-v2}{B_f_both}{generated}{3192}
\chexdefskip{settings_openi_cxr-llava-v2}{B_f_both}{length_truncated}{0}
\chexdefskip{settings_openi_cxr-llava-v2}{B_f_both}{no_frontal}{163}
\chexdefskip{settings_openi_cxr-llava-v2}{B_f_both}{non_english}{0}
\chexdefskip{settings_openi_cxr-llava-v2}{B_f_both}{refused}{0}
\chexdefskip{settings_openi_cxr-llava-v2}{B_f_both}{truncations}{0}
\chexdefskip{settings_openi_cxr-llava-v2}{B_f_findings}{empty}{0}
\chexdefskip{settings_openi_cxr-llava-v2}{B_f_findings}{empty_target}{490}
\chexdefskip{settings_openi_cxr-llava-v2}{B_f_findings}{generated}{3198}
\chexdefskip{settings_openi_cxr-llava-v2}{B_f_findings}{length_truncated}{0}
\chexdefskip{settings_openi_cxr-llava-v2}{B_f_findings}{no_frontal}{163}
\chexdefskip{settings_openi_cxr-llava-v2}{B_f_findings}{non_english}{0}
\chexdefskip{settings_openi_cxr-llava-v2}{B_f_findings}{refused}{0}
\chexdefskip{settings_openi_cxr-llava-v2}{B_f_findings}{truncations}{0}
\chexdefskip{settings_openi_cxr-llava-v2}{B_f_impression}{empty}{0}
\chexdefskip{settings_openi_cxr-llava-v2}{B_f_impression}{empty_target}{29}
\chexdefskip{settings_openi_cxr-llava-v2}{B_f_impression}{generated}{3659}
\chexdefskip{settings_openi_cxr-llava-v2}{B_f_impression}{length_truncated}{0}
\chexdefskip{settings_openi_cxr-llava-v2}{B_f_impression}{no_frontal}{163}
\chexdefskip{settings_openi_cxr-llava-v2}{B_f_impression}{non_english}{0}
\chexdefskip{settings_openi_cxr-llava-v2}{B_f_impression}{refused}{0}
\chexdefskip{settings_openi_cxr-llava-v2}{B_f_impression}{truncations}{0}
\chexdefskip{settings_openi_cxr-llava-v2}{P_f_both}{empty}{0}
\chexdefskip{settings_openi_cxr-llava-v2}{P_f_both}{empty_target}{496}
\chexdefskip{settings_openi_cxr-llava-v2}{P_f_both}{generated}{3192}
\chexdefskip{settings_openi_cxr-llava-v2}{P_f_both}{length_truncated}{0}
\chexdefskip{settings_openi_cxr-llava-v2}{P_f_both}{no_frontal}{163}
\chexdefskip{settings_openi_cxr-llava-v2}{P_f_both}{non_english}{0}
\chexdefskip{settings_openi_cxr-llava-v2}{P_f_both}{refused}{0}
\chexdefskip{settings_openi_cxr-llava-v2}{P_f_both}{truncations}{0}
\chexdefskip{settings_openi_cxr-llava-v2}{P_f_findings}{empty}{0}
\chexdefskip{settings_openi_cxr-llava-v2}{P_f_findings}{empty_target}{490}
\chexdefskip{settings_openi_cxr-llava-v2}{P_f_findings}{generated}{3198}
\chexdefskip{settings_openi_cxr-llava-v2}{P_f_findings}{length_truncated}{0}
\chexdefskip{settings_openi_cxr-llava-v2}{P_f_findings}{no_frontal}{163}
\chexdefskip{settings_openi_cxr-llava-v2}{P_f_findings}{non_english}{0}
\chexdefskip{settings_openi_cxr-llava-v2}{P_f_findings}{refused}{0}
\chexdefskip{settings_openi_cxr-llava-v2}{P_f_findings}{truncations}{0}
\chexdefskip{settings_openi_cxr-llava-v2}{P_f_impression}{empty}{0}
\chexdefskip{settings_openi_cxr-llava-v2}{P_f_impression}{empty_target}{29}
\chexdefskip{settings_openi_cxr-llava-v2}{P_f_impression}{generated}{3659}
\chexdefskip{settings_openi_cxr-llava-v2}{P_f_impression}{length_truncated}{0}
\chexdefskip{settings_openi_cxr-llava-v2}{P_f_impression}{no_frontal}{163}
\chexdefskip{settings_openi_cxr-llava-v2}{P_f_impression}{non_english}{0}
\chexdefskip{settings_openi_cxr-llava-v2}{P_f_impression}{refused}{0}
\chexdefskip{settings_openi_cxr-llava-v2}{P_f_impression}{truncations}{0}
\chexdefskip{settings_openi_cxr-llava-v2}{P_n_both}{empty}{0}
\chexdefskip{settings_openi_cxr-llava-v2}{P_n_both}{generated}{0}
\chexdefskip{settings_openi_cxr-llava-v2}{P_n_both}{length_truncated}{0}
\chexdefskip{settings_openi_cxr-llava-v2}{P_n_both}{non_english}{0}
\chexdefskip{settings_openi_cxr-llava-v2}{P_n_both}{refused}{0}
\chexdefskip{settings_openi_cxr-llava-v2}{P_n_both}{text_only_unsupported}{3851}
\chexdefskip{settings_openi_cxr-llava-v2}{P_n_both}{truncations}{0}
\chexdefskip{settings_openi_cxr-llava-v2}{P_n_findings}{empty}{0}
\chexdefskip{settings_openi_cxr-llava-v2}{P_n_findings}{generated}{0}
\chexdefskip{settings_openi_cxr-llava-v2}{P_n_findings}{length_truncated}{0}
\chexdefskip{settings_openi_cxr-llava-v2}{P_n_findings}{non_english}{0}
\chexdefskip{settings_openi_cxr-llava-v2}{P_n_findings}{refused}{0}
\chexdefskip{settings_openi_cxr-llava-v2}{P_n_findings}{text_only_unsupported}{3851}
\chexdefskip{settings_openi_cxr-llava-v2}{P_n_findings}{truncations}{0}
\chexdefskip{settings_openi_cxr-llava-v2}{P_n_impression}{empty}{0}
\chexdefskip{settings_openi_cxr-llava-v2}{P_n_impression}{generated}{0}
\chexdefskip{settings_openi_cxr-llava-v2}{P_n_impression}{length_truncated}{0}
\chexdefskip{settings_openi_cxr-llava-v2}{P_n_impression}{non_english}{0}
\chexdefskip{settings_openi_cxr-llava-v2}{P_n_impression}{refused}{0}
\chexdefskip{settings_openi_cxr-llava-v2}{P_n_impression}{text_only_unsupported}{3851}
\chexdefskip{settings_openi_cxr-llava-v2}{P_n_impression}{truncations}{0}
\chexdefskip{settings_openi_empty}{B_f_both}{empty}{3192}
\chexdefskip{settings_openi_empty}{B_f_both}{empty_target}{496}
\chexdefskip{settings_openi_empty}{B_f_both}{generated}{3192}
\chexdefskip{settings_openi_empty}{B_f_both}{length_truncated}{0}
\chexdefskip{settings_openi_empty}{B_f_both}{no_frontal}{163}
\chexdefskip{settings_openi_empty}{B_f_both}{non_english}{0}
\chexdefskip{settings_openi_empty}{B_f_both}{refused}{0}
\chexdefskip{settings_openi_empty}{B_f_both}{truncations}{0}
\chexdefskip{settings_openi_empty}{B_f_findings}{empty}{3198}
\chexdefskip{settings_openi_empty}{B_f_findings}{empty_target}{490}
\chexdefskip{settings_openi_empty}{B_f_findings}{generated}{3198}
\chexdefskip{settings_openi_empty}{B_f_findings}{length_truncated}{0}
\chexdefskip{settings_openi_empty}{B_f_findings}{no_frontal}{163}
\chexdefskip{settings_openi_empty}{B_f_findings}{non_english}{0}
\chexdefskip{settings_openi_empty}{B_f_findings}{refused}{0}
\chexdefskip{settings_openi_empty}{B_f_findings}{truncations}{0}
\chexdefskip{settings_openi_empty}{B_f_impression}{empty}{3659}
\chexdefskip{settings_openi_empty}{B_f_impression}{empty_target}{29}
\chexdefskip{settings_openi_empty}{B_f_impression}{generated}{3659}
\chexdefskip{settings_openi_empty}{B_f_impression}{length_truncated}{0}
\chexdefskip{settings_openi_empty}{B_f_impression}{no_frontal}{163}
\chexdefskip{settings_openi_empty}{B_f_impression}{non_english}{0}
\chexdefskip{settings_openi_empty}{B_f_impression}{refused}{0}
\chexdefskip{settings_openi_empty}{B_f_impression}{truncations}{0}
\chexdefskip{settings_openi_empty}{P_f_both}{empty}{3192}
\chexdefskip{settings_openi_empty}{P_f_both}{empty_target}{496}
\chexdefskip{settings_openi_empty}{P_f_both}{generated}{3192}
\chexdefskip{settings_openi_empty}{P_f_both}{length_truncated}{0}
\chexdefskip{settings_openi_empty}{P_f_both}{no_frontal}{163}
\chexdefskip{settings_openi_empty}{P_f_both}{non_english}{0}
\chexdefskip{settings_openi_empty}{P_f_both}{refused}{0}
\chexdefskip{settings_openi_empty}{P_f_both}{truncations}{0}
\chexdefskip{settings_openi_empty}{P_f_findings}{empty}{3198}
\chexdefskip{settings_openi_empty}{P_f_findings}{empty_target}{490}
\chexdefskip{settings_openi_empty}{P_f_findings}{generated}{3198}
\chexdefskip{settings_openi_empty}{P_f_findings}{length_truncated}{0}
\chexdefskip{settings_openi_empty}{P_f_findings}{no_frontal}{163}
\chexdefskip{settings_openi_empty}{P_f_findings}{non_english}{0}
\chexdefskip{settings_openi_empty}{P_f_findings}{refused}{0}
\chexdefskip{settings_openi_empty}{P_f_findings}{truncations}{0}
\chexdefskip{settings_openi_empty}{P_f_impression}{empty}{3659}
\chexdefskip{settings_openi_empty}{P_f_impression}{empty_target}{29}
\chexdefskip{settings_openi_empty}{P_f_impression}{generated}{3659}
\chexdefskip{settings_openi_empty}{P_f_impression}{length_truncated}{0}
\chexdefskip{settings_openi_empty}{P_f_impression}{no_frontal}{163}
\chexdefskip{settings_openi_empty}{P_f_impression}{non_english}{0}
\chexdefskip{settings_openi_empty}{P_f_impression}{refused}{0}
\chexdefskip{settings_openi_empty}{P_f_impression}{truncations}{0}
\chexdefskip{settings_openi_empty}{P_n_both}{empty}{3331}
\chexdefskip{settings_openi_empty}{P_n_both}{empty_target}{520}
\chexdefskip{settings_openi_empty}{P_n_both}{generated}{3331}
\chexdefskip{settings_openi_empty}{P_n_both}{length_truncated}{0}
\chexdefskip{settings_openi_empty}{P_n_both}{non_english}{0}
\chexdefskip{settings_openi_empty}{P_n_both}{refused}{0}
\chexdefskip{settings_openi_empty}{P_n_both}{truncations}{0}
\chexdefskip{settings_openi_empty}{P_n_findings}{empty}{3337}
\chexdefskip{settings_openi_empty}{P_n_findings}{empty_target}{514}
\chexdefskip{settings_openi_empty}{P_n_findings}{generated}{3337}
\chexdefskip{settings_openi_empty}{P_n_findings}{length_truncated}{0}
\chexdefskip{settings_openi_empty}{P_n_findings}{non_english}{0}
\chexdefskip{settings_openi_empty}{P_n_findings}{refused}{0}
\chexdefskip{settings_openi_empty}{P_n_findings}{truncations}{0}
\chexdefskip{settings_openi_empty}{P_n_impression}{empty}{3820}
\chexdefskip{settings_openi_empty}{P_n_impression}{empty_target}{31}
\chexdefskip{settings_openi_empty}{P_n_impression}{generated}{3820}
\chexdefskip{settings_openi_empty}{P_n_impression}{length_truncated}{0}
\chexdefskip{settings_openi_empty}{P_n_impression}{non_english}{0}
\chexdefskip{settings_openi_empty}{P_n_impression}{refused}{0}
\chexdefskip{settings_openi_empty}{P_n_impression}{truncations}{0}
\chexdefskip{settings_openi_internvl3-8b}{B_f_findings}{empty}{0}
\chexdefskip{settings_openi_internvl3-8b}{B_f_findings}{empty_target}{490}
\chexdefskip{settings_openi_internvl3-8b}{B_f_findings}{generated}{3198}
\chexdefskip{settings_openi_internvl3-8b}{B_f_findings}{length_truncated}{0}
\chexdefskip{settings_openi_internvl3-8b}{B_f_findings}{no_frontal}{163}
\chexdefskip{settings_openi_internvl3-8b}{B_f_findings}{non_english}{0}
\chexdefskip{settings_openi_internvl3-8b}{B_f_findings}{refused}{0}
\chexdefskip{settings_openi_internvl3-8b}{B_f_findings}{truncations}{0}
\chexdefskip{settings_openi_llava-med}{B_f_both}{empty}{0}
\chexdefskip{settings_openi_llava-med}{B_f_both}{empty_target}{496}
\chexdefskip{settings_openi_llava-med}{B_f_both}{generated}{3192}
\chexdefskip{settings_openi_llava-med}{B_f_both}{length_truncated}{0}
\chexdefskip{settings_openi_llava-med}{B_f_both}{no_frontal}{163}
\chexdefskip{settings_openi_llava-med}{B_f_both}{non_english}{0}
\chexdefskip{settings_openi_llava-med}{B_f_both}{refused}{0}
\chexdefskip{settings_openi_llava-med}{B_f_both}{truncations}{0}
\chexdefskip{settings_openi_llava-med}{B_f_findings}{empty}{0}
\chexdefskip{settings_openi_llava-med}{B_f_findings}{empty_target}{490}
\chexdefskip{settings_openi_llava-med}{B_f_findings}{generated}{3198}
\chexdefskip{settings_openi_llava-med}{B_f_findings}{length_truncated}{0}
\chexdefskip{settings_openi_llava-med}{B_f_findings}{no_frontal}{163}
\chexdefskip{settings_openi_llava-med}{B_f_findings}{non_english}{0}
\chexdefskip{settings_openi_llava-med}{B_f_findings}{refused}{0}
\chexdefskip{settings_openi_llava-med}{B_f_findings}{truncations}{0}
\chexdefskip{settings_openi_llava-med}{B_f_impression}{empty}{0}
\chexdefskip{settings_openi_llava-med}{B_f_impression}{empty_target}{29}
\chexdefskip{settings_openi_llava-med}{B_f_impression}{generated}{3659}
\chexdefskip{settings_openi_llava-med}{B_f_impression}{length_truncated}{0}
\chexdefskip{settings_openi_llava-med}{B_f_impression}{no_frontal}{163}
\chexdefskip{settings_openi_llava-med}{B_f_impression}{non_english}{0}
\chexdefskip{settings_openi_llava-med}{B_f_impression}{refused}{0}
\chexdefskip{settings_openi_llava-med}{B_f_impression}{truncations}{0}
\chexdefskip{settings_openi_llava-med}{P_f_both}{empty}{3}
\chexdefskip{settings_openi_llava-med}{P_f_both}{empty_target}{496}
\chexdefskip{settings_openi_llava-med}{P_f_both}{generated}{3192}
\chexdefskip{settings_openi_llava-med}{P_f_both}{length_truncated}{0}
\chexdefskip{settings_openi_llava-med}{P_f_both}{no_frontal}{163}
\chexdefskip{settings_openi_llava-med}{P_f_both}{non_english}{0}
\chexdefskip{settings_openi_llava-med}{P_f_both}{refused}{0}
\chexdefskip{settings_openi_llava-med}{P_f_both}{truncations}{0}
\chexdefskip{settings_openi_llava-med}{P_f_findings}{empty}{20}
\chexdefskip{settings_openi_llava-med}{P_f_findings}{empty_target}{490}
\chexdefskip{settings_openi_llava-med}{P_f_findings}{generated}{3198}
\chexdefskip{settings_openi_llava-med}{P_f_findings}{length_truncated}{0}
\chexdefskip{settings_openi_llava-med}{P_f_findings}{no_frontal}{163}
\chexdefskip{settings_openi_llava-med}{P_f_findings}{non_english}{0}
\chexdefskip{settings_openi_llava-med}{P_f_findings}{refused}{0}
\chexdefskip{settings_openi_llava-med}{P_f_findings}{truncations}{0}
\chexdefskip{settings_openi_llava-med}{P_f_impression}{empty}{75}
\chexdefskip{settings_openi_llava-med}{P_f_impression}{empty_target}{29}
\chexdefskip{settings_openi_llava-med}{P_f_impression}{generated}{3659}
\chexdefskip{settings_openi_llava-med}{P_f_impression}{length_truncated}{0}
\chexdefskip{settings_openi_llava-med}{P_f_impression}{no_frontal}{163}
\chexdefskip{settings_openi_llava-med}{P_f_impression}{non_english}{0}
\chexdefskip{settings_openi_llava-med}{P_f_impression}{refused}{0}
\chexdefskip{settings_openi_llava-med}{P_f_impression}{truncations}{0}
\chexdefskip{settings_openi_llava-med}{P_n_both}{empty}{0}
\chexdefskip{settings_openi_llava-med}{P_n_both}{generated}{0}
\chexdefskip{settings_openi_llava-med}{P_n_both}{length_truncated}{0}
\chexdefskip{settings_openi_llava-med}{P_n_both}{non_english}{0}
\chexdefskip{settings_openi_llava-med}{P_n_both}{refused}{0}
\chexdefskip{settings_openi_llava-med}{P_n_both}{text_only_unsupported}{3851}
\chexdefskip{settings_openi_llava-med}{P_n_both}{truncations}{0}
\chexdefskip{settings_openi_llava-med}{P_n_findings}{empty}{0}
\chexdefskip{settings_openi_llava-med}{P_n_findings}{generated}{0}
\chexdefskip{settings_openi_llava-med}{P_n_findings}{length_truncated}{0}
\chexdefskip{settings_openi_llava-med}{P_n_findings}{non_english}{0}
\chexdefskip{settings_openi_llava-med}{P_n_findings}{refused}{0}
\chexdefskip{settings_openi_llava-med}{P_n_findings}{text_only_unsupported}{3851}
\chexdefskip{settings_openi_llava-med}{P_n_findings}{truncations}{0}
\chexdefskip{settings_openi_llava-med}{P_n_impression}{empty}{0}
\chexdefskip{settings_openi_llava-med}{P_n_impression}{generated}{0}
\chexdefskip{settings_openi_llava-med}{P_n_impression}{length_truncated}{0}
\chexdefskip{settings_openi_llava-med}{P_n_impression}{non_english}{0}
\chexdefskip{settings_openi_llava-med}{P_n_impression}{refused}{0}
\chexdefskip{settings_openi_llava-med}{P_n_impression}{text_only_unsupported}{3851}
\chexdefskip{settings_openi_llava-med}{P_n_impression}{truncations}{0}
\chexdefskip{settings_openi_llava-rad}{B_f_both}{empty}{0}
\chexdefskip{settings_openi_llava-rad}{B_f_both}{empty_target}{496}
\chexdefskip{settings_openi_llava-rad}{B_f_both}{generated}{3192}
\chexdefskip{settings_openi_llava-rad}{B_f_both}{length_truncated}{0}
\chexdefskip{settings_openi_llava-rad}{B_f_both}{no_frontal}{163}
\chexdefskip{settings_openi_llava-rad}{B_f_both}{non_english}{0}
\chexdefskip{settings_openi_llava-rad}{B_f_both}{refused}{0}
\chexdefskip{settings_openi_llava-rad}{B_f_both}{truncations}{0}
\chexdefskip{settings_openi_llava-rad}{B_f_findings}{empty}{0}
\chexdefskip{settings_openi_llava-rad}{B_f_findings}{empty_target}{490}
\chexdefskip{settings_openi_llava-rad}{B_f_findings}{generated}{3198}
\chexdefskip{settings_openi_llava-rad}{B_f_findings}{length_truncated}{0}
\chexdefskip{settings_openi_llava-rad}{B_f_findings}{no_frontal}{163}
\chexdefskip{settings_openi_llava-rad}{B_f_findings}{non_english}{0}
\chexdefskip{settings_openi_llava-rad}{B_f_findings}{refused}{0}
\chexdefskip{settings_openi_llava-rad}{B_f_findings}{truncations}{0}
\chexdefskip{settings_openi_llava-rad}{B_f_impression}{empty}{0}
\chexdefskip{settings_openi_llava-rad}{B_f_impression}{empty_target}{29}
\chexdefskip{settings_openi_llava-rad}{B_f_impression}{generated}{3659}
\chexdefskip{settings_openi_llava-rad}{B_f_impression}{length_truncated}{0}
\chexdefskip{settings_openi_llava-rad}{B_f_impression}{no_frontal}{163}
\chexdefskip{settings_openi_llava-rad}{B_f_impression}{non_english}{0}
\chexdefskip{settings_openi_llava-rad}{B_f_impression}{refused}{0}
\chexdefskip{settings_openi_llava-rad}{B_f_impression}{truncations}{0}
\chexdefskip{settings_openi_llava-rad}{P_f_both}{empty}{0}
\chexdefskip{settings_openi_llava-rad}{P_f_both}{empty_target}{496}
\chexdefskip{settings_openi_llava-rad}{P_f_both}{generated}{3192}
\chexdefskip{settings_openi_llava-rad}{P_f_both}{length_truncated}{0}
\chexdefskip{settings_openi_llava-rad}{P_f_both}{no_frontal}{163}
\chexdefskip{settings_openi_llava-rad}{P_f_both}{non_english}{0}
\chexdefskip{settings_openi_llava-rad}{P_f_both}{refused}{0}
\chexdefskip{settings_openi_llava-rad}{P_f_both}{truncations}{0}
\chexdefskip{settings_openi_llava-rad}{P_f_findings}{empty}{0}
\chexdefskip{settings_openi_llava-rad}{P_f_findings}{empty_target}{490}
\chexdefskip{settings_openi_llava-rad}{P_f_findings}{generated}{3198}
\chexdefskip{settings_openi_llava-rad}{P_f_findings}{length_truncated}{0}
\chexdefskip{settings_openi_llava-rad}{P_f_findings}{no_frontal}{163}
\chexdefskip{settings_openi_llava-rad}{P_f_findings}{non_english}{0}
\chexdefskip{settings_openi_llava-rad}{P_f_findings}{refused}{0}
\chexdefskip{settings_openi_llava-rad}{P_f_findings}{truncations}{0}
\chexdefskip{settings_openi_llava-rad}{P_f_impression}{empty}{0}
\chexdefskip{settings_openi_llava-rad}{P_f_impression}{empty_target}{29}
\chexdefskip{settings_openi_llava-rad}{P_f_impression}{generated}{3659}
\chexdefskip{settings_openi_llava-rad}{P_f_impression}{length_truncated}{0}
\chexdefskip{settings_openi_llava-rad}{P_f_impression}{no_frontal}{163}
\chexdefskip{settings_openi_llava-rad}{P_f_impression}{non_english}{0}
\chexdefskip{settings_openi_llava-rad}{P_f_impression}{refused}{0}
\chexdefskip{settings_openi_llava-rad}{P_f_impression}{truncations}{0}
\chexdefskip{settings_openi_llava-rad}{P_n_both}{empty}{0}
\chexdefskip{settings_openi_llava-rad}{P_n_both}{generated}{0}
\chexdefskip{settings_openi_llava-rad}{P_n_both}{length_truncated}{0}
\chexdefskip{settings_openi_llava-rad}{P_n_both}{non_english}{0}
\chexdefskip{settings_openi_llava-rad}{P_n_both}{refused}{0}
\chexdefskip{settings_openi_llava-rad}{P_n_both}{text_only_unsupported}{3851}
\chexdefskip{settings_openi_llava-rad}{P_n_both}{truncations}{0}
\chexdefskip{settings_openi_llava-rad}{P_n_findings}{empty}{0}
\chexdefskip{settings_openi_llava-rad}{P_n_findings}{generated}{0}
\chexdefskip{settings_openi_llava-rad}{P_n_findings}{length_truncated}{0}
\chexdefskip{settings_openi_llava-rad}{P_n_findings}{non_english}{0}
\chexdefskip{settings_openi_llava-rad}{P_n_findings}{refused}{0}
\chexdefskip{settings_openi_llava-rad}{P_n_findings}{text_only_unsupported}{3851}
\chexdefskip{settings_openi_llava-rad}{P_n_findings}{truncations}{0}
\chexdefskip{settings_openi_llava-rad}{P_n_impression}{empty}{0}
\chexdefskip{settings_openi_llava-rad}{P_n_impression}{generated}{0}
\chexdefskip{settings_openi_llava-rad}{P_n_impression}{length_truncated}{0}
\chexdefskip{settings_openi_llava-rad}{P_n_impression}{non_english}{0}
\chexdefskip{settings_openi_llava-rad}{P_n_impression}{refused}{0}
\chexdefskip{settings_openi_llava-rad}{P_n_impression}{text_only_unsupported}{3851}
\chexdefskip{settings_openi_llava-rad}{P_n_impression}{truncations}{0}
\chexdefskip{settings_openi_maira-2}{B_f_both}{empty}{0}
\chexdefskip{settings_openi_maira-2}{B_f_both}{empty_target}{496}
\chexdefskip{settings_openi_maira-2}{B_f_both}{generated}{3192}
\chexdefskip{settings_openi_maira-2}{B_f_both}{length_truncated}{0}
\chexdefskip{settings_openi_maira-2}{B_f_both}{no_frontal}{163}
\chexdefskip{settings_openi_maira-2}{B_f_both}{non_english}{0}
\chexdefskip{settings_openi_maira-2}{B_f_both}{refused}{0}
\chexdefskip{settings_openi_maira-2}{B_f_both}{truncations}{0}
\chexdefskip{settings_openi_maira-2}{B_f_findings}{empty}{0}
\chexdefskip{settings_openi_maira-2}{B_f_findings}{empty_target}{490}
\chexdefskip{settings_openi_maira-2}{B_f_findings}{generated}{3198}
\chexdefskip{settings_openi_maira-2}{B_f_findings}{length_truncated}{0}
\chexdefskip{settings_openi_maira-2}{B_f_findings}{no_frontal}{163}
\chexdefskip{settings_openi_maira-2}{B_f_findings}{non_english}{0}
\chexdefskip{settings_openi_maira-2}{B_f_findings}{refused}{0}
\chexdefskip{settings_openi_maira-2}{B_f_findings}{truncations}{0}
\chexdefskip{settings_openi_maira-2}{B_f_impression}{empty}{0}
\chexdefskip{settings_openi_maira-2}{B_f_impression}{empty_target}{29}
\chexdefskip{settings_openi_maira-2}{B_f_impression}{generated}{3659}
\chexdefskip{settings_openi_maira-2}{B_f_impression}{length_truncated}{0}
\chexdefskip{settings_openi_maira-2}{B_f_impression}{no_frontal}{163}
\chexdefskip{settings_openi_maira-2}{B_f_impression}{non_english}{0}
\chexdefskip{settings_openi_maira-2}{B_f_impression}{refused}{0}
\chexdefskip{settings_openi_maira-2}{B_f_impression}{truncations}{0}
\chexdefskip{settings_openi_maira-2}{P_f_both}{empty}{0}
\chexdefskip{settings_openi_maira-2}{P_f_both}{empty_target}{496}
\chexdefskip{settings_openi_maira-2}{P_f_both}{generated}{3192}
\chexdefskip{settings_openi_maira-2}{P_f_both}{length_truncated}{0}
\chexdefskip{settings_openi_maira-2}{P_f_both}{no_frontal}{163}
\chexdefskip{settings_openi_maira-2}{P_f_both}{non_english}{0}
\chexdefskip{settings_openi_maira-2}{P_f_both}{refused}{0}
\chexdefskip{settings_openi_maira-2}{P_f_both}{truncations}{0}
\chexdefskip{settings_openi_maira-2}{P_f_findings}{empty}{0}
\chexdefskip{settings_openi_maira-2}{P_f_findings}{empty_target}{490}
\chexdefskip{settings_openi_maira-2}{P_f_findings}{generated}{3198}
\chexdefskip{settings_openi_maira-2}{P_f_findings}{length_truncated}{0}
\chexdefskip{settings_openi_maira-2}{P_f_findings}{no_frontal}{163}
\chexdefskip{settings_openi_maira-2}{P_f_findings}{non_english}{0}
\chexdefskip{settings_openi_maira-2}{P_f_findings}{refused}{0}
\chexdefskip{settings_openi_maira-2}{P_f_findings}{truncations}{0}
\chexdefskip{settings_openi_maira-2}{P_f_impression}{empty}{0}
\chexdefskip{settings_openi_maira-2}{P_f_impression}{empty_target}{29}
\chexdefskip{settings_openi_maira-2}{P_f_impression}{generated}{3659}
\chexdefskip{settings_openi_maira-2}{P_f_impression}{length_truncated}{0}
\chexdefskip{settings_openi_maira-2}{P_f_impression}{no_frontal}{163}
\chexdefskip{settings_openi_maira-2}{P_f_impression}{non_english}{0}
\chexdefskip{settings_openi_maira-2}{P_f_impression}{refused}{0}
\chexdefskip{settings_openi_maira-2}{P_f_impression}{truncations}{0}
\chexdefskip{settings_openi_maira-2}{P_n_both}{empty}{0}
\chexdefskip{settings_openi_maira-2}{P_n_both}{generated}{0}
\chexdefskip{settings_openi_maira-2}{P_n_both}{length_truncated}{0}
\chexdefskip{settings_openi_maira-2}{P_n_both}{non_english}{0}
\chexdefskip{settings_openi_maira-2}{P_n_both}{refused}{0}
\chexdefskip{settings_openi_maira-2}{P_n_both}{text_only_unsupported}{3851}
\chexdefskip{settings_openi_maira-2}{P_n_both}{truncations}{0}
\chexdefskip{settings_openi_maira-2}{P_n_findings}{empty}{0}
\chexdefskip{settings_openi_maira-2}{P_n_findings}{generated}{0}
\chexdefskip{settings_openi_maira-2}{P_n_findings}{length_truncated}{0}
\chexdefskip{settings_openi_maira-2}{P_n_findings}{non_english}{0}
\chexdefskip{settings_openi_maira-2}{P_n_findings}{refused}{0}
\chexdefskip{settings_openi_maira-2}{P_n_findings}{text_only_unsupported}{3851}
\chexdefskip{settings_openi_maira-2}{P_n_findings}{truncations}{0}
\chexdefskip{settings_openi_maira-2}{P_n_impression}{empty}{0}
\chexdefskip{settings_openi_maira-2}{P_n_impression}{generated}{0}
\chexdefskip{settings_openi_maira-2}{P_n_impression}{length_truncated}{0}
\chexdefskip{settings_openi_maira-2}{P_n_impression}{non_english}{0}
\chexdefskip{settings_openi_maira-2}{P_n_impression}{refused}{0}
\chexdefskip{settings_openi_maira-2}{P_n_impression}{text_only_unsupported}{3851}
\chexdefskip{settings_openi_maira-2}{P_n_impression}{truncations}{0}
\chexdefskip{settings_openi_medgemma-4b}{B_f_both}{empty}{0}
\chexdefskip{settings_openi_medgemma-4b}{B_f_both}{empty_target}{496}
\chexdefskip{settings_openi_medgemma-4b}{B_f_both}{generated}{3192}
\chexdefskip{settings_openi_medgemma-4b}{B_f_both}{length_truncated}{0}
\chexdefskip{settings_openi_medgemma-4b}{B_f_both}{no_frontal}{163}
\chexdefskip{settings_openi_medgemma-4b}{B_f_both}{non_english}{0}
\chexdefskip{settings_openi_medgemma-4b}{B_f_both}{refused}{0}
\chexdefskip{settings_openi_medgemma-4b}{B_f_both}{truncations}{0}
\chexdefskip{settings_openi_medgemma-4b}{B_f_findings}{empty}{0}
\chexdefskip{settings_openi_medgemma-4b}{B_f_findings}{empty_target}{490}
\chexdefskip{settings_openi_medgemma-4b}{B_f_findings}{generated}{3198}
\chexdefskip{settings_openi_medgemma-4b}{B_f_findings}{length_truncated}{2}
\chexdefskip{settings_openi_medgemma-4b}{B_f_findings}{no_frontal}{163}
\chexdefskip{settings_openi_medgemma-4b}{B_f_findings}{non_english}{0}
\chexdefskip{settings_openi_medgemma-4b}{B_f_findings}{refused}{0}
\chexdefskip{settings_openi_medgemma-4b}{B_f_findings}{truncations}{0}
\chexdefskip{settings_openi_medgemma-4b}{B_f_impression}{empty}{0}
\chexdefskip{settings_openi_medgemma-4b}{B_f_impression}{empty_target}{29}
\chexdefskip{settings_openi_medgemma-4b}{B_f_impression}{generated}{3659}
\chexdefskip{settings_openi_medgemma-4b}{B_f_impression}{length_truncated}{1}
\chexdefskip{settings_openi_medgemma-4b}{B_f_impression}{no_frontal}{163}
\chexdefskip{settings_openi_medgemma-4b}{B_f_impression}{non_english}{0}
\chexdefskip{settings_openi_medgemma-4b}{B_f_impression}{refused}{0}
\chexdefskip{settings_openi_medgemma-4b}{B_f_impression}{truncations}{0}
\chexdefskip{settings_openi_medgemma-4b}{P_f_both}{empty}{0}
\chexdefskip{settings_openi_medgemma-4b}{P_f_both}{empty_target}{496}
\chexdefskip{settings_openi_medgemma-4b}{P_f_both}{generated}{3192}
\chexdefskip{settings_openi_medgemma-4b}{P_f_both}{length_truncated}{0}
\chexdefskip{settings_openi_medgemma-4b}{P_f_both}{no_frontal}{163}
\chexdefskip{settings_openi_medgemma-4b}{P_f_both}{non_english}{0}
\chexdefskip{settings_openi_medgemma-4b}{P_f_both}{refused}{0}
\chexdefskip{settings_openi_medgemma-4b}{P_f_both}{truncations}{0}
\chexdefskip{settings_openi_medgemma-4b}{P_f_findings}{empty}{0}
\chexdefskip{settings_openi_medgemma-4b}{P_f_findings}{empty_target}{490}
\chexdefskip{settings_openi_medgemma-4b}{P_f_findings}{generated}{3198}
\chexdefskip{settings_openi_medgemma-4b}{P_f_findings}{length_truncated}{0}
\chexdefskip{settings_openi_medgemma-4b}{P_f_findings}{no_frontal}{163}
\chexdefskip{settings_openi_medgemma-4b}{P_f_findings}{non_english}{0}
\chexdefskip{settings_openi_medgemma-4b}{P_f_findings}{refused}{0}
\chexdefskip{settings_openi_medgemma-4b}{P_f_findings}{truncations}{0}
\chexdefskip{settings_openi_medgemma-4b}{P_f_impression}{empty}{0}
\chexdefskip{settings_openi_medgemma-4b}{P_f_impression}{empty_target}{29}
\chexdefskip{settings_openi_medgemma-4b}{P_f_impression}{generated}{3659}
\chexdefskip{settings_openi_medgemma-4b}{P_f_impression}{length_truncated}{1}
\chexdefskip{settings_openi_medgemma-4b}{P_f_impression}{no_frontal}{163}
\chexdefskip{settings_openi_medgemma-4b}{P_f_impression}{non_english}{0}
\chexdefskip{settings_openi_medgemma-4b}{P_f_impression}{refused}{0}
\chexdefskip{settings_openi_medgemma-4b}{P_f_impression}{truncations}{0}
\chexdefskip{settings_openi_medgemma-4b}{P_n_both}{empty}{0}
\chexdefskip{settings_openi_medgemma-4b}{P_n_both}{empty_target}{520}
\chexdefskip{settings_openi_medgemma-4b}{P_n_both}{generated}{3331}
\chexdefskip{settings_openi_medgemma-4b}{P_n_both}{length_truncated}{0}
\chexdefskip{settings_openi_medgemma-4b}{P_n_both}{non_english}{0}
\chexdefskip{settings_openi_medgemma-4b}{P_n_both}{refused}{0}
\chexdefskip{settings_openi_medgemma-4b}{P_n_both}{truncations}{0}
\chexdefskip{settings_openi_medgemma-4b}{P_n_findings}{empty}{0}
\chexdefskip{settings_openi_medgemma-4b}{P_n_findings}{empty_target}{514}
\chexdefskip{settings_openi_medgemma-4b}{P_n_findings}{generated}{3337}
\chexdefskip{settings_openi_medgemma-4b}{P_n_findings}{length_truncated}{0}
\chexdefskip{settings_openi_medgemma-4b}{P_n_findings}{non_english}{0}
\chexdefskip{settings_openi_medgemma-4b}{P_n_findings}{refused}{0}
\chexdefskip{settings_openi_medgemma-4b}{P_n_findings}{truncations}{0}
\chexdefskip{settings_openi_medgemma-4b}{P_n_impression}{empty}{0}
\chexdefskip{settings_openi_medgemma-4b}{P_n_impression}{empty_target}{31}
\chexdefskip{settings_openi_medgemma-4b}{P_n_impression}{generated}{3820}
\chexdefskip{settings_openi_medgemma-4b}{P_n_impression}{length_truncated}{2}
\chexdefskip{settings_openi_medgemma-4b}{P_n_impression}{non_english}{0}
\chexdefskip{settings_openi_medgemma-4b}{P_n_impression}{refused}{0}
\chexdefskip{settings_openi_medgemma-4b}{P_n_impression}{truncations}{0}
\chexdefskip{settings_openi_most-common-report}{B_f_both}{empty}{0}
\chexdefskip{settings_openi_most-common-report}{B_f_both}{empty_target}{496}
\chexdefskip{settings_openi_most-common-report}{B_f_both}{generated}{3192}
\chexdefskip{settings_openi_most-common-report}{B_f_both}{length_truncated}{0}
\chexdefskip{settings_openi_most-common-report}{B_f_both}{no_frontal}{163}
\chexdefskip{settings_openi_most-common-report}{B_f_both}{non_english}{0}
\chexdefskip{settings_openi_most-common-report}{B_f_both}{refused}{0}
\chexdefskip{settings_openi_most-common-report}{B_f_both}{truncations}{0}
\chexdefskip{settings_openi_most-common-report}{B_f_findings}{empty}{0}
\chexdefskip{settings_openi_most-common-report}{B_f_findings}{empty_target}{490}
\chexdefskip{settings_openi_most-common-report}{B_f_findings}{generated}{3198}
\chexdefskip{settings_openi_most-common-report}{B_f_findings}{length_truncated}{0}
\chexdefskip{settings_openi_most-common-report}{B_f_findings}{no_frontal}{163}
\chexdefskip{settings_openi_most-common-report}{B_f_findings}{non_english}{0}
\chexdefskip{settings_openi_most-common-report}{B_f_findings}{refused}{0}
\chexdefskip{settings_openi_most-common-report}{B_f_findings}{truncations}{0}
\chexdefskip{settings_openi_most-common-report}{B_f_impression}{empty}{0}
\chexdefskip{settings_openi_most-common-report}{B_f_impression}{empty_target}{29}
\chexdefskip{settings_openi_most-common-report}{B_f_impression}{generated}{3659}
\chexdefskip{settings_openi_most-common-report}{B_f_impression}{length_truncated}{0}
\chexdefskip{settings_openi_most-common-report}{B_f_impression}{no_frontal}{163}
\chexdefskip{settings_openi_most-common-report}{B_f_impression}{non_english}{0}
\chexdefskip{settings_openi_most-common-report}{B_f_impression}{refused}{0}
\chexdefskip{settings_openi_most-common-report}{B_f_impression}{truncations}{0}
\chexdefskip{settings_openi_most-common-report}{P_f_both}{empty}{0}
\chexdefskip{settings_openi_most-common-report}{P_f_both}{empty_target}{496}
\chexdefskip{settings_openi_most-common-report}{P_f_both}{generated}{3192}
\chexdefskip{settings_openi_most-common-report}{P_f_both}{length_truncated}{0}
\chexdefskip{settings_openi_most-common-report}{P_f_both}{no_frontal}{163}
\chexdefskip{settings_openi_most-common-report}{P_f_both}{non_english}{0}
\chexdefskip{settings_openi_most-common-report}{P_f_both}{refused}{0}
\chexdefskip{settings_openi_most-common-report}{P_f_both}{truncations}{0}
\chexdefskip{settings_openi_most-common-report}{P_f_findings}{empty}{0}
\chexdefskip{settings_openi_most-common-report}{P_f_findings}{empty_target}{490}
\chexdefskip{settings_openi_most-common-report}{P_f_findings}{generated}{3198}
\chexdefskip{settings_openi_most-common-report}{P_f_findings}{length_truncated}{0}
\chexdefskip{settings_openi_most-common-report}{P_f_findings}{no_frontal}{163}
\chexdefskip{settings_openi_most-common-report}{P_f_findings}{non_english}{0}
\chexdefskip{settings_openi_most-common-report}{P_f_findings}{refused}{0}
\chexdefskip{settings_openi_most-common-report}{P_f_findings}{truncations}{0}
\chexdefskip{settings_openi_most-common-report}{P_f_impression}{empty}{0}
\chexdefskip{settings_openi_most-common-report}{P_f_impression}{empty_target}{29}
\chexdefskip{settings_openi_most-common-report}{P_f_impression}{generated}{3659}
\chexdefskip{settings_openi_most-common-report}{P_f_impression}{length_truncated}{0}
\chexdefskip{settings_openi_most-common-report}{P_f_impression}{no_frontal}{163}
\chexdefskip{settings_openi_most-common-report}{P_f_impression}{non_english}{0}
\chexdefskip{settings_openi_most-common-report}{P_f_impression}{refused}{0}
\chexdefskip{settings_openi_most-common-report}{P_f_impression}{truncations}{0}
\chexdefskip{settings_openi_most-common-report}{P_n_both}{empty}{0}
\chexdefskip{settings_openi_most-common-report}{P_n_both}{empty_target}{520}
\chexdefskip{settings_openi_most-common-report}{P_n_both}{generated}{3331}
\chexdefskip{settings_openi_most-common-report}{P_n_both}{length_truncated}{0}
\chexdefskip{settings_openi_most-common-report}{P_n_both}{non_english}{0}
\chexdefskip{settings_openi_most-common-report}{P_n_both}{refused}{0}
\chexdefskip{settings_openi_most-common-report}{P_n_both}{truncations}{0}
\chexdefskip{settings_openi_most-common-report}{P_n_findings}{empty}{0}
\chexdefskip{settings_openi_most-common-report}{P_n_findings}{empty_target}{514}
\chexdefskip{settings_openi_most-common-report}{P_n_findings}{generated}{3337}
\chexdefskip{settings_openi_most-common-report}{P_n_findings}{length_truncated}{0}
\chexdefskip{settings_openi_most-common-report}{P_n_findings}{non_english}{0}
\chexdefskip{settings_openi_most-common-report}{P_n_findings}{refused}{0}
\chexdefskip{settings_openi_most-common-report}{P_n_findings}{truncations}{0}
\chexdefskip{settings_openi_most-common-report}{P_n_impression}{empty}{0}
\chexdefskip{settings_openi_most-common-report}{P_n_impression}{empty_target}{31}
\chexdefskip{settings_openi_most-common-report}{P_n_impression}{generated}{3820}
\chexdefskip{settings_openi_most-common-report}{P_n_impression}{length_truncated}{0}
\chexdefskip{settings_openi_most-common-report}{P_n_impression}{non_english}{0}
\chexdefskip{settings_openi_most-common-report}{P_n_impression}{refused}{0}
\chexdefskip{settings_openi_most-common-report}{P_n_impression}{truncations}{0}
\chexdefskip{settings_openi_qwen2.5-vl-3b}{B_f_both}{empty}{0}
\chexdefskip{settings_openi_qwen2.5-vl-3b}{B_f_both}{empty_target}{496}
\chexdefskip{settings_openi_qwen2.5-vl-3b}{B_f_both}{generated}{3192}
\chexdefskip{settings_openi_qwen2.5-vl-3b}{B_f_both}{length_truncated}{78}
\chexdefskip{settings_openi_qwen2.5-vl-3b}{B_f_both}{no_frontal}{163}
\chexdefskip{settings_openi_qwen2.5-vl-3b}{B_f_both}{non_english}{0}
\chexdefskip{settings_openi_qwen2.5-vl-3b}{B_f_both}{refused}{0}
\chexdefskip{settings_openi_qwen2.5-vl-3b}{B_f_both}{truncations}{0}
\chexdefskip{settings_openi_qwen2.5-vl-3b}{B_f_findings}{empty}{0}
\chexdefskip{settings_openi_qwen2.5-vl-3b}{B_f_findings}{empty_target}{490}
\chexdefskip{settings_openi_qwen2.5-vl-3b}{B_f_findings}{generated}{3198}
\chexdefskip{settings_openi_qwen2.5-vl-3b}{B_f_findings}{length_truncated}{71}
\chexdefskip{settings_openi_qwen2.5-vl-3b}{B_f_findings}{no_frontal}{163}
\chexdefskip{settings_openi_qwen2.5-vl-3b}{B_f_findings}{non_english}{0}
\chexdefskip{settings_openi_qwen2.5-vl-3b}{B_f_findings}{refused}{0}
\chexdefskip{settings_openi_qwen2.5-vl-3b}{B_f_findings}{truncations}{0}
\chexdefskip{settings_openi_qwen2.5-vl-3b}{B_f_impression}{empty}{0}
\chexdefskip{settings_openi_qwen2.5-vl-3b}{B_f_impression}{empty_target}{29}
\chexdefskip{settings_openi_qwen2.5-vl-3b}{B_f_impression}{generated}{3659}
\chexdefskip{settings_openi_qwen2.5-vl-3b}{B_f_impression}{length_truncated}{0}
\chexdefskip{settings_openi_qwen2.5-vl-3b}{B_f_impression}{no_frontal}{163}
\chexdefskip{settings_openi_qwen2.5-vl-3b}{B_f_impression}{non_english}{0}
\chexdefskip{settings_openi_qwen2.5-vl-3b}{B_f_impression}{refused}{0}
\chexdefskip{settings_openi_qwen2.5-vl-3b}{B_f_impression}{truncations}{0}
\chexdefskip{settings_openi_qwen2.5-vl-3b}{P_f_both}{empty}{0}
\chexdefskip{settings_openi_qwen2.5-vl-3b}{P_f_both}{empty_target}{496}
\chexdefskip{settings_openi_qwen2.5-vl-3b}{P_f_both}{generated}{3192}
\chexdefskip{settings_openi_qwen2.5-vl-3b}{P_f_both}{length_truncated}{196}
\chexdefskip{settings_openi_qwen2.5-vl-3b}{P_f_both}{no_frontal}{163}
\chexdefskip{settings_openi_qwen2.5-vl-3b}{P_f_both}{non_english}{0}
\chexdefskip{settings_openi_qwen2.5-vl-3b}{P_f_both}{refused}{0}
\chexdefskip{settings_openi_qwen2.5-vl-3b}{P_f_both}{truncations}{0}
\chexdefskip{settings_openi_qwen2.5-vl-3b}{P_f_findings}{empty}{0}
\chexdefskip{settings_openi_qwen2.5-vl-3b}{P_f_findings}{empty_target}{490}
\chexdefskip{settings_openi_qwen2.5-vl-3b}{P_f_findings}{generated}{3198}
\chexdefskip{settings_openi_qwen2.5-vl-3b}{P_f_findings}{length_truncated}{19}
\chexdefskip{settings_openi_qwen2.5-vl-3b}{P_f_findings}{no_frontal}{163}
\chexdefskip{settings_openi_qwen2.5-vl-3b}{P_f_findings}{non_english}{0}
\chexdefskip{settings_openi_qwen2.5-vl-3b}{P_f_findings}{refused}{0}
\chexdefskip{settings_openi_qwen2.5-vl-3b}{P_f_findings}{truncations}{0}
\chexdefskip{settings_openi_qwen2.5-vl-3b}{P_f_impression}{empty}{0}
\chexdefskip{settings_openi_qwen2.5-vl-3b}{P_f_impression}{empty_target}{29}
\chexdefskip{settings_openi_qwen2.5-vl-3b}{P_f_impression}{generated}{3659}
\chexdefskip{settings_openi_qwen2.5-vl-3b}{P_f_impression}{length_truncated}{3}
\chexdefskip{settings_openi_qwen2.5-vl-3b}{P_f_impression}{no_frontal}{163}
\chexdefskip{settings_openi_qwen2.5-vl-3b}{P_f_impression}{non_english}{0}
\chexdefskip{settings_openi_qwen2.5-vl-3b}{P_f_impression}{refused}{0}
\chexdefskip{settings_openi_qwen2.5-vl-3b}{P_f_impression}{truncations}{0}
\chexdefskip{settings_openi_qwen2.5-vl-3b}{P_n_both}{empty}{0}
\chexdefskip{settings_openi_qwen2.5-vl-3b}{P_n_both}{empty_target}{520}
\chexdefskip{settings_openi_qwen2.5-vl-3b}{P_n_both}{generated}{3331}
\chexdefskip{settings_openi_qwen2.5-vl-3b}{P_n_both}{length_truncated}{205}
\chexdefskip{settings_openi_qwen2.5-vl-3b}{P_n_both}{non_english}{1}
\chexdefskip{settings_openi_qwen2.5-vl-3b}{P_n_both}{refused}{0}
\chexdefskip{settings_openi_qwen2.5-vl-3b}{P_n_both}{truncations}{0}
\chexdefskip{settings_openi_qwen2.5-vl-3b}{P_n_findings}{empty}{0}
\chexdefskip{settings_openi_qwen2.5-vl-3b}{P_n_findings}{empty_target}{514}
\chexdefskip{settings_openi_qwen2.5-vl-3b}{P_n_findings}{generated}{3337}
\chexdefskip{settings_openi_qwen2.5-vl-3b}{P_n_findings}{length_truncated}{32}
\chexdefskip{settings_openi_qwen2.5-vl-3b}{P_n_findings}{non_english}{0}
\chexdefskip{settings_openi_qwen2.5-vl-3b}{P_n_findings}{refused}{0}
\chexdefskip{settings_openi_qwen2.5-vl-3b}{P_n_findings}{truncations}{0}
\chexdefskip{settings_openi_qwen2.5-vl-3b}{P_n_impression}{empty}{0}
\chexdefskip{settings_openi_qwen2.5-vl-3b}{P_n_impression}{empty_target}{31}
\chexdefskip{settings_openi_qwen2.5-vl-3b}{P_n_impression}{generated}{3820}
\chexdefskip{settings_openi_qwen2.5-vl-3b}{P_n_impression}{length_truncated}{2}
\chexdefskip{settings_openi_qwen2.5-vl-3b}{P_n_impression}{non_english}{0}
\chexdefskip{settings_openi_qwen2.5-vl-3b}{P_n_impression}{refused}{0}
\chexdefskip{settings_openi_qwen2.5-vl-3b}{P_n_impression}{truncations}{0}
\chexdefskip{settings_openi_qwen2.5-vl-7b}{B_f_both}{empty}{0}
\chexdefskip{settings_openi_qwen2.5-vl-7b}{B_f_both}{empty_target}{496}
\chexdefskip{settings_openi_qwen2.5-vl-7b}{B_f_both}{generated}{3192}
\chexdefskip{settings_openi_qwen2.5-vl-7b}{B_f_both}{length_truncated}{0}
\chexdefskip{settings_openi_qwen2.5-vl-7b}{B_f_both}{no_frontal}{163}
\chexdefskip{settings_openi_qwen2.5-vl-7b}{B_f_both}{non_english}{0}
\chexdefskip{settings_openi_qwen2.5-vl-7b}{B_f_both}{refused}{0}
\chexdefskip{settings_openi_qwen2.5-vl-7b}{B_f_both}{truncations}{0}
\chexdefskip{settings_openi_qwen2.5-vl-7b}{B_f_findings}{empty}{0}
\chexdefskip{settings_openi_qwen2.5-vl-7b}{B_f_findings}{empty_target}{490}
\chexdefskip{settings_openi_qwen2.5-vl-7b}{B_f_findings}{generated}{3198}
\chexdefskip{settings_openi_qwen2.5-vl-7b}{B_f_findings}{length_truncated}{2}
\chexdefskip{settings_openi_qwen2.5-vl-7b}{B_f_findings}{no_frontal}{163}
\chexdefskip{settings_openi_qwen2.5-vl-7b}{B_f_findings}{non_english}{0}
\chexdefskip{settings_openi_qwen2.5-vl-7b}{B_f_findings}{refused}{0}
\chexdefskip{settings_openi_qwen2.5-vl-7b}{B_f_findings}{truncations}{0}
\chexdefskip{settings_openi_qwen2.5-vl-7b}{B_f_impression}{empty}{0}
\chexdefskip{settings_openi_qwen2.5-vl-7b}{B_f_impression}{empty_target}{29}
\chexdefskip{settings_openi_qwen2.5-vl-7b}{B_f_impression}{generated}{3659}
\chexdefskip{settings_openi_qwen2.5-vl-7b}{B_f_impression}{length_truncated}{0}
\chexdefskip{settings_openi_qwen2.5-vl-7b}{B_f_impression}{no_frontal}{163}
\chexdefskip{settings_openi_qwen2.5-vl-7b}{B_f_impression}{non_english}{0}
\chexdefskip{settings_openi_qwen2.5-vl-7b}{B_f_impression}{refused}{0}
\chexdefskip{settings_openi_qwen2.5-vl-7b}{B_f_impression}{truncations}{0}
\chexdefskip{settings_openi_qwen2.5-vl-7b}{P_f_both}{empty}{0}
\chexdefskip{settings_openi_qwen2.5-vl-7b}{P_f_both}{empty_target}{496}
\chexdefskip{settings_openi_qwen2.5-vl-7b}{P_f_both}{generated}{3192}
\chexdefskip{settings_openi_qwen2.5-vl-7b}{P_f_both}{length_truncated}{2}
\chexdefskip{settings_openi_qwen2.5-vl-7b}{P_f_both}{no_frontal}{163}
\chexdefskip{settings_openi_qwen2.5-vl-7b}{P_f_both}{non_english}{0}
\chexdefskip{settings_openi_qwen2.5-vl-7b}{P_f_both}{refused}{0}
\chexdefskip{settings_openi_qwen2.5-vl-7b}{P_f_both}{truncations}{0}
\chexdefskip{settings_openi_qwen2.5-vl-7b}{P_f_findings}{empty}{0}
\chexdefskip{settings_openi_qwen2.5-vl-7b}{P_f_findings}{empty_target}{490}
\chexdefskip{settings_openi_qwen2.5-vl-7b}{P_f_findings}{generated}{3198}
\chexdefskip{settings_openi_qwen2.5-vl-7b}{P_f_findings}{length_truncated}{7}
\chexdefskip{settings_openi_qwen2.5-vl-7b}{P_f_findings}{no_frontal}{163}
\chexdefskip{settings_openi_qwen2.5-vl-7b}{P_f_findings}{non_english}{0}
\chexdefskip{settings_openi_qwen2.5-vl-7b}{P_f_findings}{refused}{0}
\chexdefskip{settings_openi_qwen2.5-vl-7b}{P_f_findings}{truncations}{0}
\chexdefskip{settings_openi_qwen2.5-vl-7b}{P_f_impression}{empty}{0}
\chexdefskip{settings_openi_qwen2.5-vl-7b}{P_f_impression}{empty_target}{29}
\chexdefskip{settings_openi_qwen2.5-vl-7b}{P_f_impression}{generated}{3659}
\chexdefskip{settings_openi_qwen2.5-vl-7b}{P_f_impression}{length_truncated}{15}
\chexdefskip{settings_openi_qwen2.5-vl-7b}{P_f_impression}{no_frontal}{163}
\chexdefskip{settings_openi_qwen2.5-vl-7b}{P_f_impression}{non_english}{0}
\chexdefskip{settings_openi_qwen2.5-vl-7b}{P_f_impression}{refused}{0}
\chexdefskip{settings_openi_qwen2.5-vl-7b}{P_f_impression}{truncations}{0}
\chexdefskip{settings_openi_qwen2.5-vl-7b}{P_n_both}{empty}{0}
\chexdefskip{settings_openi_qwen2.5-vl-7b}{P_n_both}{empty_target}{520}
\chexdefskip{settings_openi_qwen2.5-vl-7b}{P_n_both}{generated}{3331}
\chexdefskip{settings_openi_qwen2.5-vl-7b}{P_n_both}{length_truncated}{3}
\chexdefskip{settings_openi_qwen2.5-vl-7b}{P_n_both}{non_english}{0}
\chexdefskip{settings_openi_qwen2.5-vl-7b}{P_n_both}{refused}{0}
\chexdefskip{settings_openi_qwen2.5-vl-7b}{P_n_both}{truncations}{0}
\chexdefskip{settings_openi_qwen2.5-vl-7b}{P_n_findings}{empty}{0}
\chexdefskip{settings_openi_qwen2.5-vl-7b}{P_n_findings}{empty_target}{514}
\chexdefskip{settings_openi_qwen2.5-vl-7b}{P_n_findings}{generated}{3337}
\chexdefskip{settings_openi_qwen2.5-vl-7b}{P_n_findings}{length_truncated}{19}
\chexdefskip{settings_openi_qwen2.5-vl-7b}{P_n_findings}{non_english}{0}
\chexdefskip{settings_openi_qwen2.5-vl-7b}{P_n_findings}{refused}{0}
\chexdefskip{settings_openi_qwen2.5-vl-7b}{P_n_findings}{truncations}{0}
\chexdefskip{settings_openi_qwen2.5-vl-7b}{P_n_impression}{empty}{0}
\chexdefskip{settings_openi_qwen2.5-vl-7b}{P_n_impression}{empty_target}{31}
\chexdefskip{settings_openi_qwen2.5-vl-7b}{P_n_impression}{generated}{3820}
\chexdefskip{settings_openi_qwen2.5-vl-7b}{P_n_impression}{length_truncated}{2}
\chexdefskip{settings_openi_qwen2.5-vl-7b}{P_n_impression}{non_english}{0}
\chexdefskip{settings_openi_qwen2.5-vl-7b}{P_n_impression}{refused}{1}
\chexdefskip{settings_openi_qwen2.5-vl-7b}{P_n_impression}{truncations}{0}
\chexdefskip{settings_openi_raddialog}{B_f_both}{empty}{0}
\chexdefskip{settings_openi_raddialog}{B_f_both}{empty_target}{496}
\chexdefskip{settings_openi_raddialog}{B_f_both}{generated}{3192}
\chexdefskip{settings_openi_raddialog}{B_f_both}{length_truncated}{0}
\chexdefskip{settings_openi_raddialog}{B_f_both}{no_frontal}{163}
\chexdefskip{settings_openi_raddialog}{B_f_both}{non_english}{0}
\chexdefskip{settings_openi_raddialog}{B_f_both}{refused}{0}
\chexdefskip{settings_openi_raddialog}{B_f_both}{truncations}{0}
\chexdefskip{settings_openi_raddialog}{B_f_findings}{empty}{0}
\chexdefskip{settings_openi_raddialog}{B_f_findings}{empty_target}{490}
\chexdefskip{settings_openi_raddialog}{B_f_findings}{generated}{3198}
\chexdefskip{settings_openi_raddialog}{B_f_findings}{length_truncated}{0}
\chexdefskip{settings_openi_raddialog}{B_f_findings}{no_frontal}{163}
\chexdefskip{settings_openi_raddialog}{B_f_findings}{non_english}{0}
\chexdefskip{settings_openi_raddialog}{B_f_findings}{refused}{0}
\chexdefskip{settings_openi_raddialog}{B_f_findings}{truncations}{0}
\chexdefskip{settings_openi_raddialog}{B_f_impression}{empty}{0}
\chexdefskip{settings_openi_raddialog}{B_f_impression}{empty_target}{29}
\chexdefskip{settings_openi_raddialog}{B_f_impression}{generated}{3659}
\chexdefskip{settings_openi_raddialog}{B_f_impression}{length_truncated}{0}
\chexdefskip{settings_openi_raddialog}{B_f_impression}{no_frontal}{163}
\chexdefskip{settings_openi_raddialog}{B_f_impression}{non_english}{0}
\chexdefskip{settings_openi_raddialog}{B_f_impression}{refused}{0}
\chexdefskip{settings_openi_raddialog}{B_f_impression}{truncations}{0}
\chexdefskip{settings_openi_raddialog}{P_f_both}{empty}{0}
\chexdefskip{settings_openi_raddialog}{P_f_both}{empty_target}{496}
\chexdefskip{settings_openi_raddialog}{P_f_both}{generated}{3192}
\chexdefskip{settings_openi_raddialog}{P_f_both}{length_truncated}{0}
\chexdefskip{settings_openi_raddialog}{P_f_both}{no_frontal}{163}
\chexdefskip{settings_openi_raddialog}{P_f_both}{non_english}{0}
\chexdefskip{settings_openi_raddialog}{P_f_both}{refused}{0}
\chexdefskip{settings_openi_raddialog}{P_f_both}{truncations}{0}
\chexdefskip{settings_openi_raddialog}{P_f_findings}{empty}{0}
\chexdefskip{settings_openi_raddialog}{P_f_findings}{empty_target}{490}
\chexdefskip{settings_openi_raddialog}{P_f_findings}{generated}{3198}
\chexdefskip{settings_openi_raddialog}{P_f_findings}{length_truncated}{0}
\chexdefskip{settings_openi_raddialog}{P_f_findings}{no_frontal}{163}
\chexdefskip{settings_openi_raddialog}{P_f_findings}{non_english}{0}
\chexdefskip{settings_openi_raddialog}{P_f_findings}{refused}{0}
\chexdefskip{settings_openi_raddialog}{P_f_findings}{truncations}{0}
\chexdefskip{settings_openi_raddialog}{P_f_impression}{empty}{0}
\chexdefskip{settings_openi_raddialog}{P_f_impression}{empty_target}{29}
\chexdefskip{settings_openi_raddialog}{P_f_impression}{generated}{3659}
\chexdefskip{settings_openi_raddialog}{P_f_impression}{length_truncated}{0}
\chexdefskip{settings_openi_raddialog}{P_f_impression}{no_frontal}{163}
\chexdefskip{settings_openi_raddialog}{P_f_impression}{non_english}{0}
\chexdefskip{settings_openi_raddialog}{P_f_impression}{refused}{0}
\chexdefskip{settings_openi_raddialog}{P_f_impression}{truncations}{0}
\chexdefskip{settings_openi_raddialog}{P_n_both}{empty}{0}
\chexdefskip{settings_openi_raddialog}{P_n_both}{generated}{0}
\chexdefskip{settings_openi_raddialog}{P_n_both}{length_truncated}{0}
\chexdefskip{settings_openi_raddialog}{P_n_both}{non_english}{0}
\chexdefskip{settings_openi_raddialog}{P_n_both}{refused}{0}
\chexdefskip{settings_openi_raddialog}{P_n_both}{text_only_unsupported}{3851}
\chexdefskip{settings_openi_raddialog}{P_n_both}{truncations}{0}
\chexdefskip{settings_openi_raddialog}{P_n_findings}{empty}{0}
\chexdefskip{settings_openi_raddialog}{P_n_findings}{generated}{0}
\chexdefskip{settings_openi_raddialog}{P_n_findings}{length_truncated}{0}
\chexdefskip{settings_openi_raddialog}{P_n_findings}{non_english}{0}
\chexdefskip{settings_openi_raddialog}{P_n_findings}{refused}{0}
\chexdefskip{settings_openi_raddialog}{P_n_findings}{text_only_unsupported}{3851}
\chexdefskip{settings_openi_raddialog}{P_n_findings}{truncations}{0}
\chexdefskip{settings_openi_raddialog}{P_n_impression}{empty}{0}
\chexdefskip{settings_openi_raddialog}{P_n_impression}{generated}{0}
\chexdefskip{settings_openi_raddialog}{P_n_impression}{length_truncated}{0}
\chexdefskip{settings_openi_raddialog}{P_n_impression}{non_english}{0}
\chexdefskip{settings_openi_raddialog}{P_n_impression}{refused}{0}
\chexdefskip{settings_openi_raddialog}{P_n_impression}{text_only_unsupported}{3851}
\chexdefskip{settings_openi_raddialog}{P_n_impression}{truncations}{0}
\chexdefskip{settings_openi_raddialog-prefixed}{B_f_both}{empty}{0}
\chexdefskip{settings_openi_raddialog-prefixed}{B_f_both}{empty_target}{496}
\chexdefskip{settings_openi_raddialog-prefixed}{B_f_both}{generated}{3192}
\chexdefskip{settings_openi_raddialog-prefixed}{B_f_both}{length_truncated}{0}
\chexdefskip{settings_openi_raddialog-prefixed}{B_f_both}{no_frontal}{163}
\chexdefskip{settings_openi_raddialog-prefixed}{B_f_both}{non_english}{0}
\chexdefskip{settings_openi_raddialog-prefixed}{B_f_both}{refused}{0}
\chexdefskip{settings_openi_raddialog-prefixed}{B_f_both}{truncations}{0}
\chexdefskip{settings_openi_raddialog-prefixed}{B_f_findings}{empty}{0}
\chexdefskip{settings_openi_raddialog-prefixed}{B_f_findings}{empty_target}{490}
\chexdefskip{settings_openi_raddialog-prefixed}{B_f_findings}{generated}{3198}
\chexdefskip{settings_openi_raddialog-prefixed}{B_f_findings}{length_truncated}{0}
\chexdefskip{settings_openi_raddialog-prefixed}{B_f_findings}{no_frontal}{163}
\chexdefskip{settings_openi_raddialog-prefixed}{B_f_findings}{non_english}{0}
\chexdefskip{settings_openi_raddialog-prefixed}{B_f_findings}{refused}{0}
\chexdefskip{settings_openi_raddialog-prefixed}{B_f_findings}{truncations}{0}
\chexdefskip{settings_openi_raddialog-prefixed}{B_f_impression}{empty}{0}
\chexdefskip{settings_openi_raddialog-prefixed}{B_f_impression}{empty_target}{29}
\chexdefskip{settings_openi_raddialog-prefixed}{B_f_impression}{generated}{3659}
\chexdefskip{settings_openi_raddialog-prefixed}{B_f_impression}{length_truncated}{0}
\chexdefskip{settings_openi_raddialog-prefixed}{B_f_impression}{no_frontal}{163}
\chexdefskip{settings_openi_raddialog-prefixed}{B_f_impression}{non_english}{0}
\chexdefskip{settings_openi_raddialog-prefixed}{B_f_impression}{refused}{0}
\chexdefskip{settings_openi_raddialog-prefixed}{B_f_impression}{truncations}{0}
\chexdefskip{settings_openi_raddialog-prefixed}{P_f_both}{empty}{0}
\chexdefskip{settings_openi_raddialog-prefixed}{P_f_both}{empty_target}{496}
\chexdefskip{settings_openi_raddialog-prefixed}{P_f_both}{generated}{3192}
\chexdefskip{settings_openi_raddialog-prefixed}{P_f_both}{length_truncated}{0}
\chexdefskip{settings_openi_raddialog-prefixed}{P_f_both}{no_frontal}{163}
\chexdefskip{settings_openi_raddialog-prefixed}{P_f_both}{non_english}{0}
\chexdefskip{settings_openi_raddialog-prefixed}{P_f_both}{refused}{0}
\chexdefskip{settings_openi_raddialog-prefixed}{P_f_both}{truncations}{0}
\chexdefskip{settings_openi_raddialog-prefixed}{P_f_findings}{empty}{0}
\chexdefskip{settings_openi_raddialog-prefixed}{P_f_findings}{empty_target}{490}
\chexdefskip{settings_openi_raddialog-prefixed}{P_f_findings}{generated}{3198}
\chexdefskip{settings_openi_raddialog-prefixed}{P_f_findings}{length_truncated}{0}
\chexdefskip{settings_openi_raddialog-prefixed}{P_f_findings}{no_frontal}{163}
\chexdefskip{settings_openi_raddialog-prefixed}{P_f_findings}{non_english}{0}
\chexdefskip{settings_openi_raddialog-prefixed}{P_f_findings}{refused}{0}
\chexdefskip{settings_openi_raddialog-prefixed}{P_f_findings}{truncations}{0}
\chexdefskip{settings_openi_raddialog-prefixed}{P_f_impression}{empty}{0}
\chexdefskip{settings_openi_raddialog-prefixed}{P_f_impression}{empty_target}{29}
\chexdefskip{settings_openi_raddialog-prefixed}{P_f_impression}{generated}{3659}
\chexdefskip{settings_openi_raddialog-prefixed}{P_f_impression}{length_truncated}{0}
\chexdefskip{settings_openi_raddialog-prefixed}{P_f_impression}{no_frontal}{163}
\chexdefskip{settings_openi_raddialog-prefixed}{P_f_impression}{non_english}{0}
\chexdefskip{settings_openi_raddialog-prefixed}{P_f_impression}{refused}{0}
\chexdefskip{settings_openi_raddialog-prefixed}{P_f_impression}{truncations}{0}
\chexdefskip{settings_openi_raddialog-prefixed}{P_n_both}{empty}{0}
\chexdefskip{settings_openi_raddialog-prefixed}{P_n_both}{generated}{0}
\chexdefskip{settings_openi_raddialog-prefixed}{P_n_both}{length_truncated}{0}
\chexdefskip{settings_openi_raddialog-prefixed}{P_n_both}{non_english}{0}
\chexdefskip{settings_openi_raddialog-prefixed}{P_n_both}{refused}{0}
\chexdefskip{settings_openi_raddialog-prefixed}{P_n_both}{text_only_unsupported}{3851}
\chexdefskip{settings_openi_raddialog-prefixed}{P_n_both}{truncations}{0}
\chexdefskip{settings_openi_raddialog-prefixed}{P_n_findings}{empty}{0}
\chexdefskip{settings_openi_raddialog-prefixed}{P_n_findings}{generated}{0}
\chexdefskip{settings_openi_raddialog-prefixed}{P_n_findings}{length_truncated}{0}
\chexdefskip{settings_openi_raddialog-prefixed}{P_n_findings}{non_english}{0}
\chexdefskip{settings_openi_raddialog-prefixed}{P_n_findings}{refused}{0}
\chexdefskip{settings_openi_raddialog-prefixed}{P_n_findings}{text_only_unsupported}{3851}
\chexdefskip{settings_openi_raddialog-prefixed}{P_n_findings}{truncations}{0}
\chexdefskip{settings_openi_raddialog-prefixed}{P_n_impression}{empty}{0}
\chexdefskip{settings_openi_raddialog-prefixed}{P_n_impression}{generated}{0}
\chexdefskip{settings_openi_raddialog-prefixed}{P_n_impression}{length_truncated}{0}
\chexdefskip{settings_openi_raddialog-prefixed}{P_n_impression}{non_english}{0}
\chexdefskip{settings_openi_raddialog-prefixed}{P_n_impression}{refused}{0}
\chexdefskip{settings_openi_raddialog-prefixed}{P_n_impression}{text_only_unsupported}{3851}
\chexdefskip{settings_openi_raddialog-prefixed}{P_n_impression}{truncations}{0}
\chexdefskip{settings_openi_random-report}{B_f_both}{empty}{0}
\chexdefskip{settings_openi_random-report}{B_f_both}{empty_target}{496}
\chexdefskip{settings_openi_random-report}{B_f_both}{generated}{3192}
\chexdefskip{settings_openi_random-report}{B_f_both}{length_truncated}{0}
\chexdefskip{settings_openi_random-report}{B_f_both}{no_frontal}{163}
\chexdefskip{settings_openi_random-report}{B_f_both}{non_english}{0}
\chexdefskip{settings_openi_random-report}{B_f_both}{refused}{2}
\chexdefskip{settings_openi_random-report}{B_f_both}{truncations}{0}
\chexdefskip{settings_openi_random-report}{B_f_findings}{empty}{0}
\chexdefskip{settings_openi_random-report}{B_f_findings}{empty_target}{490}
\chexdefskip{settings_openi_random-report}{B_f_findings}{generated}{3198}
\chexdefskip{settings_openi_random-report}{B_f_findings}{length_truncated}{0}
\chexdefskip{settings_openi_random-report}{B_f_findings}{no_frontal}{163}
\chexdefskip{settings_openi_random-report}{B_f_findings}{non_english}{0}
\chexdefskip{settings_openi_random-report}{B_f_findings}{refused}{0}
\chexdefskip{settings_openi_random-report}{B_f_findings}{truncations}{0}
\chexdefskip{settings_openi_random-report}{B_f_impression}{empty}{0}
\chexdefskip{settings_openi_random-report}{B_f_impression}{empty_target}{29}
\chexdefskip{settings_openi_random-report}{B_f_impression}{generated}{3659}
\chexdefskip{settings_openi_random-report}{B_f_impression}{length_truncated}{0}
\chexdefskip{settings_openi_random-report}{B_f_impression}{no_frontal}{163}
\chexdefskip{settings_openi_random-report}{B_f_impression}{non_english}{0}
\chexdefskip{settings_openi_random-report}{B_f_impression}{refused}{6}
\chexdefskip{settings_openi_random-report}{B_f_impression}{truncations}{0}
\chexdefskip{settings_openi_random-report}{P_f_both}{empty}{0}
\chexdefskip{settings_openi_random-report}{P_f_both}{empty_target}{496}
\chexdefskip{settings_openi_random-report}{P_f_both}{generated}{3192}
\chexdefskip{settings_openi_random-report}{P_f_both}{length_truncated}{0}
\chexdefskip{settings_openi_random-report}{P_f_both}{no_frontal}{163}
\chexdefskip{settings_openi_random-report}{P_f_both}{non_english}{0}
\chexdefskip{settings_openi_random-report}{P_f_both}{refused}{1}
\chexdefskip{settings_openi_random-report}{P_f_both}{truncations}{0}
\chexdefskip{settings_openi_random-report}{P_f_findings}{empty}{0}
\chexdefskip{settings_openi_random-report}{P_f_findings}{empty_target}{490}
\chexdefskip{settings_openi_random-report}{P_f_findings}{generated}{3198}
\chexdefskip{settings_openi_random-report}{P_f_findings}{length_truncated}{0}
\chexdefskip{settings_openi_random-report}{P_f_findings}{no_frontal}{163}
\chexdefskip{settings_openi_random-report}{P_f_findings}{non_english}{0}
\chexdefskip{settings_openi_random-report}{P_f_findings}{refused}{0}
\chexdefskip{settings_openi_random-report}{P_f_findings}{truncations}{0}
\chexdefskip{settings_openi_random-report}{P_f_impression}{empty}{0}
\chexdefskip{settings_openi_random-report}{P_f_impression}{empty_target}{29}
\chexdefskip{settings_openi_random-report}{P_f_impression}{generated}{3659}
\chexdefskip{settings_openi_random-report}{P_f_impression}{length_truncated}{0}
\chexdefskip{settings_openi_random-report}{P_f_impression}{no_frontal}{163}
\chexdefskip{settings_openi_random-report}{P_f_impression}{non_english}{0}
\chexdefskip{settings_openi_random-report}{P_f_impression}{refused}{6}
\chexdefskip{settings_openi_random-report}{P_f_impression}{truncations}{0}
\chexdefskip{settings_openi_random-report}{P_n_both}{empty}{0}
\chexdefskip{settings_openi_random-report}{P_n_both}{empty_target}{520}
\chexdefskip{settings_openi_random-report}{P_n_both}{generated}{3331}
\chexdefskip{settings_openi_random-report}{P_n_both}{length_truncated}{0}
\chexdefskip{settings_openi_random-report}{P_n_both}{non_english}{0}
\chexdefskip{settings_openi_random-report}{P_n_both}{refused}{3}
\chexdefskip{settings_openi_random-report}{P_n_both}{truncations}{0}
\chexdefskip{settings_openi_random-report}{P_n_findings}{empty}{0}
\chexdefskip{settings_openi_random-report}{P_n_findings}{empty_target}{514}
\chexdefskip{settings_openi_random-report}{P_n_findings}{generated}{3337}
\chexdefskip{settings_openi_random-report}{P_n_findings}{length_truncated}{0}
\chexdefskip{settings_openi_random-report}{P_n_findings}{non_english}{0}
\chexdefskip{settings_openi_random-report}{P_n_findings}{refused}{0}
\chexdefskip{settings_openi_random-report}{P_n_findings}{truncations}{0}
\chexdefskip{settings_openi_random-report}{P_n_impression}{empty}{0}
\chexdefskip{settings_openi_random-report}{P_n_impression}{empty_target}{31}
\chexdefskip{settings_openi_random-report}{P_n_impression}{generated}{3820}
\chexdefskip{settings_openi_random-report}{P_n_impression}{length_truncated}{0}
\chexdefskip{settings_openi_random-report}{P_n_impression}{non_english}{0}
\chexdefskip{settings_openi_random-report}{P_n_impression}{refused}{6}
\chexdefskip{settings_openi_random-report}{P_n_impression}{truncations}{0}
\chexdefskip{settings_openi_rgrg}{B_f_findings}{empty}{0}
\chexdefskip{settings_openi_rgrg}{B_f_findings}{empty_target}{490}
\chexdefskip{settings_openi_rgrg}{B_f_findings}{generated}{3198}
\chexdefskip{settings_openi_rgrg}{B_f_findings}{length_truncated}{0}
\chexdefskip{settings_openi_rgrg}{B_f_findings}{no_frontal}{163}
\chexdefskip{settings_openi_rgrg}{B_f_findings}{non_english}{0}
\chexdefskip{settings_openi_rgrg}{B_f_findings}{refused}{0}
\chexdefskip{settings_openi_rgrg}{B_f_findings}{truncations}{0}
\chexdefskip{settings_openi_rgrg}{P_f_findings}{empty}{0}
\chexdefskip{settings_openi_rgrg}{P_f_findings}{empty_target}{490}
\chexdefskip{settings_openi_rgrg}{P_f_findings}{generated}{3198}
\chexdefskip{settings_openi_rgrg}{P_f_findings}{length_truncated}{0}
\chexdefskip{settings_openi_rgrg}{P_f_findings}{no_frontal}{163}
\chexdefskip{settings_openi_rgrg}{P_f_findings}{non_english}{0}
\chexdefskip{settings_openi_rgrg}{P_f_findings}{refused}{0}
\chexdefskip{settings_openi_rgrg}{P_f_findings}{truncations}{0}

\chexdefmeta{mimic-critical}{ncases}{75}
\chexdefmeta{mimic-critical}{nclasses}{19}
\chexdefmeta{mimic-critical}{nfamilies}{7}
\chexdefmeta{mimic-critical}{npatients}{75}
\chexdefmeta{mimic-critical}{split}{test}
\chexdefmeta{mimic-critical}{version}{0.1.0}

\chexdefmeta{openi-critical}{dataset}{OpenI}
\chexdefmeta{openi-critical}{ncases}{30}
\chexdefmeta{openi-critical}{nclasses}{12}
\chexdefmeta{openi-critical}{nfamilies}{7}
\chexdefmeta{openi-critical}{npatients}{30}
\chexdefmeta{openi-critical}{split}{all (eval-only, open access)}
\chexdefmeta{openi-critical}{version}{0.1.0}

\chexdefmeta{radiopaedia-critical}{dataset}{Radiopaedia}
\chexdefmeta{radiopaedia-critical}{nclasses}{20}
\chexdefmeta{radiopaedia-critical}{nfamilies}{7}
\chexdefmeta{radiopaedia-critical}{split}{fixed set (eval-only)}
\chexdefmeta{radiopaedia-critical}{version}{0.1.0}

\chexdefmeta{MIMIC-CXR}{splitstudies}{3269}
\chexdefmeta{MIMIC-CXR}{splitpatients}{293}

\chexdefmeta{MIMIC-critical}{splitstudies}{75}
\chexdefmeta{MIMIC-critical}{splitpatients}{75}

\chexdefmeta{OpenI}{splitstudies}{3851}
\chexdefmeta{OpenI}{splitpatients}{3851}

\chexdefmeta{OpenI-critical}{splitstudies}{30}
\chexdefmeta{OpenI-critical}{splitpatients}{30}

\newcommand{\annAgreedPct}{82}
\newcommand{\annAnatomyCompletedPct}{92.2}
\newcommand{\annAnatomyWrittenPct}{81.8}

\newcommand{\annCertaintyKappa}{0.933}
\newcommand{\annChangeKappa}{0.932}

\newcommand{\annConcepts}{173}

\newcommand{\annCorpusSentences}{22{,}763}

\newcommand{\annNativeAgreePct}{97.1}

\newcommand{\annPaired}{22{,}764}
\newcommand{\annPresenceKappa}{0.989}

\newcommand{\annSeverityQWK}{0.948}

\newcommand{\silvBest}{Qwen3.5-9B}

\newcommand{\silvNeval}{5{,}562}

\newcommand{\silvBestStrict}{0.842}

\newcommand{\silvMedStrict}{0.836}

\newcommand{\silvSmallStrict}{0.808}
\newcommand{\silvTwoStrict}{0.823}

\newcommand{\silvEnsMemMin}{0.842}
\newcommand{\silvEnsMemMax}{0.849}

\newcommand{\silvMbr}{0.856}

\newcommand{\audCorrSent}{428}

\newcommand{\audCorrSentPct}{1.9}

\providecommand{\abl}[3]{\csname abl@#1@#2@#3\endcsname}
\expandafter\def\csname abl@frozen@a2@rg\endcsname{0.155}
\expandafter\def\csname abl@frozen@a2@mi\endcsname{0.542}
\expandafter\def\csname abl@frozen@a2@ma\endcsname{0.355}
\expandafter\def\csname abl@frozen@a2-checklist@rg\endcsname{0.218}
\expandafter\def\csname abl@frozen@a2-checklist@mi\endcsname{0.539}
\expandafter\def\csname abl@frozen@a2-checklist@ma\endcsname{0.345}
\expandafter\def\csname abl@lora@a2@rg\endcsname{0.163}
\expandafter\def\csname abl@lora@a2@mi\endcsname{0.565}
\expandafter\def\csname abl@lora@a2@ma\endcsname{0.380}
\expandafter\def\csname abl@lora@a2-checklist@rg\endcsname{0.224}
\expandafter\def\csname abl@lora@a2-checklist@mi\endcsname{0.556}
\expandafter\def\csname abl@lora@a2-checklist@ma\endcsname{0.369}
\expandafter\def\csname abl@ft@a2@rg\endcsname{0.165}
\expandafter\def\csname abl@ft@a2@mi\endcsname{0.566}
\expandafter\def\csname abl@ft@a2@ma\endcsname{0.381}
\expandafter\def\csname abl@ft@a2-checklist@rg\endcsname{0.228}
\expandafter\def\csname abl@ft@a2-checklist@mi\endcsname{0.562}
\expandafter\def\csname abl@ft@a2-checklist@ma\endcsname{0.380}
\expandafter\def\csname abl@repthr@a2@rg\endcsname{0.166}
\expandafter\def\csname abl@repthr@a2@mi\endcsname{0.568}
\expandafter\def\csname abl@repthr@a2@ma\endcsname{0.386}
\expandafter\def\csname abl@repthr@a2-checklist@rg\endcsname{0.226}
\expandafter\def\csname abl@repthr@a2-checklist@mi\endcsname{0.562}
\expandafter\def\csname abl@repthr@a2-checklist@ma\endcsname{0.387}
\expandafter\def\csname abl@cap@a2@rg\endcsname{0.166}
\expandafter\def\csname abl@cap@a2@mi\endcsname{0.569}
\expandafter\def\csname abl@cap@a2@ma\endcsname{0.396}
\expandafter\def\csname abl@cap@a2-checklist@rg\endcsname{0.224}
\expandafter\def\csname abl@cap@a2-checklist@mi\endcsname{0.566}
\expandafter\def\csname abl@cap@a2-checklist@ma\endcsname{0.395}
\expandafter\def\csname abl@full@tv3@rg\endcsname{0.254}
\expandafter\def\csname abl@full@tv3@mi\endcsname{0.612}
\expandafter\def\csname abl@full@tv3@ma\endcsname{0.397}
\expandafter\def\csname abl@gold@a2@rg\endcsname{0.258}
\expandafter\def\csname abl@gold@a2@mi\endcsname{0.800}
\expandafter\def\csname abl@gold@a2@ma\endcsname{0.750}
\expandafter\def\csname abl@gold@a2-checklist@rg\endcsname{0.309}
\expandafter\def\csname abl@gold@a2-checklist@mi\endcsname{0.801}
\expandafter\def\csname abl@gold@a2-checklist@ma\endcsname{0.748}
\expandafter\def\csname abl@gold@tv3@rg\endcsname{0.360}
\expandafter\def\csname abl@gold@tv3@mi\endcsname{0.852}
\expandafter\def\csname abl@gold@tv3@ma\endcsname{0.766}
\providecommand{\corehead}[1]{\csname corehead@#1\endcsname}
\expandafter\def\csname corehead@oursparams\endcsname{110M}
\expandafter\def\csname corehead@oursmi\endcsname{0.626}
\expandafter\def\csname corehead@bestmi\endcsname{0.592}
\expandafter\def\csname corehead@minparams\endcsname{0.6B}
\expandafter\def\csname corehead@maxparams\endcsname{9.4B}
\expandafter\def\csname corehead@oursma\endcsname{0.409}
\expandafter\def\csname corehead@bestma\endcsname{0.386}
\expandafter\def\csname corehead@nothers\endcsname{ten}
\expandafter\def\csname corehead@oursmiopeni\endcsname{0.556}
\expandafter\def\csname corehead@bestmiopeni\endcsname{0.549}
\expandafter\def\csname corehead@oursgreen\endcsname{0.289}
\expandafter\def\csname corehead@writergreen\endcsname{0.320}
\expandafter\def\csname corehead@writermi\endcsname{0.606}
\expandafter\def\csname corehead@sizeratio\endcsname{85}
\expandafter\def\csname corehead@flparams\endcsname{217M}
\expandafter\def\csname corehead@flminparams\endcsname{3.1B}
\expandafter\def\csname corehead@flmaxparams\endcsname{9.4B}

\providecommand{\detabl}[1]{\csname detabl@#1\endcsname}
\expandafter\def\csname detabl@clsmi\endcsname{0.480}
\expandafter\def\csname detabl@setmi\endcsname{0.484}
\expandafter\def\csname detabl@clsma\endcsname{0.188}
\expandafter\def\csname detabl@setma\endcsname{0.172}
\providecommand{\pd}[2]{\ifcsname pd@#1@#2\endcsname\csname pd@#1@#2\endcsname\else\errmessage{paired delta not found: #1 / #2}\fi}
\providecommand{\pdv}[2]{\ifcsname pdv@#1@#2\endcsname\csname pdv@#1@#2\endcsname\else\errmessage{paired delta not found: #1 / #2}\fi}
\expandafter\def\csname pdv@img:mimic:chexone-instruct@chexbert_micro_f1_14\endcsname{+0.277}
\expandafter\def\csname pd@img:mimic:chexone-instruct@chexbert_micro_f1_14\endcsname{+0.277 [+0.255, +0.298]}
\expandafter\def\csname pdv@img:mimic:chexone-instruct@chexbert_macro_f1_14\endcsname{+0.152}
\expandafter\def\csname pd@img:mimic:chexone-instruct@chexbert_macro_f1_14\endcsname{+0.152 [+0.132, +0.173]}
\expandafter\def\csname pdv@img:mimic:chexone-instruct@radgraph_partial\endcsname{+0.065}
\expandafter\def\csname pd@img:mimic:chexone-instruct@radgraph_partial\endcsname{+0.065 [+0.059, +0.072]}
\expandafter\def\csname pdv@img:mimic:chexone-instruct@radcliq_v1\endcsname{$-$0.32}
\expandafter\def\csname pd@img:mimic:chexone-instruct@radcliq_v1\endcsname{$-$0.32 [$-$0.34, $-$0.30]}
\expandafter\def\csname pdv@img:mimic:chexone-instruct@green\endcsname{+0.098}
\expandafter\def\csname pd@img:mimic:chexone-instruct@green\endcsname{+0.098 [+0.088, +0.108]}
\expandafter\def\csname pdv@img:mimic:internvl3-8b@chexbert_micro_f1_14\endcsname{+0.096}
\expandafter\def\csname pd@img:mimic:internvl3-8b@chexbert_micro_f1_14\endcsname{+0.096 [+0.076, +0.116]}
\expandafter\def\csname pdv@img:mimic:internvl3-8b@chexbert_macro_f1_14\endcsname{$-$0.026}
\expandafter\def\csname pd@img:mimic:internvl3-8b@chexbert_macro_f1_14\endcsname{$-$0.026 [$-$0.042, $-$0.008]}
\expandafter\def\csname pdv@img:mimic:internvl3-8b@radgraph_partial\endcsname{$-$0.000}
\expandafter\def\csname pd@img:mimic:internvl3-8b@radgraph_partial\endcsname{$-$0.000 [$-$0.004, +0.004]}
\expandafter\def\csname pdv@img:mimic:internvl3-8b@radcliq_v1\endcsname{+0.00}
\expandafter\def\csname pd@img:mimic:internvl3-8b@radcliq_v1\endcsname{+0.00 [$-$0.02, +0.02]}
\expandafter\def\csname pdv@img:mimic:internvl3-8b@green\endcsname{+0.002}
\expandafter\def\csname pd@img:mimic:internvl3-8b@green\endcsname{+0.002 [$-$0.008, +0.014]}
\expandafter\def\csname pdv@img:mimic:qwen2.5-vl-7b@chexbert_micro_f1_14\endcsname{+0.052}
\expandafter\def\csname pd@img:mimic:qwen2.5-vl-7b@chexbert_micro_f1_14\endcsname{+0.052 [+0.028, +0.077]}
\expandafter\def\csname pdv@img:mimic:qwen2.5-vl-7b@chexbert_macro_f1_14\endcsname{$-$0.079}
\expandafter\def\csname pd@img:mimic:qwen2.5-vl-7b@chexbert_macro_f1_14\endcsname{$-$0.079 [$-$0.097, $-$0.058]}
\expandafter\def\csname pdv@img:mimic:qwen2.5-vl-7b@radgraph_partial\endcsname{$-$0.029}
\expandafter\def\csname pd@img:mimic:qwen2.5-vl-7b@radgraph_partial\endcsname{$-$0.029 [$-$0.032, $-$0.025]}
\expandafter\def\csname pdv@img:mimic:qwen2.5-vl-7b@radcliq_v1\endcsname{+0.05}
\expandafter\def\csname pd@img:mimic:qwen2.5-vl-7b@radcliq_v1\endcsname{+0.05 [+0.03, +0.06]}
\expandafter\def\csname pdv@img:mimic:qwen2.5-vl-7b@green\endcsname{$-$0.023}
\expandafter\def\csname pd@img:mimic:qwen2.5-vl-7b@green\endcsname{$-$0.023 [$-$0.033, $-$0.013]}
\expandafter\def\csname pdv@img:mimic:medgemma-4b@chexbert_micro_f1_14\endcsname{+0.338}
\expandafter\def\csname pd@img:mimic:medgemma-4b@chexbert_micro_f1_14\endcsname{+0.338 [+0.314, +0.362]}
\expandafter\def\csname pdv@img:mimic:medgemma-4b@chexbert_macro_f1_14\endcsname{+0.151}
\expandafter\def\csname pd@img:mimic:medgemma-4b@chexbert_macro_f1_14\endcsname{+0.151 [+0.134, +0.170]}
\expandafter\def\csname pdv@img:mimic:medgemma-4b@radgraph_partial\endcsname{+0.031}
\expandafter\def\csname pd@img:mimic:medgemma-4b@radgraph_partial\endcsname{+0.031 [+0.026, +0.037]}
\expandafter\def\csname pdv@img:mimic:medgemma-4b@radcliq_v1\endcsname{$-$0.16}
\expandafter\def\csname pd@img:mimic:medgemma-4b@radcliq_v1\endcsname{$-$0.16 [$-$0.18, $-$0.14]}
\expandafter\def\csname pdv@img:mimic:medgemma-4b@green\endcsname{+0.080}
\expandafter\def\csname pd@img:mimic:medgemma-4b@green\endcsname{+0.080 [+0.071, +0.089]}
\expandafter\def\csname pdv@main:fsp-devB-c40-f21-tv3-chexone-instruct:mimic@chexbert_micro_f1_14\endcsname{+0.019}
\expandafter\def\csname pd@main:fsp-devB-c40-f21-tv3-chexone-instruct:mimic@chexbert_micro_f1_14\endcsname{+0.019 [+0.010, +0.029]}
\expandafter\def\csname pdv@main:fsp-devB-c40-f21-tv3-chexone-instruct:mimic@chexbert_macro_f1_14\endcsname{+0.058}
\expandafter\def\csname pd@main:fsp-devB-c40-f21-tv3-chexone-instruct:mimic@chexbert_macro_f1_14\endcsname{+0.058 [+0.040, +0.075]}
\expandafter\def\csname pdv@main:fsp-devB-c40-f21-tv3-chexone-instruct:mimic@radgraph_partial\endcsname{$-$0.046}
\expandafter\def\csname pd@main:fsp-devB-c40-f21-tv3-chexone-instruct:mimic@radgraph_partial\endcsname{$-$0.046 [$-$0.052, $-$0.041]}
\expandafter\def\csname pdv@main:fsp-devB-c40-f21-tv3-chexone-instruct:mimic@radcliq_v1\endcsname{+0.18}
\expandafter\def\csname pd@main:fsp-devB-c40-f21-tv3-chexone-instruct:mimic@radcliq_v1\endcsname{+0.18 [+0.16, +0.20]}
\expandafter\def\csname pdv@main:fsp-devB-c40-f21-tv3-chexone-instruct:mimic@green\endcsname{$-$0.038}
\expandafter\def\csname pd@main:fsp-devB-c40-f21-tv3-chexone-instruct:mimic@green\endcsname{$-$0.038 [$-$0.046, $-$0.030]}
\expandafter\def\csname pdv@main:fsp-devB-c40-f21-tv3-maira-2:mimic@chexbert_micro_f1_14\endcsname{+0.061}
\expandafter\def\csname pd@main:fsp-devB-c40-f21-tv3-maira-2:mimic@chexbert_micro_f1_14\endcsname{+0.061 [+0.049, +0.071]}
\expandafter\def\csname pdv@main:fsp-devB-c40-f21-tv3-maira-2:mimic@chexbert_macro_f1_14\endcsname{+0.060}
\expandafter\def\csname pd@main:fsp-devB-c40-f21-tv3-maira-2:mimic@chexbert_macro_f1_14\endcsname{+0.060 [+0.043, +0.078]}
\expandafter\def\csname pdv@main:fsp-devB-c40-f21-tv3-maira-2:mimic@radgraph_partial\endcsname{+0.006}
\expandafter\def\csname pd@main:fsp-devB-c40-f21-tv3-maira-2:mimic@radgraph_partial\endcsname{+0.006 [$-$0.000, +0.013]}
\expandafter\def\csname pdv@main:fsp-devB-c40-f21-tv3-maira-2:mimic@radcliq_v1\endcsname{+0.02}
\expandafter\def\csname pd@main:fsp-devB-c40-f21-tv3-maira-2:mimic@radcliq_v1\endcsname{+0.02 [$-$0.00, +0.03]}
\expandafter\def\csname pdv@main:fsp-devB-c40-f21-tv3-maira-2:mimic@green\endcsname{$-$0.028}
\expandafter\def\csname pd@main:fsp-devB-c40-f21-tv3-maira-2:mimic@green\endcsname{$-$0.028 [$-$0.036, $-$0.020]}
\expandafter\def\csname pdv@main:fsp-devB-c40-f21-lora2b-all-gold-chexone-instruct:mimic@chexbert_micro_f1_14\endcsname{+0.005}
\expandafter\def\csname pd@main:fsp-devB-c40-f21-lora2b-all-gold-chexone-instruct:mimic@chexbert_micro_f1_14\endcsname{+0.005 [$-$0.006, +0.015]}
\expandafter\def\csname pdv@main:fsp-devB-c40-f21-lora2b-all-gold-chexone-instruct:mimic@chexbert_macro_f1_14\endcsname{+0.040}
\expandafter\def\csname pd@main:fsp-devB-c40-f21-lora2b-all-gold-chexone-instruct:mimic@chexbert_macro_f1_14\endcsname{+0.040 [+0.022, +0.056]}
\expandafter\def\csname pdv@main:fsp-devB-c40-f21-lora2b-all-gold-chexone-instruct:mimic@radgraph_partial\endcsname{$-$0.035}
\expandafter\def\csname pd@main:fsp-devB-c40-f21-lora2b-all-gold-chexone-instruct:mimic@radgraph_partial\endcsname{$-$0.035 [$-$0.040, $-$0.030]}
\expandafter\def\csname pdv@main:fsp-devB-c40-f21-lora2b-all-gold-chexone-instruct:mimic@radcliq_v1\endcsname{+0.11}
\expandafter\def\csname pd@main:fsp-devB-c40-f21-lora2b-all-gold-chexone-instruct:mimic@radcliq_v1\endcsname{+0.11 [+0.10, +0.13]}
\expandafter\def\csname pdv@main:fsp-devB-c40-f21-lora2b-all-gold-chexone-instruct:mimic@green\endcsname{$-$0.009}
\expandafter\def\csname pd@main:fsp-devB-c40-f21-lora2b-all-gold-chexone-instruct:mimic@green\endcsname{$-$0.009 [$-$0.016, $-$0.002]}
\expandafter\def\csname pdv@main:fsp-devB-c40-f21-lora2b-all-gold-maira-2:mimic@chexbert_micro_f1_14\endcsname{+0.046}
\expandafter\def\csname pd@main:fsp-devB-c40-f21-lora2b-all-gold-maira-2:mimic@chexbert_micro_f1_14\endcsname{+0.046 [+0.035, +0.057]}
\expandafter\def\csname pdv@main:fsp-devB-c40-f21-lora2b-all-gold-maira-2:mimic@chexbert_macro_f1_14\endcsname{+0.041}
\expandafter\def\csname pd@main:fsp-devB-c40-f21-lora2b-all-gold-maira-2:mimic@chexbert_macro_f1_14\endcsname{+0.041 [+0.023, +0.060]}
\expandafter\def\csname pdv@main:fsp-devB-c40-f21-lora2b-all-gold-maira-2:mimic@radgraph_partial\endcsname{+0.017}
\expandafter\def\csname pd@main:fsp-devB-c40-f21-lora2b-all-gold-maira-2:mimic@radgraph_partial\endcsname{+0.017 [+0.012, +0.022]}
\expandafter\def\csname pdv@main:fsp-devB-c40-f21-lora2b-all-gold-maira-2:mimic@radcliq_v1\endcsname{$-$0.05}
\expandafter\def\csname pd@main:fsp-devB-c40-f21-lora2b-all-gold-maira-2:mimic@radcliq_v1\endcsname{$-$0.05 [$-$0.06, $-$0.03]}
\expandafter\def\csname pdv@main:fsp-devB-c40-f21-lora2b-all-gold-maira-2:mimic@green\endcsname{+0.000}
\expandafter\def\csname pd@main:fsp-devB-c40-f21-lora2b-all-gold-maira-2:mimic@green\endcsname{+0.000 [$-$0.007, +0.008]}
\expandafter\def\csname pdv@main:fsp-devB-c40-f21-lora2b-all-gold-fsp-devB-c40-f21-tv3:mimic@chexbert_micro_f1_14\endcsname{$-$0.015}
\expandafter\def\csname pd@main:fsp-devB-c40-f21-lora2b-all-gold-fsp-devB-c40-f21-tv3:mimic@chexbert_micro_f1_14\endcsname{$-$0.015 [$-$0.020, $-$0.009]}
\expandafter\def\csname pdv@main:fsp-devB-c40-f21-lora2b-all-gold-fsp-devB-c40-f21-tv3:mimic@chexbert_macro_f1_14\endcsname{$-$0.019}
\expandafter\def\csname pd@main:fsp-devB-c40-f21-lora2b-all-gold-fsp-devB-c40-f21-tv3:mimic@chexbert_macro_f1_14\endcsname{$-$0.019 [$-$0.029, $-$0.009]}
\expandafter\def\csname pdv@main:fsp-devB-c40-f21-lora2b-all-gold-fsp-devB-c40-f21-tv3:mimic@radgraph_partial\endcsname{+0.011}
\expandafter\def\csname pd@main:fsp-devB-c40-f21-lora2b-all-gold-fsp-devB-c40-f21-tv3:mimic@radgraph_partial\endcsname{+0.011 [+0.006, +0.016]}
\expandafter\def\csname pdv@main:fsp-devB-c40-f21-lora2b-all-gold-fsp-devB-c40-f21-tv3:mimic@radcliq_v1\endcsname{$-$0.06}
\expandafter\def\csname pd@main:fsp-devB-c40-f21-lora2b-all-gold-fsp-devB-c40-f21-tv3:mimic@radcliq_v1\endcsname{$-$0.06 [$-$0.08, $-$0.05]}
\expandafter\def\csname pdv@main:fsp-devB-c40-f21-lora2b-all-gold-fsp-devB-c40-f21-tv3:mimic@green\endcsname{+0.028}
\expandafter\def\csname pd@main:fsp-devB-c40-f21-lora2b-all-gold-fsp-devB-c40-f21-tv3:mimic@green\endcsname{+0.028 [+0.022, +0.035]}
\expandafter\def\csname pdv@img:openi:chexone-instruct@chexbert_micro_f1_14\endcsname{+0.305}
\expandafter\def\csname pd@img:openi:chexone-instruct@chexbert_micro_f1_14\endcsname{+0.305 [+0.285, +0.326]}
\expandafter\def\csname pdv@img:openi:chexone-instruct@chexbert_macro_f1_14\endcsname{+0.164}
\expandafter\def\csname pd@img:openi:chexone-instruct@chexbert_macro_f1_14\endcsname{+0.164 [+0.136, +0.191]}
\expandafter\def\csname pdv@img:openi:chexone-instruct@radgraph_partial\endcsname{+0.019}
\expandafter\def\csname pd@img:openi:chexone-instruct@radgraph_partial\endcsname{+0.019 [+0.015, +0.023]}
\expandafter\def\csname pdv@img:openi:chexone-instruct@radcliq_v1\endcsname{$-$0.08}
\expandafter\def\csname pd@img:openi:chexone-instruct@radcliq_v1\endcsname{$-$0.08 [$-$0.10, $-$0.07]}
\expandafter\def\csname pdv@img:openi:chexone-instruct@green\endcsname{+0.062}
\expandafter\def\csname pd@img:openi:chexone-instruct@green\endcsname{+0.062 [+0.055, +0.069]}
\expandafter\def\csname pdv@img:openi:internvl3-8b@chexbert_micro_f1_14\endcsname{+0.170}
\expandafter\def\csname pd@img:openi:internvl3-8b@chexbert_micro_f1_14\endcsname{+0.170 [+0.150, +0.190]}
\expandafter\def\csname pdv@img:openi:internvl3-8b@chexbert_macro_f1_14\endcsname{+0.014}
\expandafter\def\csname pd@img:openi:internvl3-8b@chexbert_macro_f1_14\endcsname{+0.014 [$-$0.009, +0.036]}
\expandafter\def\csname pdv@img:openi:internvl3-8b@radgraph_partial\endcsname{+0.052}
\expandafter\def\csname pd@img:openi:internvl3-8b@radgraph_partial\endcsname{+0.052 [+0.049, +0.056]}
\expandafter\def\csname pdv@img:openi:internvl3-8b@radcliq_v1\endcsname{$-$0.18}
\expandafter\def\csname pd@img:openi:internvl3-8b@radcliq_v1\endcsname{$-$0.18 [$-$0.19, $-$0.16]}
\expandafter\def\csname pdv@img:openi:internvl3-8b@green\endcsname{+0.328}
\expandafter\def\csname pd@img:openi:internvl3-8b@green\endcsname{+0.328 [+0.315, +0.342]}
\expandafter\def\csname pdv@img:openi:qwen2.5-vl-7b@chexbert_micro_f1_14\endcsname{+0.022}
\expandafter\def\csname pd@img:openi:qwen2.5-vl-7b@chexbert_micro_f1_14\endcsname{+0.022 [+0.002, +0.041]}
\expandafter\def\csname pdv@img:openi:qwen2.5-vl-7b@chexbert_macro_f1_14\endcsname{$-$0.039}
\expandafter\def\csname pd@img:openi:qwen2.5-vl-7b@chexbert_macro_f1_14\endcsname{$-$0.039 [$-$0.056, $-$0.022]}
\expandafter\def\csname pdv@img:openi:qwen2.5-vl-7b@radgraph_partial\endcsname{$-$0.017}
\expandafter\def\csname pd@img:openi:qwen2.5-vl-7b@radgraph_partial\endcsname{$-$0.017 [$-$0.020, $-$0.015]}
\expandafter\def\csname pdv@img:openi:qwen2.5-vl-7b@radcliq_v1\endcsname{$-$0.02}
\expandafter\def\csname pd@img:openi:qwen2.5-vl-7b@radcliq_v1\endcsname{$-$0.02 [$-$0.04, $-$0.01]}
\expandafter\def\csname pdv@img:openi:qwen2.5-vl-7b@green\endcsname{$-$0.067}
\expandafter\def\csname pd@img:openi:qwen2.5-vl-7b@green\endcsname{$-$0.067 [$-$0.078, $-$0.057]}
\expandafter\def\csname pdv@img:openi:medgemma-4b@chexbert_micro_f1_14\endcsname{+0.068}
\expandafter\def\csname pd@img:openi:medgemma-4b@chexbert_micro_f1_14\endcsname{+0.068 [+0.052, +0.082]}
\expandafter\def\csname pdv@img:openi:medgemma-4b@chexbert_macro_f1_14\endcsname{+0.118}
\expandafter\def\csname pd@img:openi:medgemma-4b@chexbert_macro_f1_14\endcsname{+0.118 [+0.096, +0.140]}
\expandafter\def\csname pdv@img:openi:medgemma-4b@radgraph_partial\endcsname{+0.003}
\expandafter\def\csname pd@img:openi:medgemma-4b@radgraph_partial\endcsname{+0.003 [+0.000, +0.005]}
\expandafter\def\csname pdv@img:openi:medgemma-4b@radcliq_v1\endcsname{$-$0.02}
\expandafter\def\csname pd@img:openi:medgemma-4b@radcliq_v1\endcsname{$-$0.02 [$-$0.03, $-$0.01]}
\expandafter\def\csname pdv@img:openi:medgemma-4b@green\endcsname{$-$0.002}
\expandafter\def\csname pd@img:openi:medgemma-4b@green\endcsname{$-$0.002 [$-$0.007, +0.004]}
\expandafter\def\csname pdv@main:fsp-devB-c40-f21-tv3-chexone-instruct:openi@chexbert_micro_f1_14\endcsname{$-$0.009}
\expandafter\def\csname pd@main:fsp-devB-c40-f21-tv3-chexone-instruct:openi@chexbert_micro_f1_14\endcsname{$-$0.009 [$-$0.021, +0.006]}
\expandafter\def\csname pdv@main:fsp-devB-c40-f21-tv3-chexone-instruct:openi@chexbert_macro_f1_14\endcsname{+0.061}
\expandafter\def\csname pd@main:fsp-devB-c40-f21-tv3-chexone-instruct:openi@chexbert_macro_f1_14\endcsname{+0.061 [+0.032, +0.088]}
\expandafter\def\csname pdv@main:fsp-devB-c40-f21-tv3-chexone-instruct:openi@radgraph_partial\endcsname{$-$0.014}
\expandafter\def\csname pd@main:fsp-devB-c40-f21-tv3-chexone-instruct:openi@radgraph_partial\endcsname{$-$0.014 [$-$0.018, $-$0.009]}
\expandafter\def\csname pdv@main:fsp-devB-c40-f21-tv3-chexone-instruct:openi@radcliq_v1\endcsname{+0.04}
\expandafter\def\csname pd@main:fsp-devB-c40-f21-tv3-chexone-instruct:openi@radcliq_v1\endcsname{+0.04 [+0.03, +0.05]}
\expandafter\def\csname pdv@main:fsp-devB-c40-f21-tv3-chexone-instruct:openi@green\endcsname{$-$0.114}
\expandafter\def\csname pd@main:fsp-devB-c40-f21-tv3-chexone-instruct:openi@green\endcsname{$-$0.114 [$-$0.121, $-$0.107]}
\expandafter\def\csname pdv@main:fsp-devB-c40-f21-tv3-maira-2:openi@chexbert_micro_f1_14\endcsname{+0.017}
\expandafter\def\csname pd@main:fsp-devB-c40-f21-tv3-maira-2:openi@chexbert_micro_f1_14\endcsname{+0.017 [+0.003, +0.032]}
\expandafter\def\csname pdv@main:fsp-devB-c40-f21-tv3-maira-2:openi@chexbert_macro_f1_14\endcsname{+0.049}
\expandafter\def\csname pd@main:fsp-devB-c40-f21-tv3-maira-2:openi@chexbert_macro_f1_14\endcsname{+0.049 [+0.017, +0.078]}
\expandafter\def\csname pdv@main:fsp-devB-c40-f21-tv3-maira-2:openi@radgraph_partial\endcsname{$-$0.002}
\expandafter\def\csname pd@main:fsp-devB-c40-f21-tv3-maira-2:openi@radgraph_partial\endcsname{$-$0.002 [$-$0.007, +0.002]}
\expandafter\def\csname pdv@main:fsp-devB-c40-f21-tv3-maira-2:openi@radcliq_v1\endcsname{$-$0.01}
\expandafter\def\csname pd@main:fsp-devB-c40-f21-tv3-maira-2:openi@radcliq_v1\endcsname{$-$0.01 [$-$0.02, $-$0.00]}
\expandafter\def\csname pdv@main:fsp-devB-c40-f21-tv3-maira-2:openi@green\endcsname{$-$0.103}
\expandafter\def\csname pd@main:fsp-devB-c40-f21-tv3-maira-2:openi@green\endcsname{$-$0.103 [$-$0.109, $-$0.096]}
\expandafter\def\csname pdv@main:fsp-devB-c40-f21-lora2b-all-gold-chexone-instruct:openi@chexbert_micro_f1_14\endcsname{$-$0.008}
\expandafter\def\csname pd@main:fsp-devB-c40-f21-lora2b-all-gold-chexone-instruct:openi@chexbert_micro_f1_14\endcsname{$-$0.008 [$-$0.022, +0.006]}
\expandafter\def\csname pdv@main:fsp-devB-c40-f21-lora2b-all-gold-chexone-instruct:openi@chexbert_macro_f1_14\endcsname{+0.051}
\expandafter\def\csname pd@main:fsp-devB-c40-f21-lora2b-all-gold-chexone-instruct:openi@chexbert_macro_f1_14\endcsname{+0.051 [+0.021, +0.080]}
\expandafter\def\csname pdv@main:fsp-devB-c40-f21-lora2b-all-gold-chexone-instruct:openi@radgraph_partial\endcsname{$-$0.058}
\expandafter\def\csname pd@main:fsp-devB-c40-f21-lora2b-all-gold-chexone-instruct:openi@radgraph_partial\endcsname{$-$0.058 [$-$0.062, $-$0.055]}
\expandafter\def\csname pdv@main:fsp-devB-c40-f21-lora2b-all-gold-chexone-instruct:openi@radcliq_v1\endcsname{+0.12}
\expandafter\def\csname pd@main:fsp-devB-c40-f21-lora2b-all-gold-chexone-instruct:openi@radcliq_v1\endcsname{+0.12 [+0.11, +0.13]}
\expandafter\def\csname pdv@main:fsp-devB-c40-f21-lora2b-all-gold-chexone-instruct:openi@green\endcsname{$-$0.034}
\expandafter\def\csname pd@main:fsp-devB-c40-f21-lora2b-all-gold-chexone-instruct:openi@green\endcsname{$-$0.034 [$-$0.040, $-$0.026]}
\expandafter\def\csname pdv@main:fsp-devB-c40-f21-lora2b-all-gold-maira-2:openi@chexbert_micro_f1_14\endcsname{+0.017}
\expandafter\def\csname pd@main:fsp-devB-c40-f21-lora2b-all-gold-maira-2:openi@chexbert_micro_f1_14\endcsname{+0.017 [+0.003, +0.032]}
\expandafter\def\csname pdv@main:fsp-devB-c40-f21-lora2b-all-gold-maira-2:openi@chexbert_macro_f1_14\endcsname{+0.040}
\expandafter\def\csname pd@main:fsp-devB-c40-f21-lora2b-all-gold-maira-2:openi@chexbert_macro_f1_14\endcsname{+0.040 [+0.009, +0.069]}
\expandafter\def\csname pdv@main:fsp-devB-c40-f21-lora2b-all-gold-maira-2:openi@radgraph_partial\endcsname{$-$0.047}
\expandafter\def\csname pd@main:fsp-devB-c40-f21-lora2b-all-gold-maira-2:openi@radgraph_partial\endcsname{$-$0.047 [$-$0.051, $-$0.043]}
\expandafter\def\csname pdv@main:fsp-devB-c40-f21-lora2b-all-gold-maira-2:openi@radcliq_v1\endcsname{+0.07}
\expandafter\def\csname pd@main:fsp-devB-c40-f21-lora2b-all-gold-maira-2:openi@radcliq_v1\endcsname{+0.07 [+0.06, +0.08]}
\expandafter\def\csname pdv@main:fsp-devB-c40-f21-lora2b-all-gold-maira-2:openi@green\endcsname{$-$0.022}
\expandafter\def\csname pd@main:fsp-devB-c40-f21-lora2b-all-gold-maira-2:openi@green\endcsname{$-$0.022 [$-$0.029, $-$0.015]}
\expandafter\def\csname pdv@main:fsp-devB-c40-f21-lora2b-all-gold-fsp-devB-c40-f21-tv3:openi@chexbert_micro_f1_14\endcsname{+0.000}
\expandafter\def\csname pd@main:fsp-devB-c40-f21-lora2b-all-gold-fsp-devB-c40-f21-tv3:openi@chexbert_micro_f1_14\endcsname{+0.000 [$-$0.007, +0.008]}
\expandafter\def\csname pdv@main:fsp-devB-c40-f21-lora2b-all-gold-fsp-devB-c40-f21-tv3:openi@chexbert_macro_f1_14\endcsname{$-$0.010}
\expandafter\def\csname pd@main:fsp-devB-c40-f21-lora2b-all-gold-fsp-devB-c40-f21-tv3:openi@chexbert_macro_f1_14\endcsname{$-$0.010 [$-$0.027, +0.006]}
\expandafter\def\csname pdv@main:fsp-devB-c40-f21-lora2b-all-gold-fsp-devB-c40-f21-tv3:openi@radgraph_partial\endcsname{$-$0.045}
\expandafter\def\csname pd@main:fsp-devB-c40-f21-lora2b-all-gold-fsp-devB-c40-f21-tv3:openi@radgraph_partial\endcsname{$-$0.045 [$-$0.049, $-$0.041]}
\expandafter\def\csname pdv@main:fsp-devB-c40-f21-lora2b-all-gold-fsp-devB-c40-f21-tv3:openi@radcliq_v1\endcsname{+0.08}
\expandafter\def\csname pd@main:fsp-devB-c40-f21-lora2b-all-gold-fsp-devB-c40-f21-tv3:openi@radcliq_v1\endcsname{+0.08 [+0.07, +0.09]}
\expandafter\def\csname pdv@main:fsp-devB-c40-f21-lora2b-all-gold-fsp-devB-c40-f21-tv3:openi@green\endcsname{+0.080}
\expandafter\def\csname pd@main:fsp-devB-c40-f21-lora2b-all-gold-fsp-devB-c40-f21-tv3:openi@green\endcsname{+0.080 [+0.074, +0.086]}
\expandafter\def\csname pdv@abl:lora-frozen:a2@chexbert_micro_f1_14\endcsname{+0.023}
\expandafter\def\csname pd@abl:lora-frozen:a2@chexbert_micro_f1_14\endcsname{+0.023 [+0.014, +0.032]}
\expandafter\def\csname pdv@abl:lora-frozen:a2@chexbert_macro_f1_14\endcsname{+0.025}
\expandafter\def\csname pd@abl:lora-frozen:a2@chexbert_macro_f1_14\endcsname{+0.025 [+0.008, +0.043]}
\expandafter\def\csname pdv@abl:lora-frozen:a2@radgraph_partial\endcsname{+0.007}
\expandafter\def\csname pd@abl:lora-frozen:a2@radgraph_partial\endcsname{+0.007 [+0.005, +0.010]}
\expandafter\def\csname pdv@abl:lora-frozen:a2@radcliq_v1\endcsname{$-$0.05}
\expandafter\def\csname pd@abl:lora-frozen:a2@radcliq_v1\endcsname{$-$0.05 [$-$0.06, $-$0.04]}
\expandafter\def\csname pdv@abl:lora-frozen:a2-checklist@chexbert_micro_f1_14\endcsname{+0.016}
\expandafter\def\csname pd@abl:lora-frozen:a2-checklist@chexbert_micro_f1_14\endcsname{+0.016 [+0.008, +0.026]}
\expandafter\def\csname pdv@abl:lora-frozen:a2-checklist@chexbert_macro_f1_14\endcsname{+0.024}
\expandafter\def\csname pd@abl:lora-frozen:a2-checklist@chexbert_macro_f1_14\endcsname{+0.024 [+0.003, +0.047]}
\expandafter\def\csname pdv@abl:lora-frozen:a2-checklist@radgraph_partial\endcsname{+0.007}
\expandafter\def\csname pd@abl:lora-frozen:a2-checklist@radgraph_partial\endcsname{+0.007 [+0.004, +0.009]}
\expandafter\def\csname pdv@abl:lora-frozen:a2-checklist@radcliq_v1\endcsname{$-$0.05}
\expandafter\def\csname pd@abl:lora-frozen:a2-checklist@radcliq_v1\endcsname{$-$0.05 [$-$0.06, $-$0.04]}
\expandafter\def\csname pdv@abl:ft-lora:a2@chexbert_micro_f1_14\endcsname{+0.001}
\expandafter\def\csname pd@abl:ft-lora:a2@chexbert_micro_f1_14\endcsname{+0.001 [$-$0.006, +0.009]}
\expandafter\def\csname pdv@abl:ft-lora:a2@chexbert_macro_f1_14\endcsname{+0.001}
\expandafter\def\csname pd@abl:ft-lora:a2@chexbert_macro_f1_14\endcsname{+0.001 [$-$0.014, +0.018]}
\expandafter\def\csname pdv@abl:ft-lora:a2@radgraph_partial\endcsname{+0.003}
\expandafter\def\csname pd@abl:ft-lora:a2@radgraph_partial\endcsname{+0.003 [+0.000, +0.006]}
\expandafter\def\csname pdv@abl:ft-lora:a2@radcliq_v1\endcsname{$-$0.03}
\expandafter\def\csname pd@abl:ft-lora:a2@radcliq_v1\endcsname{$-$0.03 [$-$0.04, $-$0.03]}
\expandafter\def\csname pdv@abl:ft-lora:a2-checklist@chexbert_micro_f1_14\endcsname{+0.006}
\expandafter\def\csname pd@abl:ft-lora:a2-checklist@chexbert_micro_f1_14\endcsname{+0.006 [$-$0.003, +0.015]}
\expandafter\def\csname pdv@abl:ft-lora:a2-checklist@chexbert_macro_f1_14\endcsname{+0.011}
\expandafter\def\csname pd@abl:ft-lora:a2-checklist@chexbert_macro_f1_14\endcsname{+0.011 [$-$0.006, +0.029]}
\expandafter\def\csname pdv@abl:ft-lora:a2-checklist@radgraph_partial\endcsname{+0.004}
\expandafter\def\csname pd@abl:ft-lora:a2-checklist@radgraph_partial\endcsname{+0.004 [+0.001, +0.006]}
\expandafter\def\csname pdv@abl:ft-lora:a2-checklist@radcliq_v1\endcsname{$-$0.03}
\expandafter\def\csname pd@abl:ft-lora:a2-checklist@radcliq_v1\endcsname{$-$0.03 [$-$0.04, $-$0.02]}
\expandafter\def\csname pdv@abl:ft-frozen:a2@chexbert_micro_f1_14\endcsname{+0.024}
\expandafter\def\csname pd@abl:ft-frozen:a2@chexbert_micro_f1_14\endcsname{+0.024 [+0.015, +0.034]}
\expandafter\def\csname pdv@abl:ft-frozen:a2@chexbert_macro_f1_14\endcsname{+0.026}
\expandafter\def\csname pd@abl:ft-frozen:a2@chexbert_macro_f1_14\endcsname{+0.026 [+0.008, +0.047]}
\expandafter\def\csname pdv@abl:ft-frozen:a2@radgraph_partial\endcsname{+0.010}
\expandafter\def\csname pd@abl:ft-frozen:a2@radgraph_partial\endcsname{+0.010 [+0.006, +0.014]}
\expandafter\def\csname pdv@abl:ft-frozen:a2@radcliq_v1\endcsname{$-$0.09}
\expandafter\def\csname pd@abl:ft-frozen:a2@radcliq_v1\endcsname{$-$0.09 [$-$0.10, $-$0.07]}
\expandafter\def\csname pdv@abl:ft-frozen:a2-checklist@chexbert_micro_f1_14\endcsname{+0.022}
\expandafter\def\csname pd@abl:ft-frozen:a2-checklist@chexbert_micro_f1_14\endcsname{+0.022 [+0.012, +0.033]}
\expandafter\def\csname pdv@abl:ft-frozen:a2-checklist@chexbert_macro_f1_14\endcsname{+0.035}
\expandafter\def\csname pd@abl:ft-frozen:a2-checklist@chexbert_macro_f1_14\endcsname{+0.035 [+0.013, +0.058]}
\expandafter\def\csname pdv@abl:ft-frozen:a2-checklist@radgraph_partial\endcsname{+0.010}
\expandafter\def\csname pd@abl:ft-frozen:a2-checklist@radgraph_partial\endcsname{+0.010 [+0.007, +0.013]}
\expandafter\def\csname pdv@abl:ft-frozen:a2-checklist@radcliq_v1\endcsname{$-$0.08}
\expandafter\def\csname pd@abl:ft-frozen:a2-checklist@radcliq_v1\endcsname{$-$0.08 [$-$0.09, $-$0.07]}
\expandafter\def\csname pdv@abl:repthr-ft:a2@chexbert_micro_f1_14\endcsname{+0.002}
\expandafter\def\csname pd@abl:repthr-ft:a2@chexbert_micro_f1_14\endcsname{+0.002 [$-$0.002, +0.005]}
\expandafter\def\csname pdv@abl:repthr-ft:a2@chexbert_macro_f1_14\endcsname{+0.005}
\expandafter\def\csname pd@abl:repthr-ft:a2@chexbert_macro_f1_14\endcsname{+0.005 [$-$0.001, +0.012]}
\expandafter\def\csname pdv@abl:repthr-ft:a2@radgraph_partial\endcsname{+0.001}
\expandafter\def\csname pd@abl:repthr-ft:a2@radgraph_partial\endcsname{+0.001 [$-$0.000, +0.002]}
\expandafter\def\csname pdv@abl:repthr-ft:a2@radcliq_v1\endcsname{$-$0.01}
\expandafter\def\csname pd@abl:repthr-ft:a2@radcliq_v1\endcsname{$-$0.01 [$-$0.02, $-$0.01]}
\expandafter\def\csname pdv@abl:repthr-ft:a2-checklist@chexbert_micro_f1_14\endcsname{+0.001}
\expandafter\def\csname pd@abl:repthr-ft:a2-checklist@chexbert_micro_f1_14\endcsname{+0.001 [$-$0.003, +0.004]}
\expandafter\def\csname pdv@abl:repthr-ft:a2-checklist@chexbert_macro_f1_14\endcsname{+0.007}
\expandafter\def\csname pd@abl:repthr-ft:a2-checklist@chexbert_macro_f1_14\endcsname{+0.007 [+0.000, +0.015]}
\expandafter\def\csname pdv@abl:repthr-ft:a2-checklist@radgraph_partial\endcsname{$-$0.002}
\expandafter\def\csname pd@abl:repthr-ft:a2-checklist@radgraph_partial\endcsname{$-$0.002 [$-$0.003, $-$0.001]}
\expandafter\def\csname pdv@abl:repthr-ft:a2-checklist@radcliq_v1\endcsname{$-$0.01}
\expandafter\def\csname pd@abl:repthr-ft:a2-checklist@radcliq_v1\endcsname{$-$0.01 [$-$0.01, $-$0.00]}
\expandafter\def\csname pdv@abl:cap-repthr:a2@chexbert_micro_f1_14\endcsname{+0.001}
\expandafter\def\csname pd@abl:cap-repthr:a2@chexbert_micro_f1_14\endcsname{+0.001 [$-$0.007, +0.009]}
\expandafter\def\csname pdv@abl:cap-repthr:a2@chexbert_macro_f1_14\endcsname{+0.010}
\expandafter\def\csname pd@abl:cap-repthr:a2@chexbert_macro_f1_14\endcsname{+0.010 [$-$0.009, +0.024]}
\expandafter\def\csname pdv@abl:cap-repthr:a2@radgraph_partial\endcsname{$-$0.000}
\expandafter\def\csname pd@abl:cap-repthr:a2@radgraph_partial\endcsname{$-$0.000 [$-$0.002, +0.002]}
\expandafter\def\csname pdv@abl:cap-repthr:a2@radcliq_v1\endcsname{+0.01}
\expandafter\def\csname pd@abl:cap-repthr:a2@radcliq_v1\endcsname{+0.01 [+0.00, +0.02]}
\expandafter\def\csname pdv@abl:cap-repthr:a2-checklist@chexbert_micro_f1_14\endcsname{+0.003}
\expandafter\def\csname pd@abl:cap-repthr:a2-checklist@chexbert_micro_f1_14\endcsname{+0.003 [$-$0.005, +0.012]}
\expandafter\def\csname pdv@abl:cap-repthr:a2-checklist@chexbert_macro_f1_14\endcsname{+0.007}
\expandafter\def\csname pd@abl:cap-repthr:a2-checklist@chexbert_macro_f1_14\endcsname{+0.007 [$-$0.010, +0.023]}
\expandafter\def\csname pdv@abl:cap-repthr:a2-checklist@radgraph_partial\endcsname{$-$0.002}
\expandafter\def\csname pd@abl:cap-repthr:a2-checklist@radgraph_partial\endcsname{$-$0.002 [$-$0.004, $-$0.000]}
\expandafter\def\csname pdv@abl:cap-repthr:a2-checklist@radcliq_v1\endcsname{+0.01}
\expandafter\def\csname pd@abl:cap-repthr:a2-checklist@radcliq_v1\endcsname{+0.01 [+0.01, +0.02]}
\expandafter\def\csname pdv@abl:cap-repthr:tv3@chexbert_micro_f1_14\endcsname{$-$0.000}
\expandafter\def\csname pd@abl:cap-repthr:tv3@chexbert_micro_f1_14\endcsname{$-$0.000 [$-$0.008, +0.008]}
\expandafter\def\csname pdv@abl:cap-repthr:tv3@chexbert_macro_f1_14\endcsname{+0.005}
\expandafter\def\csname pd@abl:cap-repthr:tv3@chexbert_macro_f1_14\endcsname{+0.005 [$-$0.013, +0.022]}
\expandafter\def\csname pdv@abl:cap-repthr:tv3@radgraph_partial\endcsname{$-$0.000}
\expandafter\def\csname pd@abl:cap-repthr:tv3@radgraph_partial\endcsname{$-$0.000 [$-$0.003, +0.002]}
\expandafter\def\csname pdv@abl:cap-repthr:tv3@radcliq_v1\endcsname{+0.00}
\expandafter\def\csname pd@abl:cap-repthr:tv3@radcliq_v1\endcsname{+0.00 [$-$0.01, +0.01]}
\expandafter\def\csname pdv@abl:dev-cap:tv3@chexbert_micro_f1_14\endcsname{$-$0.002}
\expandafter\def\csname pd@abl:dev-cap:tv3@chexbert_micro_f1_14\endcsname{$-$0.002 [$-$0.009, +0.006]}
\expandafter\def\csname pdv@abl:dev-cap:tv3@chexbert_macro_f1_14\endcsname{+0.014}
\expandafter\def\csname pd@abl:dev-cap:tv3@chexbert_macro_f1_14\endcsname{+0.014 [+0.000, +0.030]}
\expandafter\def\csname pdv@abl:dev-cap:tv3@radgraph_partial\endcsname{+0.002}
\expandafter\def\csname pd@abl:dev-cap:tv3@radgraph_partial\endcsname{+0.002 [$-$0.001, +0.004]}
\expandafter\def\csname pdv@abl:dev-cap:tv3@radcliq_v1\endcsname{$-$0.01}
\expandafter\def\csname pd@abl:dev-cap:tv3@radcliq_v1\endcsname{$-$0.01 [$-$0.01, +0.00]}
\expandafter\def\csname pdv@abl:full-dev:tv3@chexbert_micro_f1_14\endcsname{+0.014}
\expandafter\def\csname pd@abl:full-dev:tv3@chexbert_micro_f1_14\endcsname{+0.014 [+0.008, +0.021]}
\expandafter\def\csname pdv@abl:full-dev:tv3@chexbert_macro_f1_14\endcsname{$-$0.005}
\expandafter\def\csname pd@abl:full-dev:tv3@chexbert_macro_f1_14\endcsname{$-$0.005 [$-$0.017, +0.008]}
\expandafter\def\csname pdv@abl:full-dev:tv3@radgraph_partial\endcsname{+0.006}
\expandafter\def\csname pd@abl:full-dev:tv3@radgraph_partial\endcsname{+0.006 [+0.004, +0.009]}
\expandafter\def\csname pdv@abl:full-dev:tv3@radcliq_v1\endcsname{$-$0.02}
\expandafter\def\csname pd@abl:full-dev:tv3@radcliq_v1\endcsname{$-$0.02 [$-$0.02, $-$0.01]}
\expandafter\def\csname pdv@abl:full-repthr:tv3@chexbert_micro_f1_14\endcsname{+0.012}
\expandafter\def\csname pd@abl:full-repthr:tv3@chexbert_micro_f1_14\endcsname{+0.012 [+0.004, +0.020]}
\expandafter\def\csname pdv@abl:full-repthr:tv3@chexbert_macro_f1_14\endcsname{+0.014}
\expandafter\def\csname pd@abl:full-repthr:tv3@chexbert_macro_f1_14\endcsname{+0.014 [$-$0.002, +0.030]}
\expandafter\def\csname pdv@abl:full-repthr:tv3@radgraph_partial\endcsname{+0.008}
\expandafter\def\csname pd@abl:full-repthr:tv3@radgraph_partial\endcsname{+0.008 [+0.005, +0.011]}
\expandafter\def\csname pdv@abl:full-repthr:tv3@radcliq_v1\endcsname{$-$0.02}
\expandafter\def\csname pd@abl:full-repthr:tv3@radcliq_v1\endcsname{$-$0.02 [$-$0.03, $-$0.01]}
\expandafter\def\csname pdv@abl:gold-full:tv3@chexbert_micro_f1_14\endcsname{+0.240}
\expandafter\def\csname pd@abl:gold-full:tv3@chexbert_micro_f1_14\endcsname{+0.240 [+0.228, +0.252]}
\expandafter\def\csname pdv@abl:gold-full:tv3@chexbert_macro_f1_14\endcsname{+0.370}
\expandafter\def\csname pd@abl:gold-full:tv3@chexbert_macro_f1_14\endcsname{+0.370 [+0.351, +0.390]}
\expandafter\def\csname pdv@abl:gold-full:tv3@radgraph_partial\endcsname{+0.104}
\expandafter\def\csname pd@abl:gold-full:tv3@radgraph_partial\endcsname{+0.104 [+0.098, +0.110]}
\expandafter\def\csname pdv@abl:gold-full:tv3@radcliq_v1\endcsname{$-$0.40}
\expandafter\def\csname pd@abl:gold-full:tv3@radcliq_v1\endcsname{$-$0.40 [$-$0.41, $-$0.38]}
\expandafter\def\csname pdv@abl:ft-checklist:a2@chexbert_micro_f1_14\endcsname{$-$0.005}
\expandafter\def\csname pd@abl:ft-checklist:a2@chexbert_micro_f1_14\endcsname{$-$0.005 [$-$0.009, $-$0.000]}
\expandafter\def\csname pdv@abl:ft-checklist:a2@chexbert_macro_f1_14\endcsname{$-$0.001}
\expandafter\def\csname pd@abl:ft-checklist:a2@chexbert_macro_f1_14\endcsname{$-$0.001 [$-$0.007, +0.004]}
\expandafter\def\csname pdv@abl:ft-checklist:a2@radgraph_partial\endcsname{+0.062}
\expandafter\def\csname pd@abl:ft-checklist:a2@radgraph_partial\endcsname{+0.062 [+0.056, +0.069]}
\expandafter\def\csname pdv@abl:ft-checklist:a2@radcliq_v1\endcsname{$-$0.17}
\expandafter\def\csname pd@abl:ft-checklist:a2@radcliq_v1\endcsname{$-$0.17 [$-$0.20, $-$0.14]}
\expandafter\def\csname pdv@abl:repthr-checklist:a2@chexbert_micro_f1_14\endcsname{$-$0.005}
\expandafter\def\csname pd@abl:repthr-checklist:a2@chexbert_micro_f1_14\endcsname{$-$0.005 [$-$0.010, $-$0.001]}
\expandafter\def\csname pdv@abl:repthr-checklist:a2@chexbert_macro_f1_14\endcsname{+0.001}
\expandafter\def\csname pd@abl:repthr-checklist:a2@chexbert_macro_f1_14\endcsname{+0.001 [$-$0.006, +0.009]}
\expandafter\def\csname pdv@abl:repthr-checklist:a2@radgraph_partial\endcsname{+0.060}
\expandafter\def\csname pd@abl:repthr-checklist:a2@radgraph_partial\endcsname{+0.060 [+0.053, +0.067]}
\expandafter\def\csname pdv@abl:repthr-checklist:a2@radcliq_v1\endcsname{$-$0.17}
\expandafter\def\csname pd@abl:repthr-checklist:a2@radcliq_v1\endcsname{$-$0.17 [$-$0.20, $-$0.13]}
\expandafter\def\csname pdv@abl:cap-checklist:a2@chexbert_micro_f1_14\endcsname{$-$0.003}
\expandafter\def\csname pd@abl:cap-checklist:a2@chexbert_micro_f1_14\endcsname{$-$0.003 [$-$0.007, +0.001]}
\expandafter\def\csname pdv@abl:cap-checklist:a2@chexbert_macro_f1_14\endcsname{$-$0.001}
\expandafter\def\csname pd@abl:cap-checklist:a2@chexbert_macro_f1_14\endcsname{$-$0.001 [$-$0.005, +0.003]}
\expandafter\def\csname pdv@abl:cap-checklist:a2@radgraph_partial\endcsname{+0.058}
\expandafter\def\csname pd@abl:cap-checklist:a2@radgraph_partial\endcsname{+0.058 [+0.051, +0.065]}
\expandafter\def\csname pdv@abl:cap-checklist:a2@radcliq_v1\endcsname{$-$0.16}
\expandafter\def\csname pd@abl:cap-checklist:a2@radcliq_v1\endcsname{$-$0.16 [$-$0.19, $-$0.13]}

\newcommand{\mthRsMinGate}{0.1}

\newcommand{\mthCWrDedupJac}{0.9}
\newcommand{\mthCQueries}{70}
\newcommand{\mthCDecLayers}{6}
\newcommand{\mthCDecDim}{256}
\newcommand{\mthCNames}{5{,}264}
\newcommand{\mthCLocW}{3}
\newcommand{\mthCMaskW}{5}
\newcommand{\mthCEos}{0.1}
\newcommand{\mthCCapLayers}{3}
\newcommand{\mthCCapDim}{512}
\newcommand{\mthCItersName}{20{,}000}
\newcommand{\mthCHeadsFirst}{1{,}000}

\newcommand{\mthCItersTotal}{60{,}000}
\newcommand{\mthCTrainImages}{172{,}820}
\newcommand{\mthCLrName}{$10^{-4}$}
\newcommand{\mthCLrHeads}{$5\times10^{-5}$}
\newcommand{\mthCBbLr}{$10^{-5}$}
\newcommand{\mthCLlrd}{0.8}
\newcommand{\mthCWd}{$10^{-4}$}
\newcommand{\mthCClip}{0.1}
\newcommand{\mthCBs}{32}
\newcommand{\mthCWarmup}{1{,}000}
\newcommand{\mthCItersA}{10{,}000}
\newcommand{\mthCItersB}{30{,}000}
\newcommand{\mthCImg}{518}
\newcommand{\mthCGrid}{37}
\newcommand{\mthCUnfrozen}{12}
\newcommand{\mthCDecHeads}{8}
\newcommand{\mthCDecFfn}{2{,}048}
\newcommand{\mthCWrStudies}{19{,}285}
\newcommand{\mthCWrR}{16}
\newcommand{\mthCWrLr}{$2\times10^{-4}$}
\newcommand{\mthCWrBatch}{32}
\newcommand{\mthCWrMaxlen}{1{,}280}
\newcommand{\mthCWrAlpha}{32}
\newcommand{\mthCWrSteps}{600}
\newcommand{\mthRvLatImages}{97{,}996}
\newcommand{\mthRvLatFindings}{258{,}007}
\newcommand{\mthRvLatItersName}{20{,}000}
\newcommand{\mthRvLatItersHeads}{10{,}000}
\newcommand{\mthRvTokDim}{1{,}226}

\newcommand{\mthRvNameTop}{8}
\newcommand{\mthRvK}{3}
\newcommand{\mthRvNameFeats}{17}
\newcommand{\mthRvXview}{8}
\newcommand{\mthRvXviewGate}{0.5}
\newcommand{\mthRvNameEmb}{16}

\newcommand{\mthRvLayers}{4}
\newcommand{\mthRvHidden}{256}
\newcommand{\mthRvHeads}{4}
\newcommand{\mthRvFfn}{512}
\newcommand{\mthRvDropout}{0.2}
\newcommand{\mthRvPairDim}{64}
\newcommand{\mthRvBagDim}{32}
\newcommand{\mthRvSteps}{7{,}500}
\newcommand{\mthRvBatch}{64}
\newcommand{\mthRvLr}{$5\times10^{-4}$}
\newcommand{\mthRvWd}{$10^{-2}$}
\newcommand{\mthRvClip}{1}
\newcommand{\mthRvSubsetP}{0.5}

\newcommand{\mthRvStudies}{36{,}068}
\newcommand{\mthRvCands}{1{,}554{,}789}
\newcommand{\mthRvT}{0.25}
\newcommand{\mthRvMerge}{0.3}
\newcommand{\mthCTrainFindings}{556{,}125}
\newcommand{\mthCDevNoMaskPct}{24}
\newcommand{\setupMimicStudies}{2{,}346}
\newcommand{\setupMimicSeveral}{1{,}216}
\newcommand{\setupMimicLateral}{1{,}039}
\newcommand{\setupMimicSecondFrontal}{287}
\newcommand{\setupOpeniStudies}{3{,}198}

\newcommand{\setupOpeniLateral}{2{,}942}

\newcommand{\setupMscxrImages}{1{,}047}
\newcommand{\setupMscxrTrain}{878}
\newcommand{\setupMscxrVal}{5}
\newcommand{\setupMscxrTest}{164}
\newcommand{\setupMscxrTestPatients}{96}
\newcommand{\setupVindrImages}{3{,}000}
\newcommand{\setupVindrBoxed}{949}

\newcommand{\wfStudies}{2,346}
\newcommand{\wfSentences}{12,986}
\newcommand{\wfStated}{10,125}
\newcommand{\wfBackedPct}{99.7}
\newcommand{\wfUnlisted}{17}

\newcommand{\fusSingleOneF}{.488}

\newcommand{\fusOursAllF}{.494}

\newcommand{\paxclasses}{177}

\newcommand{\paxfrontalonly}{85}

\newcommand{\paxbothviews}{80}
\newcommand{\paxrareclasses}{9}

\newcommand{\paxrefclassesfrontal}{165}
\newcommand{\paxrefclasseslateral}{82}

\expandafter\def\csname paxschema@spine region\endcsname{S01}
\expandafter\def\csname paxschema@cervical vertebra\endcsname{S02}
\expandafter\def\csname paxschema@thoracic vertebra\endcsname{S03}
\expandafter\def\csname paxschema@ribcartilage\endcsname{S04}
\expandafter\def\csname paxschema@sternum\endcsname{S05}
\expandafter\def\csname paxschema@clavicles\endcsname{S06}
\expandafter\def\csname paxschema@clavicle one side\endcsname{S07}
\expandafter\def\csname paxschema@paired structure centres\endcsname{S08}
\expandafter\def\csname paxschema@scapula one side\endcsname{S09}
\expandafter\def\csname paxschema@posterior rib\endcsname{S10}
\expandafter\def\csname paxschema@anterior rib\endcsname{S11}
\expandafter\def\csname paxschema@diaphragm\endcsname{S12}
\expandafter\def\csname paxschema@hemidiaphragm\endcsname{S13}
\expandafter\def\csname paxschema@stomach\endcsname{S14}
\expandafter\def\csname paxschema@centre only\endcsname{S15}
\expandafter\def\csname paxschema@liver\endcsname{S16}
\expandafter\def\csname paxschema@mediastinal compartment\endcsname{S17}
\expandafter\def\csname paxschema@heart\endcsname{S18}
\expandafter\def\csname paxschema@heart atrium left\endcsname{S19}
\expandafter\def\csname paxschema@heart atrium right\endcsname{S20}
\expandafter\def\csname paxschema@heart ventricle left\endcsname{S21}
\expandafter\def\csname paxschema@aorta\endcsname{S22}
\expandafter\def\csname paxschema@ascending aorta\endcsname{S23}
\expandafter\def\csname paxschema@descending aorta\endcsname{S24}
\expandafter\def\csname paxschema@aortic arch\endcsname{S25}
\expandafter\def\csname paxschema@pulmonary trunc\endcsname{S26}
\expandafter\def\csname paxschema@pulmonary artery one side\endcsname{S27}
\expandafter\def\csname paxschema@inferior vena cava\endcsname{S28}
\expandafter\def\csname paxschema@esophagus\endcsname{S29}
\expandafter\def\csname paxschema@lung\endcsname{S30}
\expandafter\def\csname paxschema@right lung\endcsname{S31}
\expandafter\def\csname paxschema@left lung\endcsname{S32}
\expandafter\def\csname paxschema@lung base\endcsname{S33}
\expandafter\def\csname paxschema@apical zone\endcsname{S34}
\expandafter\def\csname paxschema@lobe oblique fissure\endcsname{S35}
\expandafter\def\csname paxschema@lobe both fissures\endcsname{S36}
\expandafter\def\csname paxschema@trachea\endcsname{S37}
\expandafter\def\csname paxschema@tracheal bifurcation\endcsname{S38}
\expandafter\def\csname paxschema@breast one side\endcsname{S39}
\newcommand{\schemaref}[1]{\ifcsname paxschema@#1\endcsname\csname paxschema@#1\endcsname\else\textbf{??}\fi}

\newcommand{\anatnewclasses}{4}

\newcommand{\gndadded}{6}

\newcommand{\gndclasses}{187}
\newcommand{\gndprevclasses}{181}

\usepackage{iftex}
\ifpdftex\else
  \usepackage{fontspec}
  \IfFontExistsTF{DejaVu Sans Mono}{%
    \setmonofont{DejaVu Sans Mono}[Scale=MatchLowercase]}{}
\fi

\usepackage{blindtext}

\definecolor{finaldep}{HTML}{FFF1C2}
\DeclareCaptionFormat{finaldep}{\colorbox{finaldep}{\parbox{\dimexpr\linewidth-2\fboxsep\relax}{#1#2#3}}}

\definecolor{donetext}{HTML}{D9F2D9}

\DeclareCaptionFormat{donetext}{{\setlength{\fboxsep}{0pt}\colorbox{donetext}{\parbox{\linewidth}{#1#2#3}}}}

\definecolor{pendingtext}{HTML}{FFF1C2}
\DeclareCaptionFormat{pendingtext}{{\setlength{\fboxsep}{0pt}\colorbox{pendingtext}{\parbox{\linewidth}{#1#2#3}}}}

\newif\ifdonepar \newif\ifdoneparfirst
\newcommand{\doneparcolour}{donetext}

\definecolor{proptext}{HTML}{FFD3A1}

\makeatletter
\newcommand{\done@underlay}{\ifdonepar
  \setbox\@tempboxa\lastbox
  \ifvoid\@tempboxa\else
    \@tempdimb=.72\baselineskip \ifdoneparfirst \advance\@tempdimb-1pt \global\doneparfirstfalse \fi
    \nointerlineskip
    \vbox to\z@{\kern\dimexpr\ht\@tempboxa-\@tempdimb\relax
      \hbox{\textcolor{\doneparcolour}{\vrule width\columnwidth height\dimexpr\@tempdimb+.28\baselineskip\relax}}\vss}%
    \nointerlineskip \moveright\@totalleftmargin\box\@tempboxa %
  \fi\fi}
\@ifundefined{MakeLineNo}{}{%
  \patchcmd\MakeLineNo{\@tempdima\dp\z@ \unvbox\z@}{\@tempdima\dp\z@ \unvbox\z@ \done@underlay}{}%
    {\PackageWarningNoLine{preamble}{donepar: lineno's MakeLineNo has changed, finished paragraphs are not marked}}}
\makeatother
\definecolor{finaldeppred}{HTML}{E8DEF8}

\newcommand{\methodnametext}{CheXTR}
\newcommand{\methodname}{\methodnametext\xspace}
\AtBeginDocument{\ifdefined\pdfstringdefDisableCommands\pdfstringdefDisableCommands{\def\methodname{\methodnametext}}\fi}

\chexshadetrue
\chextritrue

\makeatletter
\newcommand{\chexclearpage}{\clearpage\ifdim\pagetotal=\z@\global\@colroom\@colht\global\vsize\@colroom\fi}
\makeatother

\definecolor{linkblue}{rgb}{0.21,0.49,0.74}
\usepackage[pagebackref,breaklinks,colorlinks,allcolors=linkblue]{hyperref}

\title{Radiological Report Generation as Grounded Set Prediction}

\author{
Constantin Seibold$^{1}$\quad
Alexander Jaus$^{2}$\quad
David Vinu$^{3}$\\
Nils Havemann$^{3}$\quad
Alina Roitberg$^{4}$\quad
Matthias A. Fink$^{1}$\\[0.4em]
$^{1}$Heidelberg University Hospital\quad
$^{2}$Karlsruher Institut f\"ur Technologie\\
$^{3}$Ruprecht-Karls-Universit\"at Heidelberg\quad
$^{4}$Universit\"at Hildesheim\\[0.3em]
{\tt\small constantinseibold@gmail.com}
}

\begin{document}
\maketitle
\begin{abstract}
  
  Vision--language models have made next-token prediction the de-facto
  standard for medical report generation.
  \emph{But should identifying medical findings and writing a
  well-formulated text be treated as the same task?}
  Instead, we argue that writing a report requires only an explicit set of
  medical findings, whose recognition can be separated from language
  generation. This allows us to evaluate and improve the accuracy of medical
  findings independently of report generation and trace them back to the
  image.
  
  \vspace{2pt}
  We therefore revisit medical report generation as a set prediction problem
  and introduce \methodname, a framework which predicts medical findings as a set of
  instances, each with a name, anatomical location, and attributes. A small
  resolver merges redundant finding predictions within one or multiple
  radiograph views. The final set can be inspected on its own or passed to an exchangeable report writer, e.g., a fixed template or a language model. With a fixed template, only
  \corehead{oursparams} parameters and 0.17M radiograph-report training pairs, \methodname achieves higher clinical
  efficacy scores than report generators with up to
  \corehead{sizeratio} times its parameters and seven times its training data, whereas using a 2B language model as
  writer leads to competitive writing quality while maintaining this lead.
  Through its set predictor, \methodname also localises the reported findings
  more accurately on radiologist boxes than grounded report generators. Code
  and models will be made public.%

\end{abstract}

\section{Introduction}
\label{sec:intro}
Automatically drafting the radiology report of a routine examination like a chest radiograph is a long-standing goal of medical image analysis.
In recent years, vision--language models with billions of parameters generating radiology reports end to end~\cite{bannur2024maira2,chen2024chexagent,tu2024medpalmm,chexone} have replaced captioning networks with recurrent decoders~\cite{shin2016learningtoread,jing2018coatt}.
While these models produce fluent reports, they tie the recognition and the language expression of medical findings  directly together. 

\setlength{\abovecaptionskip}{4pt}
\begin{figure}[!t]
  \centering
  \includegraphics[width=\linewidth]{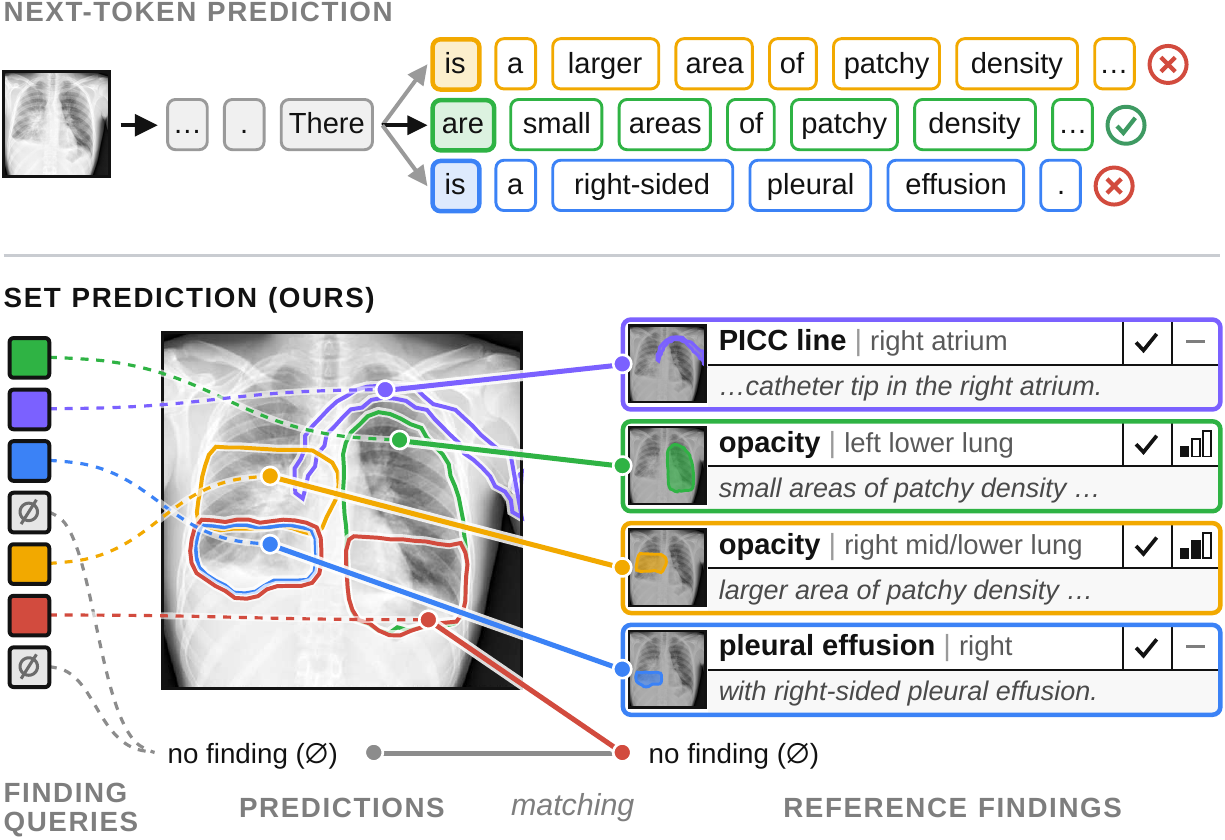}
  \caption{\textbf{Comparison of report generation approaches.} With
    next-token prediction (top), only the word the radiologist wrote next
    counts as correct. Thus, stating a different correct finding of the same
    report first is penalised. \methodname (bottom) predicts the findings as a set of
    instances, each with a name, location and attributes, and learns them by
    matching to the findings of the reference report in any order. The
    report is then written from this set by a template or a small
    language model.}
  \label{fig:teaser}
\end{figure}

A \textit{finding} is an observation, like a pneumothorax with location and
attributes such as severity.  End-to-end generators learn \textit{which findings} to state and \textit{how to word} them by predicting the next word of the reference report (\cref{fig:teaser}, top).
This formulation entangles the finding recognition with report writing.
As a result, errors are difficult to attribute to either, and improving one requires retraining the full model.

Several recent works therefore investigate how to ground the report in
intermediate structure, such as predicted
labels~\cite{jin2024promptmrg,pellegrini2025radialog}, organ
masks~\cite{gu2024comg,sharma2024mairaseg}, anatomical
regions~\cite{tanida2023rgrg,muller2024chex} or
boxes~\cite{bannur2024maira2}. 
These methods make reports more
controllable and easier to verify, but the structure states either which of a few
classes are present or where to look. 
\textit{What is found where}, and \textit{with which attributes}, is still left to the generator and learned together with the wording.

In this work, we introduce \methodname, a framework that formulates radiology report generation as a set prediction problem, separating what a report states from how it is written. 
Our starting point is that writing a report requires only its findings, which can be predicted without generating text.
 \methodname therefore begins with a \textit{set predictor} built on a radiograph encoder~\cite{carion2020detr,kerssies2025eomt}. It predicts medical findings as a set of instances, each with a name, a location among 187 anatomical regions, and attributes. During training, the predicted set is matched in any order to the findings parsed from the reference report (\cref{fig:teaser}, bottom), which requires no manual image annotation.
Subsequently, a small \textit{resolver} module reads the predicted candidates of all views of a study,  merges redundant ones and decides which to keep. A chosen report writer, such as a fixed template or a small language model, finally turns the resulting set into a report.

By modularising the report generation process, our approach allows us to evaluate and improve finding recognition separately from report writing, trace each finding back to its region, and select the writer based on the application. 
In summary, our contributions are as follows:
\begin{enumerate}
\setlength{\itemsep}{0pt}\setlength{\parskip}{0pt}\setlength{\topsep}{2pt}
 \item We frame report generation as predicting the findings of a radiograph as a set of localised instances which are used for writing the report,  separating recognition from writing and tying every stated finding to a region.
\item We present \methodname, consisting of a \textit{set predictor} that identifies candidate findings, a \textit{resolver} that merges redundant predictions within and across views, and a \textit{writer} that turns the resulting set into a report. The writer can be exchanged, and a further view adds a predictor of its own without retraining the existing one.
\item We show that \methodname with a fixed template, and hence without a language model, states findings more accurately than end-to-end generators with many times its parameters and training radiographs on MIMIC-CXR, while a small language model writing from the finding set also makes wording competitive. Its stated findings are also localised more accurately on radiologist boxes than those of grounded report generators.
\end{enumerate}

\section{Related work}
\label{sec:related}

\noindent\textbf{Report generation.}
Early methods adapt image captioning to chest radiographs, combining an encoder with a recurrent
decoder~\cite{shin2016learningtoread,wang2018tienet,jing2018coatt}.
Transformer decoders with memory modules~\cite{chen2020r2gen,chen2021r2gencmn}
and conditioning on predicted labels or knowledge
graphs~\cite{zhang2020knowledgegraph,jin2024promptmrg} improved clinical
accuracy. Recent methods build on large language models that are specialised
to
radiology~\cite{hyland2023maira1,bannur2024maira2,chen2024chexagent,chaves2025llavarad,pellegrini2025radialog},
tuned to medicine in general~\cite{tu2024medpalmm,sellergren2025medgemma}, or
general-purpose and prompted without domain
training~\cite{zhu2025internvl3,bai2025qwen25vl,qwen35}. Some add explicit
reasoning~\cite{chexone,fan2025chestxreasoner,wang2025mrgr1,gundersen2025radvlmrl}
or tool use~\cite{fallahpour2025medrax}. Across these generations, the
language decoder has grown by orders of magnitude, while the report is still
generated end-to-end. In contrast, \methodname predicts the findings of a
study explicitly and writes them with a fixed template or a language
model.

\noindent\textbf{Two-stage report generation.}
Several works separate what a report states from how it is written, filling a
report template with predicted
labels~\cite{pino2021cnntrg,kale2023replaceandreport} or retrieving a report
for them~\cite{syedamahmood2020ffl}. Such templates already stated clinical
findings more accurately than the learned decoders of their time, but for 13
abnormalities without location or severity~\cite{pino2021cnntrg}. Other
works let a language model write from classifier
or detector outputs~\cite{wang2023chatcad,danu2023collaborative} or from
sparse features of a frozen encoder~\cite{abdulaal2024saerad}. Structured
reporting fills a fixed hierarchy by question
answering~\cite{pellegrini2023radrestruct} or lets a language model emit the
structure~\cite{li2025srrgbench}. Yan et al.~\cite{yan2023styleaware} argue
that generators conflate the content of a report with its style and predict a
serialised entity graph, which is decoded word by word. There,
the content is a fixed list of labels, a sequence of words or unnamed
features. In contrast, \methodname predicts each finding as one instance with name, location and attributes.

\noindent\textbf{Set prediction.}
DETR~\cite{carion2020detr} formulates object detection as set prediction,
matching a fixed number of learned queries to the targets by bipartite
matching~\cite{stewart2016peopledet} and removing hand-crafted
post-processing such as non-maximum suppression. Later works improve
its convergence and accuracy~\cite{zhu2021deformable,zhang2023dino} and
extend it to segmentation~\cite{cheng2021maskformer,cheng2022mask2former},
to open vocabularies~\cite{minderer2022owlvit,liu2024groundingdino}, and to
plain vision transformers~\cite{kerssies2025eomt}.
Dense captioning attaches a caption to each predicted
region~\cite{johnson2016densecap,wu2022grit},
mask~\cite{huang2024sca,rasheed2024glamm}, video
object~\cite{choudhuri2024owviscap} or event~\cite{wang2021pdvc}. In report
generation, RGRG~\cite{tanida2023rgrg} describes fixed anatomical regions,
and scene-graph methods predict attributes for each
region~\cite{wu2021chestimagenome,wang2024sgrrg}.
Prior-RadGraphFormer~\cite{xiong2023priorradgraphformer} matches queries to
report graph entities. TranSQ~\cite{kong2022transq} and
RA-RRG~\cite{park2026rarrg} match them to sentences or key phrases and
retrieve these from the training reports at inference.
ChEX~\cite{muller2024chex} answers text prompts with a box and a sentence.
In contrast, \methodname formulates finding recognition
itself as set prediction, where each query predicts one finding with its name,
attributes, anatomical location and mask, so each stated finding can be
checked on the image.
Several views of a study are combined by fusing their image
features~\cite{yuan2019multiview,liu2025mlrg} or by passing all images to the
language model~\cite{bannur2024maira2,park2026rarrg}, which leaves the
correspondence of findings across views implicit. Our resolver instead merges
the finding instances of all views and can be any fusion
mechanism, such as weighted box fusion~\cite{solovyev2021wbf},
attention-based multiple-instance pooling~\cite{ilse2018attention} or a
learned set model.\looseness=-1 %

\noindent\textbf{Segmentation and grounding.}
Chest radiograph segmentation has progressed from lungs, heart and
clavicles~\cite{vanginneken2006jsrt} to large-scale anatomical
masks~\cite{gaggion2024chexmask} and to fine-grained anatomy in frontal and
lateral views obtained by projecting CT
labels~\cite{seibold2022paxray,seibold2023paxraypp},
complemented by pathology annotations~\cite{nguyen2022vindrcxr,lian2021chestxdet}.
Masks of support devices stem from annotated catheter and tube
courses~\cite{tang2021clip} or from synthetically inserted
devices~\cite{seibold2026synthfb}.
Report generators use segmentation primarily as an additional input, guiding
the visual features with organ masks~\cite{gu2024comg,gu2025orid} or the
decoder with mask-aware tokens~\cite{sharma2024mairaseg}, while phrase
grounding~\cite{boecking2022biovil} and grounded report
generation~\cite{bannur2024maira2,castro2025padchestgr} learn bounding boxes
for individual findings from radiologist annotations.
In both cases, the link between a sentence and the image is learned by the
decoder rather than given by the structure of the model. In this work, each
query predicts the location of its finding among the anatomical regions of an
extended PAX-Ray++~\cite{seibold2023paxraypp}. Its targets are the regions
the report names rather than annotated boxes, with detected device masks for
support devices.

\begin{figure*}[t]
  \centering
  \includegraphics[width=0.95\linewidth]{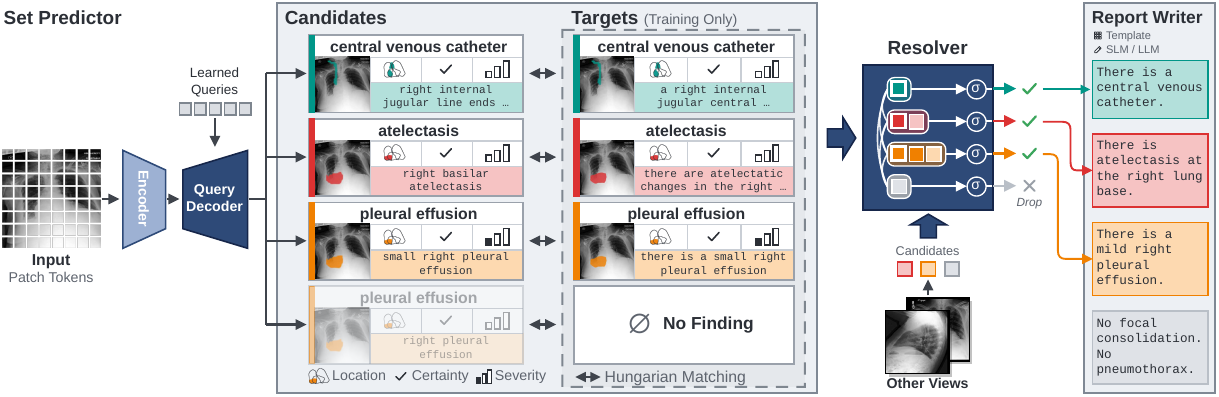}
  
  \caption%
  {\textbf{The \methodname framework}  separates diagnostics from report writing through an explicit set of medical findings. A \textbf{set predictor} uses learned queries to read image patch tokens and predict candidate findings, each with a name, anatomical location, mask, and attributes. During training, one-to-one matching assigns queries to findings parsed from the reference report, with unmatched queries being linked to \textit{no finding}. A \textbf{resolver} combines candidates within and across views, merges predictions that share a finding name, and selects which findings to retain. A \textbf{report writer}, implemented as a fixed template or language model, turns the resulting set into text.%
  }
  \label{fig:method}
\end{figure*}

\section{
 Report generation as set prediction}

\label{sec:method}

Our idea is to separate diagnostics from report writing by treating the former as a set prediction problem.
We first formulate the task (\cref{subsec:problem}) and then describe the three stages of \methodname as displayed in \cref{fig:method}: predicting the finding set (\cref{subsec:fsp}), resolving candidate findings from one or more views of a study (\cref{subsec:rescorer}), and writing the report (\cref{subsec:verbalise}).

\subsection{Problem Formulation}
\label{subsec:problem}

A study consists of one or more radiographs
$\mathcal{I}=\{I_1,\dots,I_K\}$ and a report $R$. We describe the content of
$R$ as a set $F$ of findings. A \emph{finding} $f=(n,\pi,c,s,L)$ has a name
$n$ from a vocabulary $\mathcal{V}$, a presence $\pi$, a certainty $c$, a
severity $s$ and a location $L\subseteq\mathcal{A}$ among the anatomical
regions $\mathcal{A}$ (\cref{sec:data}). Instead of generating $R$ from
$\mathcal{I}$ token by token, we factorise
\begin{equation}
  \begin{aligned}
    p(R\mid\mathcal{I})&=\textstyle\sum_{F}p_\psi(R\mid F)\,p_\theta(F\mid\mathcal{I})
    \approx p_\psi\big(R\mid\hat F\big),\\
    \hat F&=\arg\max_F\,p_\theta(F\mid\mathcal{I}).
  \end{aligned}
  \label{eq:factorisation}
\end{equation}
As the content of a report does not depend on the order in which its
findings are stated, $p_\theta(F\mid\mathcal{I})$ is a set prediction
problem~\cite{carion2020detr}. We model it with a set predictor for each
image and a resolver that merges their candidates. 

While the writer $p_\psi$ can be chosen independently of the set predictor, we keep it small (a fixed template or a small language model), as radiology reports follow a relatively constrained structure.

\subsection{Findings via the Set Predictor}
\label{subsec:fsp}

Following DETR~\cite{carion2020detr}, we predict a \textit{set of radiological findings} using $N$ learned queries and a transformer decoder that attends to the patch tokens $x_1,\dots,x_P$ of a fine-tuned image encoder~\cite{perezgarcia2025radDINO}.
Shared heads map each query $q_i$ to
\begin{equation}
  h(q_i)=\big(g_i,\;p_i,\;\hat\pi_i,\;\hat c_i,\;\hat s_i,\;\lambda_i,\;m_i\big),
  \label{eq:heads}
\end{equation}
where $g_i\in[0,1]$ is the probability that $q_i$ holds a finding, and
$p_i$, $\hat\pi_i$, $\hat c_i$, $\hat s_i$ are distributions over
$\mathcal{V}$, presence, certainty and severity. The location
$\lambda_i\in[0,1]^{|\mathcal{A}|}$ holds one probability per region, since
a finding may span several, and $m_i\in[0,1]^{P}$ is a mask over the
patches~\cite{cheng2022mask2former}. A name omits the descriptors of the
sentence that states a finding. During training, a small caption decoder that
cross-attends to $q_i$ alone~\cite{yu2022coca} therefore also writes a caption
of the finding.

\noindent\textbf{Training.}\label{subsec:training}
We assign queries to the findings of $F$ one to one with the Hungarian
algorithm~\cite{kuhn1955hungarian} under the cost
\begin{equation}
  \mathcal{M}(i,f)=-\log g_i-\log p_i(n_f)+w_L\,\ell_{\mathrm{loc}}(\lambda_i,L_f),
  \label{eq:cost}
\end{equation}
where $\ell_{\mathrm{loc}}$ is the binary cross-entropy averaged over the
regions. The attributes stay out of the cost because they qualify a finding
rather than identify it. We train every gate towards whether its query is
matched, and a matched query on the name, attributes, location, mask and
caption of its finding. Mask and caption targets are derived in
\cref{sec:data}.

\subsection{Handling Candidates via the Resolver}
\label{subsec:rescorer}

Every query with a gate above a low threshold is a
\emph{candidate}, and lateral views are read by a second
predictor of the same architecture. Since several queries of one view and
the queries of different views often name the same finding, a
\emph{resolver} $r_\phi$ merges the candidates $\mathcal{C}$ of a study into
the finding set $\hat F=r_\phi(\mathcal{C})$. In general, any mechanism that fuses
instances can fill this role (\cref{tab:core-fusion}).

\noindent\textbf{Architecture.}
We implement $r_\phi$ as a small set transformer~\cite{lee2019settransformer}
that reads each candidate as one token, marked with its view and with the
agreement its name finds among the other candidates, and returns a vector
$z_i$. For each name $n$, it pools the candidates $\mathcal{B}_n$ proposing
it into the probability
$p_n=\sigma\big(u(\sum_{i\in\mathcal{B}_n}\alpha_i z_i)\big)$ that the study
shows $n$, where $\alpha_i$ are gated attention
weights~\cite{ilse2018attention} and $u$ is an MLP. Further heads on $z_i$
score whether two candidates in $\mathcal{B}_n$ describe the same instance,
which splits them into groups, and pick one per group for its
attributes, location and mask. The group of the highest-scoring candidate
becomes a finding, and a further group only if it scores high enough itself.

\noindent\textbf{Training.}
Unlike the one-to-one matching of \cref{eq:cost}, every candidate that names
a finding of the report is assigned to it, so that the resolver learns to
merge duplicates instead of suppressing them. This assignment gives the
targets of all heads, and two candidates form a positive pair if they share
a finding. We train on the predictor's
candidates for random subsets of the images of a training study, so that one
resolver serves a single view as well as a full study.

\subsection{From Sets to Reports}
\label{subsec:verbalise}

The report writer $p_\psi$ states $\hat F$ as a report and can be chosen for the
application. A fixed template writes one sentence for each finding from its
name, attributes and location and a normal statement for each common region
without a finding, so it cannot state an unlisted finding. A small language
model~\cite{qwen35} with low-rank adapters~\cite{dettmers2023qlora} instead
reads $\hat F$ as numbered lines of name, attributes, location and
confidence, writes the FINDINGS section and cites the lines each sentence
states, so that each sentence leads back to its query, regions and mask. It
trains on reports alone, with a report's silver labels
(\cref{sec:data}) as input, each given a confidence sampled from the predictor's scores on correct findings, and sentences of listed findings as target. 

\section{Training Targets from Reports}
\label{sec:data}
\chexlatefloats

\begin{figure}[tb]
  \centering
  \includegraphics[width=0.95\linewidth, 
  ]{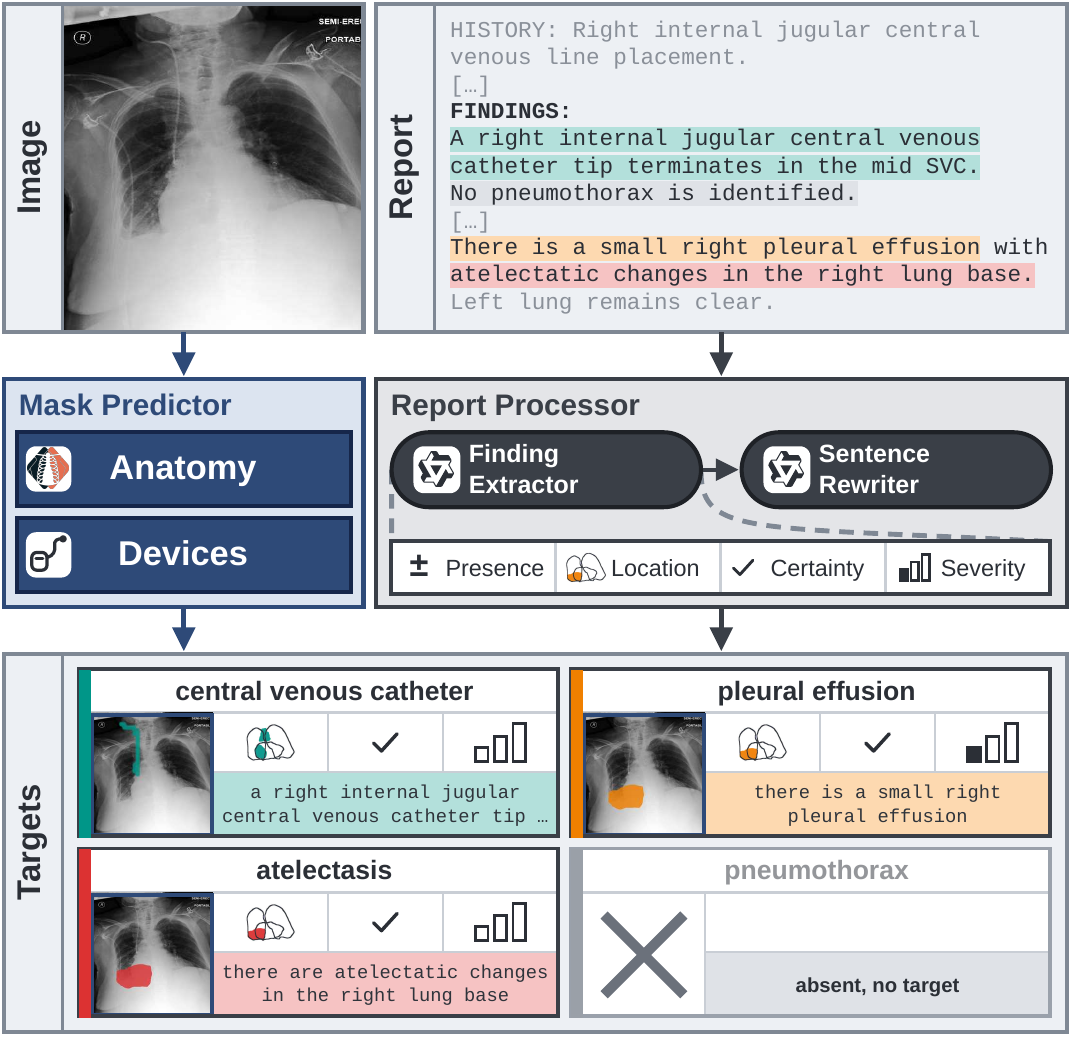}
  \caption{\textbf{Target construction.} A language model extracts findings and their attributes from a report. A rewriter produces a description per finding. Predicted anatomy or device masks provide the spatial supervision without manually annotating each finding.}
  \label{fig:data}
  \vspace{\dimexpr\chexdatafigsep-10pt\relax}%
\end{figure}

Training \methodname requires the finding set $F$ of every image and, for
each finding, a mask and a caption. We derive them from
MIMIC-CXR~\cite{johnson2019mimiccxr} by parsing each report into findings and
mapping the location of each finding to a mask (\cref{fig:data}).

\noindent\textbf{Finding sets.}\label{subsec:labels}
We parse each report sentence by sentence into core findings $f=(n,\pi,c,s,L)$, for example,\\[3pt]
\hspace*{1.5em}\emph{small bilateral pleural effusions} $\rightarrow$\\
\hspace*{1.5em}(pleural effusion, present, definite, mild,\\
\hspace*{1.5em}\{left base lung zone, right base lung zone\}).\\[3pt]
The names in the training reports form the vocabulary $\mathcal{V}$ of
\mthCNames{} names. We draw locations from a closed vocabulary of
\annConcepts{} anatomical concepts~\cite{seibold2023paxraypp}. Each concept is
a union of the 187 regions of $\mathcal{A}$, so that a segmentation model covers
every location. A location without a stated side covers both sides. A hedged
finding keeps its stated certainty as the target of the certainty head. An absent
finding (\emph{no pleural effusion}) is extracted but not added to $F$,
whereas normal statements (\emph{the lungs are clear}) yield no finding.

A general-purpose LLM (Claude
Opus~5.5~\cite{anthropic2026claudeopus,aws2026bedrock}) parses the test reports
twice, independently. Its two parses agree on \annAgreedPct{}\% of the findings in name, presence and
location. A third pass, rules and manual review reconcile the rest into a
\emph{reference parse}. As CheXbert~\cite{smit2020chexbert} learned from expert-labeled reports, a
\emph{silver labeler} (\silvBest{}~\cite{qwen35}) learns from the reference parse to parse
the training and validation reports. No test image or report enters the training of predictor, resolver or writer. Developed by cross-validation over patients, it recovers held-out patients' findings in name, presence and location with an
F1 of \silvMbr{}. The caption target of a finding is its source sentence.
Where that sentence holds several findings or change wording, a fine-tuned \emph{sentence
rewriter} turns it into a change-free sentence of
one finding.

\setlength{\abovecaptionskip}{2pt}

We assign the findings of the FINDINGS section of a study to every frontal
image and drop studies without this section. On lateral images, left and right regions become their joint region,
or both where none exists. This gives \mthCTrainFindings{} findings on
\mthCTrainImages{} frontal and \mthRvLatFindings{} findings on \mthRvLatImages{} lateral training images.

\begin{table*}[t]
\centering
\scriptsize
\chexcompactrules
\setlength{\tabcolsep}{1.5pt}
\caption{\textbf{Report generation from frontal image.} Shaded rows are not learned. \textit{CXR imgs} shows  \#radiographs seen in training, as encoder\,/\,model when trained separately (${\sim}$ derived, n/d not disclosed). \textbf{Best} and \underline{second best} per metric.}
\label{tab:core-main}
\resizebox{\linewidth}{!}{%
\begin{tabular}{@{}l r r@{\hspace{7pt}} ccccc ccccc@{}}
\toprule
 & & & \multicolumn{5}{c}{MIMIC-CXR} & \multicolumn{5}{c}{OpenI (external)} \\
\cmidrule(lr){4-8}\cmidrule(l){9-13}
 & Params & CXR imgs (M) & GREEN$\uparrow$ & RG\textsubscript{ER}$\uparrow$ & miF1$\uparrow$ & maF1$\uparrow$ & RadCliQ$\downarrow$ & GREEN$\uparrow$ & RG\textsubscript{ER}$\uparrow$ & miF1$\uparrow$ & maF1$\uparrow$ & RadCliQ$\downarrow$ \\
\midrule
\chexnorankrow Constant ``normal'' & -- & -- & \chexci{0.160}{.140}{.180} & \chexci{0.171}{.161}{.181} & \chexci{0.037}{.027}{.047} & \chexci{0.008}{.006}{.010} & \chexci{1.52}{1.47}{1.57} & \chexci{0.615}{.604}{.626} & \chexci{0.323}{.318}{.327} & \chexci{0.420}{.402}{.439} & \chexci{0.040}{.039}{.041} & \chexci{0.77}{.75}{.79} \\
\chexnorankrow Most common report & -- & -- & \chexci{0.188}{.170}{.207} & \chexci{0.163}{.155}{.171} & \chexci{0.037}{.027}{.047} & \chexci{0.008}{.006}{.010} & \chexci{1.51}{1.47}{1.55} & \chexcib{0.649}{.639}{.659} & \chexci{0.300}{.296}{.304} & \chexci{0.420}{.402}{.439} & \chexci{0.040}{.039}{.041} & \chexci{0.78}{.77}{.80} \\
\chexnorankrow Random report & -- & -- & \chexci{0.142}{.132}{.152} & \chexci{0.136}{.131}{.141} & \chexci{0.238}{.225}{.252} & \chexci{0.140}{.131}{.150} & \chexci{1.61}{1.59}{1.64} & \chexci{0.300}{.290}{.309} & \chexci{0.162}{.158}{.166} & \chexci{0.136}{.126}{.147} & \chexci{0.071}{.062}{.079} & \chexci{1.44}{1.42}{1.46} \\
\midrule
CheXagent-2~\cite{chen2024chexagent} & 3.1B & 1.08 & \chexci{0.310}{.295}{.326} & \chexci{0.261}{.254}{.269} & \chexci{0.580}{.563}{.595} & \chexci{0.386}{.363}{.404} & \chexci{1.37}{1.34}{1.40} & \chexci{0.407}{.395}{.417} & \chexci{0.222}{.216}{.227} & \chexci{0.399}{.382}{.415} & \chexci{0.277}{.252}{.298} & \chexci{1.40}{1.38}{1.42} \\
\chexrowrule{4-13}
CheXOne~\cite{chexone} & 3.8B & ${\sim}$1.2 & \chexciu{0.317}{.305}{.331} & \chexcib{0.300}{.294}{.307} & \chexci{0.592}{.577}{.607} & \chexci{0.339}{.324}{.353} & \chexcib{0.98}{.95}{1.01} & \chexci{0.608}{.599}{.617} & \chexcib{0.344}{.339}{.349} & \chexciu{0.549}{.532}{.565} & \chexci{0.268}{.245}{.291} & \chexcib{0.67}{.65}{.69} \\
\chexrowrule{4-13}
LLaVA-Rad~\cite{chaves2025llavarad} & 6.8B & ${\sim}$0.70\,/\,0.35 & \chexci{0.297}{.285}{.313} & \chexci{0.228}{.221}{.236} & \chexci{0.540}{.523}{.556} & \chexci{0.366}{.341}{.386} & \chexci{1.17}{1.14}{1.20} & \chexci{0.487}{.477}{.496} & \chexci{0.214}{.209}{.218} & \chexci{0.521}{.504}{.538} & \chexci{0.308}{.270}{.341} & \chexci{1.21}{1.20}{1.23} \\
\chexrowrule{4-13}
MAIRA-2~\cite{bannur2024maira2} & 6.9B & 1.40\,/\,${\sim}$0.44 & \chexci{0.308}{.294}{.322} & \chexci{0.248}{.240}{.256} & \chexci{0.551}{.532}{.570} & \chexci{0.337}{.318}{.353} & \chexci{1.14}{1.11}{1.17} & \chexci{0.597}{.589}{.606} & \chexci{0.333}{.328}{.338} & \chexci{0.523}{.507}{.540} & \chexci{0.280}{.250}{.307} & \chexciu{0.72}{.71}{.74} \\
\chexrowrule{4-13}
RaDialog~\cite{pellegrini2025radialog} & 6.8B & 0.17\,/\,${\le}$0.26 & \chexci{0.271}{.260}{.284} & \chexci{0.228}{.222}{.234} & \chexci{0.545}{.527}{.563} & \chexci{0.359}{.339}{.374} & \chexci{1.20}{1.17}{1.23} & \chexci{0.460}{.452}{.469} & \chexci{0.233}{.229}{.237} & \chexci{0.433}{.416}{.449} & \chexci{0.250}{.229}{.268} & \chexci{1.04}{1.02}{1.05} \\
\chexrowrule{4-13}
RGRG~\cite{tanida2023rgrg} & 0.6B & 0.17 & \chexci{0.272}{.259}{.286} & \chexci{0.228}{.222}{.234} & \chexci{0.498}{.477}{.516} & \chexci{0.278}{.264}{.288} & \chexci{1.28}{1.26}{1.31} & \chexci{0.614}{.604}{.623} & \chexci{0.269}{.265}{.273} & \chexci{0.437}{.421}{.452} & \chexci{0.181}{.163}{.197} & \chexci{0.91}{.90}{.93} \\
\chexrowrule{4-13}
ChEX~\cite{muller2024chex} & 1.1B & 0.38\,/\,0.24 & \chexci{0.197}{.188}{.205} & \chexci{0.200}{.194}{.205} & \chexci{0.504}{.486}{.520} & \chexci{0.316}{.301}{.329} & \chexci{1.42}{1.40}{1.44} & \chexci{0.230}{.224}{.236} & \chexci{0.155}{.151}{.159} & \chexci{0.344}{.330}{.357} & \chexci{0.205}{.186}{.223} & \chexci{1.37}{1.36}{1.39} \\
\midrule
InternVL3~\cite{zhu2025internvl3} & 7.9B & n/d & \chexci{0.182}{.166}{.199} & \chexci{0.142}{.137}{.147} & \chexci{0.408}{.389}{.428} & \chexci{0.184}{.172}{.195} & \chexci{1.57}{1.55}{1.58} & \chexciu{0.639}{.625}{.651} & \chexci{0.187}{.184}{.190} & \chexci{0.311}{.294}{.328} & \chexci{0.112}{.102}{.124} & \chexci{1.29}{1.28}{1.31} \\
\chexrowrule{4-13}
Qwen3.5~\cite{qwen35} & 9.4B & n/d & \chexci{0.309}{.295}{.324} & \chexci{0.219}{.213}{.226} & \chexci{0.501}{.478}{.521} & \chexci{0.276}{.262}{.289} & \chexci{1.25}{1.22}{1.28} & \chexci{0.638}{.628}{.648} & \chexci{0.306}{.302}{.310} & \chexci{0.496}{.478}{.513} & \chexci{0.185}{.166}{.203} & \chexci{0.74}{.73}{.76} \\
\chexrowrule{4-13}
MedGemma~\cite{sellergren2025medgemma} & 4.3B & ${\sim}$0.23 & \chexci{0.280}{.265}{.297} & \chexci{0.216}{.209}{.224} & \chexci{0.458}{.435}{.480} & \chexci{0.232}{.219}{.247} & \chexci{1.30}{1.26}{1.33} & \chexcib{0.649}{.640}{.658} & \chexci{0.337}{.333}{.341} & \chexci{0.491}{.474}{.509} & \chexci{0.190}{.170}{.209} & \chexci{0.76}{.75}{.78} \\
\midrule
RAD-DINO + 2B LM, end-to-end & 2.0B & 0.88\,/\,0.17 & \chexci{0.309}{.297}{.323} & \chexci{0.239}{.233}{.246} & \chexci{0.540}{.522}{.557} & \chexci{0.352}{.333}{.367} & \chexci{1.14}{1.12}{1.17} & \chexci{0.526}{.517}{.535} & \chexci{0.273}{.269}{.277} & \chexci{0.528}{.511}{.545} & \chexci{0.305}{.271}{.333} & \chexci{0.90}{.88}{.92} \\
\midrule
\textbf{Ours, template} & 110M & 0.88\,/\,0.17 & \chexci{0.289}{.278}{.300} & \chexci{0.254}{.247}{.261} & \chexcib{0.626}{.611}{.640} & \chexcib{0.409}{.389}{.426} & \chexci{1.14}{1.11}{1.17} & \chexci{0.510}{.503}{.518} & \chexciu{0.338}{.333}{.343} & \chexcib{0.556}{.541}{.572} & \chexcib{0.338}{.309}{.365} & \chexcib{0.67}{.66}{.69} \\
\chexrowrule{4-13}
\textbf{Ours, 2B writer} & 2.0B & 0.88\,/\,0.17 & \chexcib{0.320}{.307}{.334} & \chexciu{0.265}{.258}{.273} & \chexciu{0.606}{.590}{.623} & \chexciu{0.403}{.383}{.422} & \chexciu{1.07}{1.04}{1.10} & \chexci{0.586}{.577}{.594} & \chexci{0.289}{.285}{.293} & \chexciu{0.549}{.534}{.565} & \chexciu{0.317}{.290}{.340} & \chexci{0.77}{.76}{.79} \\
\bottomrule
\end{tabular}%
}
\vspace{\chexpagefivegap}%
\end{table*}

\begin{table*}[b]
\chexbottomfloat
\vspace*{\chexpagefivegap}%
\centering
\scriptsize
\chexcompactrules
\setlength{\tabcolsep}{1.5pt}
\caption{\textbf{Multi-image input.}
We evaluate the same studies as in \cref{tab:core-main} (one frontal image), using all available images. MAIRA-2 and CheXagent-2 use at most two. \textbf{Bold} and \underline{underlined} scores are the best and second-best results. \chexshadenote{ Shading marks a significant gain (\textcolor[rgb]{0.20,0.45,0.70}{\textbf{blue}}) or decline (\textcolor[rgb]{0.80,0.40,0.10}{\textbf{orange}}) relative to the  score for the same method in \cref{tab:core-main}. Darker shading indicates an abs. change of $>.02$ (.05 for RadCliQ).}}
\label{tab:core-multi-image}
\resizebox{\linewidth}{!}{%
\begin{tabular}{@{}l r r@{\hspace{7pt}} ccccc ccccc@{}}
\toprule
 & & & \multicolumn{5}{c}{MIMIC-CXR} & \multicolumn{5}{c}{OpenI (external)} \\
\cmidrule(lr){4-8}\cmidrule(l){9-13}
 & Params & CXR imgs (M) & GREEN$\uparrow$ & RG\textsubscript{ER}$\uparrow$ & miF1$\uparrow$ & maF1$\uparrow$ & RadCliQ$\downarrow$ & GREEN$\uparrow$ & RG\textsubscript{ER}$\uparrow$ & miF1$\uparrow$ & maF1$\uparrow$ & RadCliQ$\downarrow$ \\
\midrule
\chexnorankrow Constant ``normal'' & -- & -- & \chexci{0.160}{.140}{.180} & \chexci{0.171}{.161}{.181} & \chexci{0.037}{.027}{.047} & \chexci{0.008}{.006}{.010} & \chexci{1.52}{1.47}{1.57} & \chexci{0.615}{.604}{.626} & \chexci{0.323}{.318}{.327} & \chexci{0.420}{.402}{.439} & \chexci{0.040}{.039}{.041} & \chexci{0.77}{.75}{.79} \\
\chexnorankrow Most common report & -- & -- & \chexci{0.188}{.170}{.207} & \chexci{0.163}{.155}{.171} & \chexci{0.037}{.027}{.047} & \chexci{0.008}{.006}{.010} & \chexci{1.51}{1.47}{1.55} & \chexciu{0.649}{.639}{.659} & \chexci{0.300}{.296}{.304} & \chexci{0.420}{.402}{.439} & \chexci{0.040}{.039}{.041} & \chexci{0.78}{.77}{.80} \\
\chexnorankrow Random report & -- & -- & \chexci{0.142}{.132}{.152} & \chexci{0.136}{.131}{.141} & \chexci{0.238}{.225}{.252} & \chexci{0.140}{.131}{.150} & \chexci{1.61}{1.59}{1.64} & \chexci{0.300}{.290}{.309} & \chexci{0.162}{.158}{.166} & \chexci{0.136}{.126}{.147} & \chexci{0.071}{.062}{.079} & \chexci{1.44}{1.42}{1.46} \\
\midrule
CheXagent-2~\cite{chen2024chexagent} & 3.1B & 1.08 & \chexvsdown{28} \chexci{0.301}{.288}{.316} & \chexci{0.260}{.254}{.268} & \chexci{0.573}{.559}{.588} & \chexci{0.377}{.355}{.394} & \chexci{1.37}{1.35}{1.40} & \chexvsup{60} \chexci{0.458}{.446}{.470} & \chexvsup{60} \chexci{0.248}{.242}{.254} & \chexvsup{28} \chexci{0.417}{.400}{.433} & \chexci{0.293}{.268}{.315} & \chexvsup{60} \chexci{1.31}{1.30}{1.33} \\
\chexrowrule{4-13}
CheXOne~\cite{chexone} & 3.8B & ${\sim}$1.2 & \chexci{0.317}{.304}{.331} & \chexcib{0.301}{.295}{.309} & \chexvsup{28} \chexci{0.597}{.582}{.612} & \chexci{0.341}{.327}{.354} & \chexcib{0.98}{.95}{1.00} & \chexci{0.611}{.602}{.620} & \chexcib{0.346}{.341}{.351} & \chexciu{0.555}{.539}{.572} & \chexci{0.278}{.253}{.302} & \chexcib{0.67}{.65}{.68} \\
\chexrowrule{4-13}
MAIRA-2~\cite{bannur2024maira2} & 6.9B & 1.40\,/\,${\sim}$0.44 & \chexci{0.306}{.293}{.320} & \chexvsdown{28} \chexci{0.245}{.238}{.252} & \chexvsup{28} \chexci{0.557}{.538}{.576} & \chexci{0.346}{.326}{.362} & \chexci{1.14}{1.11}{1.18} & \chexvsdown{28} \chexci{0.584}{.576}{.593} & \chexvsdown{28} \chexci{0.318}{.314}{.322} & \chexci{0.524}{.507}{.541} & \chexci{0.282}{.252}{.311} & \chexvsdown{28} \chexci{0.75}{.74}{.77} \\
\midrule
InternVL3~\cite{zhu2025internvl3} & 7.9B & n/d & \chexvsdown{60} \chexci{0.134}{.125}{.143} & \chexvsdown{28} \chexci{0.124}{.120}{.128} & \chexci{0.412}{.390}{.434} & \chexci{0.186}{.176}{.197} & \chexvsdown{60} \chexci{1.63}{1.61}{1.64} & \chexvsdown{60} \chexci{0.309}{.298}{.321} & \chexvsdown{60} \chexci{0.137}{.133}{.139} & \chexvsdown{60} \chexci{0.160}{.149}{.171} & \chexvsdown{60} \chexci{0.082}{.074}{.092} & \chexvsdown{60} \chexci{1.57}{1.56}{1.58} \\
\chexrowrule{4-13}
Qwen3.5~\cite{qwen35} & 9.4B & n/d & \chexvsup{28} \chexciu{0.318}{.303}{.335} & \chexvsup{28} \chexci{0.233}{.226}{.241} & \chexvsup{28} \chexci{0.515}{.495}{.534} & \chexci{0.281}{.266}{.295} & \chexvsup{28} \chexci{1.22}{1.18}{1.25} & \chexvsup{60} \chexcib{0.669}{.660}{.679} & \chexvsup{60} \chexci{0.327}{.323}{.331} & \chexvsup{28} \chexci{0.515}{.498}{.532} & \chexvsup{60} \chexci{0.221}{.195}{.249} & \chexvsup{28} \chexci{0.71}{.69}{.72} \\
\chexrowrule{4-13}
MedGemma~\cite{sellergren2025medgemma} & 4.3B & ${\sim}$0.23 & \chexvsdown{28} \chexci{0.275}{.261}{.290} & \chexvsdown{28} \chexci{0.209}{.202}{.216} & \chexci{0.462}{.439}{.484} & \chexci{0.226}{.213}{.239} & \chexvsdown{60} \chexci{1.36}{1.33}{1.38} & \chexci{0.647}{.637}{.656} & \chexvsdown{28} \chexci{0.329}{.325}{.334} & \chexci{0.484}{.467}{.501} & \chexci{0.191}{.172}{.210} & \chexvsdown{28} \chexci{0.79}{.77}{.80} \\
\midrule
\textbf{Ours, template} & 217M & 0.88\,/\,0.27 & \chexci{0.291}{.280}{.302} & \chexvsup{28} \chexci{0.255}{.249}{.262} & \chexcib{0.628}{.613}{.641} & \chexcib{0.415}{.395}{.433} & \chexvsup{28} \chexci{1.14}{1.11}{1.16} & \chexci{0.508}{.501}{.516} & \chexciu{0.337}{.332}{.342} & \chexcib{0.559}{.544}{.574} & \chexcib{0.334}{.304}{.362} & \chexciu{0.68}{.66}{.69} \\
\chexrowrule{4-13}
\textbf{Ours, 2B writer} & 2.1B & 0.88\,/\,0.27 & \chexvsup{28} \chexcib{0.327}{.313}{.341} & \chexvsup{28} \chexciu{0.270}{.261}{.278} & \chexvsup{28} \chexciu{0.612}{.596}{.627} & \chexvsup{28} \chexciu{0.412}{.392}{.431} & \chexvsup{28} \chexciu{1.06}{1.03}{1.09} & \chexci{0.587}{.579}{.596} & \chexci{0.289}{.285}{.293} & \chexci{0.549}{.533}{.565} & \chexciu{0.314}{.285}{.338} & \chexci{0.77}{.76}{.79} \\
\bottomrule
\end{tabular}%
}
\end{table*}

\begin{figure*}[t]
  \centering
  \includegraphics[width=\textwidth]{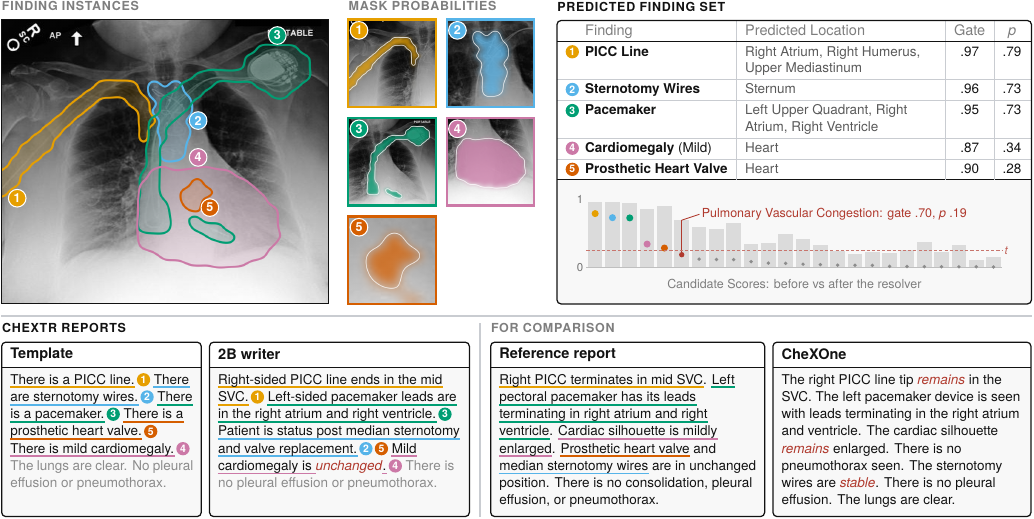}
  \caption{\textbf{Qualitative comparison.} Masks and finding set of
  \methodname{} for a test radiograph, and its two reports beside the reference
  and CheXOne. The pacemaker's mask follows the leads into the heart, and
  pulmonary vascular congestion passes the gate (bars) but its name scores
  below $t$ (dots) and is dropped, as in the reference. \underline{Underlined}
  sentences state the numbered instance, \textcolor[HTML]{8C8C8C}{grey} ones
  are normal statements without an instance, and
  \textcolor[HTML]{B03A2E}{\emph{red italics}} assert change without a prior
  image.}
  \label{fig:qualitative}
\end{figure*}

\noindent\textbf{Region masks.}\label{subsec:masks}
We obtain the mask target of \cref{subsec:training} through models trained on
PAX-Ray++~\cite{seibold2023paxraypp}, extended to the \gndclasses{} regions of
$\mathcal{A}$. A finding's target is the fraction of each patch covered by the
union of its regions (\cref{fig:data}). A device finding instead takes the mask
of a device detector~\cite{seibold2026synthfb,tang2021clip} when the detected
type matches and a learned score over confidence and region overlap accepts it.
A device finding without an accepted detection trains without a mask
(\mthCDevNoMaskPct{}\%).
Detections matching no reported device are discarded so that every mask rests
on a statement of the report. Statistics of the targets and 
construction details are listed in the supplementary.

\section{Experiments}
\label{sec:experiments}

\chexpinfloats
\subsection{Setup}
\label{subsec:setup}

\noindent\textbf{Training.}
We train on the official MIMIC-CXR train split
~\cite{johnson2019mimiccxr} and fit thresholds and template on
the validation split.

\noindent\textbf{Metrics.}
Every model is passed pure images as input to
write the FINDINGS section, which is scored against that of the original report
(\cref{tab:core-main,tab:core-multi-image,tab:core-verbaliser}). We
report GREEN~\cite{ostmeier2024green}, an LLM judge that counts clinical errors, and RadCliQ-v1~\cite{yu2023radcliq} (lower is better).
RadGraph-F1~\cite{jain2021radgraph} (RG\textsubscript{ER}~\cite{delbrouck2022semanticrewards}) matches clinical entities and whether each takes part in a relation.
Clinical efficacy~\cite{chen2020r2gen} is CheXbert micro- and macro-F1 over 14
classes (miF1/maF1)~\cite{smit2020chexbert}, the latter weighing a rare finding as much as a common one.\\
Finding sets (\cref{tab:core-det-ablation,tab:core-fusion}) are scored
against the reference parse of the test split (\cref{sec:data}) by miF1 and maF1
over the finding names of a study.

\noindent\textbf{Statistics.}
Intervals are 95\% bootstraps over patients. The interval of a difference $\Delta$
comes from a paired bootstrap, with both models scored on the same resamples.
We call a difference significant when this interval excludes zero ($p<0.05$,
two-sided). In tables, intervals are
distances to their bounds, and the best is \textbf{bold}, the second \underline{underlined}.

\noindent\textbf{Baselines and floors.}
We compare against seven specialist report generators and three
general-purpose or medically tuned VLMs, each run through its authors'
interface and decoded greedily unless they prescribe otherwise. Three \emph{floors} use no image or
language model and write a constant normal report, the most common or a
random training report.

\subsection{Comparison with report generators}
\label{subsec:main-results}
We, first, investigate whether reports written from an explicit finding set can compete with those produced by end-to-end  generators.
\Cref{tab:core-main} compares the models from one frontal image of each
study, on the \setupMimicStudies{} studies of the MIMIC-CXR test split with a
FINDINGS section and on
OpenI~\cite{demnerfushman2016openi}, an external test set of
\setupOpeniStudies{} studies on which nothing is trained or fitted.

\noindent\textbf{MIMIC-CXR.}
Both \methodname rows have higher CheXbert micro- and macro-F1 than every
compared model while using a seventh of CheXOne's training radiographs. The template reaches this without a language model and with
the fewest parameters of the learned rows. As it can state nothing but the
predicted findings, this lead is that of the finding set. On RG\textsubscript{ER} and RadCliQ, which score the full text, CheXOne leads. From the same
findings, the 2B writer is level with CheXOne on GREEN and narrows the gap on
the other two, so part of this gap is wording.
The end-to-end control shares our encoder and language model, yet falls below the 2B writer on every metric of both datasets.

\noindent\textbf{OpenI.}
On this external test set, the template again has the highest CheXbert micro-
and macro-F1, and its gap on RG\textsubscript{ER} and RadCliQ closes without a writer,
as it ties CheXOne for the best RadCliQ and is second on RG\textsubscript{ER}. The 2B
writer raises GREEN over the template but worsens for RG\textsubscript{ER} and RadCliQ, as
it keeps the wording of the MIMIC-CXR reports it was trained on. Even where
the finding set is empty and both write one normal report with identical
labels, the writer is performing worse on both. On GREEN, no model exceeds the
most common training report, so this score alone does not separate the
generators here.

\subsection{Multi-image input}
\label{subsec:multi-image}

It is commonly assumed that additional views improve the diagnostic
ability of a radiologist~\cite{feigin2010lateral}. We test whether this holds
for report generation by comparing one frontal image per study
(\cref{tab:core-main}) with all images of the same studies
(\cref{tab:core-multi-image}).
Of the MIMIC-CXR test studies, \setupMimicSeveral{} hold more than one image, including
\setupMimicLateral{} a lateral and \setupMimicSecondFrontal{} a second
frontal. Of the OpenI studies, \setupOpeniLateral{} hold a lateral.\\
The assumption does not hold for every model, as most compared models lose
significantly on at least one metric and only Qwen3.5 gains on both test sets.
In contrast, \methodname gains from the additional views on MIMIC-CXR and holds
its scores on OpenI. With \corehead{flparams} parameters and a fixed template
it keeps the highest micro- and macro-F1 on both test sets, and with the 2B
writer it reaches the highest GREEN on MIMIC-CXR, while the compared models use
\corehead{flminparams} to \corehead{flmaxparams} parameters. This suggests that
the benefit of additional views depends on how their information is combined.

\noindent\textbf{Fusing finding instances.}
\Cref{tab:core-fusion} validates the resolver against rule-based and learned
fusions of the same candidates, scored by detection F1 against the reference
parse. Reading the lateral instead of the frontal, concatenation and union all
lower both F1, and no other rule-based
fusion~\cite{solovyev2021wbf,chen2022proben} is significantly better than the
single image. Of the learned fusions, stacking~\cite{wolpert1992stacked} and
attention MIL~\cite{ilse2018attention} gain on macro-F1 alone, while learned
NMS~\cite{hosang2017nms,hu2018relation} and
linking~\cite{varlamova2024mammnet} lose micro-F1. The resolver, row Ours, is
the only fusion significantly better than the single image on both F1, and on
a single frontal image it is level with the predictor alone.

\subsection{Localised findings}
\label{subsec:grounding}

For humans, a report is easier to verify when each finding points to its image
region~\cite{tanida2023rgrg,bannur2024maira2}. We test whether \methodname
localises the findings it reports correctly by comparing the region of each
finding (\cref{subsec:fsp}) with the boxes radiologists drew for it
(\cref{tab:core-grounding}). MS-CXR~\cite{boecking2022biovil} adds finding boxes to
\setupMscxrImages{} MIMIC-CXR images. Of these, \setupMscxrTrain{} lie in the
official training and \setupMscxrVal{} in the validation split. We test on
the \setupMscxrTest{} images of the test split.
VinDr-CXR~\cite{nguyen2022vindrcxr} is external. Of its \setupVindrImages{}
test images, \setupVindrBoxed{} carry radiologist boxes. It releases no
patient identifiers, so its intervals resample images. The findings a
report names are assigned to the expert categories in the same way for every
model. We report the share of boxed findings a report names
(Stated), the share of named ones whose region holds the box centre (Hit), and
Acc, the share named with an IoU of at least $t$, averaged over
$t \in [0, .95]$. On VinDr-CXR, which boxes every finding of its classes, F1
compares the named with the boxed classes over all test images. RGRG~\cite{tanida2023rgrg},
MAIRA-2~\cite{bannur2024maira2} and ChEX~\cite{muller2024chex}, which is trained on VinDr-CXR boxes, write a report with boxes.
Since CheXagent-2~\cite{chen2024chexagent} and CheXOne~\cite{chexone} have no explicit localized reporting task, we locate each sentence of their
findings with their own phrase grounding. An oracle is told the
categories and places each at its usual location.

\methodname achieves the highest score on every metric of both datasets. On
the external VinDr-CXR, Acc stays low for every model. These results support
using the predicted regions to guide inspection of reported findings, although
the masks indicate anatomical regions rather than precise lesion boundaries.

\subsection{Qualitative results}
\label{subsec:qualitative}

\Cref{fig:qualitative} follows how \methodname gets from its findings to a report.
Every sentence that states a finding carries its number leading to its associated mask. Parsed back, we find that \wfBackedPct{}\% of the predicted findings for
the test reports appear in a cited line. We further see that the resolver keeps the prosthetic heart valve but
drops pulmonary vascular congestion despite its high gate. The writer places the pacemaker leads in the right atrium and right
ventricle, the regions predicted for this finding. Like
CheXOne~\cite{chexone}, the writer asserts change without a prior image.

\subsection{Ablations}
\label{subsec:ablations}

\noindent\textbf{Predictor.}
We test whether predicting findings as a set costs accuracy against a
multi-label classifier trained by the same recipe
(\cref{tab:core-det-ablation}), scored against the reference parse. Parsed back into findings, the reports of the end-to-end control of \cref{tab:core-main} stay below retrieving the most similar training image in micro-F1, although its GREEN on MIMIC-CXR is close to that of CheXOne. With names alone, the set predictor matches the classifier in
micro-F1, \detabl{setmi} against \detabl{clsmi}, at a lower macro-F1.
Location and attribute heads lower the
classifier's macro-F1 and rare-name recall, while both rise for the set
predictor. With the same heads, the set predictor is above the classifier on
every score. The mask head raises macro-F1, strict matching and rare-name
recall, and longer training raises strict matching further.

\noindent\textbf{Writer.}
To measure how much the writer matters, \cref{tab:core-verbaliser} exchanges
it on the same findings, from the reference parse and from the predicted ones. A checklist of
common negatives raises GREEN and RG\textsubscript{ER} of the template but not its
CheXbert F1, and tuning its normals, phrasing and hedges on validation reports raises every score. Of its inputs, the location carries most
of the RG\textsubscript{ER}, certainty and severity little. From the
reference parse, the tuned template leads on CheXbert F1 and the writer
trained on reports alone on GREEN, RG\textsubscript{ER} and RadCliQ. Training it on
predicted findings as well raises its CheXbert F1 on predicted findings and
lowers its GREEN. Alternative names, their confidence and captions change
little beyond that. As training on predicted findings also ties the writer to
one predictor, we train it on reports alone. On CheXbert F1, the gap between
the reference parse and the predicted findings is far larger than any gap
between writers.
\section{Discussion and Conclusion}
\label{sec:limitations}
\label{sec:discussion}
\label{sec:conclusion}

In this work, we asked whether recognising findings and writing reports should be treated as the same task. Our results indicate they need not be. 

We presented \methodname, which separates them through an explicit set of
findings, predicted as instances with a name, location, attributes and mask,
resolved across the views of a study and stated by an exchangeable writer.
With a fixed template and \corehead{oursparams} parameters, \methodname states
findings more accurately on MIMIC-CXR and OpenI than every compared generator,
from a seventh of CheXOne's training radiographs. An end-to-end generator
with the encoder, data and language model size of our 2B writer stays below it
on every metric, and the choice of writer moves clinical efficacy far less
than the findings it is given. Of its stated findings, \wfBackedPct{}\% trace
to a predicted region, which radiologist boxes confirm more often than for
other generators.

Its training targets are parsed from reports and thus silver, yet this lets its supervision grow with every report archive. With
its thresholds fitted on validation, the resolver states each finding name
once, although its groups allow several instances of one name, and the writer sees a single study, so the change it words has no
prior to rest on.

We argue that this modular design holds considerable potential for accurate
report generation. Every gain of the
predictor reaches the report without retraining a writer, and the same set can
in future work extend to longitudinal analysis, where prior findings are read
beside the current ones to report change.

\noindent\textbf{Ethics.}
Since the data is de-identified and
public~\cite{johnson2019mimiccxr,demnerfushman2016openi,nguyen2022vindrcxr},
additional ethics review and consent were not required.

\section*{Acknowledgments}
C.S. is supported by the Medical Data Scientist Program of the Medical Faculty Heidelberg, Heidelberg University.

\section*{LLM Usage Statement}

The authors used Claude (Anthropic) to assist with code, experiments, figures
and tables, literature search, and drafting and editing of the manuscript. The
authors reviewed all content and take full responsibility for it. Language
models are also part of the research itself, as report parser, writer,
evaluation judge and compared models.

{
    \small
    \bibliographystyle{ieeenat_fullname}
    \bibliography{main}
}

\chexclearpage
\maketitlesupplementary
\appendix
\setcounter{topnumber}{1}
\setcounter{bottomnumber}{1}
\setcounter{topnumber}{2}
\setcounter{dbltopnumber}{3}
\setcounter{totalnumber}{4}
\renewcommand{\topfraction}{0.96}
\renewcommand{\dbltopfraction}{0.96}
\renewcommand{\textfraction}{0.04}
\renewcommand{\floatpagefraction}{0.55}
\renewcommand{\dblfloatpagefraction}{0.55}

\newcommand{\tgtFImages}{172{,}820}
\newcommand{\tgtFWithPct}{78}
\newcommand{\tgtFEmptyPct}{22}
\newcommand{\tgtFFindings}{556{,}125}
\newcommand{\tgtFPerImage}{3.2}
\newcommand{\tgtFNames}{3{,}156}
\newcommand{\tgtFRegions}{146}
\newcommand{\tgtFUncPct}{18}
\newcommand{\tgtFSevPct}{24}
\newcommand{\tgtFRepeatPct}{18}
\newcommand{\tgtFTopTenPct}{50}
\newcommand{\tgtFCoverHiNames}{97}
\newcommand{\tgtFCoverTopNames}{705}
\newcommand{\tgtFCoverTailNames}{2{,}451}
\newcommand{\tgtFFreqNames}{74}
\newcommand{\tgtFFreqPct}{87}
\newcommand{\tgtFRareNames}{2{,}453}
\newcommand{\tgtFRarePct}{1.0}
\newcommand{\tgtFDevMaskPct}{76}
\newcommand{\tgtLImages}{97{,}996}
\newcommand{\tgtLWithPct}{72}
\newcommand{\tgtLEmptyPct}{28}
\newcommand{\tgtLFindings}{258{,}007}
\newcommand{\tgtLPerImage}{2.6}
\newcommand{\tgtLNames}{2{,}316}
\newcommand{\tgtLRegions}{109}
\newcommand{\tgtLUncPct}{18}
\newcommand{\tgtLSevPct}{27}
\newcommand{\tgtLRepeatPct}{13}
\newcommand{\tgtLTopTenPct}{48}
\newcommand{\tgtLCoverHiNames}{104}
\newcommand{\tgtLCoverTopNames}{736}
\newcommand{\tgtLCoverTailNames}{1{,}580}
\newcommand{\tgtLFreqNames}{42}
\newcommand{\tgtLFreqPct}{77}
\newcommand{\tgtLRareNames}{1{,}803}
\newcommand{\tgtLRarePct}{1.6}
\newcommand{\tgtLDevMaskPct}{64}
\newcommand{\tgtCoverHi}{90}
\newcommand{\tgtCoverTop}{99}
\newcommand{\tgtCoverRest}{1}
\newcommand{\tgtFreqMin}{1{,}000}
\newcommand{\tgtRareMax}{10}
\newcommand{\tgtTopK}{20}
\newcommand{\tgtRegionsAdded}{10}

\newcommand{\supBootN}{1{,}000}
\newcommand{\supBootSeed}{42}
\newcommand{\supMimicPatients}{289}
\newcommand{\supOpeniPatients}{3{,}198}
\newcommand{\supVerRadgraph}{0.1.18}
\newcommand{\supVerGreen}{0.0.12}
\newcommand{\supVerBertscore}{0.3.13}
\newcommand{\supVerCoco}{1.2}
\newcommand{\supVerVllm}{0.25.1}
\newcommand{\supGreenMaxLen}{2{,}048}
\newcommand{\supFsBootSeed}{0}
\newcommand{\supRefStudies}{2{,}346}
\newcommand{\supRefFindings}{10{,}410}
\newcommand{\supRefEmpty}{272}
\newcommand{\supRefNames}{442}
\newcommand{\supMacroNames}{112}
\newcommand{\supKMacro}{10}
\newcommand{\supTailN}{1{,}000}
\newcommand{\supTailNames}{361}
\newcommand{\supTailFindings}{1{,}328}
\newcommand{\supTailPct}{13}
\newcommand{\supRepStudies}{591}
\newcommand{\supRepPct}{25}
\newcommand{\supRepDistinct}{360}
\newcommand{\supRepDistinctPct}{15}
\newcommand{\supSurplus}{852}
\newcommand{\supSurplusPct}{8}
\newcommand{\supRecallCapPct}{92}
\newcommand{\supGridLo}{0.05}
\newcommand{\supGridHi}{0.95}
\newcommand{\supGridStep}{0.025}
\newcommand{\supGndStep}{.05}
\newcommand{\supGndMax}{.95}
\newcommand{\supVindrBoxes}{2{,}697}
\newcommand{\supGndMinFracPct}{0.5}
\newcommand{\supGndMaskThr}{0.5}
\newcommand{\supFloorPool}{3{,}000}
\newcommand{\supFloorFindings}{2{,}070}
\newcommand{\supAdjSentences}{1{,}885}
\newcommand{\supAdjFindings}{2{,}807}
\newcommand{\supProvAgreed}{19{,}563}
\newcommand{\supProvAgreedPct}{80}
\newcommand{\supProvAdj}{2{,}001}
\newcommand{\supProvAdjPct}{8}
\newcommand{\supProvRule}{1{,}010}
\newcommand{\supProvRulePct}{4}
\newcommand{\supProvConv}{1{,}444}
\newcommand{\supProvConvPct}{6}
\newcommand{\supProvHuman}{363}
\newcommand{\supProvHumanPct}{1}
\newcommand{\supProvTotal}{24{,}381}
\newcommand{\supProvDropped}{1{,}165}
\newcommand{\supRevAgreedSampled}{50}
\newcommand{\supRevAgreedKept}{48}
\newcommand{\supBranchAgreed}{19{,}613}
\newcommand{\supBranchAgreedPct}{80}
\newcommand{\supRevOwn}{313}
\newcommand{\supRevOwnPct}{1}
\newcommand{\supAdjLow}{243}
\newcommand{\supRevLow}{207}
\newcommand{\supGoldRenamed}{1{,}025}
\newcommand{\supGoldFinal}{24{,}402}
\newcommand{\supCapPairs}{725{,}326}
\newcommand{\supCapFRewritePct}{63}
\newcommand{\supCapFFlaggedPct}{19}
\newcommand{\supCapFVerbatimPct}{29}
\newcommand{\supCapFSourcePct}{8}
\newcommand{\supCapLRewritePct}{19}
\newcommand{\supCapLFlaggedPct}{4}
\newcommand{\supCapLVerbatimPct}{13}
\newcommand{\supCapLSourcePct}{68}
\newcommand{\supCapMaxLen}{48}
\newcommand{\supRwTrain}{2{,}700}
\newcommand{\supRwEpochs}{3}
\newcommand{\supRwVal}{300}
\newcommand{\supRwGoodPct}{90}
\newcommand{\supRwBaseGoodPct}{69}
\newcommand{\supDevClasses}{19}
\newcommand{\supDevQueries}{200}
\newcommand{\supDevUnfrozen}{2}
\newcommand{\supDevSteps}{6{,}000}
\newcommand{\supDevImages}{1{,}122}
\newcommand{\supDevMasks}{4{,}027}
\newcommand{\supDevMimicImages}{780}
\newcommand{\supDevRadioImages}{175}
\newcommand{\supDevOtherImages}{167}
\newcommand{\supDevLatAdded}{800}
\newcommand{\supDevMinScore}{0.2}
\newcommand{\supDevEps}{0.05}
\newcommand{\supDevAccept}{0.7}
\newcommand{\supCtlIters}{20{,}000}
\newcommand{\supCtlBatch}{32}
\newcommand{\supCtlLrLora}{$2\times10^{-4}$}
\newcommand{\supCtlLrProj}{$10^{-4}$}
\newcommand{\supCtlLrEnc}{$10^{-5}$}
\newcommand{\supCtlLlrd}{0.8}
\newcommand{\supCtlWarmup}{1{,}000}
\newcommand{\supCtlRank}{16}
\newcommand{\supCtlMaxNew}{172}
\newcommand{\supCtlHours}{29}
\newcommand{\supPredHours}{15}
\newcommand{\supClsFlatLo}{.485}
\newcommand{\supClsFlatHi}{.489}
\newcommand{\supCtlBestIters}{10{,}000}
\newcommand{\supWrPredRows}{23{,}307}
\newcommand{\supWrMaxNew}{512}
\newcommand{\supWrSentences}{141{,}340}
\newcommand{\supWrNormalPct}{55}
\newcommand{\supGpu}{NVIDIA RTX PRO 4500 Blackwell}
\newcommand{\supParPredictor}{107M}
\newcommand{\supParResolver}{3.4M}
\newcommand{\supParWriterLm}{1.88B}
\newcommand{\supParCtlEncoder}{86.6M}
\newcommand{\supParCtlProjector}{5.8M}
\newcommand{\supParTemplate}{110M}
\newcommand{\supParTemplateMulti}{217M}
\newcommand{\supParWriter}{2.0B}
\newcommand{\supParWriterMulti}{2.1B}
\newcommand{\supParControl}{2.0B}
\newcommand{\supAugDeg}{5}
\newcommand{\supAugShiftPct}{2}
\newcommand{\supAugScalePct}{3}
\newcommand{\supSoftWeight}{0.5}
\newcommand{\supRvTokMlp}{512}
\newcommand{\supRvAgrMlp}{64}
\newcommand{\supRvHeadMlp}{256}
\newcommand{\supWrMinPool}{30}
\newcommand{\supChangeStudiesPct}{40}
\newcommand{\supChangeSentPct}{13}
\newcommand{\supWfSideComparable}{8{,}650}
\newcommand{\supWfSideConflict}{6}
\newcommand{\supMimicLat}{1{,}039}
\newcommand{\supMimicNoLat}{1{,}307}
\newcommand{\supOpeniLat}{2{,}942}
\newcommand{\supOpeniNoLat}{256}
\newcommand{\supOeN}{3{,}198}
\newcommand{\supOeEmpty}{1{,}389}
\newcommand{\supOeEmptyPct}{43}
\newcommand{\supOeEmptyRg}{0.077}
\newcommand{\supOeEmptyRc}{0.13}
\newcommand{\supOeEmptyGreen}{0.098}
\newcommand{\supOeRestRg}{0.029}
\newcommand{\supOeRestRc}{0.07}
\newcommand{\supOeRestGreen}{0.058}
\newcommand{\supFsN}{1{,}216}
\newcommand{\supFsBoth}{110}
\newcommand{\supFsFrontalOnly}{177}
\newcommand{\supFsSingleF}{.456}
\newcommand{\supFsSingleMa}{.203}
\newcommand{\supFsOursAllF}{.467}
\newcommand{\supFsOursAllMa}{.217}
\newcommand{\supThrP}{.480}
\newcommand{\supThrR}{.499}
\newcommand{\supThrN}{4.6}
\newcommand{\supThrTLow}{.20}
\newcommand{\supThrTHigh}{.30}
\newcommand{\supThrFMin}{.484}
\newcommand{\supThrFMax}{.489}
\newcommand{\supThrPLow}{.439}
\newcommand{\supThrPHigh}{.516}
\newcommand{\supThrRLow}{.540}
\newcommand{\supThrRHigh}{.459}
\newcommand{\supAreaMsOurs}{21}
\newcommand{\supAreaMsGold}{17}
\newcommand{\supAreaVdOurs}{17}
\newcommand{\supAreaVdGold}{5}
\newcommand{\supEncImages}{882{,}775}
\newcommand{\supBothViewImages}{270{,}816}

\noindent
This supplement follows the order in which a result of the main paper comes
about. \Cref{sec:targets} describes how the training targets and the
reference of the evaluation are built, as every later part depends on them.
\Cref{sec:implementation} lists the settings of the predictor, the resolver,
the writers and the end-to-end control. \Cref{sec:compared-models} states how
each compared model is run, and \cref{sec:evaluation-details} how each table
of the main paper is scored. \Cref{sec:additional-results} closes with
further measurements and qualitative examples.

\section{Training targets}
\label{sec:targets}

This section describes how the targets of \cref{sec:data} are built and gives
their statistics. Three counts of anatomy recur in it.
PAX-Ray++~\cite{seibold2023paxraypp} annotates \paxclasses{} region classes,
and we add \tgtRegionsAdded{} regions, so the set $\mathcal{A}$ holds
\gndclasses{} regions, each with a mask. A report is not parsed into these
regions directly but into \annConcepts{} anatomical concepts, which are the
terms the parse prompt offers for a location, and a fixed map gives the
regions of each concept on frontal and on lateral images.

\subsection{Reference parse}
\label{sec:annotation-scheme}

The reference parse is the structured form of the MIMIC-CXR test reports
against which finding sets are scored. To obtain an evaluation set that is as
accurate as possible, each test sentence is parsed twice by Claude
Opus~5.5~\cite{anthropic2026claudeopus}, a third pass of the same model
adjudicates disagreements, and an expert reviewer decides the cases that
remain.

\noindent\textbf{Unit.}
The parser reads one sentence at a time. The \annCorpusSentences{} unique
sentences of the MIMIC-CXR test reports are parsed in isolation, without the
rest of the report, the image or the indication, which keeps boilerplate from
dominating the agreement between parses. Reports were processed via Amazon
Bedrock~\cite{aws2026bedrock} with zero data retention, in accordance with
PhysioNet's guidance on the use of LLMs with credentialed
data~\cite{physionet2023gpt}.

\noindent\textbf{Finding.}
A sentence yields a list of findings. Each consists of a name, which is the
bare condition or device, its anatomical concepts and four attribute axes with
closed value sets (\cref{tab:annotation-axes}). Keeping every modifier on its
axis prevents \emph{small left effusion} and \emph{effusion} from becoming
different names. The change axis is parsed but not used as a target in this
work, which considers a single study without its priors.

\begin{table}[ht]
  \centering
  \footnotesize
  \setlength{\tabcolsep}{4pt}
  \begin{tabular}{@{}lp{0.62\linewidth}c@{}}
    \toprule
    Axis & Values & Target \\
    \midrule
    Presence  & present, absent, uncertain & \checkmark \\
    Certainty & definite, probable, possible, cannot exclude & \checkmark \\
    Severity  & none, minimal, mild, moderate, severe, large & \checkmark \\
    Change    & new, increased, decreased, improved, worsened, stable, resolved & \\
    \bottomrule
  \end{tabular}
  \caption{Attribute axes of a finding with their closed value sets. Certainty, severity and change may also
    be unstated. The last column marks the axes that are targets of \methodname.}
  \label{tab:annotation-axes}
\end{table}

\noindent\textbf{Anatomy.}
A location is a set of the \annConcepts{} concepts. Structures without a
concept of their own, such as the superior vena cava or the costophrenic
angle, are assigned to the nearest concept by a fixed table in the prompt.

\noindent\textbf{Prompt.}
\Cref{tab:annotation-prompt} reproduces the prompt in full. It has no rules
for single categories, since the possible findings are too many to list, and
instead applies seven principles to every finding.
\begin{itemize}[leftmargin=2.2em, labelwidth=1.6em, labelsep=0.6em, itemsep=0.15em, topsep=0.3em]
  \item[\textbf{P1}] \emph{Faithful epistemic status.} Presence and certainty
    transcribe what the sentence asserts, and a hedge is not resolved.
  \item[\textbf{P2}] \emph{The finding's own anatomy.} A finding is located by
    its own site and not by other nouns of the sentence.
  \item[\textbf{P3}] \emph{One concept from the vocabulary}, at the grain the
    text gives.
  \item[\textbf{P4}] \emph{Normal is not a finding, a named negative is.}
    \textsl{``lungs are clear''} yields nothing and \textsl{``no
    pneumothorax''} an absent finding.
  \item[\textbf{P5}] \emph{The bare name only}, with every modifier on its
    axis.
  \item[\textbf{P6}] \emph{Differentials keep the sign.} The sign is present
    and each candidate cause an uncertain finding at the same site.
  \item[\textbf{P7}] \emph{Completeness.} Where the extent is unstated, a
    finding takes every plausibly affected region, so an unstated side means
    both sides and a device its whole course.
\end{itemize}
By the last principle, the regions of a finding are a superset of what the
text strictly states. They mark where a finding may lie rather than its
extent.

\begin{table*}[p]
  \centering
  \begin{minipage}{\textwidth}
  \fontsize{7.2}{8.6}\selectfont
\begin{verbatim}
# Gold EXTRACT task -- chest-radiograph report annotator (principled, vocabulary-grounded)

You are an expert chest-radiograph annotator building the GOLD reference parse of isolated,
de-identified report sentences. You extract the correct parse from scratch -- you are not grading anyone.

**COMPLIANCE (hard):** each item is ONE isolated, de-identified sentence with NO patient/case context.
Annotate only. Never try to identify anyone; never invent context. Output only item_id-keyed annotations.

## What you produce -- for each sentence, every finding / observation / device:
```
{"finding_text":"<the CORE finding / condition / device ONLY -- see P5>",
 "anatomy":[{"canonical":"<ONE concept copied from the CANONICAL VOCABULARY below>",
             "region_freetext":"<the precise structure in your own words>",
             "laterality":"left|right|bilateral|midline|none",
             "role":"primary|device_body|device_tip",
             "no_exact_mask":true|false}],
 "presence":"present|absent|uncertain",
 "certainty":"definite|probable|possible|cannot_exclude|na",
 "severity":"none|minimal|mild|moderate|severe|large|na",
 "change":"new|increased|decreased|improved|worsened|stable|resolved|na"}
```

## Four principles -- apply to EVERY finding of EVERY category. There are no per-category rules.

**P1 * Faithful epistemic status.** `presence` / `certainty` encode exactly what the sentence's LANGUAGE
asserts -- no more. Plainly stated -> `present` / `definite`. Hedged, differential ("A vs B", "likely A",
"possibly", "could represent", "?A") -> `uncertain`, with `certainty` matching the hedge (*likely*->probable;
*possible* / *vs* / *versus*->possible; *cannot exclude*->cannot_exclude). Explicitly denied ("no", "without",
"free of", "resolved") -> `absent`. `severity` and `change` likewise ONLY as the sentence states them (else
  `na`).

**P2 * Ground to the finding's OWN anatomy, at the sentence's specificity -- never finer, never coarser.**
Use your anatomical knowledge to find the structure the finding is truly about (aortic tortuosity -> the
aorta; flattened diaphragm -> the diaphragm; a device's tip -> the vessel it ends in -- not merely the other
nouns in the sentence). Specificity mirrors the text: it names an exact rib / vertebra / lobe / chamber /
vessel -> use that; it gives only a region -> use the region (do NOT invent a level or number it didn't
state); it is unlocalized -> span the whole organ. A finding may occupy SEVERAL structures -- a device
(its body where it sits + its tip/leads at their organs), a diffuse process (the whole organ), a
differential (each candidate at its own site). List every structure as its own anatomy entry with `role`.

**P3 * Express each anatomy target as ONE canonical concept from the vocabulary below** (copied verbatim),
at the grain the text gives (a zone if it says "base/apex/upper/lower zone", a lobe if it says "lobe", the
whole lung if unlocalized -- pick the concept, do not mix grains). If NO concept exactly fits the structure,
choose the NEAREST concept, set `no_exact_mask:true`, and record the precise structure in `region_freetext`.
This one instruction covers every structure the vocabulary lacks -- never skip a finding, never invent a
concept outside the list. (`region_freetext` is ALWAYS filled with the precise anatomy in words.)

**P4 * Normal != finding; named negatives ARE findings.** A structure merely called normal / clear /
unremarkable / patent / intact, with no finding, is skipped. A NAMED negative ("no pneumothorax",
"no effusion", "no free air") IS a finding with `presence:absent` at that finding's own site.

**P5 * `finding_text` = the CORE finding only.** Strip everything that already lives in another field --
anatomy, laterality, severity/size, change, and a device's position/course. Give the bare condition or
device, lower-case, singular, canonical wording. The stripped modifiers are NOT lost; they are in
`anatomy` / `severity` / `laterality` / `change`. Examples:
- "small bilateral pleural effusions" -> `pleural effusion`  (size->severity, side->laterality)
- "heart size slightly increased (cardiomegaly)" -> `cardiomegaly`  (change->change)
- "right upper extremity PICC, tip in mid SVC" -> `PICC line`  (body/tip->anatomy)
- "clips projecting over the left upper quadrant" -> `surgical clips`  ; "levoscoliosis of the thoracic spine"
  -> `scoliosis`
- "opacity at the left costophrenic angle" -> `opacity`  ; "mild interval improvement in atelectasis" ->
  `atelectasis`
Prefer the standard radiographic term and merge trivial variants (pleural fluid->pleural effusion,
opacification->opacity, enlarged heart->cardiomegaly). Keep a device as one concept (PICC line, chest port,
\end{verbatim}
  \end{minipage}
  \caption{The parse prompt in full (1 of 3). Both parses of a sentence use this text and see nothing else of its report. Unicode arrows are written in ASCII and long lines are wrapped.}
  \label{tab:annotation-prompt}
\end{table*}
\begin{table*}[p]
  \centering
  \begin{minipage}{\textwidth}
  \fontsize{7.2}{8.6}\selectfont
\begin{verbatim}
pacemaker, NG tube, ET tube, tracheostomy tube, surgical clips, sternotomy wires, IVC filter).

**P7 * COMPLETENESS -- ground to ALL plausibly-affected instances (superset, never a minimal subset).**
- **Laterality:** if the sentence does NOT state a side (or says bilateral / both), ground to BOTH sides.
  NEVER default to one side. ("bilateral effusion" -> both bases; "central/perihilar" -> both mid zones;
  "a nodule" with no side -> both lungs.)
- **Devices -- whole course:** include EVERY organ along the device's path (body + all segments + tip):
  NG / feeding / enteric tube -> esophagus + stomach + below BOTH hemidiaphragms; PICC / central line ->
  entry vein + SVC (`upper mediastinum`) + right atrium; pacemaker / ICD -> chest-wall pocket + right
  atrium + right ventricle; ET tube -> trachea; spinal fixation hardware -> spine.
- **Compound / coarse regions -- all parts:** shoulder -> humerus + scapula + clavicle; "breasts" -> both;
  rib(s) with no level stated -> all ribs of that side (NEVER a guessed number); diffuse osseous -> all bones.
- **Whole-compartment (gas / diffuse) findings:** pneumomediastinum -> the WHOLE mediastinum;
  pneumopericardium -> heart/pericardium; free air -> below both hemidiaphragms.

**P6 * Differentials -- KEEP THE SIGN, fan the candidates.** When a sentence offers differential
interpretations of an observed sign ("opacity, likely atelectasis vs pleural fluid"; "blunting consistent
with effusion"; "haziness, could be edema vs pneumonia"):
- emit the observed SIGN itself (opacity, density, blunting, obscuration, haziness, lucency, silhouette
  loss) as ONE finding -- `presence:present`, `certainty:definite` -- at its site (it WAS seen);
- emit EACH named candidate diagnosis as its OWN finding -- `presence:uncertain`, `certainty` from the
  hedge (likely->probable; vs / versus / possible / could / suggests / consistent with->possible; cannot
  exclude->cannot_exclude) -- at the SAME site.
Do NOT fold the candidates into one finding, and do NOT drop the sign. If the sentence asserts a diagnosis
with NO hedge and NO separately-named sign, that is simply one `present` finding (P1) -- do not invent a sign.

## CANONICAL VOCABULARY -- the ONLY allowed `canonical` values (copy verbatim; nothing else is valid)

**LUNG (zones)** -- {left, right, bilateral} x {apical lung zone, upper lung zone, mid lung zone, base lung
  zone}
**LUNG (lobes)** -- left upper lobe, left lower lobe, right upper lobe, right middle lobe, right lower lobe
**LUNG (whole)** -- left lung (whole), right lung (whole), bilateral lung (whole)
**DIAPHRAGM** -- left hemidiaphragm, right hemidiaphragm, below left hemidiaphragm, below right hemidiaphragm,
  lateral diaphragm
**HEART** -- heart, heart atrium left, heart atrium right, heart ventricle left, heart ventricle right, heart
  myocardium
**VESSELS** -- aorta, ascending aorta, aortic arch, descending aorta, pulmonary artery, pulmonary trunc, left
  pulmonary artery, right pulmonary artery, inferior vena cava
**MEDIASTINUM / AIRWAY** -- cardiomediastinum, upper mediastinum, lower mediastinum, lateral anterior
  mediastinum, lateral middle mediastinum, lateral posterior mediastinum, trachea, tracheal bifurcation
**SPINE** -- spine, cervical spine, thoracic spine, lumbar spine, and per-level `vertebrae <C3..C7 | T1..T12 |
  L1..L5>` (e.g. `vertebrae t11`)
**RIBS** -- `<anterior | posterior | lateral anterior | lateral posterior> <N>th rib <left | right>` for
  N=1..12 (e.g. `posterior 7th rib left`); `rib_cartilage`
**OTHER BONES** -- clavicle left, clavicle right, lateral clavicles, scapula left, scapula right, lateral
  scapulas, left humerus, right humerus, sternum
**ABDOMEN** -- esophagus, stomach, duodenum, small bowel, liver, kidney left, kidney right, pancreas, below
  left hemidiaphragm, below right hemidiaphragm; and the four torso tiles: upper left quadrant, upper right
  quadrant, lower left quadrant, lower right quadrant. For any sub-diaphragmatic / abdominal finding PREFER
  the named organ or `below <side> hemidiaphragm` (precisely placed). The quadrants are whole-IMAGE tiles
  ("upper" = upper chest, "lower" = lower chest + upper abdomen) -- use them only as a COARSE fallback, and
  remember the abdomen sits in the LOWER quadrants.
**NECK** -- head-neck-region
**SOFT TISSUE** -- left breast, right breast, breasts

Use the COARSE concept when the text is coarse and the SPECIFIC one when the text is specific: "lower lung"
-> `left/right base lung zone` (a zone); "lower lobe" -> `left/right lower lobe` (a lobe); unlocalized "left
lung" -> `left lung (whole)`; "the aorta" -> `aorta`, but "aortic arch" -> `aortic arch`.

Coarse bone concepts for when the text does NOT give a level/number (NEVER guess a specific rib or
vertebra): `ribs`, `left ribs`, `right ribs`, `osseous structures (all bones)`, `spine`, `thoracic spine`,
`lumbar spine`, `cervical spine`.

### P3 nearest-mask table -- structures with NO dedicated mask (use EXACTLY these so annotators converge)
\end{verbatim}
  \end{minipage}
  \caption{The parse prompt, continued (2 of 3).}
  \label{tab:annotation-prompt-2}
\end{table*}
\begin{table*}[t]
  \centering
  \begin{minipage}{\textwidth}
  \fontsize{7.2}{8.6}\selectfont
\begin{verbatim}
| structure named in the sentence | canonical concept (set `no_exact_mask:true`, keep the real structure in
  `region_freetext`) |
|---|---|
| costophrenic angle | `<side> base lung zone` (bilateral -> both bases) |
| pleura / pleural space (general) | `<side> lung (whole)` |
| perihilar / hilar / central lung | `<side> mid lung zone` |
| superior vena cava (SVC) | `upper mediastinum` |
| cavoatrial junction | `heart atrium right` |
| subclavian / brachiocephalic / innominate vein | `upper mediastinum` |
| internal jugular vein / carotid | `head-neck-region` |
| mitral valve | `heart atrium left` |  * | tricuspid valve | `heart atrium right` |
| aortic valve / aortic root | `ascending aorta` |  * | pulmonic valve | `pulmonary trunc` |
| pericardium | `heart` |
| chest-wall / subcutaneous soft tissue | `upper <side> quadrant` |
| gastroesophageal junction | `stomach` | * | epigastric / hypochondrium / upper abdomen | `upper <side>
  quadrant` / `below <side> hemidiaphragm` |
| rib(s), level/number NOT stated | `<side> ribs` (or `ribs`) -- NEVER a numbered rib |
| diffuse / osseous, no bone named | `osseous structures (all bones)` |

## Output
Write JSONL, ONE line per input item (in order), to the path you are given:
`{"item_id":"g0000","findings":[ ... ]}`  -- a sentence with no real finding ->
  `{"item_id":...,"findings":[]}`.
Return only a one-line summary (items done, total findings). All detail goes in the file.
\end{verbatim}
  \end{minipage}
  \caption{The parse prompt, continued (3 of 3).}
  \label{tab:annotation-prompt-3}
\end{table*}

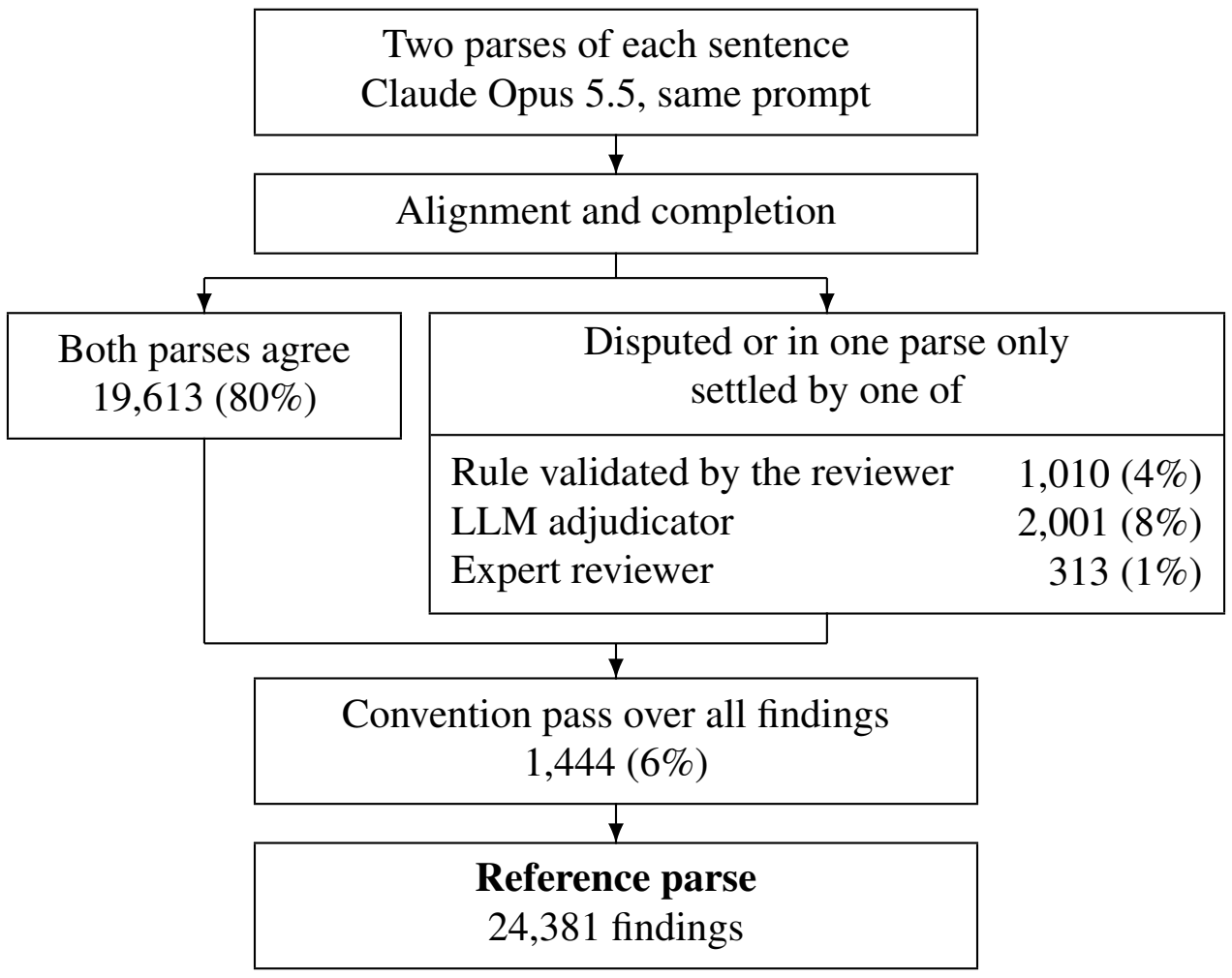
\begin{figure}[b]
  \centering
  \footnotesize
  \setlength{\unitlength}{1pt}
  \newcommand{\rectwo}[5]{\put(#1,#2){\framebox(#3,24){\parbox{#3pt}{\centering #4\\#5}}}}
  \newcommand{\recrow}[3]{\put(86,#1){\makebox(0,0)[bl]{#2}}\put(232,#1){\makebox(0,0)[br]{#3}}}
  \begin{picture}(236,186)
    \rectwo{48}{162}{140}{Two parses of each sentence}{Claude Opus~5.5, same prompt}
    \put(118,162){\vector(0,-1){8}}
    \put(48,139){\framebox(140,15){Alignment and completion}}
    \put(118,139){\line(0,-1){5}}
    \put(38,134){\line(1,0){121}}
    \put(38,134){\vector(0,-1){7}}
    \put(159,134){\vector(0,-1){7}}
    \rectwo{0}{103}{76}{Both parses agree}{\supBranchAgreed{} (\supBranchAgreedPct\%)}
    \put(82,69){\framebox(154,58){}}
    \put(159,117.5){\makebox(0,0)[b]{Disputed or in one parse only}}
    \put(159,108){\makebox(0,0)[b]{settled by one of}}
    \put(82,103.5){\line(1,0){154}}
    \recrow{92}{Rule validated by the reviewer}{\supProvRule{} (\supProvRulePct\%)}
    \recrow{82.5}{LLM adjudicator}{\supProvAdj{} (\supProvAdjPct\%)}
    \recrow{73}{Expert reviewer}{\supRevOwn{} (\supRevOwnPct\%)}
    \put(38,103){\line(0,-1){40}}
    \put(159,69){\line(0,-1){6}}
    \put(38,63){\line(1,0){121}}
    \put(118,63){\vector(0,-1){7}}
    \rectwo{48}{32}{140}{Convention pass over all findings}{\supProvConv{} (\supProvConvPct\%)}
    \put(118,32){\vector(0,-1){8}}
    \rectwo{48}{0}{140}{\textbf{Reference parse}}{\supProvTotal{} findings}
  \end{picture}
  \caption{From the two parses of a sentence to the reference parse. The reviewer validated each rule on a set of
    samples and decided the disputes on which the adjudicator was least confident. Each share counts the findings
    whose final form that step decided, without the \supProvDropped{} dropped candidates.}
  \label{fig:reconciliation}
\end{figure}

\noindent\textbf{Reconciliation.}
Every sentence is parsed twice, independently, with the same prompt, and
\cref{fig:reconciliation} lists the steps from the two parses to the
reference. A finding of one parse is paired with the finding of the other
parse that has the same normalised name and, among several of them, shares
the most concepts. Where no name is equal, it is paired with the finding
whose name shares the most words, provided that the Jaccard index of the two
word sets reaches one half, and otherwise it stays unpaired. A deterministic
completion then applies the conventions that no parse should derive anew,
namely the nearest-concept table, both sides where no side is stated, the
standard course of each device family and sites that belong to a finding,
such as the lung bases for a pleural effusion. Over the \annPaired{} findings
that both parses name, the name agrees for \annNativeAgreePct{}\%, and
completion raises the exact agreement on the location from
\annAnatomyWrittenPct{}\% to \annAnatomyCompletedPct{}\%. Cohen's $\kappa$ is
\annPresenceKappa{} for presence, \annCertaintyKappa{} for certainty and
\annChangeKappa{} for change, and the quadratically weighted $\kappa$ of the
stated severities is \annSeverityQWK{}.

A paired finding on which both parses agree in name, presence and location
enters the reference unchanged. An expert reviewer with eight years of
experience in medical imaging checked \supRevAgreedSampled{} sampled findings
of this kind and left \supRevAgreedKept{} unchanged. The other findings are
disputed or named by one parse only. Disputes of a recurring shape, such as
one location that contains the other, are settled by a rule after the
reviewer checked a set of samples of the group. A third
pass of Claude Opus adjudicates the remaining disputes and returns a verdict
with a confidence. The reviewer went through the adjudications in order of
rising confidence, which covers \supRevLow{} of the \supAdjLow{} verdicts
with low confidence. A last pass applies the conventions to all findings and repairs
the course of devices. \supProvDropped{} candidate findings are dropped, and
a reviewed name map finally merges synonymous names, which renames
\supGoldRenamed{} findings. The parse is faithful to the report and not to
the image, so a finding the report omits is unmentioned rather than absent.

\subsection{Silver labeler}
\label{sec:silver-labeler}

The reference parse covers the test split only. The sentences of the
training and validation reports are parsed by a silver labeler, a small
language model that is fine-tuned on the reference parse, runs locally and
returns the same structured parse. We first describe how it is trained and then how it
labels the training reports.

\noindent\textbf{Training.}
The input is one sentence after the fixed instruction \emph{You annotate one
sentence of a chest X-ray report. Return a JSON array with one object per
finding, condition or device the sentence asserts (present, absent or
uncertain); return [] if it asserts none (headers, history, indication,
recommendations). Each object: finding (the finding itself, concise), anatomy
(canonical anatomy concepts), presence, certainty, severity, change.} The
target is the reference parse of the sentence as a compact JSON array, without
in-context examples. Each model is fine-tuned with
QLoRA~\cite{dettmers2023qlora}. The name is free text, so the label space
stays open, while anatomy and attributes are constrained to their vocabularies
by schema-guided decoding. Sentences with an empty parse, such as headers or
the history, are kept, so that the labeler learns when to return nothing. The
labeler thus learns from sentences of test reports. It maps one sentence to
its parse and sees neither an image nor a study, so no test image and no
pairing of a test image with its report reaches it.

\noindent\textbf{Model selection.}
We divide the reference sentences by patient into five folds, with a sentence
shared by several patients in exactly one of them. One fold of \silvNeval{}
sentences is held out from every model, and all scores of this subsection are
measured on it. A predicted finding is strictly correct if it matches a
reference finding in name, presence and exact anatomy. We compare four base
models, each trained on three of the other folds. Qwen3.5~\cite{qwen35}
reaches a strict F1 of \silvSmallStrict{} with 0.8B, \silvTwoStrict{} with 2B
and \silvBestStrict{} with 9B parameters, and
MedGemma-4B~\cite{sellergren2025medgemma} reaches \silvMedStrict{}.

\noindent\textbf{Ensemble.}
The labeler consists of four \silvBest{} members, each trained on three of the
four remaining folds, with a strict F1 between \silvEnsMemMin{} and
\silvEnsMemMax{}. For each sentence, we keep the member parse that agrees most
with the other three. This minimum-Bayes-risk selection reaches a strict F1 of
\silvMbr{} and always returns a parse that one model produced.

\noindent\textbf{Audit of the reference.}
The members also serve to check the reference parse. Every reference sentence
has four predictions of members that did not train on it, and a sentence is
flagged where they agree against the reference or where a text check fires,
such as a negation before a finding that is marked present. Claude Opus judges
every flagged sentence, and a proposed correction enters the reference only
after the reviewer checked it or a sample of its group. The audit changed
\audCorrSent{} sentences (\audCorrSentPct{}\% of the reference), and every
score against the reference parse uses the audited version.

\noindent\textbf{Labeling the training reports.}
The ensemble parses every sentence of the training and validation reports.
Rules derived from the audit are also applied to the labeler output. A
training sentence that occurs verbatim in the reference takes its reference
parse, and a reviewed name map merges spelling variants and synonyms into the
vocabulary $\mathcal{V}$. The report metrics of
\cref{tab:core-main,tab:core-multi-image} and the localisation scores of
\cref{tab:core-grounding} are computed from the original reports and the
radiologist boxes and use neither parse.

\subsection{Caption targets}
\label{subsec:caption-targets}

The caption target of a finding is a sentence that states this finding
alone. A report sentence often states several findings or compares with a
prior study, so it is rewritten per finding. A sentence that states a single
finding without change wording is copied verbatim. Every other sentence is
rewritten for each of its findings by Qwen3.5-9B~\cite{qwen35} with a QLoRA
adapter~\cite{dettmers2023qlora}, trained for \supRwEpochs{} epochs on
\supRwTrain{} rewrites by Claude Opus~\cite{anthropic2026claudeopus} that keep
only the words of the target finding and drop change wording. In a blind
comparison on \supRwVal{} held-out sentences of unseen patients,
\supRwGoodPct{}\% of the rewrites of the adapter are judged correct, against
\supRwBaseGoodPct{}\% for the base model with a prompt. A caption target has
at most \supCapMaxLen{} tokens.

On frontal images, \supCapFRewritePct{}\% of the caption targets are rewrites
by the model and \supCapFVerbatimPct{}\% are verbatim copies
(\cref{tab:target-stats}). The remaining \supCapFSourcePct{}\% are the whole
report sentence, which is the fallback where no rewrite exists.

\subsection{Region masks}
\label{sec:anatomy-scheme}

\begin{figure*}[t]
  \centering
  \includegraphics[width=0.94\linewidth]{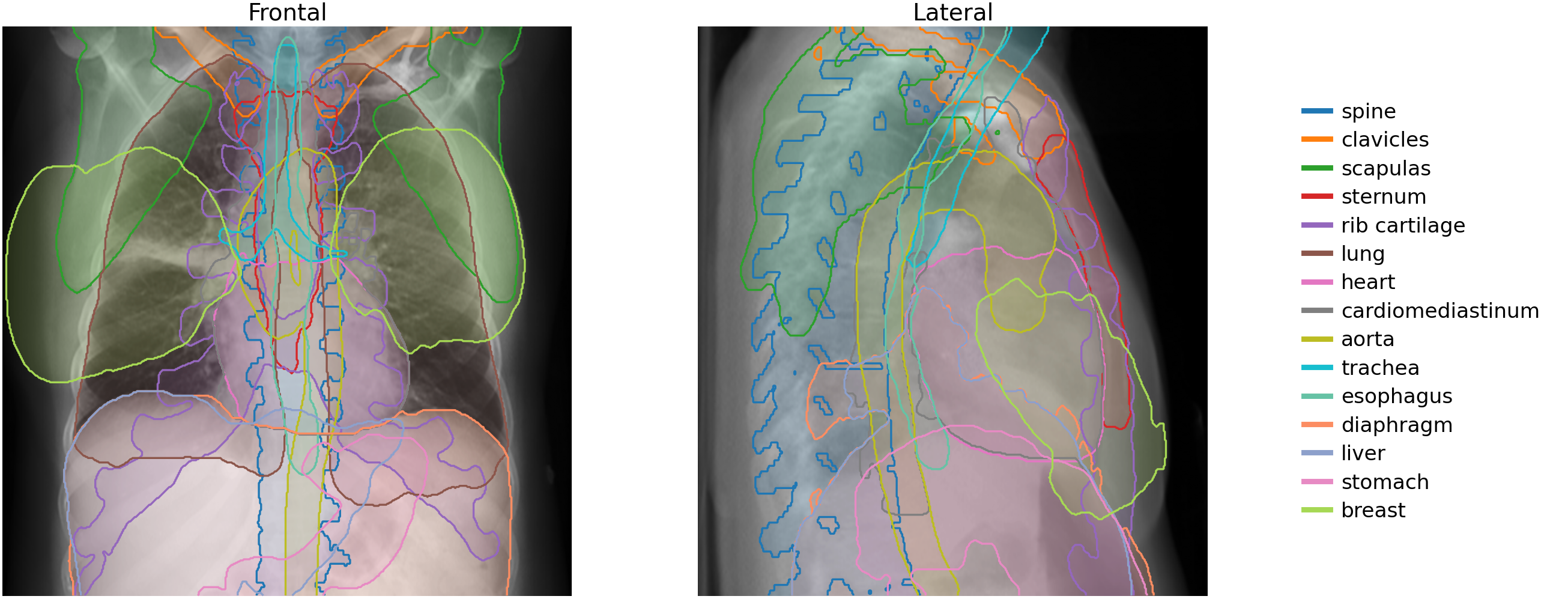}
  \caption{Coarse classes of the PAX-Ray++ label space on a frontal and a lateral case, each outlined in its
    own colour. The frontal case carries \paxrefclassesfrontal{} of the \paxclasses{} classes and the lateral
    case \paxrefclasseslateral{}.}
  \label{fig:paxray-overview}
\end{figure*}

\noindent\textbf{Label space.}
\methodname locates a finding among the \gndclasses{} regions of
$\mathcal{A}$ (\cref{tab:regions}). They are the \paxclasses{} classes of the
released PAX-Ray++ annotations (\cref{fig:paxray-overview}) and the
\tgtRegionsAdded{} regions we add below. The classes form a hierarchy and nest
rather than partition, as the lung contains the left lung, which contains the
left lung base. Of the \paxclasses{} classes, \paxfrontalonly{} are annotated
on frontal projections only, among them every sided class and every lung zone
and lobe, three on laterals only, namely the anterior, middle and posterior
mediastinum, and \paxbothviews{} routinely on both. The remaining
\paxrareclasses{} are rare on either view.

\begin{table*}[t]
  \centering
  \scriptsize
  \setlength{\tabcolsep}{2.5pt}
  \renewcommand{\arraystretch}{0.86}
  \begin{tabular}{@{}lcc@{\hspace{1.6em}}lcc@{\hspace{1.6em}}lcc@{\hspace{1.6em}}lcc@{}}
  \toprule
  Region & F & L & Region & F & L & Region & F & L & Region & F & L \\
  \midrule
spine & $\bullet$ & $\bullet$ & posterior 9th rib right & $\bullet$ &  & anterior 6th rib &  & $\bullet$ & mid zone lung &  & $\bullet$ \\
cervical spine & $\bullet$ & $\bullet$ & anterior 9th rib left & $\bullet$ &  & posterior 5th rib &  & $\bullet$ & upper zone lung &  & $\bullet$ \\
thoracic spine & $\bullet$ & $\bullet$ & posterior 9th rib left & $\bullet$ &  & anterior 5th rib &  & $\bullet$ & apical zone lung &  & $\bullet$ \\
lumbar spine & $\bullet$ & $\bullet$ & anterior 8th rib right & $\bullet$ &  & posterior 4th rib &  & $\bullet$ & right upper zone lung & $\bullet$ &  \\
vertebrae c1 & $\bullet$ & $\bullet$ & posterior 8th rib right & $\bullet$ &  & anterior 4th rib &  & $\bullet$ & right mid zone lung & $\bullet$ &  \\
vertebrae c2 & $\bullet$ & $\bullet$ & anterior 8th rib left & $\bullet$ &  & posterior 3rd rib &  & $\bullet$ & right lung base & $\bullet$ &  \\
vertebrae c3 & $\bullet$ & $\bullet$ & posterior 8th rib left & $\bullet$ &  & anterior 3rd rib &  & $\bullet$ & right apical zone lung & $\bullet$ &  \\
vertebrae c4 & $\bullet$ & $\bullet$ & anterior 7th rib right & $\bullet$ &  & posterior 2nd rib &  & $\bullet$ & left upper zone lung & $\bullet$ &  \\
vertebrae c5 & $\bullet$ & $\bullet$ & posterior 7th rib right & $\bullet$ &  & anterior 2nd rib &  & $\bullet$ & left mid zone lung & $\bullet$ &  \\
vertebrae c6 & $\bullet$ & $\bullet$ & anterior 7th rib left & $\bullet$ &  & posterior 1st rib &  & $\bullet$ & left lung base & $\bullet$ &  \\
vertebrae c7 & $\bullet$ & $\bullet$ & posterior 7th rib left & $\bullet$ &  & anterior 1st rib &  & $\bullet$ & left apical zone lung & $\bullet$ &  \\
vertebrae t1 & $\bullet$ & $\bullet$ & anterior 6th rib right & $\bullet$ &  & diaphragm &  & $\bullet$ & lung lower lobe left & $\bullet$ & $\bullet$ \\
vertebrae t2 & $\bullet$ & $\bullet$ & posterior 6th rib right & $\bullet$ &  & left hemidiaphragm & $\bullet$ &  & lung upper lobe left & $\bullet$ & $\bullet$ \\
vertebrae t3 & $\bullet$ & $\bullet$ & anterior 6th rib left & $\bullet$ &  & right hemidiaphragm & $\bullet$ &  & lung lower lobe right & $\bullet$ & $\bullet$ \\
vertebrae t4 & $\bullet$ & $\bullet$ & posterior 6th rib left & $\bullet$ &  & stomach & $\bullet$ & $\bullet$ & lung middle lobe right & $\bullet$ & $\bullet$ \\
vertebrae t5 & $\bullet$ & $\bullet$ & anterior 5th rib right & $\bullet$ &  & small bowel & $\bullet$ & $\bullet$ & lung upper lobe right & $\bullet$ & $\bullet$ \\
vertebrae t6 & $\bullet$ & $\bullet$ & posterior 5th rib right & $\bullet$ &  & duodenum & $\bullet$ & $\bullet$ & less obstructed lung &  & $\bullet$ \\
vertebrae t7 & $\bullet$ & $\bullet$ & anterior 5th rib left & $\bullet$ &  & liver & $\bullet$ & $\bullet$ & less obstructed right lung & $\bullet$ &  \\
vertebrae t8 & $\bullet$ & $\bullet$ & posterior 5th rib left & $\bullet$ &  & pancreas & $\bullet$ & $\bullet$ & less obstructed left lung & $\bullet$ &  \\
vertebrae t9 & $\bullet$ & $\bullet$ & anterior 4th rib right & $\bullet$ &  & kidney left & $\bullet$ & $\bullet$ & less obstructed lung base &  & $\bullet$ \\
vertebrae t10 & $\bullet$ & $\bullet$ & posterior 4th rib right & $\bullet$ &  & kidney right & $\bullet$ & $\bullet$ & less obstructed mid zone lung &  & $\bullet$ \\
vertebrae t11 & $\bullet$ & $\bullet$ & anterior 4th rib left & $\bullet$ &  & cardiomediastinum & $\bullet$ & $\bullet$ & less obstructed upper zone lung &  & $\bullet$ \\
vertebrae t12 & $\bullet$ & $\bullet$ & posterior 4th rib left & $\bullet$ &  & upper mediastinum & $\bullet$ & $\bullet$ & less obstructed apical zone lung &  & $\bullet$ \\
vertebrae l1 & $\bullet$ & $\bullet$ & anterior 3rd rib right & $\bullet$ &  & lower mediastinum & $\bullet$ & $\bullet$ & less obstructed right upper zone lung & $\bullet$ &  \\
vertebrae l2 & $\bullet$ & $\bullet$ & posterior 3rd rib right & $\bullet$ &  & anterior mediastinum &  & $\bullet$ & less obstructed right mid zone lung & $\bullet$ &  \\
vertebrae l3 & $\bullet$ & $\bullet$ & anterior 3rd rib left & $\bullet$ &  & middle mediastinum &  & $\bullet$ & less obstructed right lung base & $\bullet$ &  \\
vertebrae l4 & $\bullet$ & $\bullet$ & posterior 3rd rib left & $\bullet$ &  & posterior mediastinum &  & $\bullet$ & less obstructed right apical zone lung & $\bullet$ &  \\
vertebrae l5 & $\bullet$ & $\bullet$ & anterior 2nd rib right & $\bullet$ &  & heart & $\bullet$ & $\bullet$ & less obstructed left upper zone lung & $\bullet$ &  \\
rib cartilage & $\bullet$ & $\bullet$ & posterior 2nd rib right & $\bullet$ &  & heart atrium left & $\bullet$ & $\bullet$ & less obstructed left mid zone lung & $\bullet$ &  \\
sternum & $\bullet$ & $\bullet$ & anterior 2nd rib left & $\bullet$ &  & heart atrium right & $\bullet$ & $\bullet$ & less obstructed left lung base & $\bullet$ &  \\
clavicles &  & $\bullet$ & posterior 2nd rib left & $\bullet$ &  & heart myocardium & $\bullet$ & $\bullet$ & less obstructed left apical zone lung & $\bullet$ &  \\
clavicle left & $\bullet$ &  & anterior 1st rib right & $\bullet$ &  & heart ventricle left & $\bullet$ & $\bullet$ & trachea & $\bullet$ & $\bullet$ \\
clavicle right & $\bullet$ &  & posterior 1st rib right & $\bullet$ &  & heart ventricle right & $\bullet$ & $\bullet$ & tracheal bifurcation & $\bullet$ & $\bullet$ \\
scapulas &  & $\bullet$ & anterior 1st rib left & $\bullet$ &  & aorta & $\bullet$ & $\bullet$ & breast &  & $\bullet$ \\
scapula left & $\bullet$ &  & posterior 1st rib left & $\bullet$ &  & ascending aorta & $\bullet$ & $\bullet$ & breast left & $\bullet$ &  \\
scapula right & $\bullet$ &  & 12th rib & $\bullet$ & $\bullet$ & descending aorta & $\bullet$ & $\bullet$ & breast right & $\bullet$ &  \\
posterior 12th rib right & $\bullet$ &  & posterior 11th rib &  & $\bullet$ & aortic arch & $\bullet$ & $\bullet$ & head$^{\dagger}$ &  & $\bullet$ \\
posterior 12th rib left & $\bullet$ &  & anterior 11th rib & $\bullet$ & $\bullet$ & pulmonary artery &  & $\bullet$ & left humerus$^{\dagger}$ & $\bullet$ &  \\
anterior 11th rib right & $\bullet$ &  & posterior 10th rib &  & $\bullet$ & pulmonary trunc & $\bullet$ & $\bullet$ & right humerus$^{\dagger}$ & $\bullet$ &  \\
posterior 11th rib right & $\bullet$ &  & anterior 10th rib &  & $\bullet$ & left pulmonary artery & $\bullet$ &  & humerus$^{\dagger}$ &  & $\bullet$ \\
anterior 11th rib left & $\bullet$ &  & posterior 9th rib &  & $\bullet$ & right pulmonary artery & $\bullet$ &  & below right hemidiaphragm$^{\dagger}$ & $\bullet$ & $\bullet$ \\
posterior 11th rib left & $\bullet$ &  & anterior 9th rib &  & $\bullet$ & inferior vena cava & $\bullet$ & $\bullet$ & below left hemidiaphragm$^{\dagger}$ & $\bullet$ & $\bullet$ \\
anterior 10th rib right & $\bullet$ &  & posterior 8th rib &  & $\bullet$ & esophagus & $\bullet$ & $\bullet$ & upper right quadrant$^{\dagger}$ & $\bullet$ & $\bullet$ \\
posterior 10th rib right & $\bullet$ &  & anterior 8th rib &  & $\bullet$ & lung &  & $\bullet$ & upper left quadrant$^{\dagger}$ & $\bullet$ & $\bullet$ \\
anterior 10th rib left & $\bullet$ &  & posterior 7th rib &  & $\bullet$ & right lung & $\bullet$ &  & lower right quadrant$^{\dagger}$ & $\bullet$ & $\bullet$ \\
posterior 10th rib left & $\bullet$ &  & anterior 7th rib &  & $\bullet$ & left lung & $\bullet$ &  & lower left quadrant$^{\dagger}$ & $\bullet$ & $\bullet$ \\
anterior 9th rib right & $\bullet$ &  & posterior 6th rib &  & $\bullet$ & lung base &  & $\bullet$ &  & &  \\
  \bottomrule
  \end{tabular}
  \caption{The 187 regions of $\mathcal{A}$, which are the 177 classes of PAX-Ray++ and the 10 regions that we add, marked by a dagger. A bullet marks a region that lies in the location of at least one training finding on frontal (F) or lateral (L) images, which holds for every region in at least one view.}
  \label{tab:regions}
\end{table*}

\noindent\textbf{Added regions.}\label{subsec:anatomy-extension}
PAX-Ray++ is projected from chest CT and lacks structures that a radiograph
shows and reports name. A first step appends \anatnewclasses{} classes for the
head and the humeri, which gives \gndprevclasses{} classes. A second step
appends the \gndadded{} regions in which device bodies lie, the two regions
below the diaphragm and the four abdominal quadrants, which gives
\gndclasses{} classes.

\noindent\textbf{Segmentation.}
The frontal masks come from a
Mask2Former~\cite{cheng2022mask2former} with \gndclasses{} classes, trained on
frontal MIMIC-CXR radiographs with pseudo-labels, which are the predictions of
the \gndprevclasses{}-class model for the inherited classes and a
pseudo-label set for the \gndadded{} added regions. Its masks are thus
consistent with their teachers rather than verified by radiologists.
Lateral images are segmented by a separate Mask2Former with the
\paxclasses{} PAX-Ray++ classes. A lateral projection superimposes both
sides, so a sided location maps to the joint class where one exists and to
the union of both sides otherwise.

\noindent\textbf{Mask target.}
The mask target of a finding is the union of the regions of its location,
average-pooled to the patch grid, so that each patch holds the fraction of its
area that the union covers.

\subsection{Device masks}
\label{subsec:device-masks}

Devices lie in no specific anatomical region, so their mask targets come from a
seperate model. A Mask2Former~\cite{cheng2022mask2former} with \supDevQueries{}
queries segments \supDevClasses{} device groups, among them airway and enteric
tubes, vascular catheters, chest drains, cardiac rhythm devices, sternotomy
wires, valves and external objects such as ECG leads. We train it for
\supDevSteps{} steps on \supDevImages{} radiographs with \supDevMasks{} masks,
of which \supDevMimicImages{} come from MIMIC-CXR, \supDevRadioImages{} from
Radiopaedia~\cite{radiopaedia} and \supDevOtherImages{} from smaller
collections. The detector for lateral images follows the same recipe with
\supDevLatAdded{} additional lateral radiographs.

The detector runs on every training image. Each of its masks takes the device
group that is most probable given the devices the report names, and a logistic
score over the probability of this group, the probability of an external
object, the detection score and the fraction of the mask inside the stated
location decides whether the mask belongs to a finding. The mask target of a
device finding is the union of its accepted masks, pooled to the patch grid as
for the regions. An accepted mask exists for \tgtFDevMaskPct{}\% of the
frontal and \tgtLDevMaskPct{}\% of the lateral device findings, and the others
receive no mask loss.

\subsection{Assembled targets}
\label{subsec:target-stats}
\begin{table}[b]
  \centering
  \small
  \setlength{\tabcolsep}{5pt}
  \begin{tabular}{@{}lrr@{}}
  \toprule
   & Frontal & Lateral \\
  \midrule
Images & 172{,}820 & 97{,}996 \\
\quad with at least one finding & 134{,}802 & 70{,}528 \\
\quad with a repeated name & 30{,}726 & 12{,}902 \\
Studies & 151{,}990 & 89{,}155 \\
Patients & 59{,}747 & 48{,}526 \\
\midrule
Findings & 556{,}125 & 258{,}007 \\
\quad per image (mean) & 3.22 & 2.63 \\
\quad uncertain (\%) & 18.1 & 18.4 \\
\quad with a severity (\%) & 24.2 & 27.3 \\
\quad without a location & 2{,}370 & 0 \\
\quad regions per finding (median) & 4 & 3 \\
Distinct regions & 146 & 109 \\
\midrule
Device findings & 103{,}919 & 31{,}469 \\
\quad with an accepted mask (\%) & 75.9 & 64.3 \\
\midrule
Caption targets (\%) &  &  \\
\quad rewritten & \supCapFRewritePct & \supCapLRewritePct \\
\quad copied verbatim & \supCapFVerbatimPct & \supCapLVerbatimPct \\
\quad whole report sentence & \supCapFSourcePct & \supCapLSourcePct \\
\midrule
Distinct names & 3{,}156 & 2{,}316 \\
\quad $\geq$\,1{,}000 findings & 74 (86.7) & 42 (76.9) \\
\quad 100 to 999 findings & 149 (9.5) & 131 (17.5) \\
\quad 10 to 99 findings & 480 (2.9) & 340 (4.1) \\
\quad 1 to 9 findings & 2{,}453 (1.0) & 1{,}803 (1.6) \\
  \bottomrule
  \end{tabular}
  \caption{Training targets of the frontal and the lateral predictor. Absent findings are removed. The last rows count the names by their number of findings, with the share of all findings in percent in brackets.}
  \label{tab:target-stats}
\end{table}

We assign the findings of the FINDINGS section of a report to every frontal
image of its study and, for the lateral predictor, to every lateral image, and
remove the absent ones (\cref{tab:target-stats}). A target is thus present or
uncertain, and the presence head separates these two values. An image whose
report states no finding keeps an empty set, which applies to
\tgtFEmptyPct{}\% of the frontal and \tgtLEmptyPct{}\% of the lateral images.
Of the frontal findings, \tgtFUncPct{}\% are uncertain and \tgtFSevPct{}\%
carry a severity. A name occurs more than once on \tgtFRepeatPct{}\% of the
frontal images.

\begin{figure}[t]
  \centering
  \includegraphics[width=\linewidth]{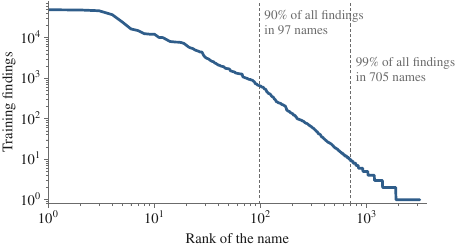}
  \caption{Number of frontal training findings per name, with the names ordered by frequency on logarithmic
    axes. The dashed lines mark the ranks up to which the names hold \tgtCoverHi{}\% and
    \tgtCoverTop{}\% of all findings.}
  \label{fig:names-rank}
\end{figure}

The names follow a long-tailed distribution (\cref{fig:names-rank}). On
frontal images, the \tgtFCoverHiNames{} most frequent names hold
\tgtCoverHi{}\% of all findings, and the ten most frequent names alone hold
\tgtFTopTenPct{}\% (\cref{tab:target-names}). The \tgtFCoverTopNames{} most
frequent names hold \tgtCoverTop{}\%, so the remaining \tgtFCoverTailNames{}
names share \tgtCoverRest{}\% of the findings. The vocabulary $\mathcal{V}$ of \mthCNames{} names exceeds
the \tgtFNames{} names of these targets, as it also holds the names that the
training reports state only as absent or outside the FINDINGS section.

\begin{table}[b]
  \centering
  \small
  \setlength{\tabcolsep}{5pt}
  \begin{tabular}{@{}rlrr@{}}
  \toprule
   & Name & Frontal & Lateral \\
  \midrule
1 & atelectasis & 49{,}544 & 22{,}078 \\
2 & opacity & 48{,}889 & 21{,}508 \\
3 & pleural effusion & 46{,}820 & 18{,}714 \\
4 & cardiomegaly & 37{,}407 & 18{,}793 \\
5 & low lung volumes & 24{,}377 & 11{,}941 \\
6 & enteric tube & 16{,}626 & 451 \\
7 & pulmonary edema & 15{,}125 & 4{,}761 \\
8 & central venous catheter & 12{,}723 & 1{,}822 \\
9 & endotracheal tube & 12{,}336 & 7 \\
10 & pulmonary vascular congestion & 12{,}294 & 5{,}475 \\
11 & sternotomy wires & 10{,}263 & 5{,}982 \\
12 & pneumonia & 10{,}148 & 4{,}717 \\
13 & vascular tortuosity & 8{,}990 & 6{,}326 \\
14 & hyperinflation & 8{,}135 & 5{,}927 \\
15 & degenerative change & 7{,}979 & 6{,}726 \\
16 & pacemaker & 7{,}854 & 4{,}541 \\
17 & surgical clips & 7{,}833 & 5{,}390 \\
18 & consolidation & 7{,}632 & 2{,}974 \\
19 & PICC line & 7{,}256 & 1{,}639 \\
20 & scarring & 6{,}560 & 4{,}903 \\
  \bottomrule
  \end{tabular}
  \caption{The 20 most frequent names of the frontal training targets with their number of findings on frontal and lateral images.}
  \label{tab:target-names}
\end{table}

\section{Implementation}
\label{sec:implementation}

\subsection{Finding-set predictor}
\label{subsec:impl-predictor}

\Cref{tab:impl-settings} lists the settings of the predictor.

\noindent\textbf{Architecture.}
The image encoder is RAD-DINO~\cite{perezgarcia2025radDINO}. The query
decoder has learned positional embeddings for queries and patches. The
queries are initialised from the FINCH~\cite{sarfraz2019finch} cluster
centroids of the text embeddings~\cite{zhang2025qwen3embedding} of the
training names. The heads are shared across layers and applied after every
layer, as in DETR~\cite{carion2020detr}. Name, presence, certainty and
severity end in a softmax, and the mask is
$m_{ip}=\sigma\big(\mathrm{MLP}(q_i)^{\top}x_p\big)$.

\noindent\textbf{Training.}
We fine-tune all blocks of the encoder with AdamW, with a learning rate that
decays layer-wise below the top block. Training has two stages, and the
second starts from the last checkpoint of the first without warm-up. During
its first \mthCHeadsFirst{} iterations, only the new heads learn and the
matching ignores location, since untrained heads would otherwise reassign
queries that already name their finding. Images are drawn uniformly with
replacement from the \mthCTrainImages{} frontal training images whose report
has a FINDINGS section. Each image is rotated by at most \supAugDeg{}
degrees, shifted by at most \supAugShiftPct\% of its size and scaled by at
most \supAugScalePct\%. We use the last checkpoint.

\noindent\textbf{Loss and candidates.}
In the cost of \cref{eq:cost}, $w_L=\mthCLocW$, and averaging over the
regions rather than summing keeps the location from outweighing the name. A
matched query receives cross-entropies on name and attributes, the location
term, a binary cross-entropy and a Dice loss on its mask, and the token
cross-entropy of the caption, which is computed after the last layer only.
Unmatched queries enter
the gate loss with weight \mthCEos{} as in DETR~\cite{carion2020detr}, and
the two mask losses have weight \mthCMaskW{} each. A finding from a sentence
on which the members of the labeler disagree enters every term with a weight
of \supSoftWeight{}. All terms are normalised by the number of target
findings in the batch. At test time, a query is a candidate if its gate
exceeds \mthRsMinGate{}. It carries its three best names and the regions
$\{a:\lambda_{ia}>0.5\}$, at least the most likely one.

\begin{table*}[t]
\footnotesize
\chexcompactrules
\setlength{\tabcolsep}{5pt}
\begin{minipage}[t]{0.49\linewidth}
\centering
\begin{tabular}[t]{@{}l>{\raggedright\arraybackslash}p{0.68\linewidth}@{}}
\toprule
\multicolumn{2}{@{}l}{\textbf{Finding-set predictor}} \\
\midrule
Image encoder & RAD-DINO~\cite{perezgarcia2025radDINO}, all \mthCUnfrozen{} blocks fine-tuned \\
Input & $\mthCImg\times\mthCImg$ pixels, $P=\mthCGrid^2$ patch tokens \\
Queries & $N=\mthCQueries$ \\
Query decoder & \mthCDecLayers{} layers, width \mthCDecDim{}, \mthCDecHeads{} heads, feed-forward \mthCDecFfn{} \\
Caption decoder & \mthCCapLayers{} layers, width \mthCCapDim{} \\
\midrule
Optimiser & AdamW, weight decay \mthCWd{}, clipping at \mthCClip{} \\
Batch size & \mthCBs{} \\
Learning rate & \\
\quad top encoder block & \mthCBbLr{}, layer-wise decay \mthCLlrd{} \\
\quad decoder and heads & \mthCLrName{} in the first stage, \mthCLrHeads{} in the second \\
First stage & \mthCItersName{} iterations, \mthCWarmup{} of them warm-up, cosine decay \\
Second stage & cosine cycles of \mthCItersA{} and \mthCItersB{} iterations \\
Total & \mthCItersTotal{} iterations \\
\midrule
\multicolumn{2}{@{}l}{\textbf{Fine-tuned writer}} \\
\midrule
Adapter & QLoRA~\cite{dettmers2023qlora} on Qwen3.5-2B~\cite{qwen35}, rank \mthCWrR{}, $\alpha=\mthCWrAlpha$ \\
Learning rate & \mthCWrLr{} \\
Training & \mthCWrSteps{} steps at an effective batch size of \mthCWrBatch{} \\
Example length & at most \mthCWrMaxlen{} tokens \\
\bottomrule
\end{tabular}
\end{minipage}
\hfill
\begin{minipage}[t]{0.49\linewidth}
\centering
\begin{tabular}[t]{@{}l>{\raggedright\arraybackslash}p{0.68\linewidth}@{}}
\toprule
\multicolumn{2}{@{}l}{\textbf{Resolver}} \\
\midrule
Token width & \mthRvHidden{} \\
Set transformer & \mthRvLayers{} pre-norm layers, \mthRvHeads{} heads, feed-forward \mthRvFfn{}, dropout \mthRvDropout{} \\
Name embedding & \mthRvNameEmb{} dimensions \\
Pairwise projection & \mthRvPairDim{} dimensions \\
Gated attention & width \mthRvBagDim{} \\
MLPs & hidden width \supRvTokMlp{} for the token, \supRvAgrMlp{} for the agreement, \supRvHeadMlp{} in the heads \\
\midrule
Optimiser & AdamW, weight decay \mthRvWd{}, clipping at \mthRvClip{} \\
Learning rate & \mthRvLr{}, one-cycle schedule \\
Training & \mthRvSteps{} steps of \mthRvBatch{} studies \\
\midrule
\multicolumn{2}{@{}l}{\textbf{End-to-end control}} \\
\midrule
Optimiser & AdamW \\
Learning rate & \\
\quad top encoder block & \supCtlLrEnc{}, layer-wise decay \supCtlLlrd{} \\
\quad projector & \supCtlLrProj{} \\
\quad LoRA adapter & \supCtlLrLora{}, rank \supCtlRank{} \\
Training & \supCtlIters{} steps at batch size \supCtlBatch{} \\
Schedule & \supCtlWarmup{} warm-up steps, cosine decay \\
\bottomrule
\end{tabular}
\end{minipage}
\caption{\textbf{Settings} of the trained components of \methodname and of the end-to-end control.}
\label{tab:impl-settings}
\end{table*}

\subsection{Resolver}
\label{subsec:impl-resolver}

A second predictor of the same architecture reads the lateral images. It is
trained like the first on the \mthRvLatImages{} lateral training images, with
\mthRvLatItersName{} iterations in the first stage and a single cycle of
\mthRvLatItersHeads{} iterations in the second. The resolver reads the
candidates of all images of a study, each from the predictor of its view.

\noindent\textbf{Model.}
Each candidate enters as one token (\cref{tab:resolver-inputs}), whose values
are standardised over the training candidates and to which a learned
embedding of the view is added. A set
transformer~\cite{lee2019settransformer} without positional encoding returns
a vector $z_i$ for each candidate, on which four heads operate
(\cref{tab:resolver-heads}). The name head pools the candidates that share
the best name $n$ and are not predicted absent with gated
attention~\cite{ilse2018attention} and returns $p_n$. The keep head and the
representative head return $k_i$ and $\rho_i$ for one candidate, and the
pairwise head returns the probability that two candidates state the same
finding.

\noindent\textbf{Targets.}
The targets of all heads follow from one assignment of the candidates to the
findings of the report. A candidate that is not predicted absent and whose
best name occurs in the report is assigned to the finding of this name whose
regions overlap its own the most, compared without their body side, so that
duplicates of a finding are positives. Each finding that is still uncovered
then takes the first unassigned candidate, in order of the gate, that lists
its name among its alternatives.

\noindent\textbf{Training.}
Every head is trained with a binary cross-entropy, and the loss is their
unweighted sum, together with the cross-entropy of a further head that may
swap the best name of a candidate for an alternative and is not used in
decoding. Each step draws \mthRvBatch{} of the \mthRvStudies{} training
studies, which hold \mthRvCands{} candidates. With probability
\mthRvSubsetP{}, a study with several images is reduced to a random subset of
its images, so that one resolver serves a single image as well as a full
study. The candidates come from the two predictors, which have seen these
images in training, whereas the thresholds below are fitted on validation
images that neither predictor has seen.

\begin{table}[b]
\footnotesize
\chexcompactrules
\centering
\setlength{\tabcolsep}{5pt}
\begin{tabular}{@{}lr>{\raggedright\hangindent=0.9em\hangafter=1\arraybackslash}p{0.65\linewidth}@{}}
\toprule
Part & Values & Content \\
\midrule
Candidate & \mthRvTokDim{} & query token of the predictor \\
 & & mean text embedding~\cite{zhang2025qwen3embedding} of its \mthRvNameTop{} most probable names, weighted by their probabilities \\
 & & certainty, severity and location probabilities \\
 & & area, peak and mean of the mask \\
\midrule
Per name & \mthRvNameFeats{} & gate, probability of the name and presence probabilities \\
 & & rank of the gate within the image \\
 & & number of higher-gated candidates of the image with this name \\
 & & training frequency of the name \\
\midrule
Agreement & \mthRvXview{} & highest gate of a candidate with best name $n_i$ in the other view \\
 & & the same in other images of the same view \\
 & & the same in the same image \\
 & & highest gate of a candidate of the other view that lists $n_i$ among its \mthRvK{} names \\
 & & number of candidates of the other view with best name $n_i$ and a gate of at least \mthRvXviewGate{} \\
 & & highest gate of a candidate of the other view that predicts $n_i$ as absent \\
 & & whether the input holds another view \\
 & & whether the input holds another image of the same view \\
\bottomrule
\end{tabular}
\caption{\textbf{Token of a candidate $i$ in the resolver.} The part \emph{Per name} is repeated for each of the \mthRvK{} best names of the candidate and lists the main ones of its values. \emph{Agreement} refers to the best name $n_i$, and its first five values range over the other candidates of the study that are not predicted absent.}
\label{tab:resolver-inputs}
\end{table}

\begin{table}[b]
\footnotesize
\chexcompactrules
\centering
\setlength{\tabcolsep}{4pt}
\begin{tabular}{@{}lll@{}}
\toprule
Head & Output & Target is 1 if \\
\midrule
Name & $p_n$ & the report states $n$ \\
Keep & $k_i$ & $i$ is assigned \\
Pairwise & pair probability & $i$ and $j$ share a finding \\
Representative & $\rho_i$ & $i$ fits its finding best \\
\bottomrule
\end{tabular}
\caption{\textbf{Heads of the resolver} with their binary targets, which follow from the assignment of the candidates to the findings of the report. The best fit is the assigned candidate that agrees most with the finding in location, presence and certainty.}
\label{tab:resolver-heads}
\end{table}

\noindent\textbf{Decoding.}
A name $n$ is kept if $p_n\geq t$. Its candidates with a pairwise
probability of at least $t_m$ are linked, and the linked group that holds the
candidate with the highest $k_i$ becomes one finding with name $n$ and score
$p_n$. The finding takes presence, certainty, severity and location from its
member with the highest $\rho_i$. A further group of the same name becomes a
finding only if its highest $k_i$ reaches a third threshold. The thresholds
are chosen on the validation split for the highest detection F1, which gives
$t=\mthRvT$ and $t_m=\mthRvMerge$ for a single frontal image as well as for
all images of a study. For the third threshold the fit selects a value that
no group reaches, as a further finding of a name does not raise the detection
F1 on validation. The resolver thus returns one finding for each kept name.

\subsection{Report writers}
\label{subsec:impl-writers}

\noindent\textbf{Template.}
The template writes one sentence for each present or uncertain finding. The
sentence joins the severity, the name and a location phrase that is built from
at most two regions of the location, such as \emph{at the left lung base} or
\emph{bilaterally}. Devices and cardiac findings carry no location phrase,
and a pleural finding takes its side as an adjective. A finding that is
uncertain or whose certainty is below definite is
hedged with \emph{likely}, \emph{possible} or \emph{cannot be excluded}
according to its certainty. The sentences are ordered by region, from the
devices over the lungs, the pleura and the heart to the bones and the abdomen,
and by score within a region. Three normal statements follow. The first
calls the lungs clear, or denies a focal consolidation where the set holds a
lung finding other than a consolidation or a pneumonia. The second denies a
pleural effusion and a pneumothorax unless the set names them, and the third
calls the cardiomediastinal silhouette normal unless the set holds a cardiac
finding. An empty set thus yields these three statements. We chose the
sentence patterns, the order and the normal statements over several rounds of
variants on the validation split of MIMIC-CXR, with the findings of an earlier
predictor as input, and apply the result unchanged to both test sets.

\noindent\textbf{Fine-tuned writer.}
The fine-tuned writer is a QLoRA~\cite{dettmers2023qlora} adapter on
Qwen3.5-2B~\cite{qwen35}, trained on \mthCWrStudies{} training studies with
the labeler parse of the report as input (\cref{tab:impl-settings}). It
writes the FINDINGS and the IMPRESSION section, of which the FINDINGS section
is the report. The confidence on each input line is the score $p_n$ of the
finding. In training, the parse carries no score, so each finding takes a
confidence drawn from the scores of correct predicted findings of the same
name. The writer decodes greedily. A sentence is then removed as a duplicate
of an earlier sentence of the report if the Jaccard index of their sets of
content words, which are the lowercased words without stop words, is at least
\mthCWrDedupJac{}.

\subsection{End-to-end control}
\label{subsec:impl-control}

The end-to-end control of \cref{tab:core-main} uses the image encoder and the
language model of \methodname without a finding set between them.
RAD-DINO~\cite{perezgarcia2025radDINO} encodes a frontal image at \mthCImg{}
pixels, a two-layer MLP projects its $\mthCGrid\times\mthCGrid$ patch tokens
into the embedding space of Qwen3.5-2B~\cite{qwen35}, and the language model
writes the FINDINGS section. We fine-tune the whole encoder, the projector
and a LoRA adapter on the language model for \supCtlIters{} steps
(\cref{tab:impl-settings}). The training images are
the \mthCTrainImages{} frontal images of the predictor, and the target is the
FINDINGS section of the original report. The validation loss is lowest
after \supCtlBestIters{} steps and stays above this value until the end, so
the schedule does not cut the control short. Decoding is greedy with at most
\supCtlMaxNew{} new tokens, the 99th percentile of the target lengths, and we
report the last checkpoint.

\subsection{Parameters and compute}
\label{subsec:impl-params}

\noindent\textbf{Parameters.}
The parameter counts of \cref{tab:core-main} hold the tensors that run at
inference. The predictor has \supParPredictor{} parameters without its
caption decoder, which serves training only, and the resolver has
\supParResolver{}, which gives \supParTemplate{} for \methodname with the
template. The fine-tuned writer adds the \supParWriterLm{} parameters of its
language model, \supParWriter{} in total. With several images, the lateral
predictor adds as many parameters as the frontal one, which gives
\supParTemplateMulti{} and \supParWriterMulti{}. The control has
\supParCtlEncoder{} parameters in the encoder, \supParCtlProjector{} in the
projector and the same language model, \supParControl{} in total. The name
embeddings are computed once in advance, so the text encoder does not run at
test time.

\noindent\textbf{Training data.}
The image counts of \cref{tab:core-main} are unique chest radiographs, for the
image encoder and for the model. The released RAD-DINO encoder was trained on
\supEncImages{} radiographs without reports~\cite{perezgarcia2025radDINO}.
The predictor adds the \mthCTrainImages{} frontal training images with their
reports, and the lateral predictor of \cref{tab:core-multi-image} a further
\mthRvLatImages{} lateral images, \supBothViewImages{} in total. The resolver
reads the candidates of these images and the writer reads text, so neither
adds an image. The models that build the training targets are not counted,
as none of them runs when a report is written. They are the two anatomy
segmenters, the device detector and the language models that parse and
rewrite the reports.

\noindent\textbf{Compute.}
We train and evaluate every model of this work on one \supGpu{} GPU. Training
the frontal predictor takes \supPredHours{} hours and the control
\supCtlHours{} hours.

\section{Compared models and prompts}
\label{sec:compared-models}

Every compared model runs through the inference interface that its authors
document. Where the authors provide an instruction for report generation, a
structured input or their own decoding, we use it unchanged. Our own
instruction is used only for the models whose authors provide none, and the
column \emph{Instruction} of \cref{tab:supp-models} states which of the two
each model receives. No model receives the
indication, a prior study or a prior report. RA-RRG~\cite{park2026rarrg},
which retrieves key phrases for the images of a study and lets a language
model write the report from them, has no released code or weights and is
therefore absent from the comparison. The same holds for
MAIRA-Seg~\cite{sharma2024mairaseg} and MRG-R1~\cite{wang2025mrgr1}. The
reinforcement-learning code of Gundersen et al.~\cite{gundersen2025radvlmrl}
is public, but its trained weights are not.

\begin{table*}[t]
\centering
\scriptsize
\chexcompactrules
\setlength{\tabcolsep}{2.7pt}
\begin{tabular}{@{}llllccl@{}}
\toprule
Model & Weights & Decoding & Tokens & Images & Cut & Instruction \\
\midrule
CheXagent-2~\cite{chen2024chexagent} & \texttt{StanfordAIMI/CheXagent-2-3b} & greedy & 512 & 2 & 265\,/\,165 & authors' task prompt \\
CheXOne~\cite{chexone} & \texttt{StanfordAIMI/CheXOne} & greedy, rep.\ penalty 1.05 & 2{,}048 & 8 & 0\,/\,0 & authors' task prompt \\
LLaVA-Rad~\cite{chaves2025llavarad} & \texttt{microsoft/llava-rad} & greedy & 512 & 1 & -- & authors' trained prompt \\
MAIRA-2~\cite{bannur2024maira2} & \texttt{microsoft/maira-2} & greedy & 512 & 2 & 420\,/\,169 & authors' structured input \\
RaDialog~\cite{pellegrini2025radialog} & \begin{tabular}[t]{@{}l@{}}\texttt{ChantalPellegrini/RaDialog-interactive-}\\\texttt{radiology-report-generation}\end{tabular} & greedy & 512 & 1 & -- & authors' instruction prompt after label input \\
RGRG~\cite{tanida2023rgrg} & authors' release & beam search, 4 beams & 300 & 1 & -- & none, image only \\
ChEX~\cite{muller2024chex} & authors' release & greedy & 128 & 1 & -- & authors' region prompts \\
\midrule
InternVL3~\cite{zhu2025internvl3} & \texttt{OpenGVLab/InternVL3-8B} & greedy & 768 & 4 & 1\,/\,1 & ours \\
Qwen3.5~\cite{qwen35} & \texttt{Qwen/Qwen3.5-9B} & greedy & 768 & 8 & 0\,/\,0 & ours \\
MedGemma~\cite{sellergren2025medgemma} & \texttt{google/medgemma-4b-it} & greedy & 768 & 8 & 0\,/\,0 & ours \\
\bottomrule
\end{tabular}
\caption{\textbf{Interface of the compared models.} \emph{Instruction} names the text that a model receives, which is that of its authors wherever they provide one and ours otherwise. \emph{Tokens} is the cap on new tokens of one generation call, which writes one region or prompt for RGRG and ChEX and the whole section otherwise. \emph{Images} is the number of images a model reads per study and \emph{Cut} the number of MIMIC-CXR\,/\,OpenI studies with an image that it does not read in \cref{tab:core-multi-image}.}
\label{tab:supp-models}
\end{table*}

\subsection{Shared harness}
\label{subsec:harness}

\noindent\textbf{Instruction.}
A model with an instruction of its authors receives exactly that instruction
(\cref{tab:supp-models}). Our instruction serves the three general VLMs,
which have no prescribed report prompt, and RaDialog, where it follows the
labels that the prompt of its authors starts with. It is a single user
message that holds the images followed by the text \emph{You are an expert radiologist generating a chest X-ray report.
Examine the provided chest radiograph(s). Write the FINDINGS section only: a
systematic description of the radiographic findings. Do not include an
impression, any section header, or preamble; output only the findings text.}

\noindent\textbf{Backends.}
CheXOne and the three general VLMs run offline on vLLM \supVerVllm{}, which
applies the chat template of each model to the message unaltered.
CheXagent-2, LLaVA-Rad, MAIRA-2 and RaDialog run on HuggingFace
\texttt{transformers}, some with compatibility patches because their
modelling code predates current releases. RGRG and ChEX run in the code of
their authors.

\noindent\textbf{Decoding.}
Decoding is greedy with a fixed seed unless the authors prescribe otherwise,
which applies to the repetition penalty of CheXOne and the beam search of
RGRG. For CheXOne, InternVL3 and MedGemma, vLLM runs with batch-invariant
kernels and without prefix caching, so the composition of a batch cannot
change an output. Qwen3.5 and the \texttt{transformers} models decode
greedily with a fixed seed, without this guarantee. The outputs of all
models pass through the same post-processing.

\noindent\textbf{Several images.}
In \cref{tab:core-multi-image}, a model that accepts several images receives
the images of a study in a fixed order, with PA before AP before lateral views
and by acquisition time within a view. A study with more images than the
model accepts is cut to that number (\cref{tab:supp-models}). MAIRA-2 fills
its own frontal and lateral slots, so a second frontal image never reaches it.
The
single-image models are absent from \cref{tab:core-multi-image}.

\noindent\textbf{Parameters and training images.}
The parameter counts of \cref{tab:core-main} hold the unique tensors of the
released weights that run at inference. The training radiographs are the
numbers the authors state, and a tilde marks a number that we derive from
their dataset tables.

\subsection{Per-model interface}
\label{subsec:per-model}

\noindent\textbf{CheXagent-2.}
CheXagent-2~\cite{chen2024chexagent} receives the task prompt of its
repository, \emph{Given the indication: ``\ldots'', write a structured
findings section for the CXR.}, with an empty indication, since the authors
ship no image-only variant. Its output is scored as produced, including the
anatomy tags in square brackets that the prompt asks for and, on OpenI, a
small share of outputs that are bounding boxes without prose.

\noindent\textbf{CheXOne.}
CheXOne~\cite{chexone} receives \emph{Write an example findings section for
the CXR.} Decoding follows the evaluation script of its authors, with an image
size of at most $512\times512$ pixels and the repetition penalty that their
inference engine reads from the generation configuration of the checkpoint.

\noindent\textbf{LLaVA-Rad.}
LLaVA-Rad~\cite{chaves2025llavarad} receives the instruction it was
fine-tuned on, \emph{Provide a description of the findings in the radiology
image.}, since other phrasings degrade its output.

\noindent\textbf{MAIRA-2.}
MAIRA-2~\cite{bannur2024maira2} receives separate fields for the frontal
image, the lateral image, the indication, the technique, the comparison and
the prior report, from which it composes its own instruction. For every study
the technique is the fixed string \emph{PA and lateral views of the chest.}
and the comparison is \emph{None.} 

\noindent\textbf{RaDialog.}
RaDialog~\cite{pellegrini2025radialog} uses an instruction prompt preceded by
the positive CheXpert labels of its own image classifier as it
was trained on. The classifier runs with the checkpoint of the authors at a
threshold of 0.5.

\noindent\textbf{RGRG.}
RGRG~\cite{tanida2023rgrg} receives the image alone, as its interface has no
text input. It detects 29 anatomical regions, selects the salient ones and
writes one sentence per region. Decoding is that of the authors, a beam search
with early stopping followed by their removal of duplicate sentences. Their
code and released checkpoint run on a current stack, since the library
versions they pin do not install on our GPU, and the checkpoint loads without
missing or unexpected tensors.

\noindent\textbf{ChEX.}
ChEX~\cite{muller2024chex} receives the 14 pathology and 15 anatomy prompts of
the report task of its authors and writes one sentence per prompt. Decoding
keeps at most ten sentences after their removal of synonymous sentences. The
code is theirs at a fixed commit with their stage-three checkpoint and their
preprocessing at 224 pixels, on a newer PyTorch release.

\noindent\textbf{General VLMs.}
InternVL3~\cite{zhu2025internvl3}, Qwen3.5~\cite{qwen35} and
MedGemma~\cite{sellergren2025medgemma} receive our instruction through their
own chat templates, as none documents a prompt for report generation.
Qwen3.5 runs with its reasoning mode switched off.

\subsection{Reference floors}
\label{subsec:floors}

The three floors of \cref{tab:core-main} see no image. The constant floor
writes one fixed normal FINDINGS section for every study. The other two draw
on \supFloorPool{} randomly sampled training studies of MIMIC-CXR, of which
\supFloorFindings{} have a FINDINGS section. The most common report is the
most frequent of these sections, and the random report is one of them drawn
anew for every study. Both test sets use the same pool.

\section{Evaluation details}
\label{sec:evaluation-details}

\subsection{Report metrics}
\label{subsec:report-metrics}

\noindent\textbf{Scored text.}
We score the report of a model against the FINDINGS section of the original
report. Before scoring, the same step cleans the output of every model. It
removes the text around the report, namely a reasoning channel, a leading
line that announces the report, an echoed section header and markdown markup,
and normalises white space as in the references. The report itself is left
as written, with no word reordered or dropped, and a refusal or an empty
output is scored as it stands.

\noindent\textbf{Metrics.}
GREEN~\cite{ostmeier2024green} is computed with the released judge model and
package of its authors (version \supVerGreen{}), and a judgment that cannot
be parsed scores zero. RadGraph-F1~\cite{jain2021radgraph} uses the original
RadGraph checkpoint (package version \supVerRadgraph{}) with the partial
reward of~\cite{delbrouck2022semanticrewards}. The CheXbert
labeler~\cite{smit2020chexbert} assigns 14 labels to each report, and a label
counts as positive only where the labeler states it as present. We
reimplement RadCliQ-v1~\cite{yu2023radcliq} from its four published
components, BLEU-2~\cite{papineni2002bleu},
BERTScore~\cite{zhang2020bertscore}, the cosine similarity of CheXbert
embeddings and RadGraph-F1.

\noindent\textbf{Statistics.}
The MIMIC-CXR test set holds \setupMimicStudies{} studies of
\supMimicPatients{} patients, so we resample patients and keep all studies of
a drawn patient together. OpenI has one study per patient. Each interval is
the 95\% percentile interval of \supBootN{} resamples. For a difference
between two rows, both are scored on the same resamples, and the interval is
that of their difference.

\subsection{Finding sets}
\label{subsec:finding-set-scoring}

\noindent\textbf{Reference.}
Finding sets are scored against the present and uncertain findings that the
reference parse holds for the FINDINGS section of a test report. On the
MIMIC-CXR test set these are \supRefFindings{} findings with \supRefNames{}
names in \supRefStudies{} studies, \supRefEmpty{} of which have none. The
parse follows the report and not the image, so
a predicted finding that the image shows and the report omits counts as a
false positive for every row.

\noindent\textbf{Matching.}
A predicted finding can only match a reference finding of the same name. If
a name occurs more than once in either set, the Hungarian
algorithm~\cite{kuhn1955hungarian} pairs its findings by the Jaccard index of
their locations. A matched pair is a true positive, an unmatched prediction
a false positive and an unmatched reference finding a false negative.

\noindent\textbf{Metrics.}
miF1 is the F1 score over the findings of all studies. maF1 is the mean of
the per-name F1 scores over the \supMacroNames{} names with at least
\supKMacro{} reference findings. Strict counts a matched pair as a true
positive only if its presence and its location set equal the reference
exactly. R$_{\mathrm{rare}}$ is the recall over the reference
findings of the \supTailNames{} names with fewer than \supTailN{} mentions
in the training reports, which hold \supTailPct{}\% of the reference
findings. Intervals
follow the bootstrap over patients of the report metrics.

\subsection{Predictor ablation}
\label{subsec:rows-predictor}

All rows of \cref{tab:core-det-ablation} are scored on the frontal image of
\cref{tab:core-main}. Each learned row keeps a finding whose score exceeds a
threshold fitted on the validation split, and a second finding of the same
name only above a higher threshold. The row + caption takes the thresholds of
the row above it.

\noindent\textbf{Rows that are not trained.}
The image-blind prior returns the $k$ most frequent training names for every
image, each with its most frequent location and attributes. The retrieval
row embeds the image with the frozen RAD-DINO
encoder~\cite{perezgarcia2025radDINO}, retrieves its nearest training images
of other patients by cosine similarity and returns every name that enough of
them hold, with the most frequent location and attributes among these
neighbours. The number of names, the number of neighbours and the required
votes are fitted on the validation split.

\noindent\textbf{End-to-end control, parsed.}
The report of the control is turned into a finding set sentence by sentence
with the labeler that parses the training reports, so this row also contains
the errors of the labeler.

\noindent\textbf{Multi-label classifier.}
The classifier shares the encoder, the training data and the schedule of the
predictor but has no queries and no matching. An MLP on the pooled image
tokens scores every name of $\mathcal{V}$, and a name above the threshold is
returned once with the most frequent location and attributes of that name.
The next row replaces these by the predictions of location and attribute
heads per name. Like every row, the classifier is scored at its last
checkpoint. Its detection F1 on the validation split stays between
\supClsFlatLo{} and \supClsFlatHi{} from 8k iterations on, so the schedule
does not cut it short.

\noindent\textbf{Set predictor.}
The first row trains the name loss alone, and each following row adds the loss
of the heads it names. Strict is undefined for the rows that predict no
location. Every row at 30k iterations, the classifier included, trains for
20k iterations and continues for 10k at half the learning rate with the losses
of its row. The last two rows continue the row + mask to 40k and 60k
iterations.

\subsection{Fusion methods}
\label{subsec:rows-fusion}

Every row of \cref{tab:core-fusion} starts from the candidates of the same
frontal and lateral predictor. The thresholds of every row are fitted per
column on the validation split, for the rules over a grid from \supGridLo{} to
\supGridHi{} in steps of \supGridStep{}. The learned rows train on the
training split with the labeler parse as target.

\noindent\textbf{Without fusion.}
Single image thresholds the findings of the frontal image. Lateral if
available takes the findings of the lateral predictor alone where a study has
a lateral image.

\noindent\textbf{Rules.}
Concatenation keeps the findings of all images. Union collapses findings that
agree in name, presence and location. Union per name keeps the
highest-scoring finding of each name. Voting keeps a name that a fitted share
of the images predicts. WBF~\cite{solovyev2021wbf} averages the scores of a
name over the images, scales the mean by the share of images that predict it
and keeps the regions with a weighted majority.
ProbEn~\cite{chen2022proben} sums the log-odds of a name over the images and
subtracts a fitted prior.

\noindent\textbf{Learned.}
Stacking~\cite{wolpert1992stacked} classifies each name of a study with
gradient-boosted trees on features of its candidates. Learned
NMS~\cite{hosang2017nms,hu2018relation} is the network of the resolver trained
to keep one candidate per finding, without merging. Attention
MIL~\cite{ilse2018attention} pools the candidates of a name with gated
attention. Linking~\cite{varlamova2024mammnet} classifies pairs of findings
as the same finding and returns the top finding of each connected component.
The row Ours is the resolver.

\subsection{Localised findings}
\label{subsec:rows-grounding}

\noindent\textbf{Unit.}
\Cref{tab:core-grounding} is scored per pair of an image and a category with
at least one radiologist box. The radiologist boxes of a pair form one
region, and so do the boxes that a model gives for this category. MS-CXR has
8 categories, and the VinDr-CXR test set has 22 classes and \supVindrBoxes{}
boxes.

\noindent\textbf{Categories of a sentence.}
A box of a model belongs to a sentence, so the categories of each sentence
are needed. On MS-CXR, whose categories are CheXbert labels, the CheXbert
labeler~\cite{smit2020chexbert} assigns them, and only a positive label
counts. The classes of VinDr-CXR go beyond these labels, so Claude
Opus~\cite{anthropic2026claudeopus} assigns them with one fixed prompt, and a
class counts where the sentence asserts it or calls it probable.

\noindent\textbf{Metrics.}
Acc averages over the IoU thresholds from 0 to \supGndMax{} in steps of
\supGndStep{}, and a pair that the report does not name counts as a miss at
every threshold. The precision of the F1 on VinDr-CXR is the share of named
classes that have a radiologist box, over all \setupVindrImages{} test images.
Intervals resample the \setupMscxrTestPatients{} patients of MS-CXR and the
images of VinDr-CXR.

\noindent\textbf{Rows.}
The oracle averages the radiologist boxes of a category over other images,
which are the MS-CXR images outside the test split and, on VinDr-CXR, all
other test images, and keeps the area above half of the maximum. RGRG gives
each sentence the box of its region. MAIRA-2 runs the grounded reporting mode
of its authors, and ChEX returns a box with each sentence of its report mode.
For CheXagent-2 and CheXOne, every sentence of the report of
\cref{tab:core-main} is sent to the phrase grounding prompt of the authors.
Both datasets are scored one image at a time, and the \methodname row is the
predictor on this image with a resolver that is trained on the candidates of
single images and the template as writer.
The region of a sentence is the union of the masks of the findings it
states. The mask of a finding is the mask $m_i$ of the candidate of this
image that carries its name, presence and location, upsampled to the image
and cut at the fixed value of \supGndMaskThr{}. The boxes of a region
enclose its connected components that cover at least
\supGndMinFracPct{}\% of the image. The row is thus scored like every other
row, with boxes on both sides, and runs no model besides the predictor and
the resolver on a test image. The mask is trained on the anatomical regions
of the location of a finding.

\subsection{Writer ablation}
\label{subsec:rows-writer}

\noindent\textbf{Input.}
The columns Predicted findings of \cref{tab:core-verbaliser} receive the
finding set of the predictor and the resolver on the frontal image of
\cref{tab:core-main}. The columns Reference parse receive the present and
uncertain findings that the reference parse holds for the whole report. The
reference parse has no confidence, so each of its findings takes a confidence
drawn from the scores of correct predicted findings of the same name, or of
all names where a name has fewer than \supWrMinPool{} such scores.

\noindent\textbf{Template rows.}
Template writes one sentence per finding in a fixed pattern. The row +
negatives adds a fixed negative sentence for each of six frequently negated
findings that the set does not hold. Tuned is the template of \methodname. The
indented rows keep this template and remove the location, the certainty and
severity, or all of them from its input.

\noindent\textbf{Writer rows.}
All writers share the language model, the adapter and the schedule of the
fine-tuned writer (\cref{tab:impl-settings}). Ref.\ input is trained with the parsed report as
its only input and writes the FINDINGS and the IMPRESSION section, of which
the FINDINGS section is scored. The rows below it train on \supWrPredRows{}
examples from studies without a prior study or change wording, write the
FINDINGS section alone and decode with a repetition penalty of 1.05. A part
of their training inputs are predicted finding sets, in which a finding
without support in the report keeps its line and is answered by a template
sentence. The row +
alts.\ adds to each finding the other names of its candidate with their
probabilities, and + captions adds the caption of the predictor. The reference
parse has neither, so these two rows receive the plain reference parse in the
columns Reference parse. The indented rows retrain + alts.\ without one
input field. The predicted finding sets for training come from an earlier
predictor on training images that it has not seen, and all writers run
unchanged on the final finding sets.

\subsection{Cited findings}
\label{subsec:cited-findings}

The fine-tuned writer ends a sentence with the numbers of the input lines
whose findings it states, and \cref{subsec:qualitative} reports how often a
stated finding appears in a cited line. For this measurement, the silver labeler
parses each of the \wfSentences{} sentences that the
writer of \cref{tab:core-main} writes for the \wfStudies{} MIMIC-CXR test
studies. A stated finding is one that this parse holds as present or
uncertain, and it is backed if a cited line carries its name or a name of the
same finding category. Of the \wfStated{} stated findings, \wfBackedPct{}\%
are backed and \wfUnlisted{} appear in no line of the input. Where a stated
finding and the cited line of its name both carry a body side, the sides
differ in \supWfSideConflict{} of \supWfSideComparable{} cases. The
measurement compares names and sides. It covers neither the change wording
of a sentence nor a detail beyond the
finding set, such as a measured distance (\cref{fig:supp-qual-devices}).

\newcommand{\lexOpeniFloorRouge}{0.351}
\newcommand{\lexOpeniBestRouge}{0.353}
\newcommand{\lexMimicTplBleu}{0.067}
\newcommand{\lexMimicWriterBleu}{0.090}
\newcommand{\lexMimicBestBleu}{0.101}

\section{Additional results}
\label{sec:additional-results}

\subsection{Further measurements}
\label{subsec:further-measurements}

\noindent\textbf{Studies with several images.}
\Cref{tab:core-fusion} scores every test study, and \supFsN{} of them hold
more than one image. Of these, \setupMimicLateral{} hold a lateral image,
\setupMimicSecondFrontal{} a second frontal image and \supFsBoth{} both.
\Cref{tab:fusion-subset} repeats the table on these studies alone. The
resolver then reaches \supFsOursAllF{} with all images against
\supFsSingleF{} with one frontal image, where \cref{tab:core-fusion} shows
\fusOursAllF{} against \fusSingleOneF{}. The difference is larger here, as a
study with one image cannot gain. The resolver remains the only method with a
significant gain in detection F1, while stacking and attention MIL gain in
macro F1 only.

\begin{table}[b]
\centering
\scriptsize
\chexcompactrules
\setlength{\tabcolsep}{2.5pt}
\begin{tabular}{@{}l cc cc cc@{}}
\toprule
 & \multicolumn{2}{c}{Single frontal} & \multicolumn{2}{c}{Frontal+lateral} & \multicolumn{2}{c}{All images} \\
\cmidrule(lr){2-3}\cmidrule(lr){4-5}\cmidrule(l){6-7}
Fusion & miF1$\uparrow$ & maF1$\uparrow$ & miF1$\uparrow$ & maF1$\uparrow$ & miF1$\uparrow$ & maF1$\uparrow$ \\
\midrule
Single image & \underline{.456} & \underline{.203} & .456 & .203 & .456 & .203 \\
Lateral if available & \underline{.456} & \underline{.203} & \chexvsdown{60} .431 & \chexvsdown{60} .169 & \chexvsdown{60} .431 & \chexvsdown{60} .169 \\
\midrule
Concatenation & \underline{.456} & \underline{.203} & \chexvsdown{60} .380 & \chexvsdown{60} .147 & \chexvsdown{60} .365 & \chexvsdown{60} .138 \\
Union & \underline{.456} & \underline{.203} & \chexvsdown{60} .414 & \chexvsdown{60} .174 & \chexvsdown{60} .421 & \chexvsdown{60} .163 \\
Union per name & .455 & \underline{.203} & .453 & .198 & .455 & .204 \\
Voting & .455 & \underline{.203} & .451 & \chexvsdown{28} .194 & .455 & .199 \\
WBF~\cite{solovyev2021wbf} & .455 & .201 & .451 & \chexvsdown{28} .192 & .455 & .197 \\
ProbEn~\cite{chen2022proben} & .455 & \underline{.203} & .453 & \chexvsdown{28} .192 & .456 & .199 \\
\midrule
Stacking~\cite{wolpert1992stacked} & .455 & .197 & .457 & \underline{.211} & \underline{.460} & \chexvsup{28} \underline{.219} \\
Learned NMS~\cite{hosang2017nms,hu2018relation} & \chexvsdown{28} .447 & \underline{.203} & \chexvsdown{60} .434 & .204 & \chexvsdown{60} .432 & .196 \\
Attention MIL~\cite{ilse2018attention} & \textbf{.457} & \textbf{.204} & \underline{.460} & .207 & \underline{.460} & \chexvsup{28} \textbf{.221} \\
Linking~\cite{varlamova2024mammnet} & \textbf{.457} & .201 & \chexvsdown{28} .443 & \chexvsdown{60} .169 & \chexvsdown{28} .446 & \chexvsdown{60} .175 \\
\midrule
\textbf{Ours} & .455 & \textbf{.204} & \chexvsup{28} \textbf{.463} & \textbf{.212} & \chexvsup{28} \textbf{.467} & \chexvsup{28} .217 \\
\bottomrule
\end{tabular}
\caption{\textbf{Fusion on the studies with several images.} The rows and inputs of \cref{tab:core-fusion}, scored on the 1{,}216 test studies with more than one image. Shading marks a significant \chexshadeword{chexbetter}{gain} or \chexshadeword{chexworse}{loss} against the row Single image on the same studies, darker from .02.}
\label{tab:fusion-subset}
\end{table}

\begin{table}[b]
\centering
\scriptsize
\chexcompactrules
\setlength{\tabcolsep}{4pt}
\begin{tabular}{@{}l cc cc@{}}
\toprule
 & \multicolumn{2}{c}{MS-CXR} & \multicolumn{2}{c}{VinDr-CXR} \\
\cmidrule(lr){2-3}\cmidrule(l){4-5}
 & Region & Boxes & Region & Boxes \\
\midrule
\textcolor{gray}{Oracle, prior box} & 18.0 & 16.0 & 8.6 & 4.2 \\
\midrule
RGRG~\cite{tanida2023rgrg} & 14.1 & 17.7 & 9.4 & 6.2 \\
MAIRA-2~\cite{bannur2024maira2} & 17.4 & 17.6 & 13.6 & 5.7 \\
CheXagent-2~\cite{chen2024chexagent} & 13.9 & 17.0 & 7.9 & 5.6 \\
CheXOne~\cite{chexone} & 17.2 & 18.0 & 8.8 & 5.6 \\
ChEX~\cite{muller2024chex} & 12.3 & 16.7 & 9.3 & 6.1 \\
\textbf{Ours} & 20.6 & 17.1 & 17.0 & 5.4 \\
\bottomrule
\end{tabular}
\caption{\textbf{Area of the localised regions.} Mean share of the image, in percent, that the region of a row of \cref{tab:core-grounding} covers for a stated category, next to the radiologist boxes of the same pairs.}
\label{tab:grounding-area}
\end{table}

\noindent\textbf{Empty finding sets on OpenI.}
On OpenI, the fine-tuned writer scores below the template on RG\textsubscript{ER} and
RadCliQ and above it on GREEN (\cref{tab:core-main}). The predicted set is
empty for \supOeEmpty{} of the \supOeN{} studies (\supOeEmptyPct{}\%), and
each of the two then returns its one fixed normal report. On these studies
the rows differ in the wording of this report alone, and the writer is lower
by \supOeEmptyRg{} on RG\textsubscript{ER} and higher by \supOeEmptyRc{} on RadCliQ and
by \supOeEmptyGreen{} on GREEN. On the remaining studies the differences keep
their sign and shrink to \supOeRestRg{}, \supOeRestRc{} and
\supOeRestGreen{}.

\noindent\textbf{Area of the localised regions.}
A larger region contains the centre of a radiologist box more easily, so
\cref{tab:grounding-area} gives the area behind the Hit column of
\cref{tab:core-grounding}. On MS-CXR the regions of \methodname cover
\supAreaMsOurs{}\% of the image on average where the radiologist boxes cover
\supAreaMsGold{}\%, and on VinDr-CXR \supAreaVdOurs{}\% against
\supAreaVdGold{}\%. They are the largest of all rows on both datasets, as the
mask of a finding is trained on the anatomical regions of its location,
which are coarser than a box around the finding.
Acc requires an overlap with the box that grows with its threshold, so a
region far larger than its box lowers it, and \methodname also reaches the
highest Acc on both datasets (\cref{tab:core-grounding}).

\noindent\textbf{Resolver threshold.}
The threshold $t$ of the resolver is fitted for the detection F1 on the
validation split. On the frontal image of the
MIMIC-CXR test studies, it gives a precision of \supThrP{} and a recall of
\supThrR{} with \supThrN{} findings per study. From $t=\supThrTLow$ to
$t=\supThrTHigh$ the precision rises from \supThrPLow{} to \supThrPHigh{} and
the recall falls from \supThrRLow{} to \supThrRHigh{}, while the detection F1
stays between \supThrFMin{} and \supThrFMax{}. The threshold thus trades
recall for precision at little cost in detection F1.

\noindent\textbf{Repeated names.}
The reference parse holds a name more than once in \supRepPct{}\% of the
MIMIC-CXR test studies and at distinct locations in \supRepDistinctPct{}\%.
These repetitions make up \supSurplusPct{}\% of the reference findings. The
resolver groups the candidates of a name and can return a second finding of
this name from a further group. With the thresholds fitted on the validation
split no further group passes, so the predicted sets hold one finding per
name, which bounds the detection recall at \supRecallCapPct{}\%.

\noindent\textbf{Change wording.}
The original reports compare with prior studies, and the fine-tuned writer
adopts their wording although it sees a single study. On the MIMIC-CXR test
set, \supChangeSentPct{}\% of its sentences and \supChangeStudiesPct{}\% of
its reports contain change wording. This wording comes from the training
reports of the writer and not from the finding set, so it concerns the writer
alone. The writers below Ref.\ input in \cref{tab:core-verbaliser} are
trained on studies without a prior study or change wording and replace it without a change to the predictor
or the resolver.

\subsection{Qualitative examples}
\label{subsec:qualitative-supp}

\begin{figure*}[t]
  \centering
  \includegraphics[width=\textwidth]{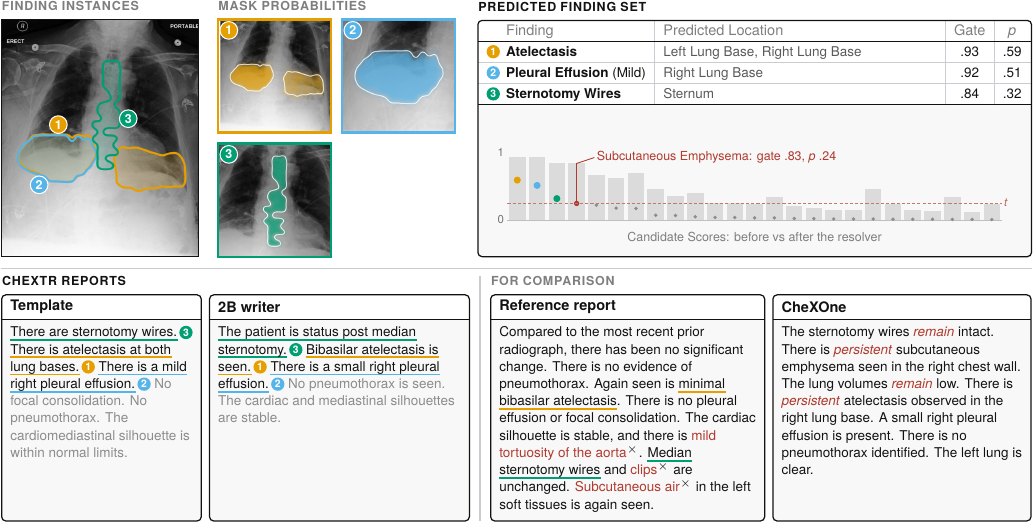}
  \caption{\textbf{Minor findings missed.} Atelectasis and the sternotomy wires match the reference. The pleural effusion is a false positive that CheXOne shares, and subcutaneous emphysema, surgical clips and aortic tortuosity are missed, the first with a resolver score just below $t$. \underline{Underlined} sentences state the numbered instance, \textcolor[HTML]{8C8C8C}{grey} ones are normal statements without an instance, \textcolor[HTML]{B03A2E}{\emph{red italics}} assert change without a prior image and \textcolor[HTML]{B03A2E}{red}$^{\times}$ marks a reference finding that the set does not hold.}
  \label{fig:supp-qual-typical}
\end{figure*}

\begin{figure*}[t]
  \centering
  \includegraphics[width=\textwidth]{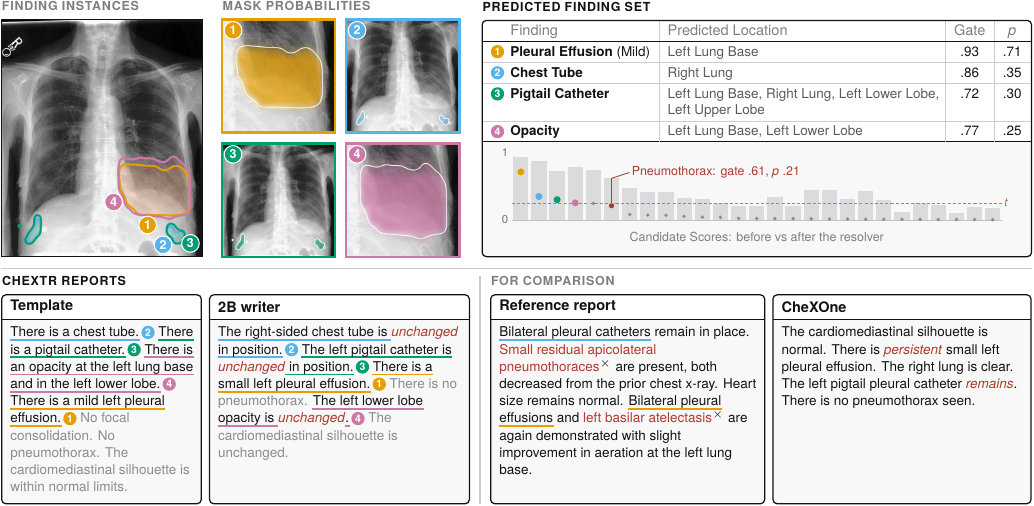}
  \caption{\textbf{Missed finding.} The reference describes small residual pneumothoraces beside two pleural catheters. The candidate for pneumothorax passes the gate, but the resolver scores it just below $t$ and drops it, so the template and the writer both state that there is no pneumothorax, as does CheXOne. The writer also calls three findings unchanged although it sees no prior image. Marks are as in \cref{fig:supp-qual-typical}.}
  \label{fig:supp-qual-miss}
\end{figure*}

\begin{figure*}[t]
  \centering
  \includegraphics[width=\textwidth]{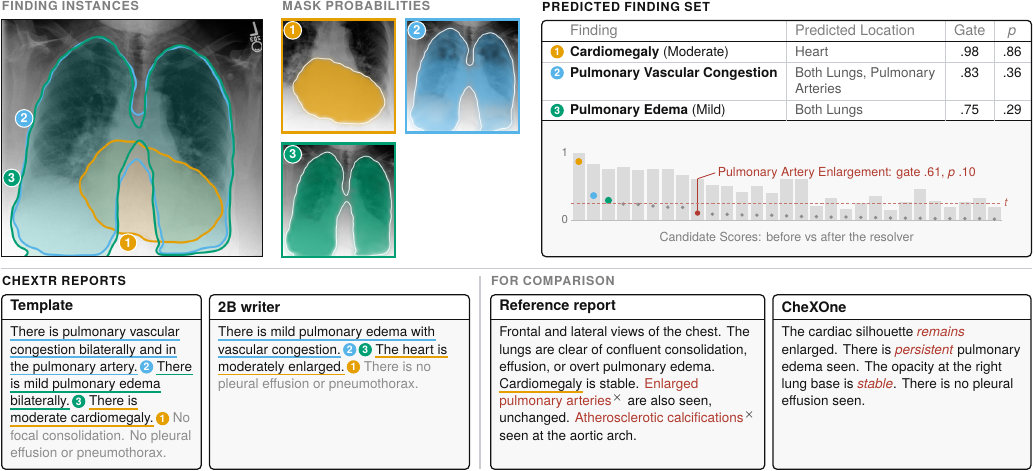}
  \caption{\textbf{Confused finding.} The reference reports enlarged pulmonary arteries and no pulmonary edema. Pulmonary artery enlargement is among the candidates, but the resolver scores it far below $t$ and keeps pulmonary vascular congestion and mild pulmonary edema, which both reports then state. CheXOne reports pulmonary edema as well. Marks are as in \cref{fig:supp-qual-typical}.}
  \label{fig:supp-qual-confused}
\end{figure*}

\begin{figure*}[t]
  \centering
  \includegraphics[width=\textwidth]{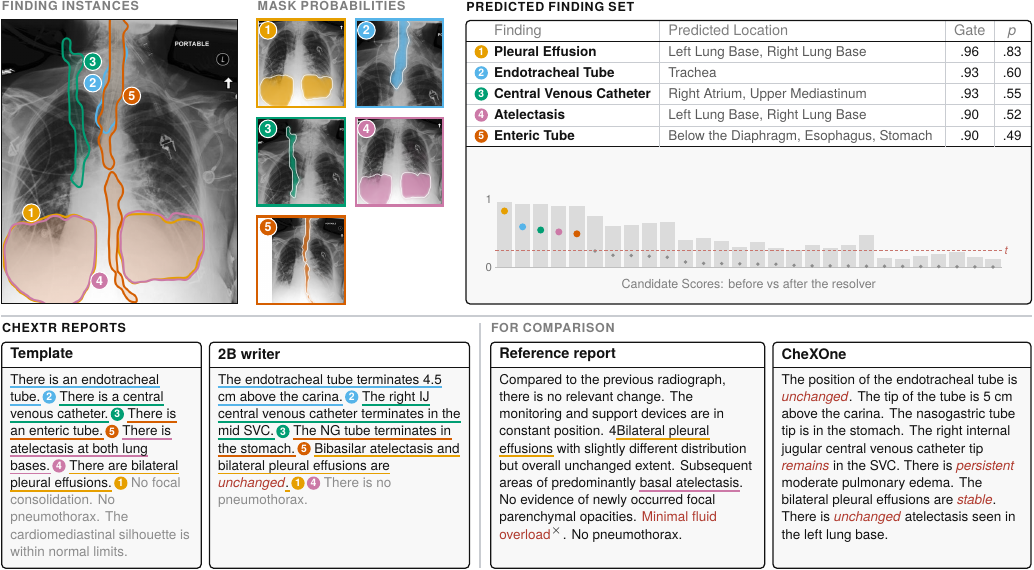}
  \caption{\textbf{Devices beyond the reference parse.} The reference mentions the monitoring and support devices only collectively, so its parse holds no device and the three predicted devices count as false positives, although CheXOne describes the same three. Fluid overload is missed. The writer states a tip distance for the endotracheal tube that its input does not hold and calls two findings unchanged. Marks are as in \cref{fig:supp-qual-typical}.}
  \label{fig:supp-qual-devices}
\end{figure*}

\Cref{fig:supp-qual-typical,fig:supp-qual-miss,fig:supp-qual-confused,fig:supp-qual-devices}
show four further MIMIC-CXR test studies in the layout of
\cref{fig:qualitative}. We picked them by hand to show errors, among the test
studies with three to six reference findings and two to six predicted findings.
The F1 between the predicted set and the reference parse of a study is at the
median of the test studies for three of the cases and in the lowest quarter for
the case of \cref{fig:supp-qual-confused}. \Cref{fig:supp-qual-openi} adds an
OpenI study, also picked by hand. In three of the cases the resolver scores
a correct candidate below $t$ and drops it, which is the recall that the
fitted threshold gives up for precision.

\twocolumn[{%
  \centering
  \includegraphics[width=\textwidth]{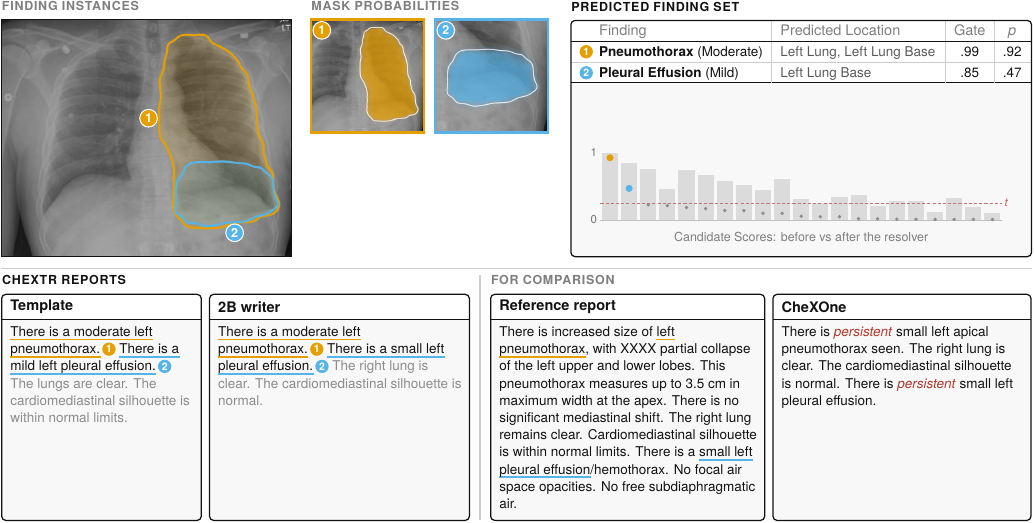}
  \captionof{figure}{\textbf{Transfer to OpenI.} An OpenI study, on which the predictor, the resolver and the writer run unchanged. Both predicted findings appear in the reference, which measures the pneumothorax at 3.5\,cm. The predicted set grades it as moderate, whereas CheXOne calls it small and apical and calls both findings persistent without a prior image. Marks are as in \cref{fig:supp-qual-typical}.}
  \label{fig:supp-qual-openi}
  \vspace{\dbltextfloatsep}
}]

\end{document}